\documentclass{article}
\usepackage{arxiv,times}

\usepackage{amsmath,amsfonts,bm}

\def\eqref#1{equation~\ref{#1}}

\def\1{\bm{1}}

\DeclareMathAlphabet{\mathsfit}{\encodingdefault}{\sfdefault}{m}{sl}
\SetMathAlphabet{\mathsfit}{bold}{\encodingdefault}{\sfdefault}{bx}{n}

\newcommand{\E}{\mathbb{E}}

\newcommand{\R}{\mathbb{R}}

\newcommand{\KL}{D_{\mathrm{KL}}}

\usepackage{setspace}
\usepackage{amsfonts}
\usepackage{bbm}
\usepackage{booktabs}
\usepackage{enumitem}
\usepackage[T1]{fontenc}
\usepackage[utf8]{inputenc}
\usepackage{makecell}
\usepackage{microtype}
\usepackage{multicol,multirow}
\usepackage{nicefrac}
\usepackage{tabularx}
\usepackage{url}

\usepackage{longtable}
\usepackage{catchfile,pgffor}

\usepackage{environ}
\usepackage[percent]{overpic}
\usepackage[dvipsnames]{xcolor}

\usepackage{ragged2e}
\usepackage{float}
\usepackage[ruled,vlined,linesnumbered]{algorithm2e}
\usepackage{placeins}

\newcolumntype{L}{>{\RaggedRight\arraybackslash}X}

\usepackage{amsmath}
\usepackage{amssymb}
\usepackage{booktabs}
\usepackage{graphicx}
\usepackage{makecell}
\usepackage{multirow}
\usepackage{pifont}%
\usepackage{tabularx}
\usepackage{caption}
\usepackage{soul}
\usepackage{changepage,threeparttable}
\usepackage{colortbl}
\usepackage[accsupp]{axessibility}
\usepackage{mathtools}
\usepackage{xfrac}
\usepackage{enumitem}
\usepackage{catchfile}
\usepackage{wrapfig}
\usepackage{needspace}
\newsavebox{\experimentinsetbox}
\newenvironment{experimentinset}[3]{%
    \par
    \sbox{\experimentinsetbox}{%
        \begin{minipage}{#1}
            \captionsetup{type=table,skip=6pt,justification=raggedright,singlelinecheck=false}
            \centering
            #2%
        \end{minipage}%
    }%
    \Needspace{\dimexpr\ht\experimentinsetbox+\dp\experimentinsetbox+3\baselineskip\relax}%
    #3%
    \begin{wraptable}{r}{#1}
        \vspace{-\intextsep}
        \usebox{\experimentinsetbox}%
        \vspace{-\intextsep}
    \end{wraptable}%
    \ignorespaces
}{%
    \par
    \ifnum\value{WF@wrappedlines}>1
        \vspace{\dimexpr\value{WF@wrappedlines}\baselineskip-\baselineskip\relax}%
    \fi
    \WFclear
    \ignorespacesafterend
}

\newcommand{\tabcite}[1]{{\scriptsize\citep{#1}}}

\DeclareRobustCommand{\checkpointmark}{\textsuperscript{\textdagger}}
\DeclareRobustCommand{\reportedmark}{\textsuperscript{\textdaggerdbl}}

\usepackage{longtable}
\usepackage{tabu}
\usepackage{tcolorbox}
\usepackage{supertabular}
\usepackage{caption}
\usepackage{subcaption}
\usepackage{pifont}
\usepackage{adjustbox}
\usepackage{titletoc}
\usepackage{multicol}
\newcolumntype{Y}{>{\centering\arraybackslash}X}
\DeclarePairedDelimiter{\norm}{\lVert}{\rVert}

\usepackage{hyperref}
\usepackage[capitalize,nameinlink]{cleveref}

\DeclareMathOperator*{\argmin}{arg\,min}

\usepackage{amsthm}
\newtheorem{theorem}{Theorem}
\newtheorem{proposition}{Proposition}

\newtheorem{corollary}{Corollary}

\crefname{equation}{Eq.}{Eqs.}
\Crefname{equation}{Eq.}{Eqs.}
\crefname{figure}{Fig.}{Figs.}
\Crefname{figure}{Fig.}{Figs.}
\crefname{table}{Tab.}{Tabs.}
\Crefname{table}{Tab.}{Tabs.}
\crefname{section}{Sec.}{Secs.}
\Crefname{section}{Sec.}{Secs.}
\Crefname{section}{Sec.}{Secs.}
\crefname{algorithm}{Alg.}{Algs.}
\Crefname{algorithm}{Alg.}{Algs.}
\crefname{algocf}{Alg.}{Algs.}
\Crefname{algocf}{Alg.}{Algs.}
\crefname{appendix}{Appendix}{Appendices}
\Crefname{appendix}{Appendix}{Appendices}

\newcommand{\Ours}[0]{\texttt{sDMD}}
\newcommand{\Oursbf}[0]{\fontfamily{lmtt}\fontseries{b}\selectfont \Ours{}}
\newcommand{\Oursu}[0]{\texttt{sDMD-U}}

\newcommand{\ie}{i.e.,}

\definecolor{color_1}{RGB}{255,0,128}
\definecolor{color_2}{RGB}{128,128,0}
\definecolor{color_3}{RGB}{0,128,0}
\definecolor{color_4}{RGB}{128,0,0}
\definecolor{color_5}{RGB}{128,0,128}

\definecolor{Goldenrod}{RGB}{218, 165, 32}
\definecolor{ForestGreen}{RGB}{34, 139, 34}
\definecolor{velvet}{RGB}{170, 60, 220}

\definecolor{highlight_color}{RGB}{255,0,128}
\definecolor{blue_highlight_color}{RGB}{0,128,255}
\definecolor{green_highlight_color}{RGB}{51,145,40}
\definecolor{orange_highlight_color}{RGB}{255,160,0}
\definecolor{purple_highlight_color}{RGB}{128,0,128}

\title{sDMD: Sinkhorn Distribution Matching Distillation}

\author{%
  \textbf{Jaihoon~Kim}$^{1,2\dagger}$,
  \textbf{Lingxiao~Li}$^{2}$,
  \textbf{Soshi~Shimada}$^{2}$, \\
  \textbf{Roshni~Cooper}$^{2}$,
  \textbf{Minhyuk~Sung}$^{1\ddagger}$,
  \textbf{Yiwei~Zhao}$^{2\ddagger}$ \\
  [0.2em]
  $^{1}$KAIST, $^{2}$Netflix \\
  [0.2em]
  \texttt{\{jh27kim, mhsung\}@kaist.ac.kr} \\
  \texttt{\{lingxiaol, soshis, roshnic, yiweiz\}@netflix.com} \\[0.2em]
  \textcolor{black}{Project Page:} \href{https://sinkhorn-dmd.github.io/}{\textcolor{magenta}{\texttt{https://sinkhorn-dmd.github.io/}}}
}

\newif\ifpaper
\paperfalse

\begin{document}

\maketitle

\begingroup
\renewcommand{\thefootnote}{}
\footnotetext{$^{\dagger}$Work done during the internship at Netflix.}
\footnotetext{$^{\ddagger}$Corresponding authors.}
\endgroup

\begin{abstract}
Distribution Matching Distillation (DMD) distills multi-step diffusion models into few-step generators by minimizing reverse KL divergence.
Despite its strong performance, DMD guides samples by local density alone, which introduces two limitations: teacher scores can be unreliable in regions of low teacher density, and the mode-seeking behavior of reverse KL can collapse the student onto a subset of the teacher modes.
We first connect DMD to Wasserstein gradient flow, showing that its gradient follows from a fixed-point formulation with integral KL (IKL) as the energy functional.
This perspective motivates augmenting IKL with an optimal transport term that addresses these limitations by guiding outlier samples toward the target and promoting mode coverage.
We realize this term using Sinkhorn divergence estimated from finite batches.
In practice, however, errors in the estimated Sinkhorn velocity can perturb in-domain samples for which IKL guidance is already accurate.
To this end, we introduce \Ours{}, which adaptively activates the Sinkhorn correction based on discriminator outputs, focusing on outlier samples where IKL may be unreliable.
With 2--4 sampling steps on FLUX.1-dev, Qwen-Image-20B, and Wan2.1-14B, \Ours{} outperforms distillation baselines on GenEval, DPG, WISE, and VBench while improving sample diversity over DMD2.
\end{abstract}

\vspace{-4pt}\section{Introduction}\vspace{-4pt}
\label{sec:introduction}

\begin{figure}[p]
    \vspace*{-0.6in}
    \centering
    \captionsetup{font=normalsize}
    \caption{
        \textbf{Teaser.}
        4-step generated images using \Ours{} with FLUX.1 and Qwen-Image.
    }
    \label{fig:qwen_sdmd_teaser}
    \makebox[\textwidth][c]{\includegraphics[height=0.98\textheight]{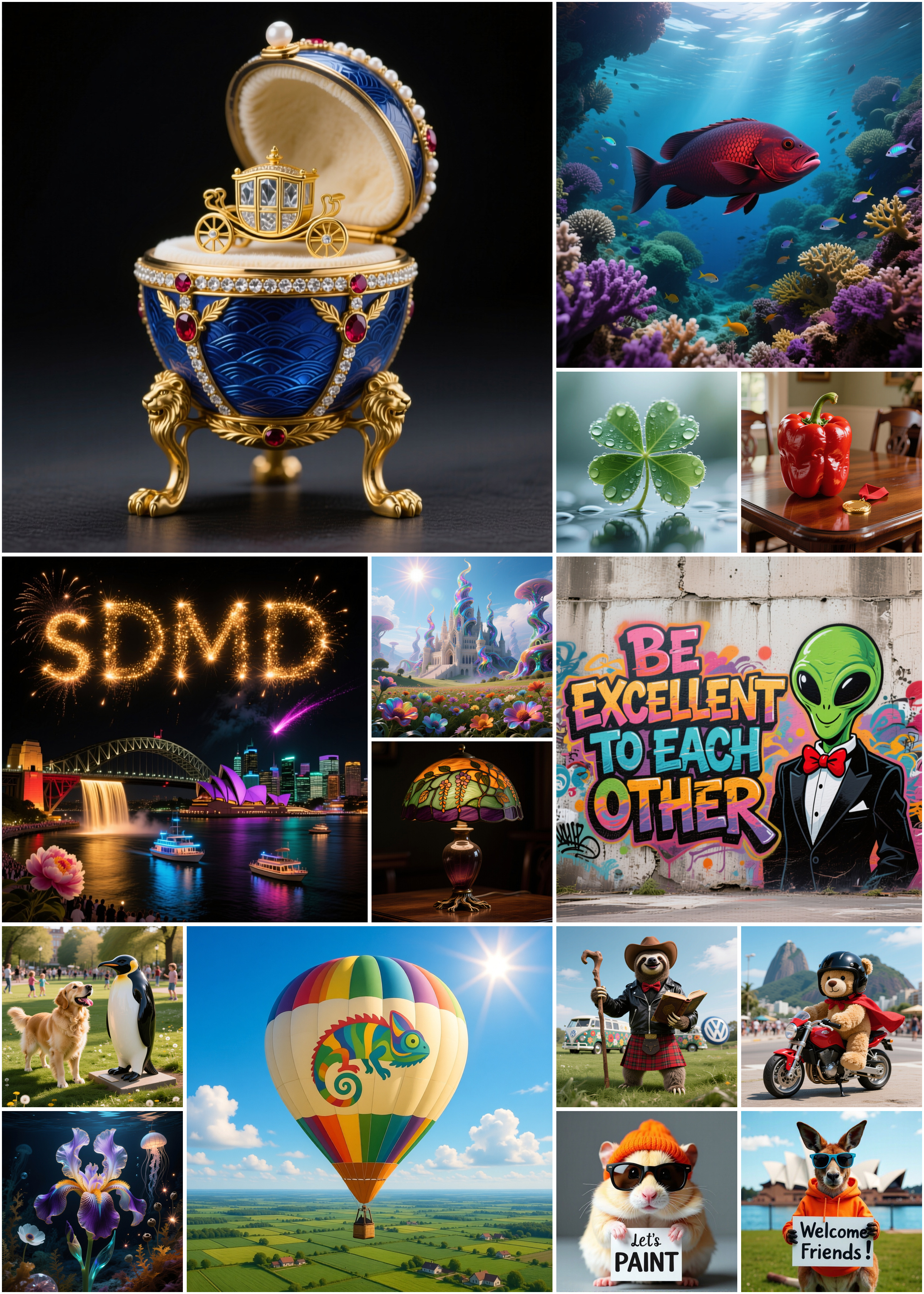}}
\end{figure}

Diffusion models~\citep{sohl2015:deep,song2021:sde,ho2020:ddpm} generate high-quality images and videos but require many sampling steps.
Diffusion distillation reduces this cost by transferring knowledge from a pretrained teacher to a student that generates samples in one or a few steps.
Existing approaches have advanced few-step generation by learning mappings along teacher trajectories~\citep{geng2026:meanflow,song2023:consistency}, using adversarial supervision~\citep{sauer2024:ladd,lin2025:apt}, or directly matching student and teacher distributions~\citep{xu2025:fdistill,zhou2024:sid}.
Among these, a representative approach is DMD~\citep{yin2024:dmd}, which optimizes a reverse KL objective using the difference between teacher and student scores.
It has been widely adopted due to its high sample fidelity and scalability in large-scale image and video generation~\citep{fan2025:qwenlightning,zheng2026:rcm,ge2025:senseflow}.

To better understand and improve the training dynamics of DMD, we view diffusion distillation from the perspective of Wasserstein gradient flow (WGF)~\citep{ambrosio2008:gradient}.
WGF evolves a probability distribution by steepest descent of an energy functional in Wasserstein space, with the corresponding velocity field specifying the update direction for each sample.
While recent works~\citep{han2026:wflow,he2026:sinkhorndrifting} use WGF to train one-step generators from scratch, we establish its connection to diffusion distillation by recovering the DMD gradient through fixed-point regression under the integral KL (IKL) energy, defined as reverse KL averaged over diffusion noise levels (\cref{sec:dmd_wgf}).
This identifies the energy underlying DMD and the velocity field that guides generated samples, allowing us to analyze the behavior and limitations of DMD and explore complementary energy functionals to improve distillation.

The IKL velocity of DMD reveals two practical limitations (\cref{fig:motivation}).
First, learned teacher scores can be unreliable for outlier samples, and the resulting updates may fail to guide them toward target modes.
Although adding diffusion noise can improve distributional overlap, outliers may remain poorly supported at low noise levels.
Higher noise levels can improve their support but blur the local structure of target modes.
Second, the mode-seeking behavior of reverse KL can concentrate the student on a subset of target modes, potentially reducing the diversity of generated samples.

To address these limitations, we augment IKL with an optimal transport (OT) term.
OT couples student and target samples globally, providing robust guidance for outlier samples and promoting mode coverage without relying on learned teacher scores.
To exploit their complementary strengths, we propose combining global OT guidance with local IKL score updates that help capture fine details.
Our analysis shows that sufficiently strong exact OT regularization ensures convergence of the combined flow to the target under regularity assumptions (\cref{sec:motivation}).
For efficient computation, we use Sinkhorn divergence, a debiased form of entropic OT.
While Sinkhorn does not automatically inherit this guarantee, we establish sufficient conditions on its approximation error and regularization strength for the combined flow to converge to the target (\cref{sec:sdmd}).

In practice, the transport maps used to compute Sinkhorn updates are estimated from finite samples, with larger batches generally reducing estimation error at greater computational cost~\citep{genevay2019:sample}.
Notably, our experiments show that even with relatively few samples, Sinkhorn can provide more informative guidance than IKL for outlier samples where learned teacher scores are unreliable.
However, for well-supported samples, estimation errors in Sinkhorn can instead interfere with the accurate guidance already provided by IKL.
To this end, we introduce \Ours{}, which adaptively activates Sinkhorn guidance for samples with relatively weak teacher support based on discriminator outputs.
We reuse the discriminator trained for adversarial regularization in DMD2~\citep{yin2024:dmd2}, without training a separate model.

We evaluate \Ours{} using 2--4 sampling steps on FLUX.1-dev~\citep{flux2024} and Qwen-Image-20B~\citep{wu2025qwenimage} for text-to-image generation (\cref{fig:qwen_sdmd_teaser}) and Wan2.1~\citep{wan2025} (1.3B and 14B) for text-to-video generation.
\Ours{} outperforms the distillation baselines in most metrics on GenEval~\citep{ghosh2023:geneval}, DPG~\citep{hu2024:ella}, and WISE~\citep{niu2025:wise}, achieves the highest overall VBench scores on both Wan2.1 models, and improves all reported diversity metrics over the strong DMD2~\citep{yin2024:dmd2} baseline.

\vspace{-2pt}\section{Related Work}\vspace{-2pt}
\label{sec:related_work}

\paragraph{Diffusion Distillation.}

To reduce the sampling cost of multi-step diffusion models, trajectory-based distillation methods enable one or few-step generation by learning shortcuts along denoising trajectories~\citep{song2023:consistency,boffi2025:flowmaps,frans2025:shortcut}.
Continuous-time formulations use JVPs~\citep{lu2025:scm,geng2026:meanflow}, whose efficient computation at scale requires specialized implementations~\citep{zheng2026:rcm}.
Recent works aim to avoid JVP computation through specialized architectures that decouple sampling substeps from backbone evaluations~\citep{chen2026:piflow,shaul2026:pdd}.
Adversarial approaches use discriminator-based supervision for distillation~\citep{sauer2023:add,sauer2024:ladd} with potential training instability~\citep{lin2025:apt}.

Distribution-matching distillation methods align generated and target distributions through $f$-divergence minimization~\citep{xu2025:fdistill,xie2024:emdistill} or score matching~\citep{zhou2024:sid,luo2024:SiM}.
Among these, DMD~\citep{yin2024:dmd, luo2023:diffinstruct} minimizes the reverse KL divergence using the difference between teacher and student scores.
With their high sample fidelity and scalability, DMD-based methods have been widely adopted for high-resolution image synthesis and video generation~\citep{yin2025:causvid,huang2026:self-forcing,zheng2026:rcm,yin2024:dmd2}.
However, DMD remains sensitive to outliers and the mode-seeking behavior of reverse KL~\citep{bai2026:amd, luo2025:tdm}.
Our work augments DMD with a complementary velocity field to address these limitations and improve sample quality and diversity.

\paragraph{Wasserstein Gradient Flow for Generative Modeling.}

WGF~\citep{jordan1998:jko,ambrosio2008:gradient} evolves a probability distribution by steepest descent of an energy functional under the 2-Wasserstein metric.
When a divergence to the target distribution is chosen as the energy, the flow reduces the discrepancy between the two distributions, providing a framework for generative modeling.
For example, MMD flows~\citep{arbel2019:maximum} depend on kernel gradients and can provide weak guidance when generated and target samples are far apart relative to the kernel bandwidth.
This requires careful kernel design, such as noise-adaptive kernels~\citep{galashov2025:deepmmd} or kernels built from multiple feature encoders and scales~\citep{feng2026:representation,deng2026:drifting}.

Sinkhorn divergence provides transport-based guidance~\citep{genevay2018:learning,feydy2019:interpolating}, and recent works~\citep{han2026:wflow,he2026:sinkhorndrifting} use its gradient flow to train one-step generators.
However, estimating transport from finite batches can introduce error into the resulting velocity field~\citep{pooladian2021:entropic,genevay2019:sample}.
Flows based on reverse KL between clean distributions use local score information to guide samples toward the target distribution~\citep{li2023:scvm,liu2017:stein,he2025:diffusivekl}.
However, these approaches require known target densities or score estimates obtained through methods such as kernel density estimation, which can limit scalability in high-dimensional generation~\citep{he2025:diffusivekl,han2026:wflow}.
In contrast, DMD offers a scalable diffusion distillation approach by applying reverse KL to noise-perturbed distributions with noise annealing.
In this work, we connect DMD to WGF and propose \Ours{}, which incorporates an optimal transport term to improve guidance for outlier samples and promote mode coverage.

\vspace{-4pt}\section{Preliminaries}\vspace{-2pt}
\label{sec:background}

\vspace{-2pt}\paragraph{Problem Setup and Notation.}
\label{sec:prob_def}
We work on $\R^d$ and denote by $\mathcal{P}_2(\R^d)$ the set of probability distributions.
Let $p \in \mathcal{P}_2(\R^d)$ denote the target distribution and let $p_\mathrm{ref} = \mathcal{N}(0, I_d)$ be the reference distribution.
We assume access to a pretrained multi-step teacher for $p$ that gives scores at different noise levels.
Our goal is to distill this teacher into a student generator $f_\theta: \R^d \to \R^d$ that generates a sample in one or a few steps, such that its output distribution $q = (f_\theta)_\# p_\mathrm{ref}$ approximates $p$.\vspace{-4pt}

\vspace{-2pt}\paragraph{Distribution Matching Distillation (DMD).}
DMD~\citep{yin2024:dmd} distills a multi-step diffusion teacher into a few-step generator by matching the student distribution $q$ to the teacher distribution $p$.
The original formulation starts from reverse KL between clean distributions, whose gradient involves the difference between the student and target scores.
To approximate the gradient, it introduces diffusion annealing, replacing the clean-distribution scores with scores of noisy distributions.
For a few-step generated sample $x=f_\theta(z)$ with $z\sim p_\mathrm{ref}$, the forward diffusion process gives $x_t=\alpha_t x+\sigma_t\epsilon$, where $\epsilon\sim\mathcal{N}(0,I_d)$ and $(\alpha_t,\sigma_t)$ defines the noise schedule.
We denote the corresponding perturbed marginals by $q_t$ and $p_t$, then the distribution-matching gradient takes the form:
\begin{equation}
    \nabla_\mathrm{DMD} (\theta) = - \E_{t, z, \epsilon} \!\left[ \alpha_t \left( \nabla \log p_t(x_t) - \nabla \log q_t(x_t) \right) \frac{\partial f_\theta(z)}{\partial \theta} \right].
    \label{eq:dmd}
\end{equation}
In practice, the pretrained teacher provides target scores, while an auxiliary diffusion model learns the student score online from generated samples using a score-matching objective~\citep{vincent2011:dsm}.\vspace{-4pt}

\vspace{-2pt}\paragraph{Wasserstein Gradient Flow (WGF).}
Let $\mathcal{F}: \mathcal{P}_2(\R^d) \to \R$ be an energy functional on distributions that satisfies $\argmin_q \mathcal{F}(q) = p$.
WGF~\citep{ambrosio2008:gradient,jordan1998:jko} describes the evolution of the probability density $\{q_\tau\}_{\tau \geq 0}$, where $\tau$ denotes gradient-flow time.
The evolution is characterized by the following partial differential equation~\citep{villani2009:optimal}:
\begin{equation}
    \partial_\tau q_\tau + \nabla \cdot (q_\tau V_\tau) = 0,
    \qquad
    V_\tau(x) = -\nabla \frac{\delta \mathcal{F}}{\delta q}(q_\tau)(x),
    \label{eq:wgf}
\end{equation}
where the velocity field $V_\tau$ is the negative gradient of the first variation of $\mathcal{F}$.
Discretizing this flow in time, each iteration displaces every particle $x_\tau \sim q_\tau$ by an Euler step, \ie~$x_{\tau+1} = x_\tau + V_\tau(x_\tau)$.
At the equilibrium $q_\tau = p$, the velocity vanishes and each step leaves every particle in place.
Writing the particles as generator output $x = f_\theta(z)$, the fixed-point relation gives the following regression objective on generator parameters~\citep{deng2026:drifting}:
\begin{equation}
    \mathcal{L}(\theta) = \E_{z \sim p_\mathrm{ref}} \norm{ f_\theta(z) - \mathrm{sg}\!\left( f_\theta(z) + V_\tau(f_\theta(z)) \right) }^2 ,
    \label{eq:fixed_point}
\end{equation}
where $\mathrm{sg}(\cdot)$ denotes the stop-gradient, and the loss decreases as $V_\tau \to 0$.
Note that different energy functionals induce different velocity fields and shape the training dynamics.
Therefore, we connect DMD to WGF to explore the design space of energy functionals for distillation and introduce a complementary transport term to address the limitations of DMD.\vspace{-4pt}

\vspace{-2pt}\paragraph{Entropic Optimal Transport and Sinkhorn Divergence.}
Entropic optimal transport (EOT) compares $q$ and $p$ via a regularized transport cost which can be computed using the Sinkhorn algorithm~\citep{cuturi2013:sinkhorn,knight2008:sinkhorn}.
For the quadratic cost $c(x,y)=\tfrac{1}{2}\norm{x-y}^2$ and regularization strength $\beta>0$, EOT is defined as:
\begin{equation}
    \mathrm{OT}_\beta(q,p)
    = \min_{\pi\in\Pi(q,p)} \int c(x,y)\,d\pi(x,y)
    + \beta\KL(\pi\,\|\,q\otimes p),
    \label{eq:eot}
\end{equation}
where $\Pi(q,p)$ denotes the couplings with marginals $q$ and $p$.
The debiased Sinkhorn divergence removes entropic bias by subtracting the self-transport costs~\citep{genevay2018:learning}:
\begin{equation}
    S_\beta(q,p)
    = \mathrm{OT}_\beta(q,p)
    - \tfrac{1}{2}\mathrm{OT}_\beta(q,q)
    - \tfrac{1}{2}\mathrm{OT}_\beta(p,p).
    \label{eq:sinkhorn}
\end{equation}
$S_\beta$ is nonnegative and vanishes if and only if $q=p$~\citep{feydy2019:interpolating}.
Let $\pi_{q,p}$ denote the optimal entropic transport plan in \cref{eq:eot}, and define its barycentric projection as $T_{q,p}(x)=\int y\,d\pi_{q,p}(y\mid x)$, the weighted average of target points under the transport plan; $T_{q,q}(x)$ denotes the corresponding projection under the self-transport plan.
Then, the WGF velocity can be written as the difference between the two barycentric projections~\citep{genevay2018:learning}:
\begin{equation}
    V_{S_\beta}(x)
    = -\nabla\frac{\delta S_\beta(q,p)}{\delta q}(x)
    = T_{q,p}(x)-T_{q,q}(x).
    \label{eq:sinkhorn_velocity}
\end{equation}\vspace{-4pt}

\begin{figure}[t]\vspace{-3pt}
    \centering
    \caption{
        \textbf{IKL and Sinkhorn Velocity Fields.}
        We visualize velocity fields from a 2D Gaussian mixture experiment.
        Black circles mark student particles and contours show target density. Solid (dashed) arrows use exact (learned) teacher scores for IKL and large (small) batches for Sinkhorn.
        (a) For in-domain student samples, the IKL velocity carries the particles onto the target.
        (b) Outlier samples under poor initialization, the IKL velocity can be inaccurate in regions of low teacher density.
        (c) The mode-seeking behavior of IKL can concentrate particles on a subset of the target modes.
        (d) The Sinkhorn velocity guides particles toward the target modes through the transport plan, remaining robust even for outlier samples (\colorbox{blue_highlight_color!15}{\textcolor{black}{blue box}}).
        In practice, finite-sample errors in the Sinkhorn velocity can perturb samples for which IKL already provides reliable guidance (\colorbox{red!15}{\textcolor{black}{red boxes}}).
        We address this using our gated scheduler in \cref{subsec:lambda}.
    }
    \label{fig:motivation}
    \scriptsize
    \begin{subfigure}[t]{0.24\textwidth}
        \centering
        \includegraphics[width=\linewidth]{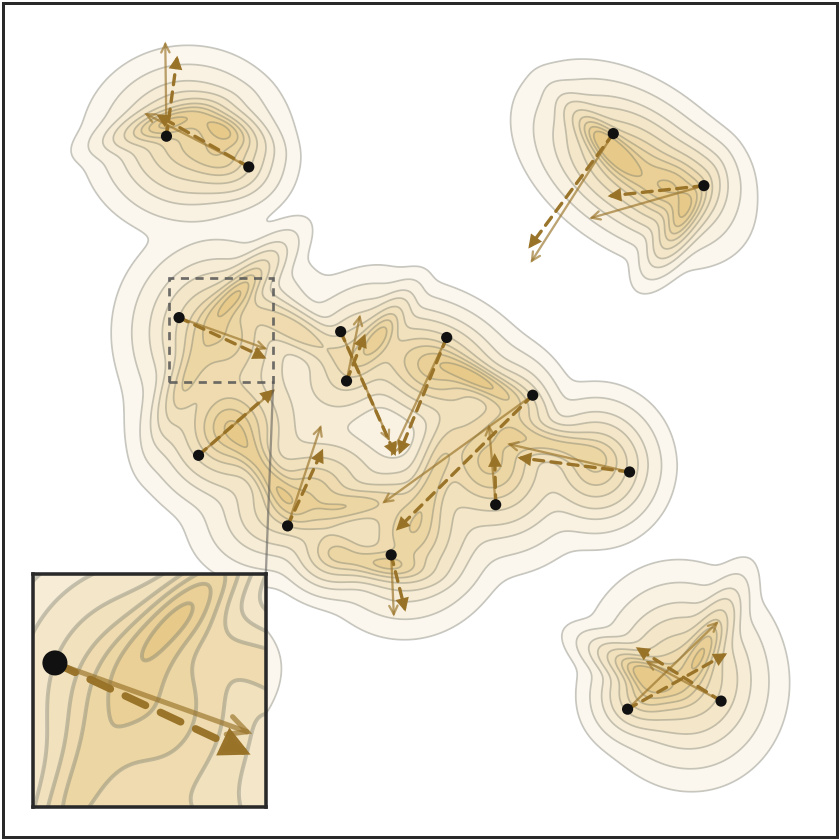}
        \caption{IKL: aligned samples.}
        \label{fig:mot_aligned}
    \end{subfigure}
    \hfill
    \begin{subfigure}[t]{0.24\textwidth}
        \centering
        \includegraphics[width=\linewidth]{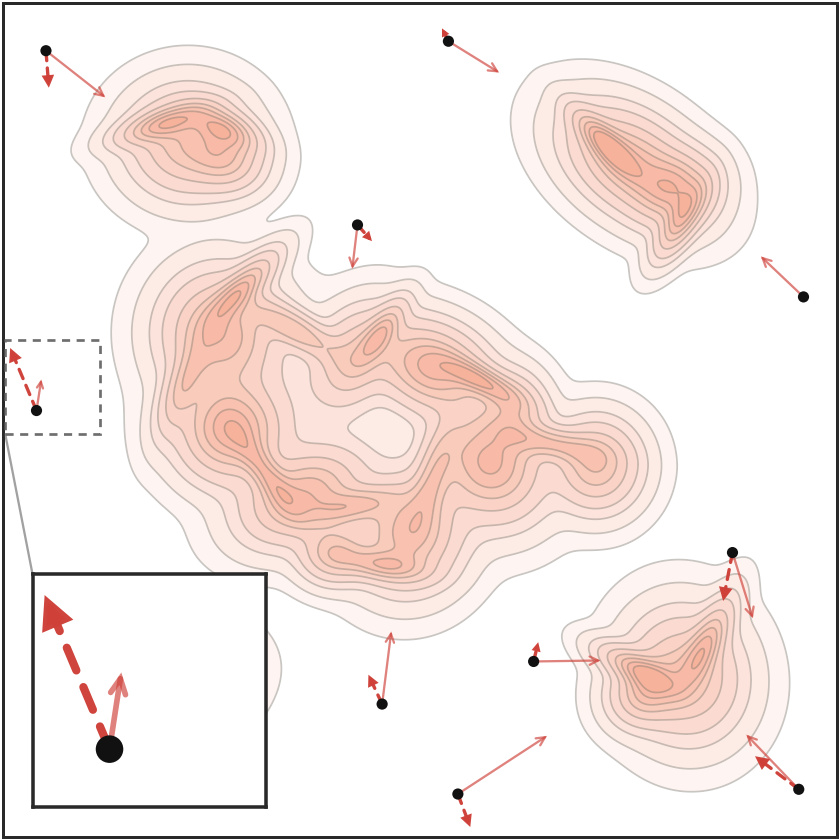}
        \caption{IKL: outlier samples.}
        \label{fig:mot_disjoint}
    \end{subfigure}
    \hfill
    \begin{subfigure}[t]{0.24\textwidth}
        \centering
        \includegraphics[width=\linewidth]{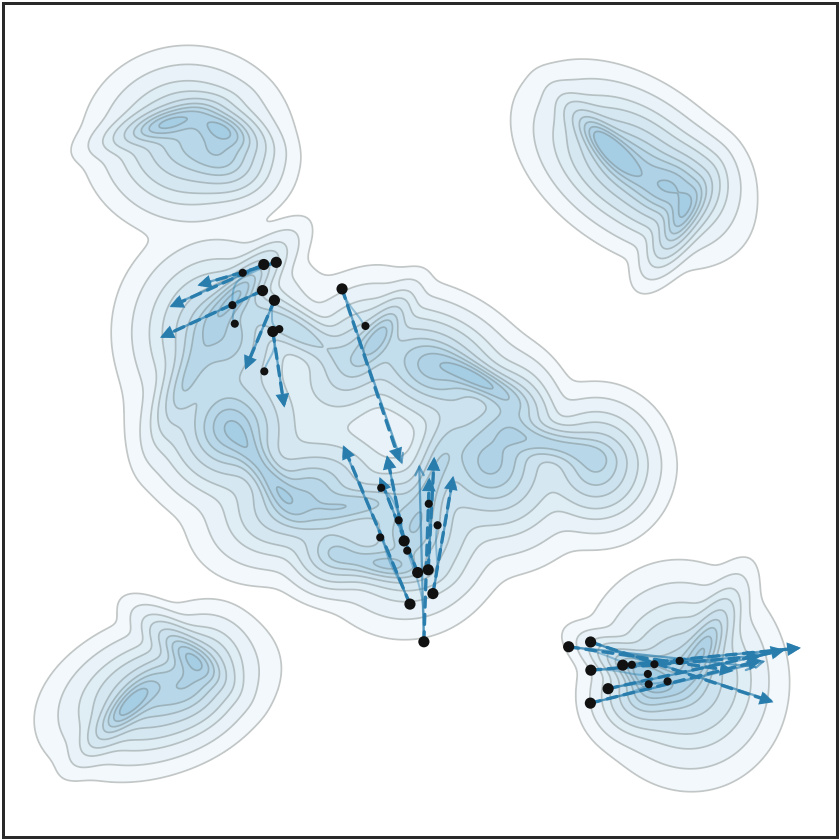}
        \caption{IKL: mode seeking.}
        \label{fig:mot_collapse}
    \end{subfigure}
    \hfill
    \begin{subfigure}[t]{0.24\textwidth}
        \centering
        \includegraphics[width=\linewidth]{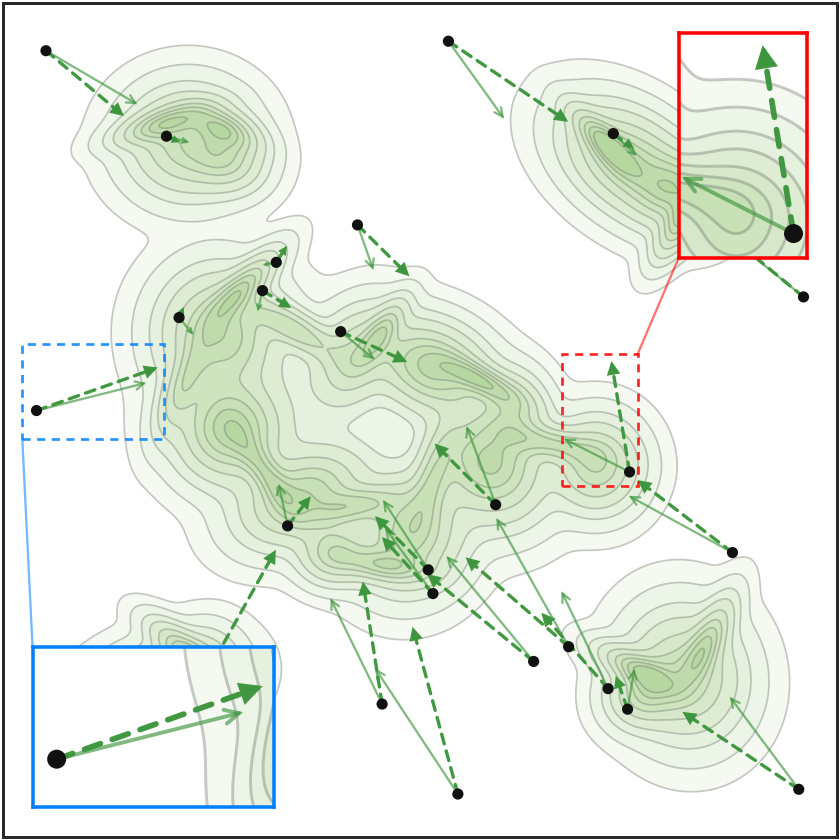}
        \caption{Sinkhorn transport.}
        \label{fig:mot_sinkhorn}
    \end{subfigure}
\vspace{-3pt}\end{figure}

\vspace{-6pt}
\section{Analyzing DMD through Wasserstein Gradient Flow}
\vspace{-2pt}
\label{sec:method}

\subsection{Connection of DMD to WGF}\vspace{-2pt}
\label{sec:dmd_wgf}
DMD~\citep{yin2024:dmd} has shown strong empirical performance in few-step generation, but the original derivation of the gradient introduces diffusion noise without a formal connection to WGF.
We show that the DMD gradient in \cref{eq:dmd} follows from the WGF fixed-point relation in \cref{eq:fixed_point}.
Specifically, the DMD update can be viewed through the WGF framework by choosing the integral KL (IKL) divergence~\citep{luo2023:diffinstruct},
\ie~reverse KL averaged over diffusion noise levels, as the energy functional:
\begin{align}
    \mathcal{F}_\mathrm{IKL}(q) = \E_t \!\left[ \KL(q_t \,\|\, p_t) \right],
    \quad
    V_\mathrm{IKL}(x) \overset{\mathrm{(i)}}{=} \E_{t,\epsilon} \!\left[ \alpha_t \left( \nabla \log p_t(x_t) - \nabla \log q_t(x_t) \right) \right],
    \label{eq:dmd_wgf}
\end{align}
where (i) follows by substituting $\mathcal{F}_\mathrm{IKL}$ into the WGF velocity in \cref{eq:wgf}.
Plugging this velocity into the fixed-point relation in \cref{eq:fixed_point} and taking the gradient gives:
\begin{equation}
    \nabla_\mathrm{IKL}(\theta) \coloneqq \nabla_\theta \E_z \!\left[ \norm{f_\theta(z) - \mathrm{sg}\!\left(f_\theta(z) + V_\mathrm{IKL}(f_\theta(z))\right)}^2 \right] = -2\E_z \!\left[ V_\mathrm{IKL}(f_\theta(z)) \frac{\partial f_\theta(z)}{\partial \theta} \right],
    \label{eq:ikl_gradient}
\end{equation}
which recovers the DMD gradient in \cref{eq:dmd} up to a constant after substituting $V_\mathrm{IKL}$ from \cref{eq:dmd_wgf}.
We present the detailed derivation in \cref{app:dmd_gradient}.
This connection allows us to analyze the behavior and limitations of DMD through its velocity field and explore synergistic energies to improve distillation.

\vspace{-2pt}\paragraph{Connection to Prior and Concurrent Work.}
Diff-Instruct~\citep{luo2023:diffinstruct} derives the generator score-difference update directly from IKL, but does not explicitly connect it to WGF.
ProlificDreamer~\citep{wang2023:prolificdreamer} derives a WGF update for text-to-3D particles, but does not consider distillation into few-step image or video generators.
Concurrent analyses~\citep{gretton2026:wgfdrifting,turan2026:scorematching} study WGF fixed-point regression for drifting models~\citep{deng2026:drifting} using kernel-smoothed KL rather than IKL, and also do not explicitly connect DMD to WGF.
In this work, we connect DMD to WGF under diffusion-noise-averaged KL from fixed-point regression and use this interpretation to introduce complementary Sinkhorn guidance for diffusion distillation.\vspace{-4pt}

\subsection{Limitations of IKL}\vspace{-2pt}
\label{sec:motivation}
DMD~\citep{yin2024:dmd} and its follow-up methods~\cite{fan2026:phased,ge2025:senseflow} have demonstrated empirical success in capturing fine details and scaling to high-resolution image generation~\citep{fan2025:qwenlightning, yin2024:dmd2} and video generation~\citep{gu2026:anyflow,zheng2026:rcm,huang2026:self-forcing}.
However, the gradient of IKL in DMD guides particles by local density alone, which brings two practical limitations.
First, the clean reverse KL gradient flow is ill-defined when $p$ and $q$ are mutually singular~\citep{glaser2021:kale}.
Although forward diffusion gives $p_t$ and $q_t$ full support, in practice the learned teacher score may remain unreliable for generated samples with low teacher density, particularly at low noise levels.
Second, the mode-seeking behavior of reverse KL can cause the student to cover only a subset of the teacher modes.

We illustrate these limitations using an experiment with a 2D Gaussian mixture (\cref{fig:motivation}).
When the two distributions are well aligned, the predicted score is accurate and the particles converge onto the target (\cref{fig:motivation}(a)).
However, when a generated sample lies in a low-density region of the teacher distribution, the predicted score becomes inaccurate, and the resulting velocity fails to carry the particles toward the target under weak diffusion annealing (\cref{fig:motivation}(b)).
With higher noise levels, the sample can be better supported, but the velocity can ignore the local structure of the target distribution, and the mode-seeking behavior of reverse KL can collapse the particles onto a single mode (\cref{fig:motivation}(c)).

\vspace{-2pt}\paragraph{Convergence with Exact OT.}
These limitations motivate augmenting IKL with an optimal transport (OT) regularizer.
We show that IKL alone does not guarantee convergence from arbitrary initializations, whereas the WGF of exact OT contracts toward the target.
Under regularity assumptions, a sufficiently strong exact OT term counteracts the negative curvature of IKL and ensures convergence of the combined flow to the target in Wasserstein distance (\cref{thm:exact_ot_contraction}).
We provide a failure case for IKL and a sufficient condition on the OT regularization strength for convergence in \cref{app:ot_regularization}.
In the next section, we introduce a practical transport correction that complements IKL.

\vspace{-6pt}\section{\Ours{}: Sinkhorn Distribution Matching Distillation}\vspace{-2pt}
\label{sec:sdmd}

\subsection{Complementing IKL with Sinkhorn Divergence}\vspace{-2pt}
\label{subsec:complementarity}

Despite the favorable convergence properties of exact OT, it can be computationally prohibitive.
Instead, we use Sinkhorn divergence, a computationally tractable, debiased form of entropic OT with more favorable sample complexity~\citep{cuturi2013:sinkhorn,genevay2019:sample}.
We compare this choice with other energy functionals, including forward KL and MMD, in \cref{sec:exp_ablation}.

IKL and Sinkhorn induce complementary velocity fields that are useful in different regimes.
IKL captures fine details through local score guidance in regions well supported by the teacher, whereas Sinkhorn uses a global transport plan to guide particles with weak teacher support toward the target and promote mode coverage (\cref{fig:motivation}(d)).

Combining the two gives \Oursu{}, which augments $\mathcal{F}_\mathrm{IKL}$ with a Sinkhorn divergence term:
\begin{equation}
    \mathcal{F}_\text{sDMD-U}(q) = \E_t \!\left[ \KL(q_t \,\|\, p_t) \right] + w \, S_\beta(q, p) ,
    \label{eq:sdmd_energy}
\end{equation}
where the hyperparameter $w$ weights the transport term and recovers $\mathcal{F}_\mathrm{IKL}$ at $w = 0$.
Both terms are nonnegative, vanish only at $q = p$, and \cref{eq:sdmd_energy} remains a valid energy functional for \cref{eq:wgf}.
By linearity of the first variation in the energy, the velocity field of the combined energy yields the weighted sum of the two fields:
\begin{equation}
\begin{aligned}
    V_\text{sDMD-U}(x) &= V_\mathrm{IKL}(x) + w \, V_{S_\beta}(x) \\
    &= \E_{t, \epsilon} \!\left[ \alpha_t \left( \nabla \log p_t(x_t) - \nabla \log q_t(x_t) \right) \right] + w \left( T_{q,p}(x) - T_{q,q}(x) \right) .
\end{aligned}
    \label{eq:sdmd_velocity}
\end{equation}
In \cref{app:identifiability}, we show that the velocity field is identifiable, \ie~$V_\text{sDMD-U}\equiv0$ if and only if $q=p$.
The score difference is obtained by passing the perturbed student samples $x_t$ through the teacher and student score networks.
Additionally, the transport term is computed using batches of student and teacher samples, \ie~$\{ x^{(i)} \}_{i=1}^{M_q} \sim q$ and $\{ y^{(i)} \}_{i=1}^{M_p} \sim p$, on which the Sinkhorn-Knopp algorithm returns the two projections of \cref{eq:sinkhorn_velocity}.
We discuss the space on which the transport is computed in \cref{sec:practical}, and study the role of $M_q$ and $M_p$ in \cref{sec:exp_ablation}.

Lastly, substituting \cref{eq:sdmd_velocity} into \cref{eq:fixed_point} gives the training objective of \Oursu{}:
\begin{equation}
    \mathcal{L}_\text{sDMD-U}(\theta) = \E_{z \sim p_\mathrm{ref}} \norm{ f_\theta(z) - \mathrm{sg}\!\left( f_\theta(z) + V_\mathrm{IKL}(x) + w \, V_{S_\beta}(x) \right) }^2 ,
    \qquad
    x = f_\theta(z).
    \label{eq:sdmd_loss}
\end{equation}

\paragraph{Convergence with Sinkhorn.}
We note that Sinkhorn does not automatically inherit the convergence guarantees of exact OT.
However, under the assumptions in \cref{app:sinkhorn_proxy}, a sufficiently large transport weight and a suitable bound on the Sinkhorn approximation error show convergence of the combined IKL and Sinkhorn flow to the target (\cref{thm:sinkhorn_relative_contraction}).
Empirically, \Oursu{} benefits from the complementary guidance of IKL and Sinkhorn.
As shown by the \colorbox{blue_highlight_color!15}{\textcolor{black}{blue box}} in \cref{fig:motivation}(d), Sinkhorn provides more informative guidance than IKL for outliers.

In practice, however, the transport plan is computed on finite batches of student and teacher samples, which introduces error in the resulting velocity~\citep{genevay2019:sample, pooladian2021:entropic}.
As a result, applying Sinkhorn at every iteration can perturb well-supported samples for which IKL already provides reliable guidance, as highlighted by the \colorbox{red!15}{\textcolor{black}{red boxes}} in \cref{fig:motivation}(d), while adding computational cost.
We therefore seek a criterion that adaptively activates the Sinkhorn correction for samples with relatively weak teacher support, for which IKL guidance may be unreliable.

\subsection{Gated Sinkhorn Scheduler}
\label{subsec:lambda}

In this section, we propose a practical criterion for identifying samples with relatively weak teacher support that can benefit from Sinkhorn transport.
During training, the few-step generator produces oversmoothed or distorted samples that are outside the typical regions of the target distribution.
We assume that these outlier samples have low density under the perturbed target distribution $p_t$ but higher density under the perturbed (fake) distribution $q_t$ of student outputs, \ie~$q_t(x_t) \gg p_t(x_t)$.
\footnote{This assumption is expected to hold primarily at low to moderate noise levels, as $p_t$ and $q_t$ approach the same prior at high noise levels.}

However, estimating these densities directly is challenging, and we instead exploit the ratio $q_t(x_t)/p_t(x_t)$, with larger values indicating relatively weak teacher support.
Specifically, we assess this density imbalance using a discriminator $D_\phi : \R^d \times [0, 1] \to (0, 1)$ trained to distinguish perturbed target samples from perturbed student samples:
\begin{equation}
    \mathcal{L}_\mathrm{adv}(\phi) = -\E_{t,\epsilon}\!\left[\E_{y\sim p}\log D_\phi(y_t,t)
    +\E_{x\sim q}\log\!\left(1-D_\phi(x_t,t)\right)\right],
    \label{eq:discriminator}
\end{equation}
where $y_t$ and $x_t$ are perturbed target and generated samples, respectively.
The discriminator is then used to define the following gating criterion:
\begin{equation}
    \lambda_\phi(x_t, t) = 1 - D_\phi(x_t, t) \approx \frac{q_t(x_t)}{p_t(x_t) + q_t(x_t)} \in (0, 1),
    \label{eq:lambda}
\end{equation}
where the approximation becomes exact for the optimal discriminator $D^\star(x_t,t)=p_t(x_t)/(p_t(x_t)+q_t(x_t))$.
A high value of $\lambda_\phi(x_t,t)$ suggests that $x_t$ has lower density under the target distribution than under the student distribution.
We assume that these samples are more likely to be out-of-distribution samples, where the learned teacher score and the resulting IKL (DMD) velocity may be unreliable.
We therefore use $\lambda$ as a threshold and activate the Sinkhorn term when $\lambda_\phi(x_t,t)>\lambda$:
\begin{equation}
\begin{aligned}
    V_\mathrm{sDMD}(x) &= V_\mathrm{IKL}(x) + w \cdot \mathbbm{1}\!\left[ \lambda_\phi(x_t, t) > \lambda \right] V_{S_\beta}(x) , \\
    \mathcal{L}_\mathrm{sDMD}(\theta, \phi) &= \E_{z \sim p_\mathrm{ref}} \norm{ f_\theta(z) - \mathrm{sg}\!\left( f_\theta(z) + V_\mathrm{sDMD}(x) \right) }^2 + \mathcal{L}_\mathrm{adv}(\phi,\theta),
\end{aligned}
    \label{eq:sdmd_gated}
\end{equation}
where $\mathbbm{1}[\cdot]$ denotes the indicator function.
Instead of training a separate discriminator, we incorporate the GAN objective of DMD2~\citep{yin2024:dmd2} and reuse its discriminator.
Here, $\mathcal{L}_\mathrm{adv}(\phi,\theta)$ denotes adversarial training with alternating updates to the generator and discriminator.
We denote the gated method by \Ours{} and the ungated variant in \cref{eq:sdmd_energy} by \Oursu{}, and compare them in \cref{sec:exp_ablation}.

\subsection{Practical Considerations}\vspace{0pt}
\label{sec:practical}

\paragraph{Choice of Transport Space.}
The transport in \cref{eq:sinkhorn_velocity} is computed on the sampled latents $x \in \mathbb{R}^d$ of the student and the teacher.
Instead, it can be computed in a feature space given a network $\psi: \mathbb{R}^d \rightarrow \mathbb{R}^n$, replacing the samples with the feature vectors $\psi(x) \in \mathbb{R}^n$.
We empirically find that Sinkhorn transport computed in the clean latent space already provides effective guidance for outlier samples, avoiding the additional cost of a feature encoder.
With this choice, \Ours{} incurs only marginal computational overhead compared with DMD2~\citep{yin2024:dmd2}.
We present an ablation of the transport space and an analysis of computational cost in \cref{sec:exp_ablation} and \cref{sec:additional_results}, respectively.\vspace{-4pt}

\vspace{-2pt}\paragraph{Conditional Generation.}
For conditional generation, we compute Sinkhorn transport between $M_q$ student samples and $M_p$ teacher samples generated under the same conditioning input.
To reduce training cost, we pre-generate the teacher samples for each conditioning input and store them in a buffer.
While previous works applied Sinkhorn divergence to class-conditioned generation~\citep{han2026:wflow,he2026:sinkhorndrifting}, we extend its use to text-conditioned generation.

\vspace{-4pt}\section{Experiments}\vspace{-5pt}
\label{sec:experiments}
We evaluate \Ours{} for text-to-image generation with Qwen-Image~\citep{wu2025qwenimage} and text-to-video generation with Wan2.1~\citep{wan2025}.
In Appendix, we provide implementation details (\cref{sec:impl_detail}), additional experimental results (\cref{sec:additional_results}), and qualitative results (\cref{sec:additional_qualitative}).

\begin{figure}[t]\vspace{-5pt}
    \centering
    \begin{threeparttable}
    \captionsetup{font=normalsize}
    \caption{
        \textbf{Qualitative Comparison on Text-to-Image Distillation.}
        \Ours{} more closely matches the diversity of the teacher than DMD-based Qwen-Lightning while preserving visual quality.
        \checkpointmark{} denotes an officially released checkpoint.
        See \cref{sec:additional_qualitative} for additional results.
    }
    \label{fig:qual_image}
    \setlength{\tabcolsep}{1.5pt}
    \renewcommand{\arraystretch}{1.05}
    \scriptsize
    \begin{tabular}{@{}>{\centering\arraybackslash}p{0.31\textwidth}@{\hspace{0.015\textwidth}}>{\centering\arraybackslash}p{0.31\textwidth}@{\hspace{0.015\textwidth}}>{\centering\arraybackslash}p{0.31\textwidth}@{}}
        \includegraphics[width=0.155\textwidth]{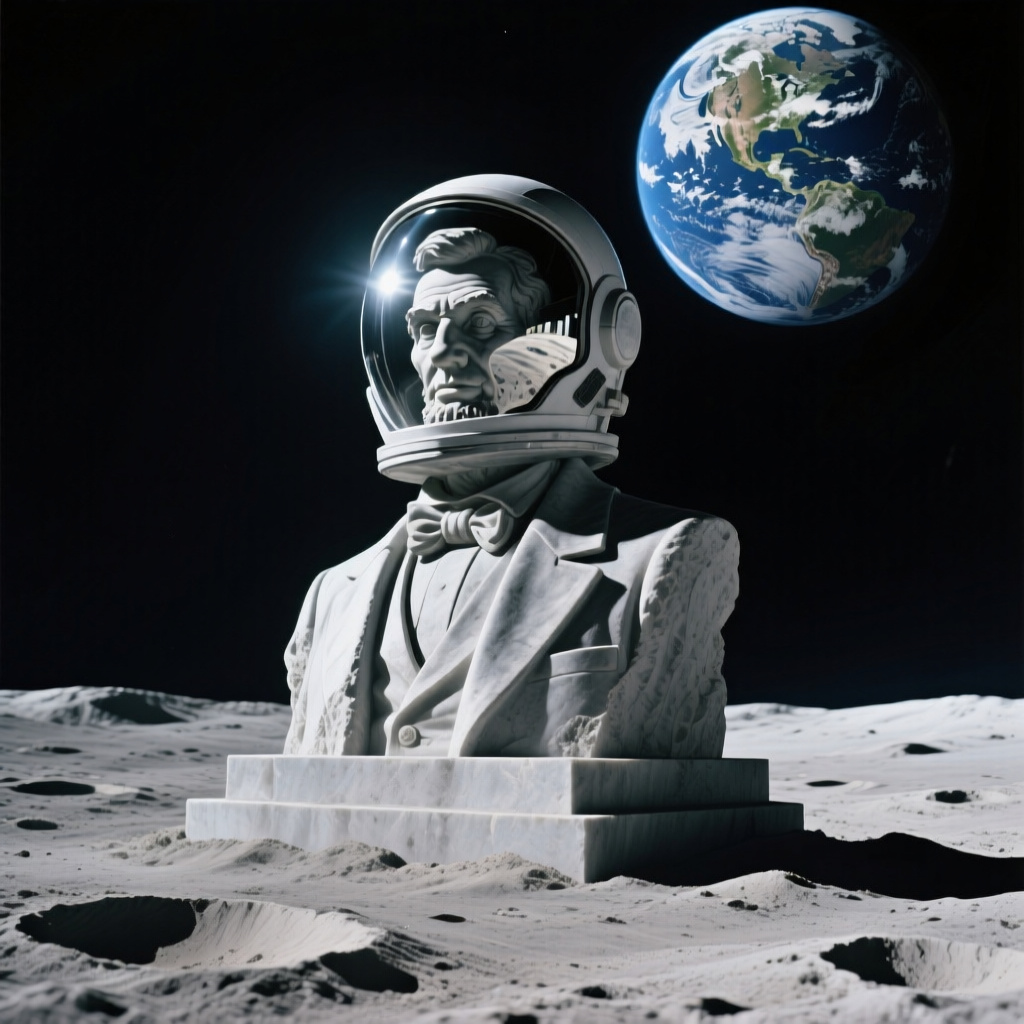}\includegraphics[width=0.155\textwidth]{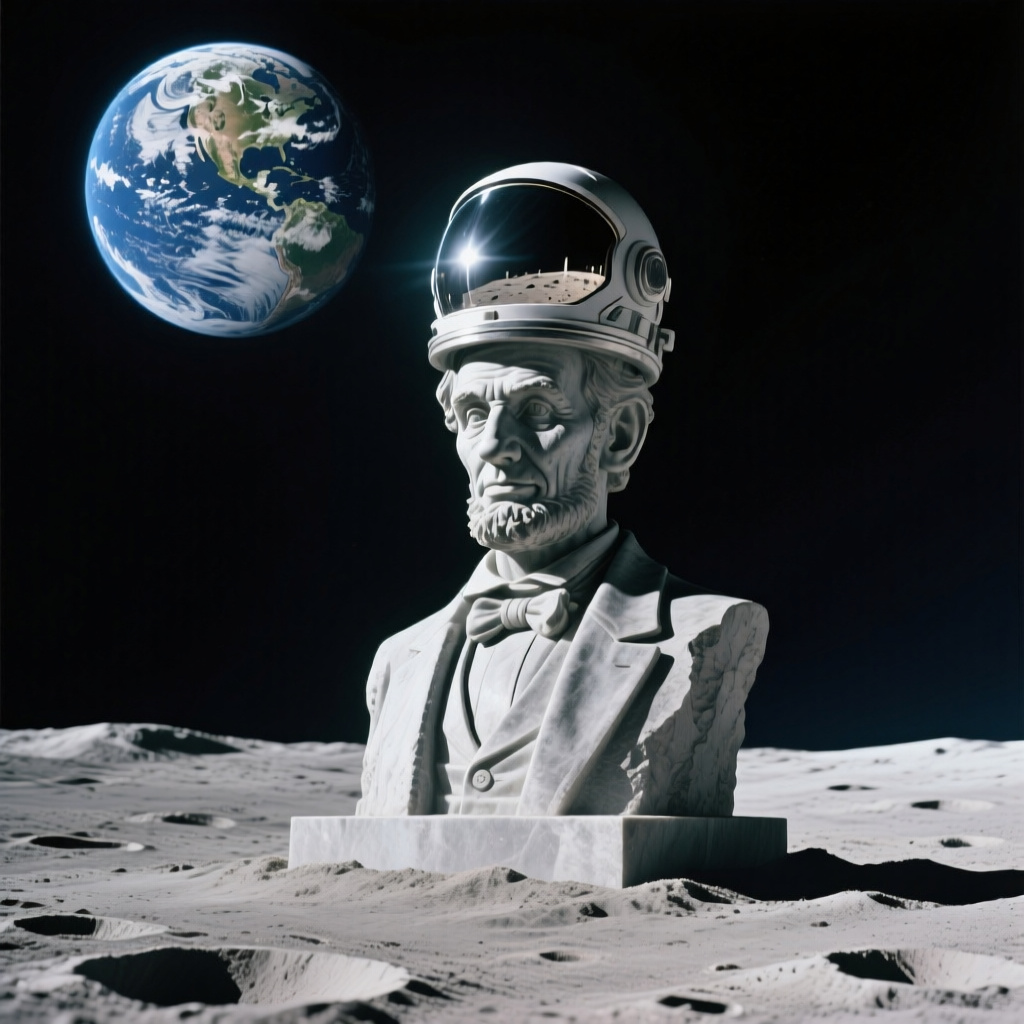} &
        \includegraphics[width=0.155\textwidth]{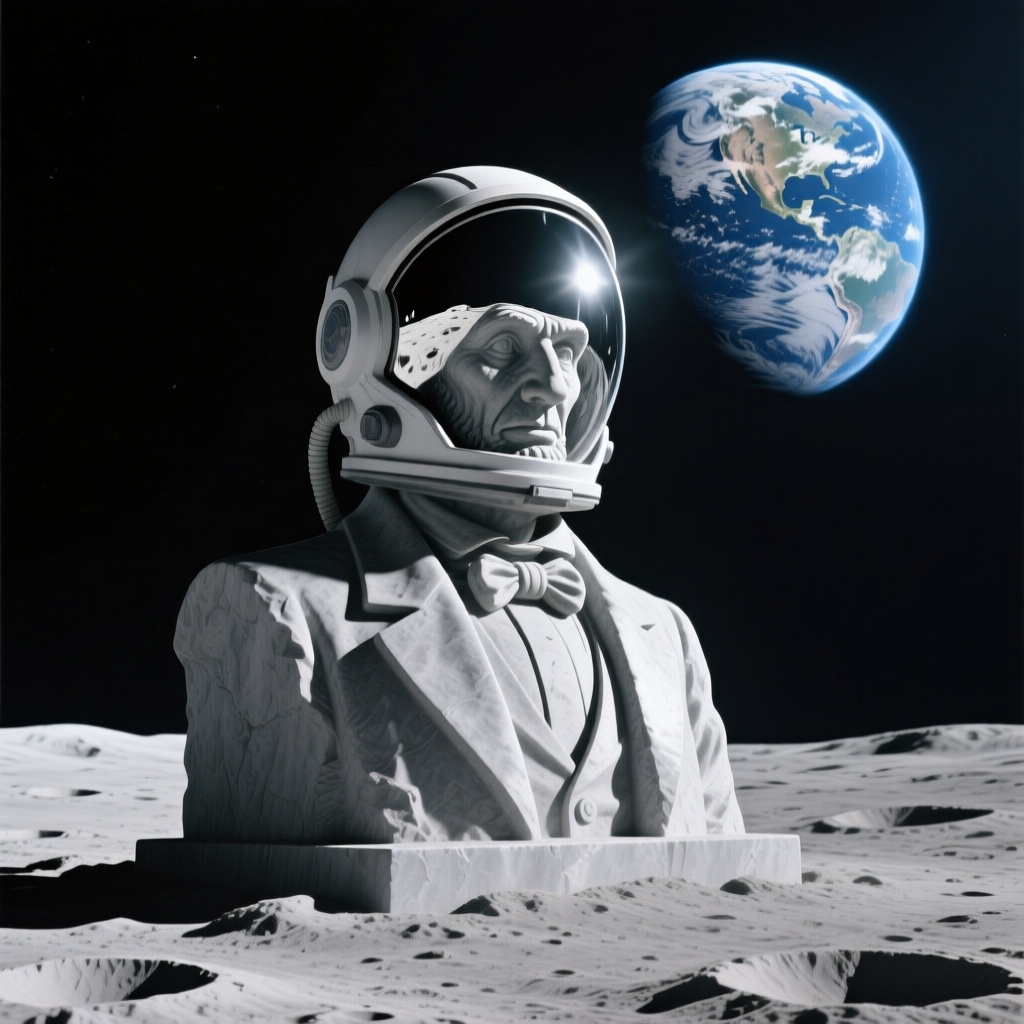}\includegraphics[width=0.155\textwidth]{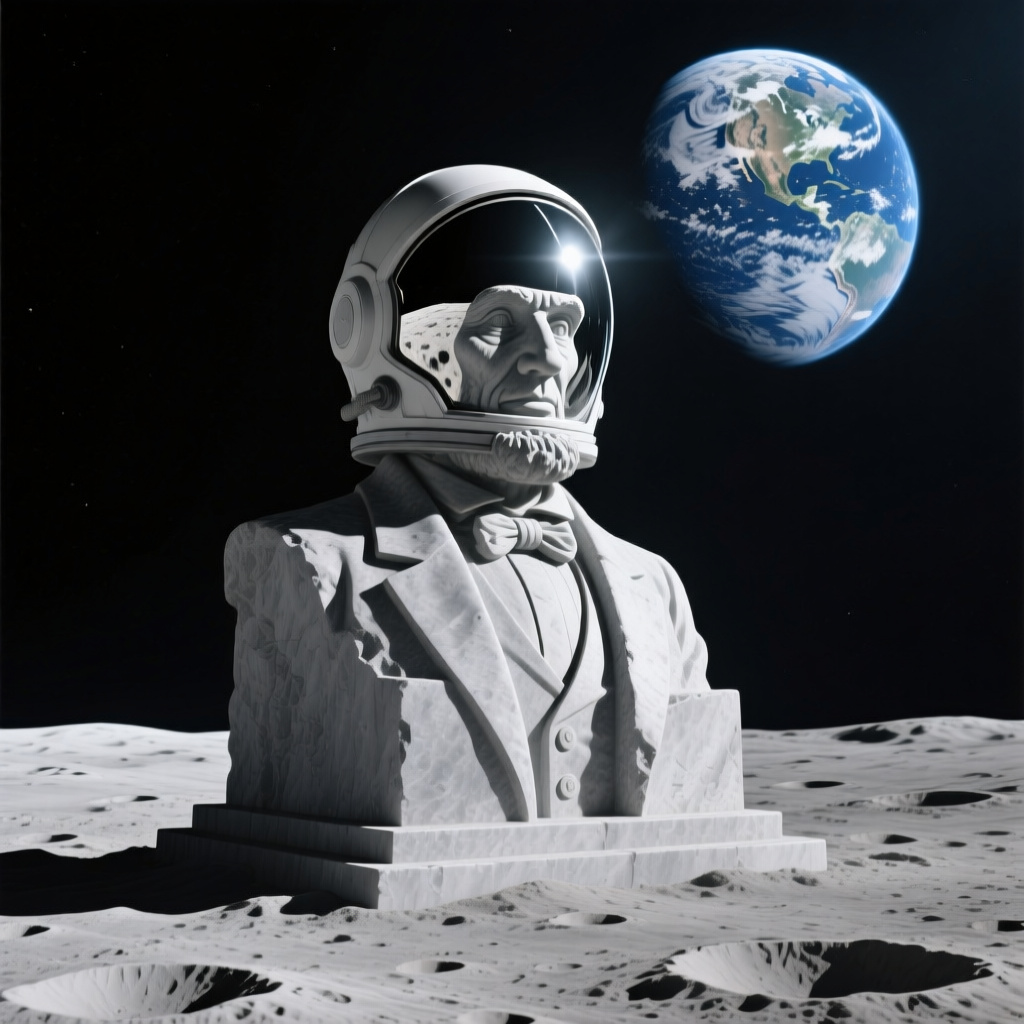} &
        \includegraphics[width=0.155\textwidth]{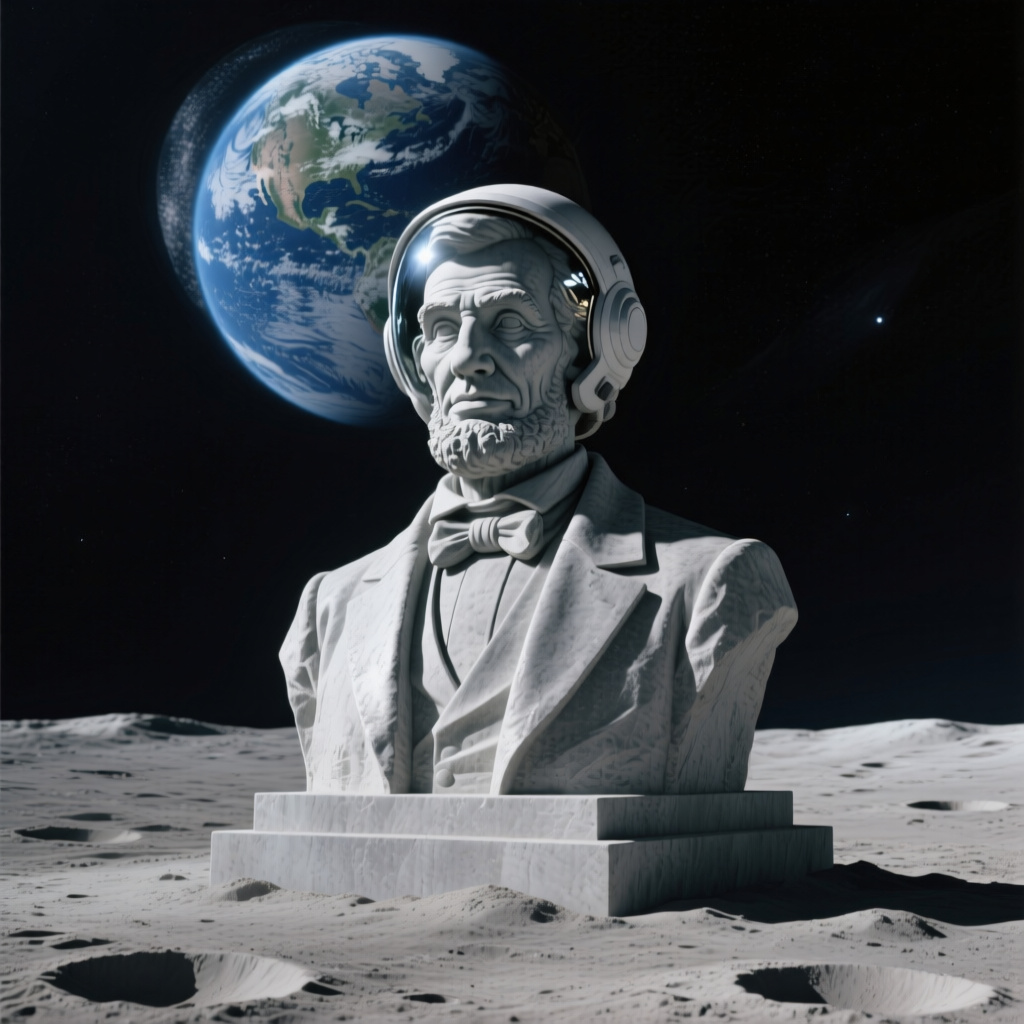}\includegraphics[width=0.155\textwidth]{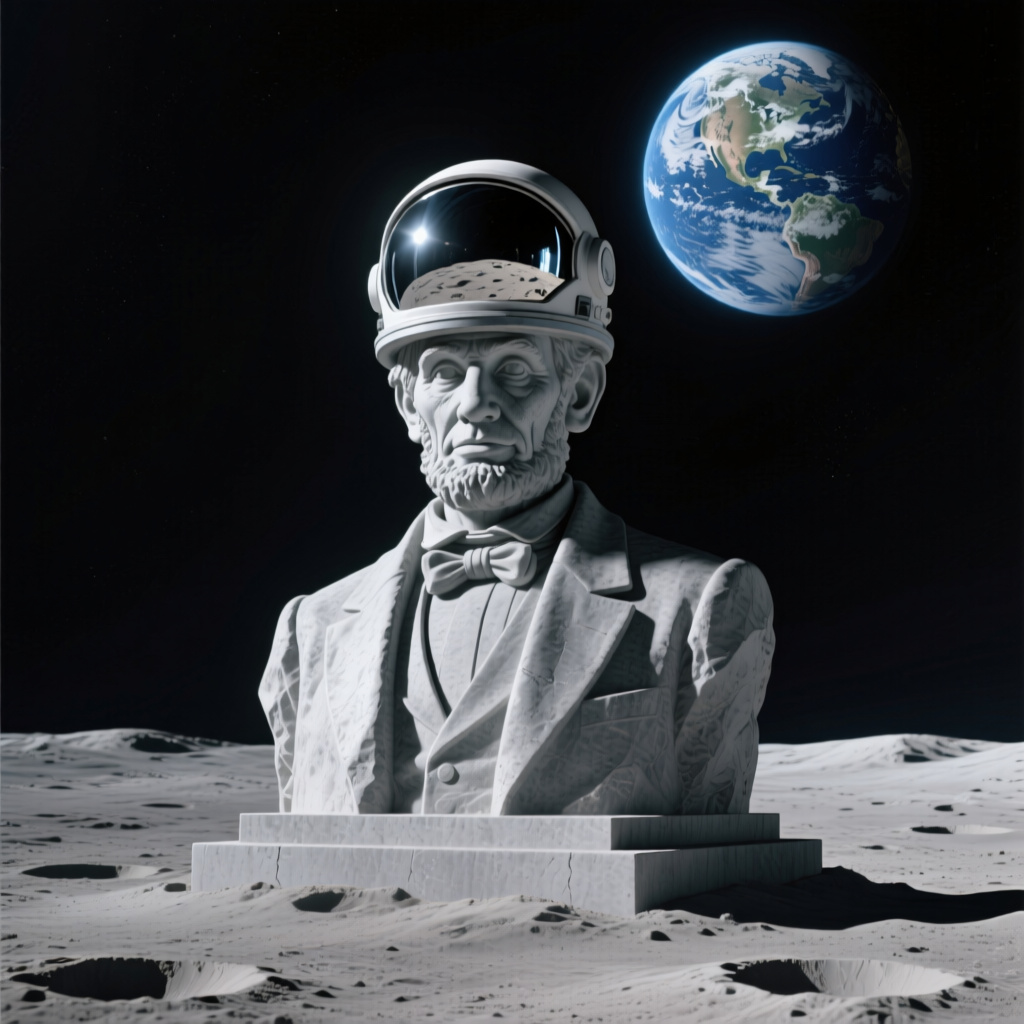} \\
        Teacher (Qwen-Image)~\citep{wu2025qwenimage} & Lightning\checkpointmark~\citep{fan2025:qwenlightning} & {\Oursbf{}} (Ours) \\[4pt]
        \multicolumn{3}{@{}c@{}}{\parbox{0.96\textwidth}{\centering\tiny\itshape
        \makecell{``A towering statue of Abraham Lincoln, cast in a silvery-gray stone, is adorned with a gleaming, opaque astronaut's helmet \\
        that reflects the barren lunar landscape. The statue is positioned on the moon's surface, with its craters and dust visible around the base. \\
        In the dark sky above, the planet Earth looms large, a swirl of blue and white against the blackness of space.''}
        }} \\[9pt]
        \includegraphics[width=0.155\textwidth]{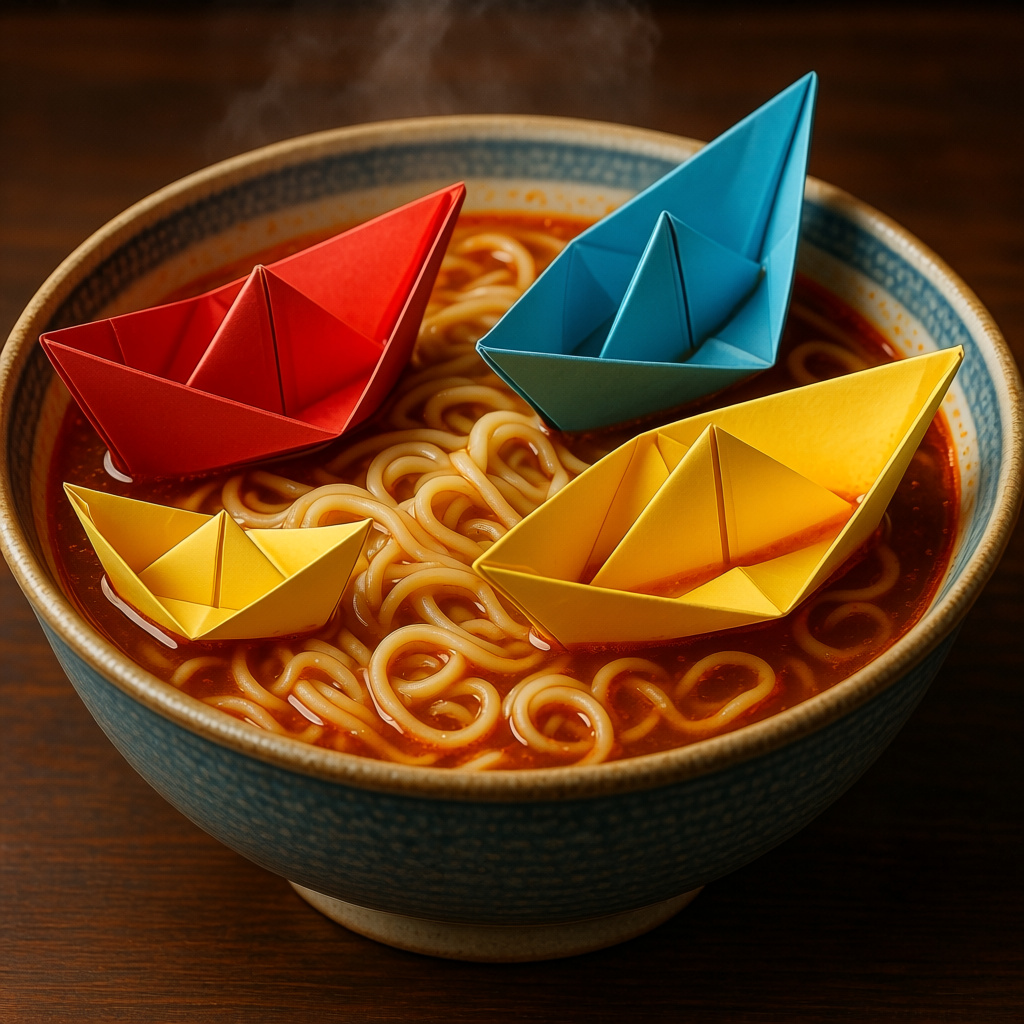}\includegraphics[width=0.155\textwidth]{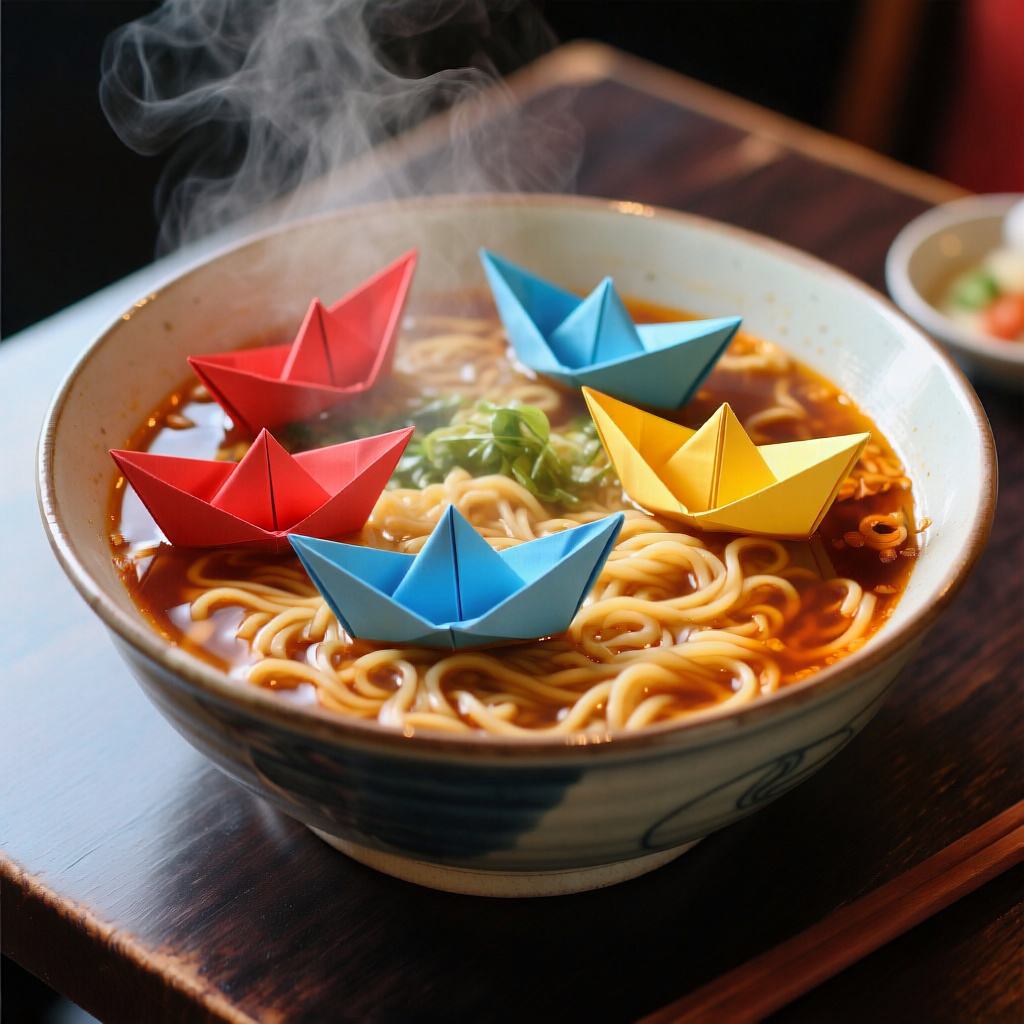} &
        \includegraphics[width=0.155\textwidth]{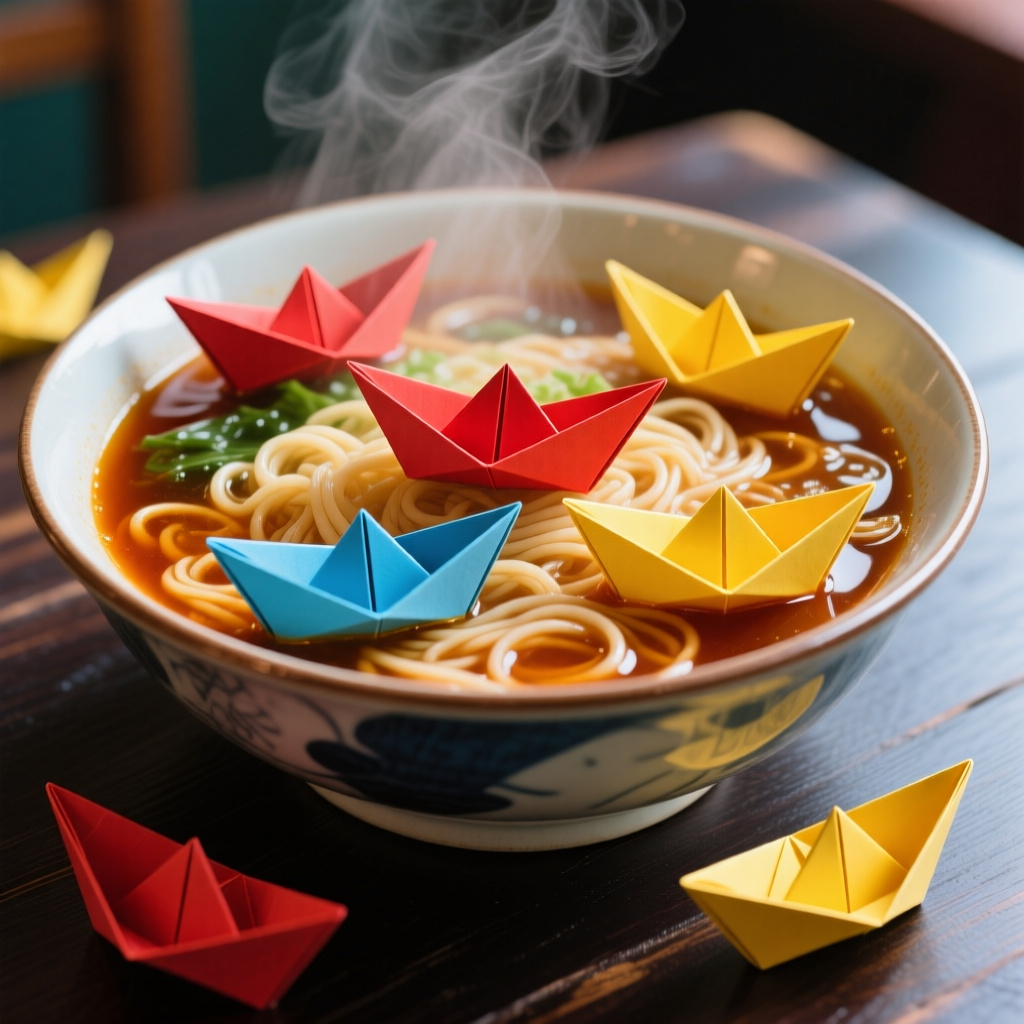}\includegraphics[width=0.155\textwidth]{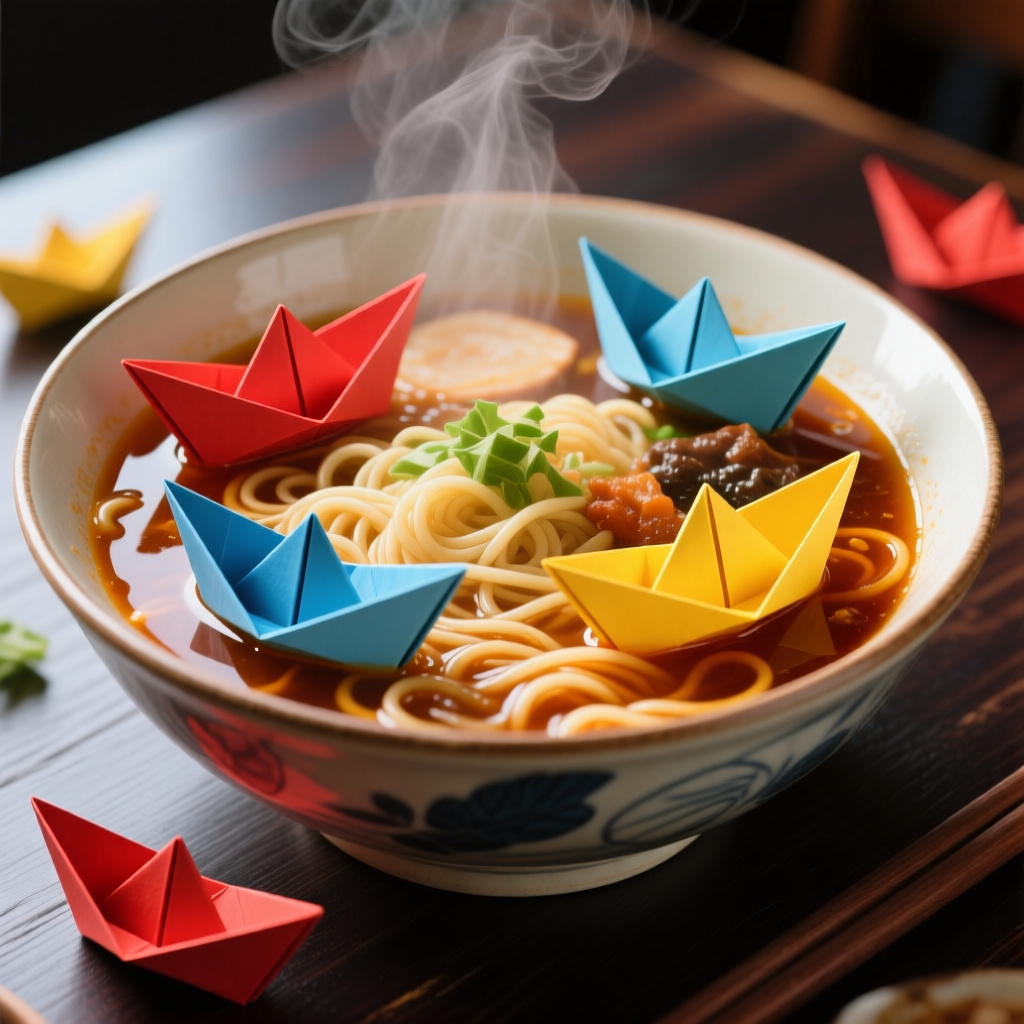} &
        \includegraphics[width=0.155\textwidth]{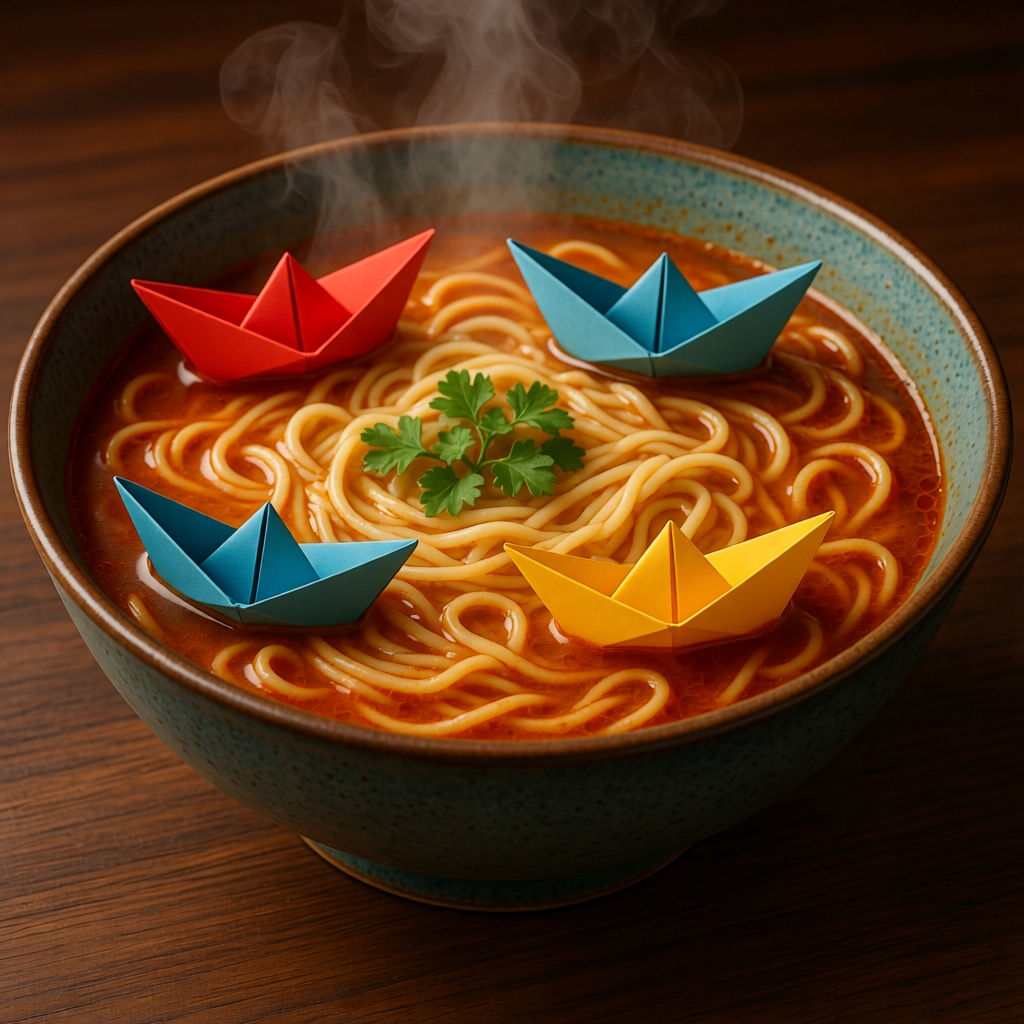}\includegraphics[width=0.155\textwidth]{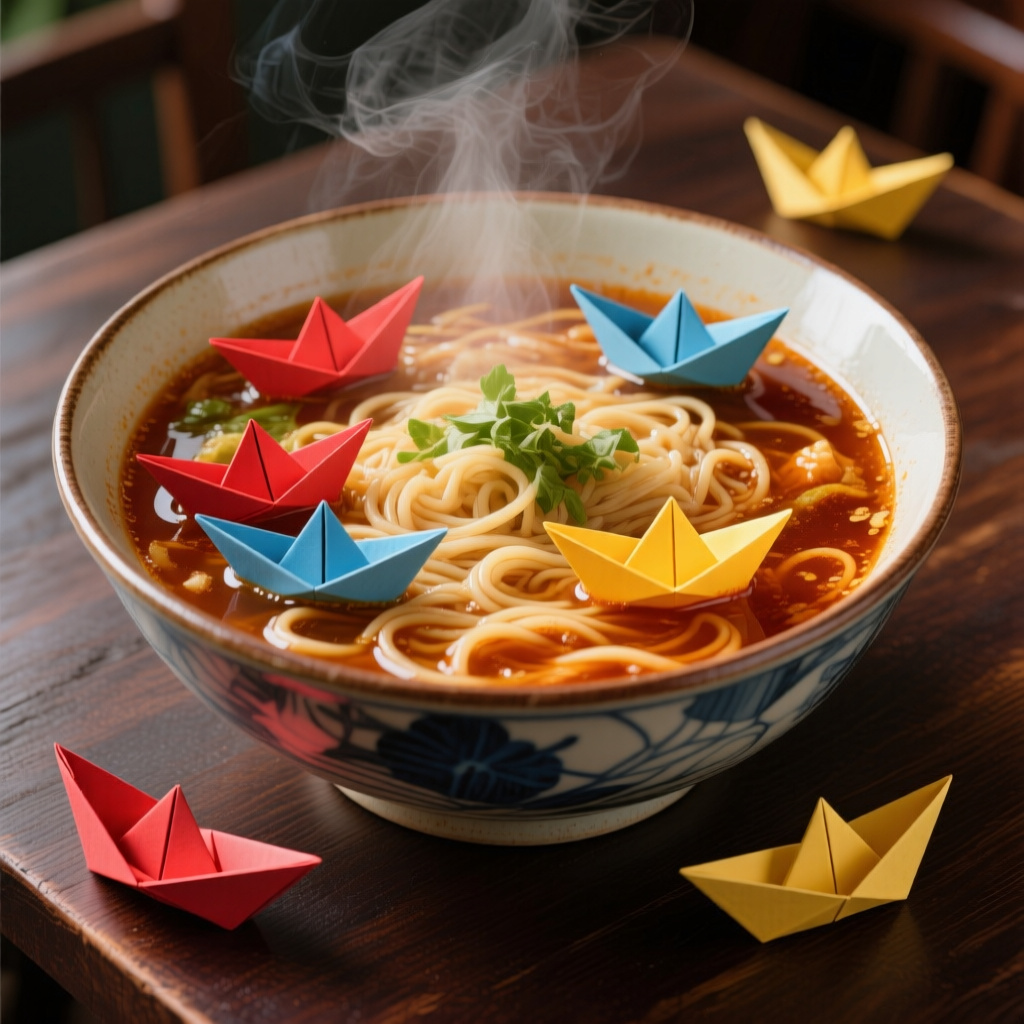} \\
        Teacher (Qwen-Image)~\citep{wu2025qwenimage} & Lightning\checkpointmark~\citep{fan2025:qwenlightning} & {\Oursbf{}} (Ours) \\[4pt]
        \multicolumn{3}{@{}c@{}}{\parbox{0.96\textwidth}{\centering\tiny\itshape ``A vibrant and detailed image showcasing a large ceramic bowl filled with steaming ramen, the noodles and broth glistening under the light. Floating amidst the savory dish are several origami boats, each crafted from paper in hues of red, blue, and yellow, adding a whimsical touch to the culinary scene. The bowl sits on a dark wooden table, contrasting with the bright colors of the paper crafts and the rich tones of the ramen ingredients.''}} \\[9pt]
    \end{tabular}
    \end{threeparttable}
\vspace{-5pt}\end{figure}

\begin{table}[t]\vspace{-5pt}
    \centering
    \begin{subtable}[b]{0.49\textwidth}
        \centering
        \renewcommand{\arraystretch}{1.39}
        \setlength{\tabcolsep}{1.25pt}
        \scriptsize
        \begin{tabularx}{\linewidth}{@{}L c c c c@{}}
            \toprule
            Method & NFE & GenEval $\uparrow$ & DPG $\uparrow$ & WISE $\uparrow$ \\
            \midrule
            Teacher & 50 $\times$ 2 & 0.860 & 88.70 & 0.54 \\
            \midrule
            TwinFlow\checkpointmark~\tabcite{cheng2026:twinflow} & 2 & 0.827 & 86.65 & 0.50 \\
            DMD2~\tabcite{yin2024:dmd2} & 2 & 0.856 & \textbf{87.08} & 0.53 \\
            \rowcolor{gray!10}
            {\Oursbf{}} (Ours) & 2 & \textbf{0.862} & 87.03 & \textbf{0.54} \\
            \bottomrule
        \end{tabularx}
        \caption{Two-Step Generation.}
        \label{tab:qwen_2step}
    \end{subtable}
    \hfill
    \begin{subtable}[b]{0.49\textwidth}
        \centering
        \renewcommand{\arraystretch}{1.15}
        \setlength{\tabcolsep}{1.25pt}
        \scriptsize
        \begin{tabularx}{\linewidth}{@{}L c c c c@{}}
            \toprule
            Method & NFE & GenEval $\uparrow$ & DPG $\uparrow$ & WISE $\uparrow$ \\
            \midrule
            Teacher & 50 $\times$ 2 & 0.860 & 88.70 & 0.54 \\
            \midrule
            Lightning\checkpointmark~\tabcite{fan2025:qwenlightning} & 4 & 0.864 & 88.41 & 0.51 \\
            TwinFlow\checkpointmark~\tabcite{cheng2026:twinflow} & 4 & 0.817 & 86.24 & 0.49 \\
            $\pi$-Flow\checkpointmark~\tabcite{chen2026:piflow} & 4 & 0.863 & 87.90 & 0.53 \\
            \rowcolor{gray!10}
            {\Oursbf{}} (Ours) & 4 & \textbf{0.867} & \textbf{88.45} & \textbf{0.53} \\
            \bottomrule
        \end{tabularx}
        \caption{Four-Step Generation.}
        \label{tab:qwen_4step}
    \end{subtable}
    \caption{
        \textbf{Qwen-Image Distillation.}
        Two- and four-step text-to-image generation with Qwen-Image 20B~\citep{wu2025qwenimage}, evaluated on GenEval~\citep{ghosh2023:geneval}, DPG-Bench~\citep{hu2024:ella}, and WISE~\citep{niu2025:wise}.
        \checkpointmark{} denotes results obtained using officially released checkpoints.
    }
    \label{tab:qwen}
\vspace{-6pt}\end{table}

\vspace{-2pt}\subsection{Text-to-Image Diffusion Distillation}\vspace{-4pt}
\label{sec:exp_t2i}

We distill Qwen-Image~\citep{wu2025qwenimage} (20B) into two- and four-step students at $1024 \times 1024$ resolution using teacher samples generated using HPSv2~\citep{wu2023:hpsv2} prompts.
Following the previous work~\citep{cheng2026:twinflow}, we evaluate on GenEval~\citep{ghosh2023:geneval}, DPG-Bench~\citep{hu2024:ella}, and WISE~\citep{niu2025:wise}.
We compare our method with state-of-the-art baselines, including TwinFlow~\citep{cheng2026:twinflow}, $\pi$-Flow~\citep{chen2026:piflow}, and the DMD2-based Qwen-Lightning~\citep{fan2025:qwenlightning}, using their official checkpoints.
Two-step DMD2~\citep{yin2024:dmd2} is trained using FastGen~\citep{fastgen2026} framework.

\vspace{-3pt}\paragraph{Results.}
As shown in \cref{tab:qwen}, \Ours{} matches or outperforms the evaluated baselines across all metrics in both two- and four-step generation, except for a marginally lower DPG-Bench score than DMD2 in the two-step setting.
The qualitative comparisons in \cref{fig:qual_image,sec:additional_qualitative} suggest that \Ours{} more closely matches the diversity of the teacher than the strong Qwen-Lightning baseline~\citep{fan2025:qwenlightning}, without degrading visual quality.
The quantitative results in \cref{sec:sample_diversity} further show improved diversity over Qwen-Lightning (DMD2) across all reported metrics.

\providecommand{\QualRowLabel}[1]{\adjustbox{valign=m}{\rotatebox[origin=c]{90}{\scriptsize #1}}}
\begin{figure}[t]\vspace{-5pt}
    \centering
    \begin{threeparttable}
    \captionsetup{font=normalsize}
    \caption{
        \textbf{Qualitative Comparison on Text-to-Video Distillation.}
        We compare \Ours{} with DMD2~\citep{yin2024:dmd2} and AnyFlow~\citep{gu2026:anyflow} using different seeds, with uniformly sampled frames.
        \checkpointmark{} denotes an officially released checkpoint.
        See \cref{sec:additional_qualitative} for additional results.
    }
    \label{fig:qual_video}
    \setlength{\tabcolsep}{1.5pt}
    \renewcommand{\arraystretch}{1.05}
    \scriptsize
    \begin{tabular}{@{}c ccc@{\hspace{8pt}}ccc@{}}
        \QualRowLabel{DMD2} &
        \adjustbox{valign=m}{\includegraphics[width=0.145\textwidth]{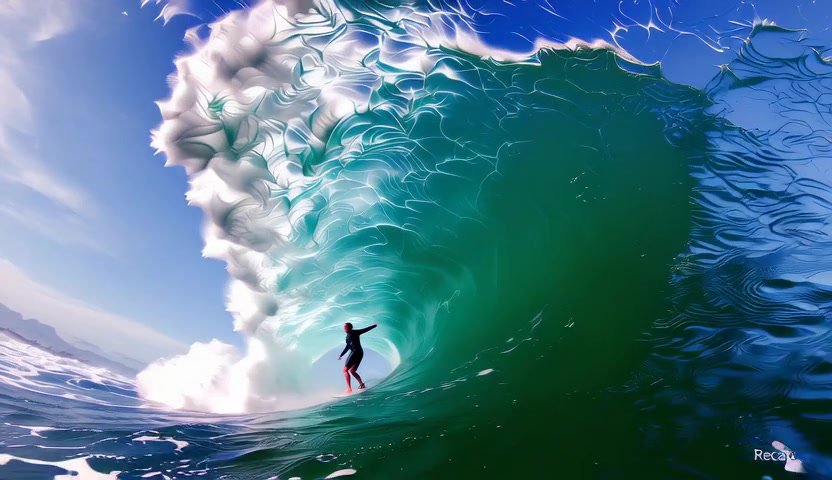}} & \adjustbox{valign=m}{\includegraphics[width=0.145\textwidth]{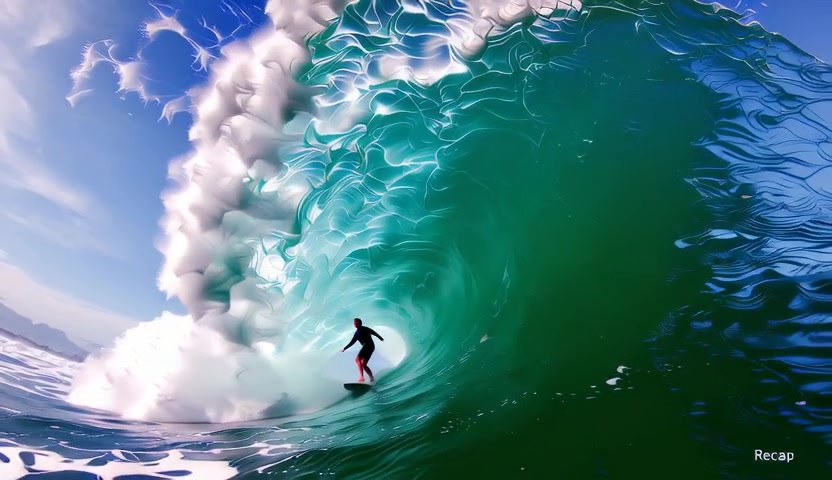}} & \adjustbox{valign=m}{\includegraphics[width=0.145\textwidth]{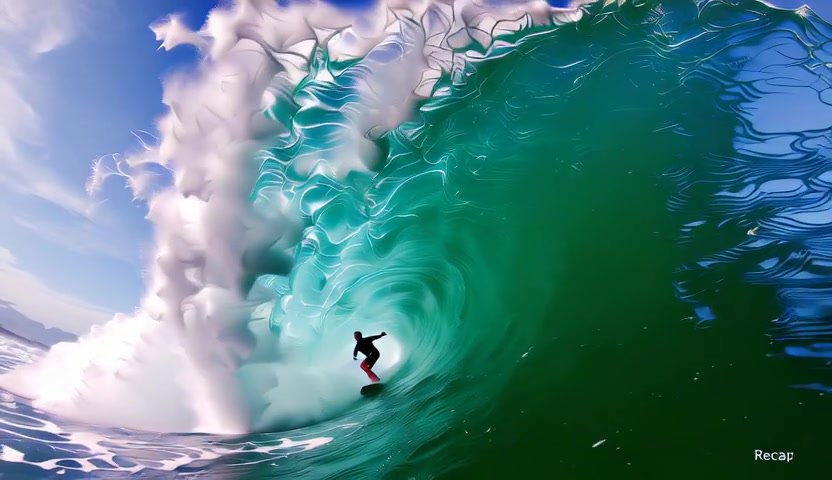}} & \adjustbox{valign=m}{\includegraphics[width=0.145\textwidth]{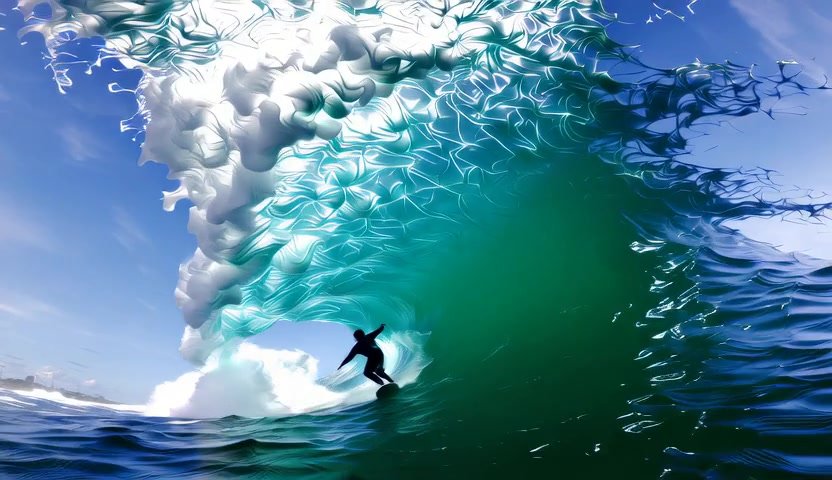}} & \adjustbox{valign=m}{\includegraphics[width=0.145\textwidth]{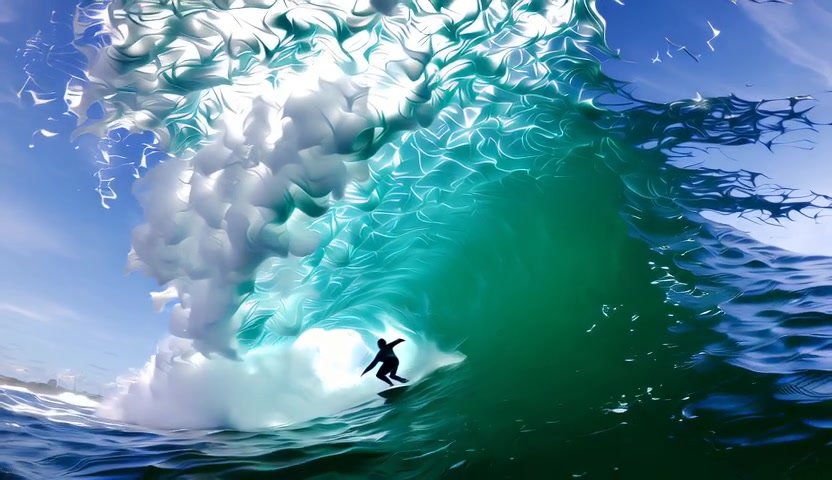}} & \adjustbox{valign=m}{\includegraphics[width=0.145\textwidth]{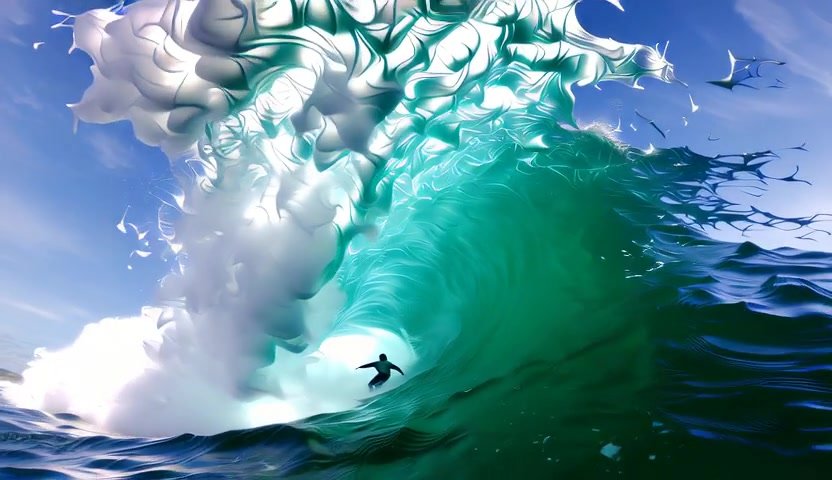}} \\[2pt]
        \QualRowLabel{AnyFlow\checkpointmark} &
        \adjustbox{valign=m}{\includegraphics[width=0.145\textwidth]{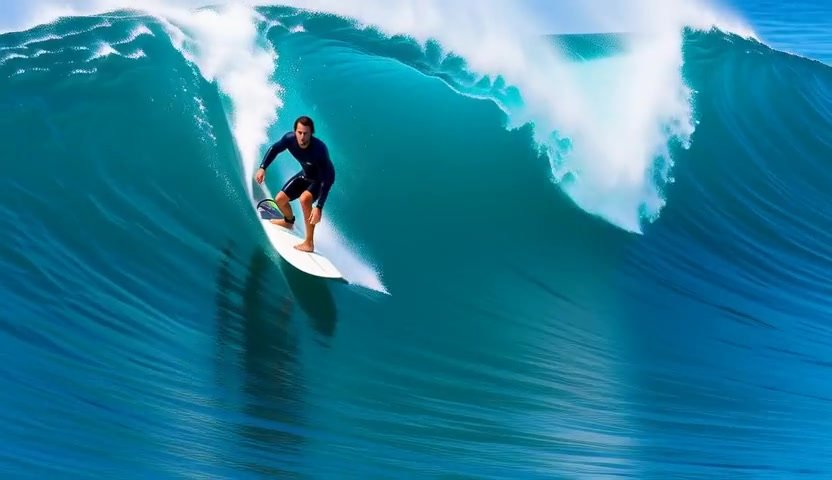}} & \adjustbox{valign=m}{\includegraphics[width=0.145\textwidth]{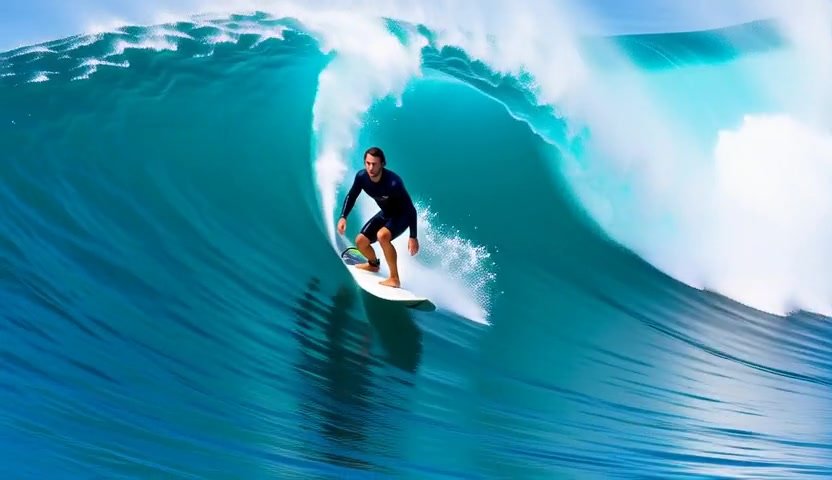}} & \adjustbox{valign=m}{\includegraphics[width=0.145\textwidth]{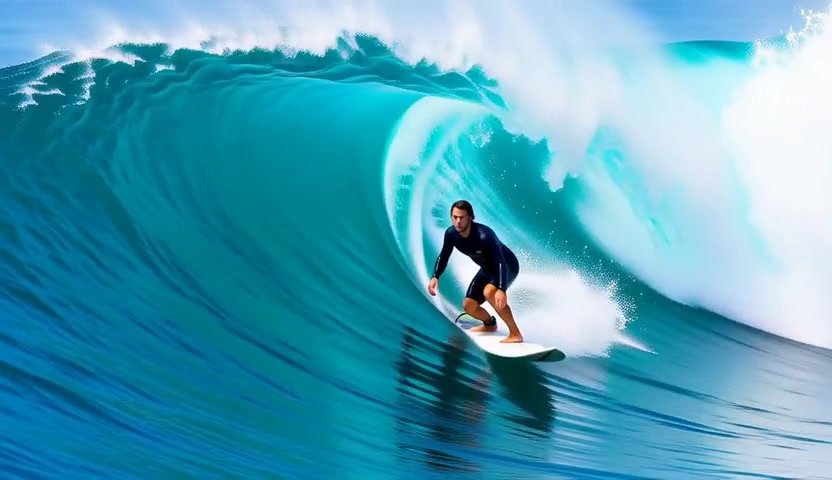}} & \adjustbox{valign=m}{\includegraphics[width=0.145\textwidth]{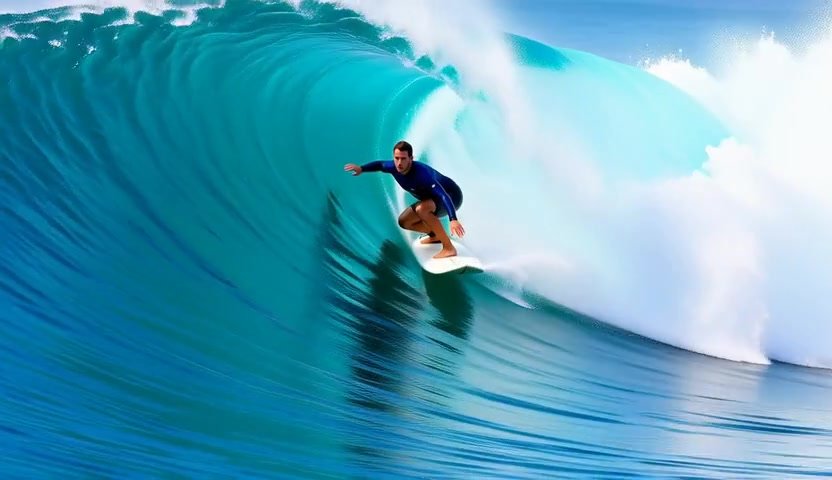}} & \adjustbox{valign=m}{\includegraphics[width=0.145\textwidth]{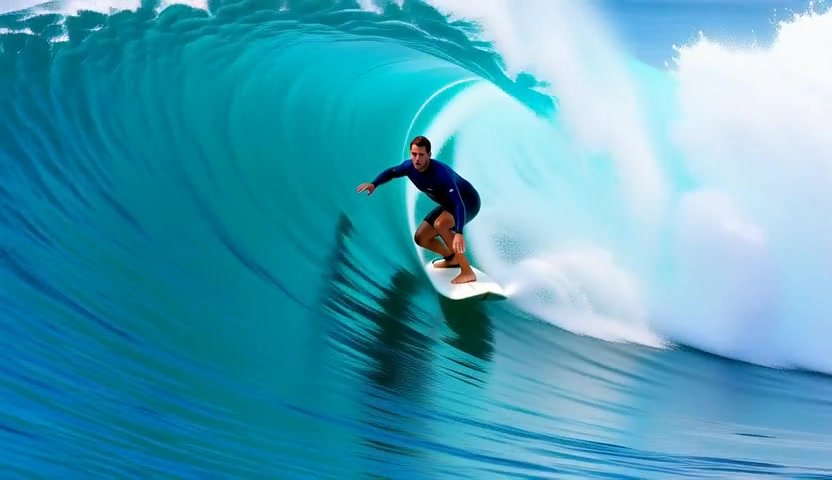}} & \adjustbox{valign=m}{\includegraphics[width=0.145\textwidth]{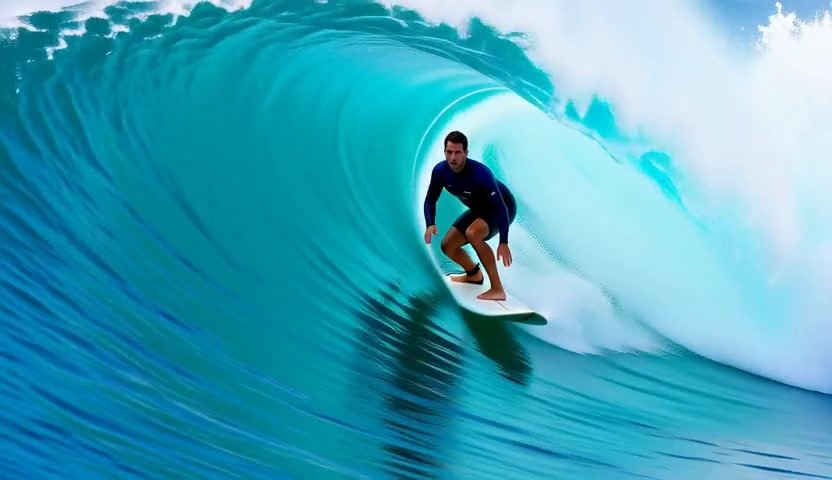}} \\[2pt]
        \QualRowLabel{\makecell{{\Oursbf{}} \\ {(Ours)}}} &
        \adjustbox{valign=m}{\includegraphics[width=0.145\textwidth]{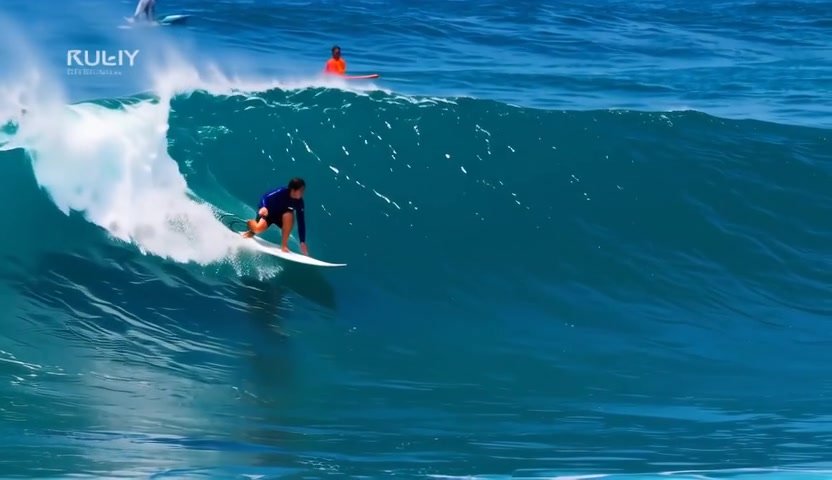}} & \adjustbox{valign=m}{\includegraphics[width=0.145\textwidth]{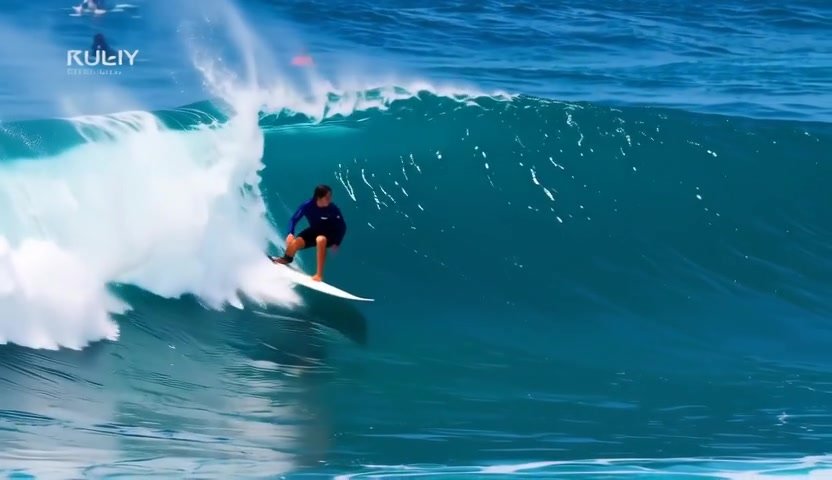}} & \adjustbox{valign=m}{\includegraphics[width=0.145\textwidth]{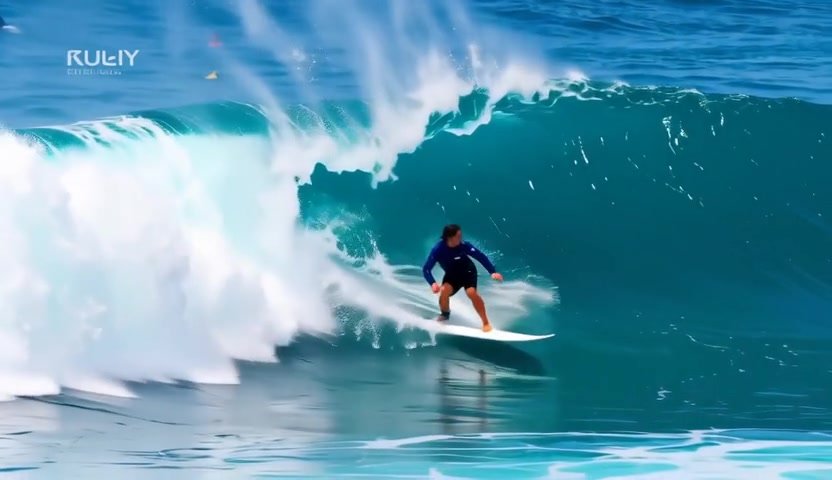}} & \adjustbox{valign=m}{\includegraphics[width=0.145\textwidth]{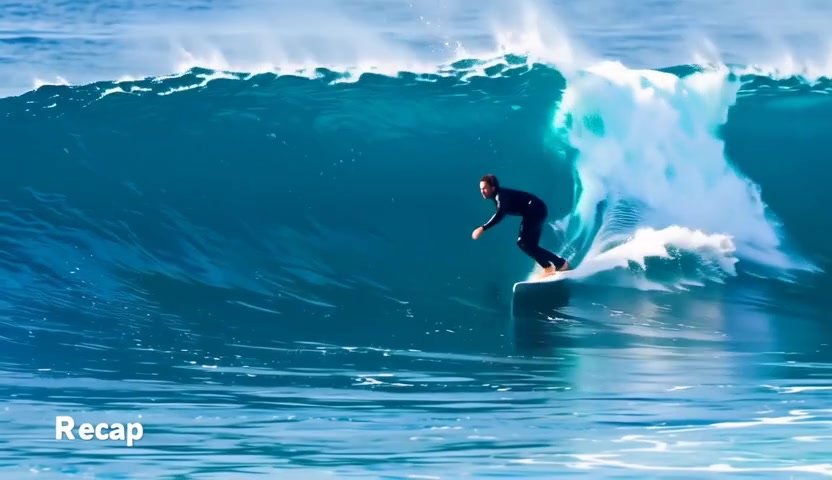}} & \adjustbox{valign=m}{\includegraphics[width=0.145\textwidth]{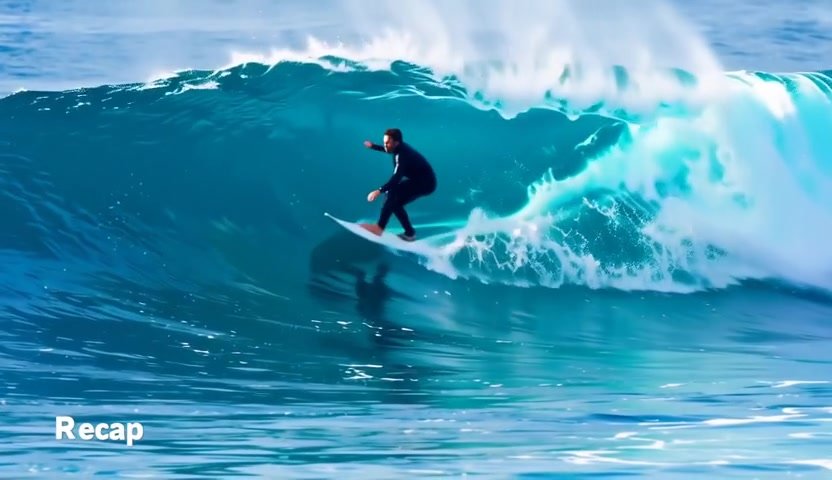}} & \adjustbox{valign=m}{\includegraphics[width=0.145\textwidth]{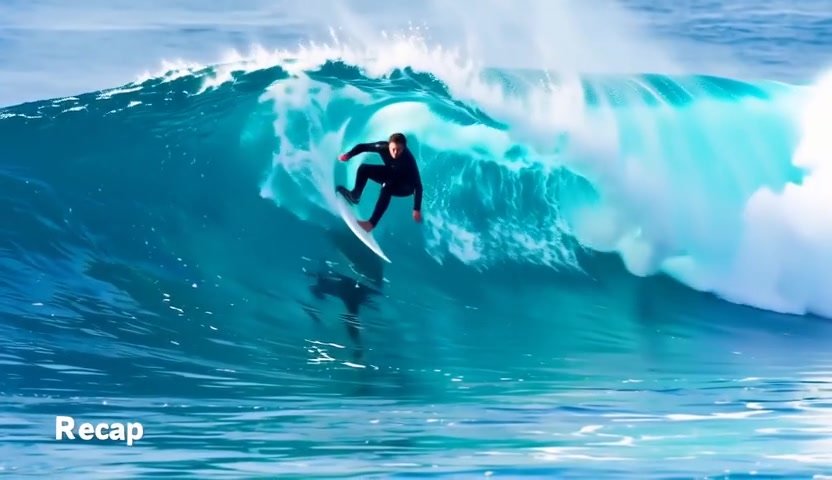}} \\[12pt]
        & \multicolumn{6}{c}{\parbox{0.90\textwidth}{\centering\tiny\itshape ``In this dynamic scene, a surfer expertly maneuvers a wave, showcasing impressive skill and balance. The surfer, clad in a wetsuit, is positioned on a sleek, white surfboard, riding the crest of a powerful, turquoise wave. The wave's size and power are evident from its towering height and the frothy white foam at its peak. The surfer's body leans forward, arms extended for stability, as they navigate the wave's curve with precision. The ocean's deep blue-green hues contrast sharply with the white foam, emphasizing the intensity of the moment. The "Recap" text in the corner suggests this is part of a larger surfing event, highlighting the surfer's performance within a competitive context.''}} \\[16pt]
        \QualRowLabel{DMD2} &
        \adjustbox{valign=m}{\includegraphics[width=0.145\textwidth]{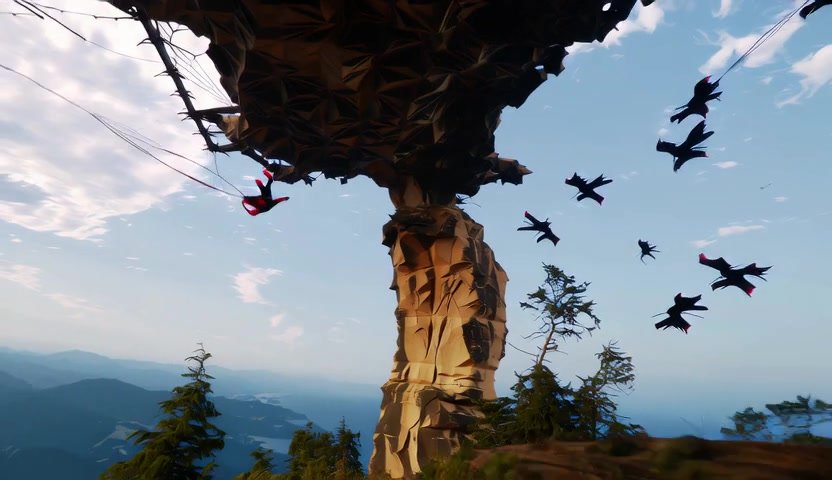}} & \adjustbox{valign=m}{\includegraphics[width=0.145\textwidth]{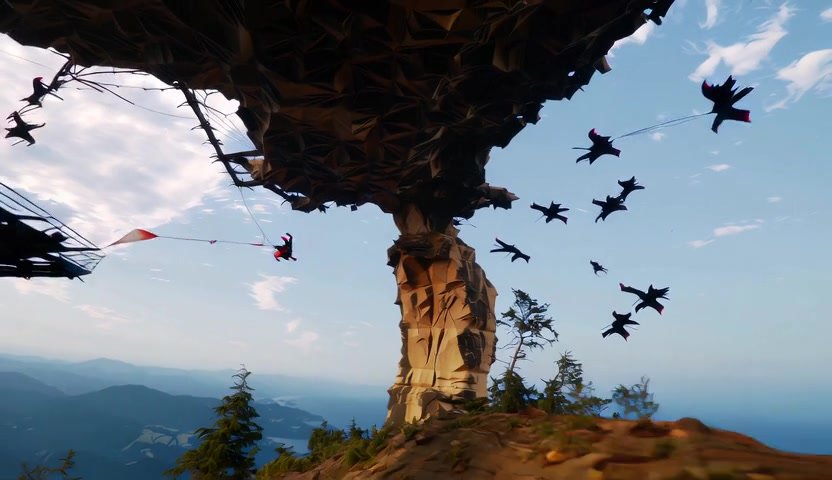}} & \adjustbox{valign=m}{\includegraphics[width=0.145\textwidth]{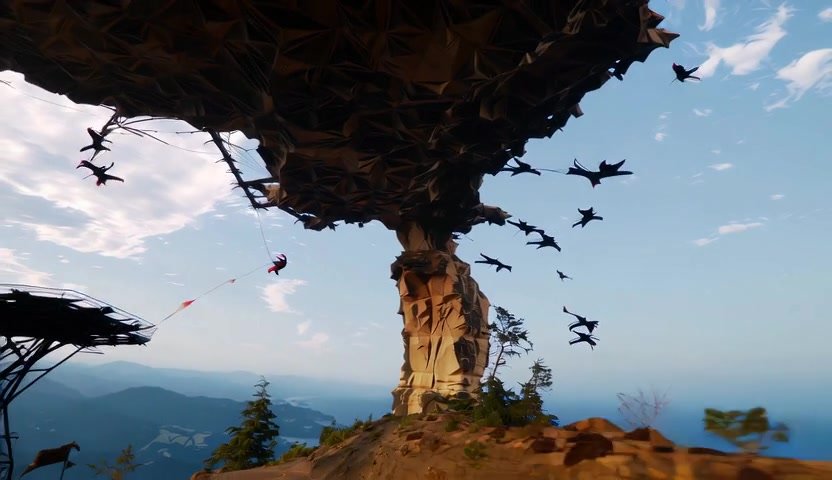}} & \adjustbox{valign=m}{\includegraphics[width=0.145\textwidth]{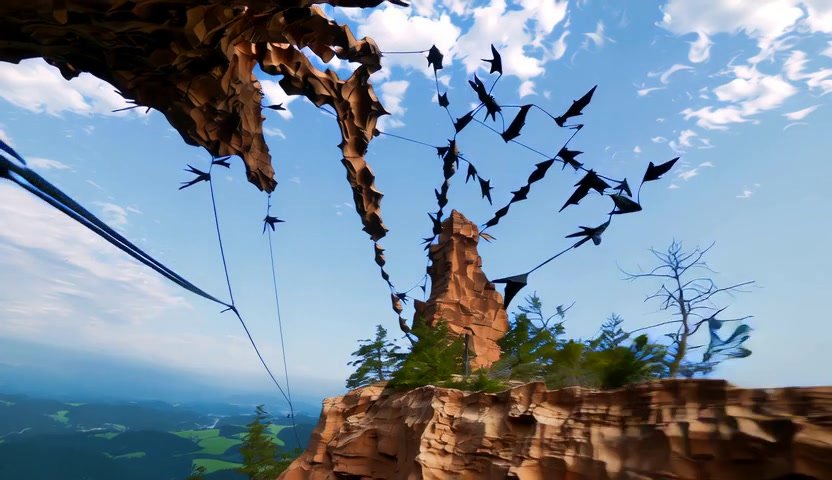}} & \adjustbox{valign=m}{\includegraphics[width=0.145\textwidth]{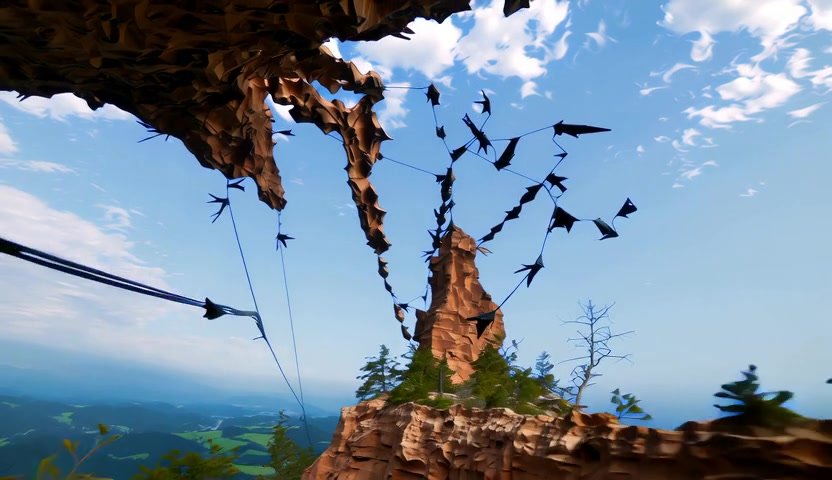}} & \adjustbox{valign=m}{\includegraphics[width=0.145\textwidth]{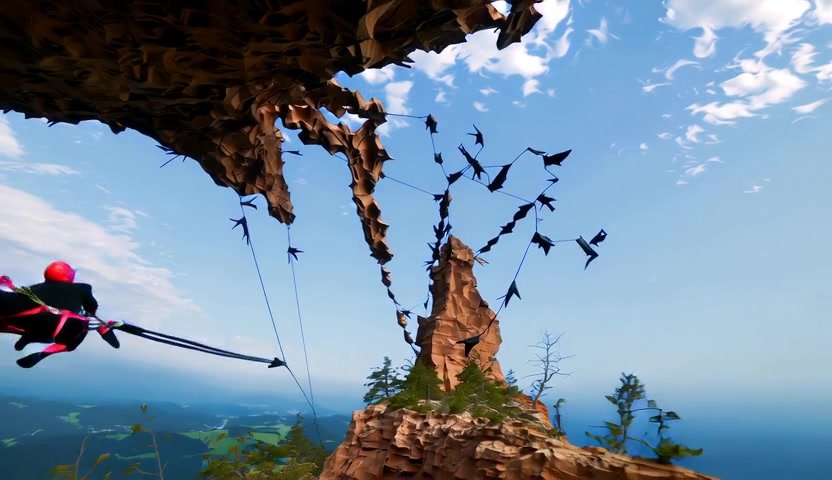}} \\[2pt]
        \QualRowLabel{AnyFlow\checkpointmark} &
        \adjustbox{valign=m}{\includegraphics[width=0.145\textwidth]{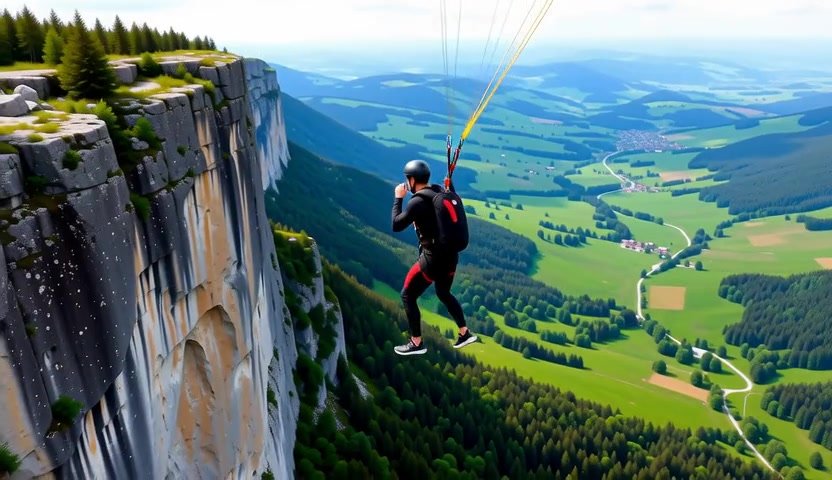}} & \adjustbox{valign=m}{\includegraphics[width=0.145\textwidth]{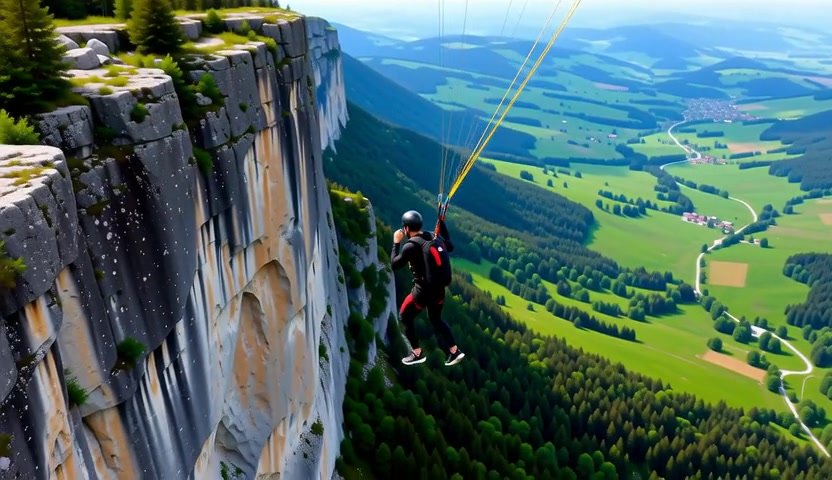}} & \adjustbox{valign=m}{\includegraphics[width=0.145\textwidth]{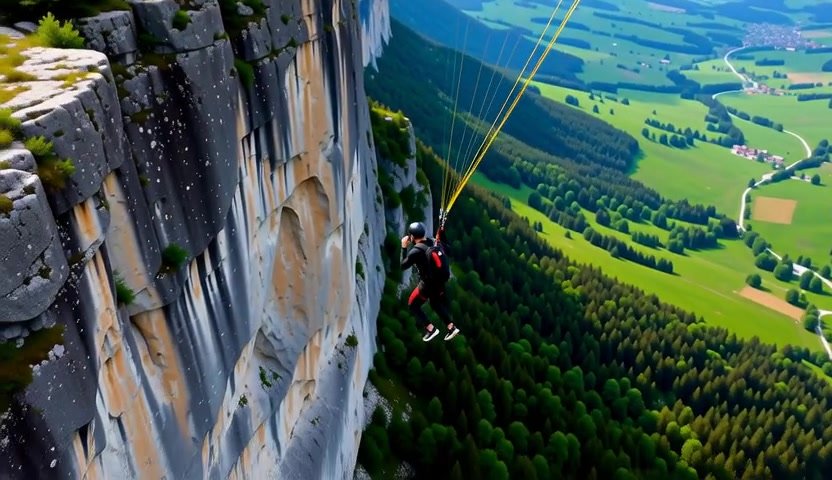}} & \adjustbox{valign=m}{\includegraphics[width=0.145\textwidth]{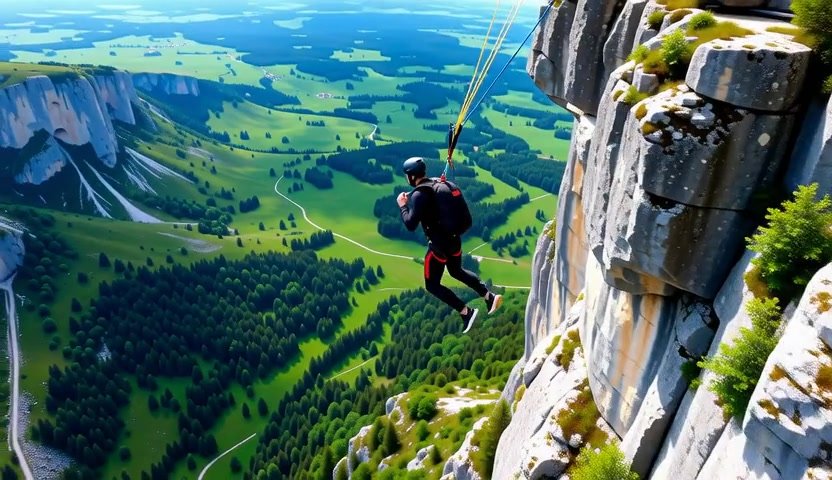}} & \adjustbox{valign=m}{\includegraphics[width=0.145\textwidth]{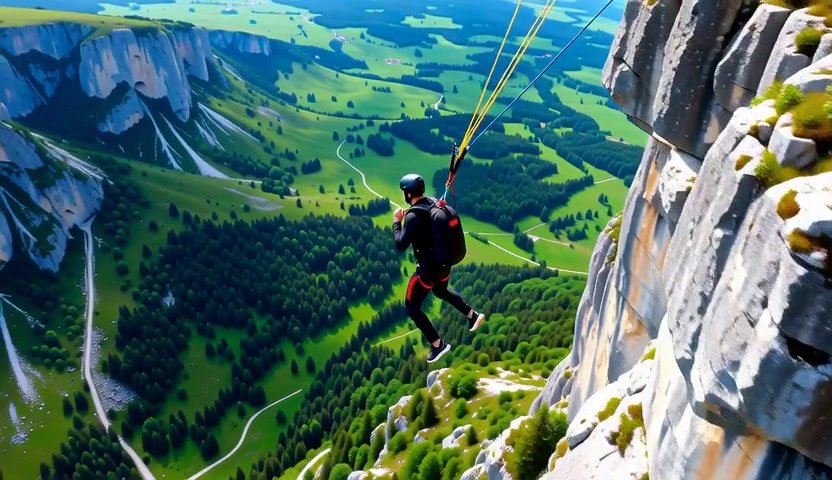}} & \adjustbox{valign=m}{\includegraphics[width=0.145\textwidth]{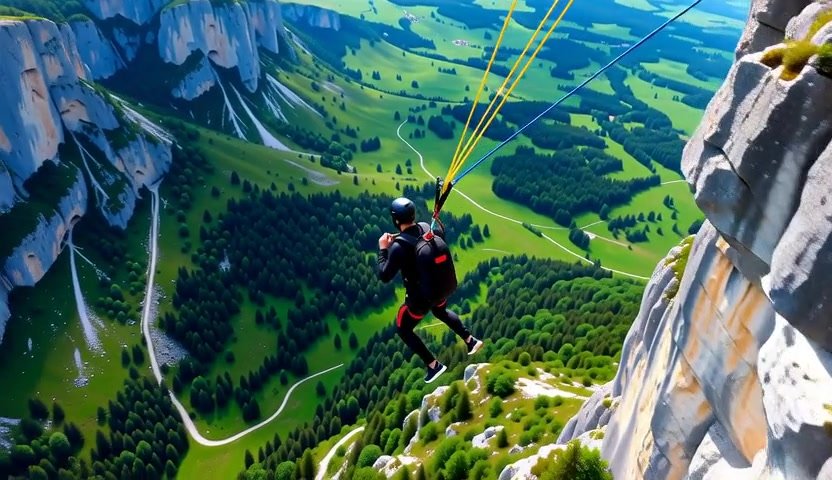}} \\[2pt]
        \QualRowLabel{\makecell{{\Oursbf{}} \\ {(Ours)}}} &
        \adjustbox{valign=m}{\includegraphics[width=0.145\textwidth]{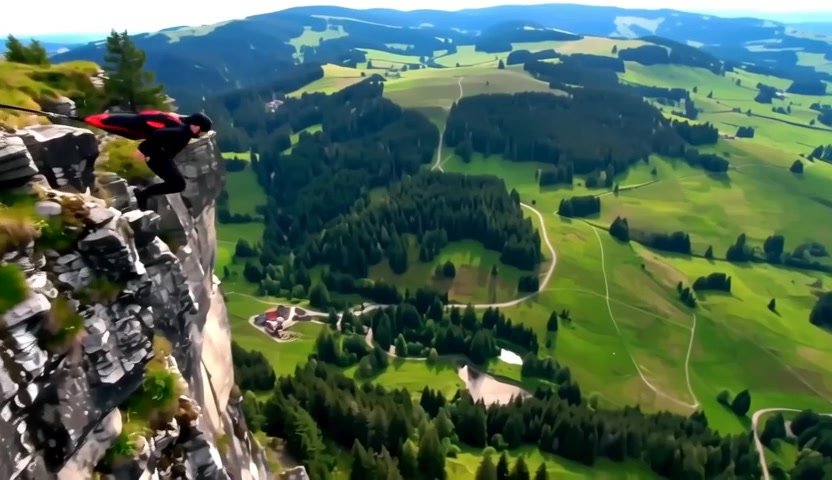}} & \adjustbox{valign=m}{\includegraphics[width=0.145\textwidth]{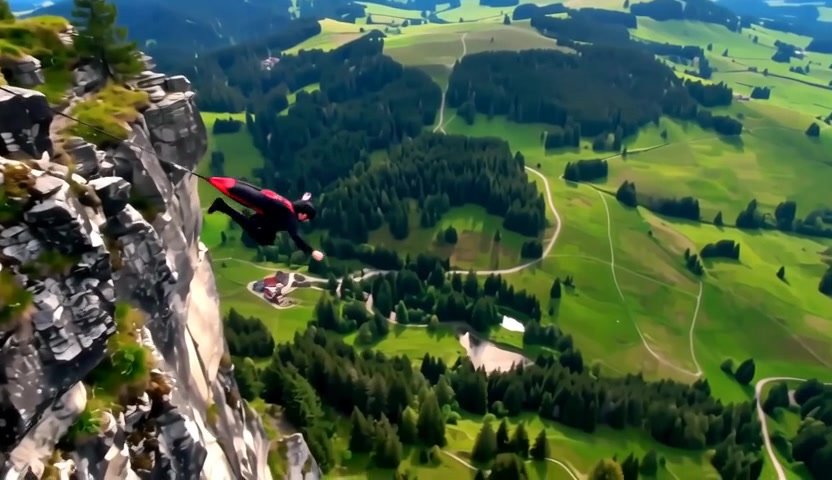}} & \adjustbox{valign=m}{\includegraphics[width=0.145\textwidth]{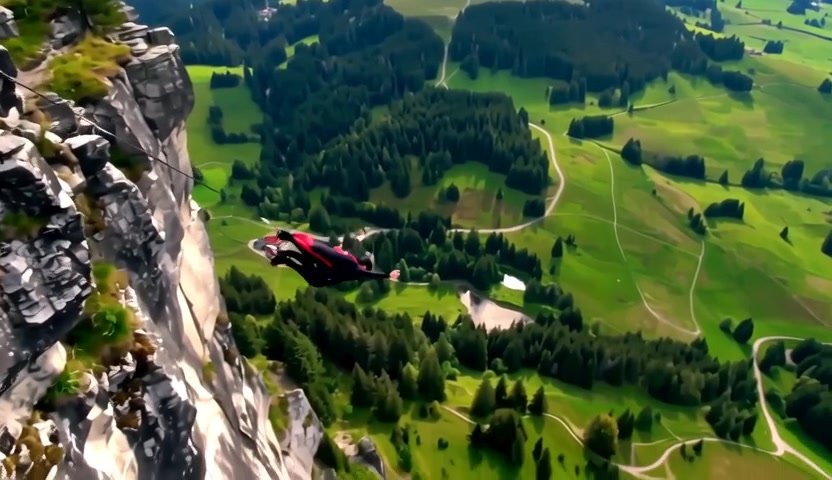}} & \adjustbox{valign=m}{\includegraphics[width=0.145\textwidth]{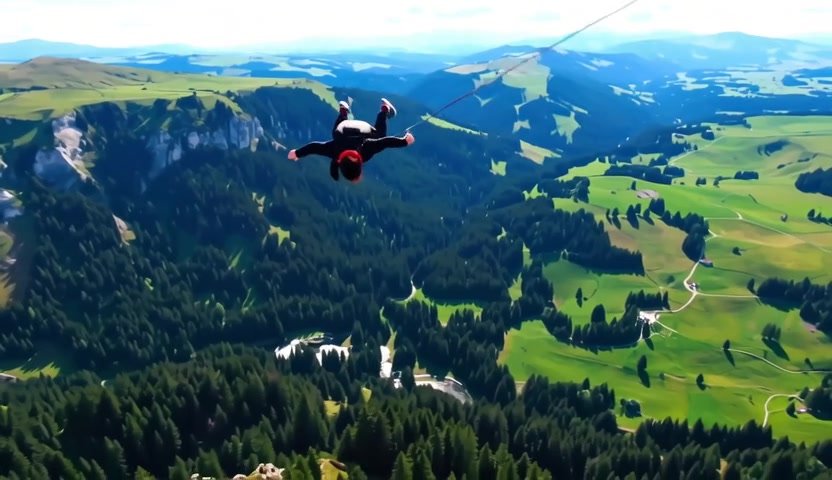}} & \adjustbox{valign=m}{\includegraphics[width=0.145\textwidth]{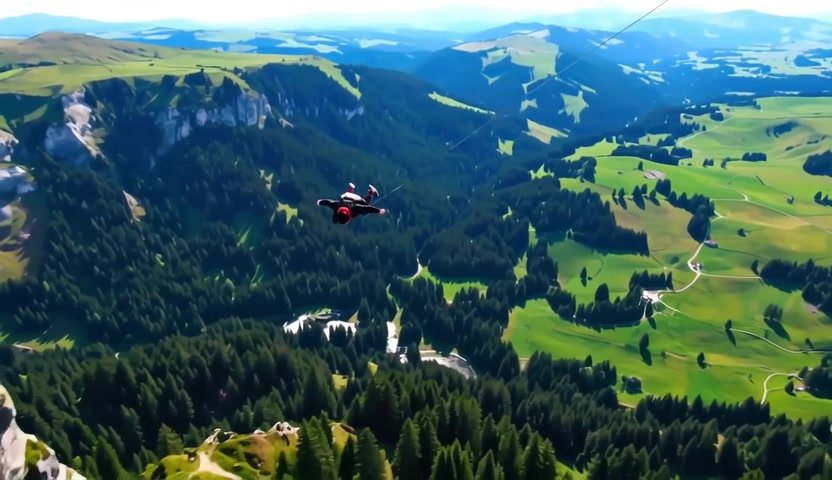}} & \adjustbox{valign=m}{\includegraphics[width=0.145\textwidth]{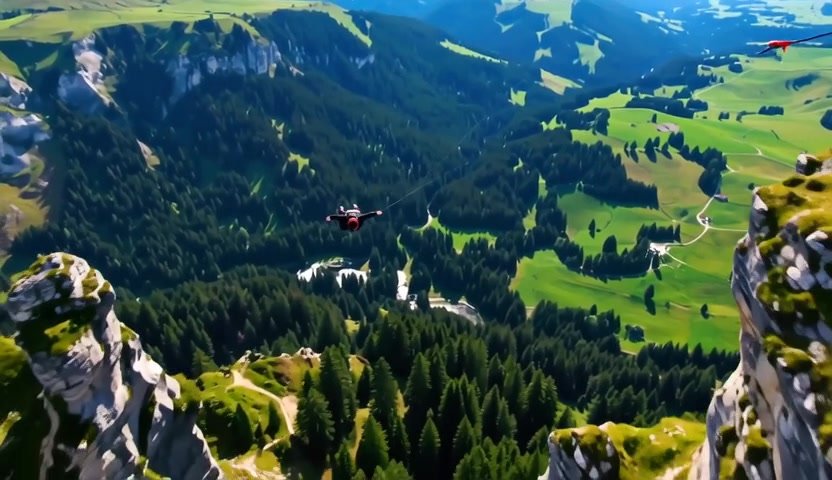}} \\[12pt]
        & \multicolumn{6}{c}{\parbox{0.90\textwidth}{\centering\tiny\itshape ``The video captures an exhilarating bungee jump from a towering cliff. A person, clad in a black outfit with red accents, is seen descending rapidly towards the valley below. The camera follows the jumper's descent, revealing the rugged, rocky cliff face and the expansive green valley dotted with patches of trees. The bungee cord stretches taut, absorbing the impact as the jumper descends. The scene emphasizes the height and scale of the jump, with the vast landscape stretching out beneath the cliff. The dynamic movement of the jumper contrasts sharply with the stillness of the surrounding terrain, highlighting the thrill and adventure of the activity.''}} \\[16pt]
    \end{tabular}
    \end{threeparttable}
\vspace{-5pt}\end{figure}

\vspace{-3pt}\subsection{Text-to-Video Diffusion Distillation}\vspace{-4pt}
\label{sec:exp_video}
We distill the 1.3B and 14B variants of Wan2.1~\citep{wan2025} into four-step students for text-to-video generation.
We use teacher samples generated from MovieGenVideoBench prompts~\citep{polyak2024:movie} for training and evaluate the distilled models on VBench~\citep{huang2024:vbench}.
We compare \Ours{} against the official checkpoints of rCM~\citep{zheng2026:rcm} and AnyFlow~\citep{gu2026:anyflow}, as well as DMD2~\citep{yin2024:dmd2} trained on FastGen framework~\citep{fastgen2026}.

\begin{table}[t]\vspace{-5pt}
    \centering
    \begin{subtable}[t]{0.48\textwidth}
        \centering
        \renewcommand{\arraystretch}{1.15}
        \setlength{\tabcolsep}{2pt}
        \scriptsize
        \begin{tabularx}{\linewidth}{@{}L c c c c@{}}
            \toprule
            Method & NFE & Overall $\uparrow$ & Quality $\uparrow$ & Sem. $\uparrow$ \\
            \midrule
            Teacher & 50$\times$2 & 83.03 & 84.29 & 78.00 \\
            \midrule
            DMD2~\tabcite{yin2024:dmd2} & 4 & 83.65 & 85.73 & 75.32 \\
            rCM\checkpointmark~\tabcite{zheng2026:rcm} & 4 & 82.78 & 84.13 & 77.31 \\
            AnyFlow\checkpointmark~\tabcite{gu2026:anyflow} & 4 & 83.86 & 85.15 & \textbf{78.71} \\
            \rowcolor{gray!10}
            {\Oursbf{}} (Ours) & 4 & \textbf{84.12} & \textbf{85.94} & 76.81 \\
            \bottomrule
        \end{tabularx}
        \caption{Wan2.1 T2V (1.3B).}
        \label{tab:wan_1p3b}
    \end{subtable}
    \hfill
    \begin{subtable}[t]{0.48\textwidth}
        \centering
        \renewcommand{\arraystretch}{1.15}
        \setlength{\tabcolsep}{2pt}
        \scriptsize
        \begin{tabularx}{\linewidth}{@{}L c c c c@{}}
            \toprule
            Method & NFE & Overall $\uparrow$ & Quality $\uparrow$ & Sem. $\uparrow$ \\
            \midrule
            Teacher & 50$\times$2 & 83.71 & 85.13 & 78.02 \\
            \midrule
            DMD2~\tabcite{yin2024:dmd2} & 4 & 84.69 & \textbf{86.67} & 76.75 \\
            rCM\checkpointmark~\tabcite{zheng2026:rcm} & 4 & 83.97 & 85.40 & 78.28 \\
            AnyFlow\checkpointmark~\tabcite{gu2026:anyflow} & 4 & 84.38 & 85.70 & 79.11 \\
            \rowcolor{gray!10}
            {\Oursbf{}} (Ours) & 4 & \textbf{84.87} & 85.96 & \textbf{80.51} \\
            \bottomrule
        \end{tabularx}
        \caption{Wan2.1 T2V (14B).}
        \label{tab:wan_14b}
    \end{subtable}
    \caption{
        \textbf{Wan Text-to-Video Distillation.}
        We evaluate the generated video samples using VBench~\citep{huang2024:vbench}.
        \checkpointmark{} denotes results obtained using officially released checkpoints.
        DMD2~\citep{yin2024:dmd2} is trained using FastGen~\citep{fastgen2026}.
    }
    \label{tab:video}
\vspace{-6pt}\end{table}

\vspace{-2pt}\paragraph{Results.}
As shown in \cref{tab:wan_14b}, \Ours{} achieves the highest overall and semantic scores on Wan2.1-14B, with a quality score comparable to DMD2.
Similarly, \cref{tab:wan_1p3b} shows that \Ours{} achieves the highest overall and quality scores on Wan2.1-1.3B.
The qualitative results in \cref{fig:qual_video} and the additional examples in \cref{sec:additional_qualitative} show that \Ours{} generates high-quality videos that closely follow the text prompts with diversity.
Consistent with the text-to-image results in \cref{sec:exp_t2i}, the quantitative evaluation in \cref{sec:sample_diversity} shows that \Ours{} also produces more diverse samples than DMD2 on Wan2.1.\vspace{-4pt}

\begin{experimentinset}{0.38\textwidth}{\centering
\renewcommand{\arraystretch}{1.15}
\setlength{\tabcolsep}{3pt}
\scriptsize
\begingroup
\newcolumntype{X}{>{\centering\arraybackslash}p{0.34\linewidth}}
\newcolumntype{Y}{>{\centering\arraybackslash}p{0.40\linewidth}}
\begin{tabular}{X Y c}
    \toprule
     & Energy $(\mathcal{F})$ & FID $\downarrow$ \\
    \midrule
    \multirow{2}{*}{Score-based} & Forward KL & 3.23 \\
    & IKL & 3.14 \\
    \midrule
    \multirow{2}{*}{Sample-based} & MMD & 6.21 \\
    & Sinkhorn & 5.09 \\
    \midrule
    \multirow{2}{*}{Combined} & IKL + MMD & 3.01 \\
    & \cellcolor{gray!10}{\Oursbf{}} (Ours) & \cellcolor{gray!10}\textbf{2.83} \\
    \bottomrule
\end{tabular}
\endgroup
\normalsize
\caption{
    \textbf{Energy Functional.}
    ImageNet-256~\citep{deng2009:imagenet} distillation with different energy functionals. Combined guidance uses both score- and sample-based updates.
}
\label{tab:ablation_energy}
\vspace{-3pt}
}{%
    \subsection{Ablation Study}
    \label{sec:exp_ablation}
    \paragraph{Energy Functional.}}
\Cref{tab:ablation_energy} reports FID on ImageNet-256 for models distilled with different energy functionals.
IKL achieves the lowest FID among the individual energy functionals, in line with previous observations that it can help preserve fine details~\citep{zheng2026:rcm, huang2026:self-forcing}.

Complementing IKL with Sinkhorn divergence in \Ours{} further reduces FID.
The global transport plan guides samples with weak teacher support toward the target distribution and can alleviate the mode collapse associated with reverse KL, as illustrated in \cref{fig:motivation}.
While MMD can also complement IKL, it yields worse performance than Sinkhorn, consistent with observations in previous work~\citep{han2026:wflow,gretton2026:wgfdrifting,he2026:sinkhorndrifting}.
\end{experimentinset}

\begin{experimentinset}{0.50\textwidth}{\centering
\renewcommand{\arraystretch}{1.15}
\setlength{\tabcolsep}{3pt}
\scriptsize
\begin{tabular}{c c c ccc}
    \toprule
    \multirow{2}{*}{Space} & \multirow{2}{*}{Scheduler} & ImageNet-256 & \multicolumn{3}{c}{FLUX} \\
    \cmidrule(lr){3-3} \cmidrule(l){4-6}
     & & FID $\downarrow$ & FID $\downarrow$ & CMMD $\downarrow$ & CLIP $\uparrow$ \\
    \midrule
    \rowcolor{gray!10}
    $\psi(x)$ & Gated & \textbf{2.83} & \textbf{13.78} & \textbf{0.187} & \textbf{27.46} \\
    $\psi(x)$ & Ungated & 2.88 & 13.89 & 0.201 & \textbf{27.46} \\
    $x$ & Gated & 2.84 & 14.48 & 0.211 & 27.45 \\
    \bottomrule
\end{tabular}
\normalsize
\caption{
    \textbf{Transport Space and Sinkhorn Scheduler.}
    Transport in the learned feature space $\psi(x)$ or clean latent space $x$, with gated (\Ours{}) or ungated (\Oursu{}) Sinkhorn updates.
}
\label{tab:ablation_transport}
\vspace{-2.0pt}
}{%
    \vspace{-4pt}\paragraph{Transport Space and Sinkhorn Scheduler.}}
\Cref{tab:ablation_transport} evaluates the effects of the transport space and the gated Sinkhorn scheduler on \Ours{}.
Applying Sinkhorn at every iteration (ungated) yields worse performance with more computational overhead.
Finite-sample Sinkhorn errors can disrupt reliable IKL guidance (\cref{subsec:complementarity}), motivating our gated scheduler to activate transport only for samples with weak teacher support.\vspace{-4pt}

Computing transport in the learned feature space $\psi(x)$ yields better performance than computing it in the clean latent space $x$, consistent with observations in previous work~\citep{han2026:wflow,feng2026:representation,deng2026:drifting}.
However, feature-space transport requires additional forward passes and backpropagation through the feature encoder, whereas clean-latent transport offers competitive quality at a lower computational cost.
We therefore compute the Sinkhorn transport in the clean latent space for the large scale generative models~\citep{wan2025,wu2025qwenimage} to balance sample quality and computational cost.
\end{experimentinset}

\vspace{-2pt}\subsection{Sample Diversity}\vspace{-4pt}
\label{sec:sample_diversity}
We compare the diversity of samples generated by DMD2~\citep{yin2024:dmd2} and \Ours{}.
For images, following DAS~\citep{kim2025:das}, we report truncated CLIP entropy (TCE) and mean pairwise LPIPS distance (LPIPS MPD).
TCE measures the spread of generated images in CLIP feature space, while LPIPS MPD measures the average perceptual difference between image pairs.
For videos, following PDD~\citep{shaul2026:pdd}, we generate five videos per prompt and compute mean pairwise $\ell_2$ distances in the feature spaces of V-JEPA 2~\citep{assran2025:vjepa2} and VideoMAE V2~\citep{wang2023:videomaev2}.\vspace{-2pt}

As shown in \cref{tab:diversity}, \Ours{} achieves higher scores than DMD2 on all reported diversity metrics across applications.
These results are consistent with the toy example in \cref{fig:motivation} and suggest that the global transport of Sinkhorn can alleviate the mode collapse associated with reverse-KL-based distribution matching.

\begin{table}[t]\vspace{-3pt}
    \centering
    \begin{threeparttable}
    \renewcommand{\arraystretch}{1.15}
    \setlength{\tabcolsep}{6pt}
    \scriptsize
    \begin{tabular}{l cc cc cc}
        \toprule
        & \multicolumn{2}{c}{FLUX} & \multicolumn{2}{c}{Qwen-Image} & \multicolumn{2}{c}{Wan 14B} \\
        \cmidrule(lr){2-3} \cmidrule(lr){4-5} \cmidrule(lr){6-7}
        Method & TCE $\uparrow$ & LPIPS MPD $\uparrow$ & TCE $\uparrow$ & LPIPS MPD $\uparrow$ & V-JEPA2 $\uparrow$ & VideoMAE $\uparrow$ \\
        \midrule
        Teacher & 7.25 & 0.469 & 5.86 & 0.397 & 28.91 & 18.36 \\
        \midrule
        DMD2~\tabcite{yin2024:dmd2} & 5.93 & 0.206 & 4.72 & 0.161 & 16.78 & 13.74 \\
        \rowcolor{gray!10}
        {\Oursbf{}} (Ours) & 6.75 & 0.348 & 5.43 & 0.301 & 20.68 & 13.85 \\
        \bottomrule
    \end{tabular}
    \normalsize
    \caption{
        \textbf{Diversity Comparison.}
        Across the applications, \Ours{} achieves higher diversity than DMD2, suggesting that Sinkhorn global transport helps mitigate mode collapse.
    }
    \label{tab:diversity}
    \end{threeparttable}
\vspace{-3pt}\end{table}

\vspace{-6pt}\section{Conclusion}\vspace{-4pt}
\label{sec:conclusion}

In this work, we proposed \Ours{}, a diffusion distillation method that combines local score guidance of DMD with global transport.
We first connected DMD to WGF by showing that the fixed-point relation with IKL as the energy functional recovers the DMD gradient.
This connection enabled us to explore energy functionals that address unreliable guidance for outliers and limited mode coverage in DMD.
Specifically, we augmented IKL with Sinkhorn divergence and introduced an adaptive scheduler to focus transport guidance on samples with relatively weak teacher support.
Experiments showed that \Ours{} improves sample quality and diversity over strong diffusion distillation baselines in image and video generation with only marginal training overhead than DMD2.
We discuss limitations and future work in \cref{sec:limitations}.

\begingroup\small
\bibliography{main}
\bibliographystyle{arxiv}
\endgroup

\ifpaper
\else
    \clearpage
    \newpage
    \appendix
    \crefalias{section}{appendix}
    \crefalias{subsection}{appendix}
    \vspace{-2pt}\section*{Appendix}\vspace{-2pt}
    \vspace{-2pt}\section{Algorithm}\vspace{-2pt}
\label{app:algorithm}

\Cref{alg:sdmd} presents \Ours{} training, with Sinkhorn barycentric projections computed by \cref{alg:sinkhorn}.
We alternate between updating the generator and optimizing the discriminator and student score network.
Transport can optionally be computed in the feature space $\psi(x)$ (\cref{sec:practical}).

\begin{algorithm}[H]
\small
\DontPrintSemicolon
\SetAlgoNlRelativeSize{-1}
\caption{\Ours{} training}
\label{alg:sdmd}
\KwIn{Teacher score $\nabla \log p_t$, generator $f_\theta$, discriminator $D_\phi$, transport weight $w$, entropic regularization $\beta$, gate threshold $\lambda$, batch sizes $M_q$ and $M_p$, Sinkhorn iterations $L$.}
\KwOut{Few-step generator $f_\theta$.}
\While{not converged}{
    Sample $\{z^{(i)}\}_{i=1}^{M_q} \sim p_\mathrm{ref}$\;
    Sample $\{y^{(j)}\}_{j=1}^{M_p} \sim p$\;
    $\{x^{(i)}\}_{i=1}^{M_q} \leftarrow \{f_\theta(z^{(i)})\}_{i=1}^{M_q}$ \tcp*{Few-step sampling}
    Sample noise levels $\{t_i\}_{i=1}^{M_q}$ and $\{\epsilon^{(i)}\}_{i=1}^{M_q}\sim\mathcal{N}(0,I_d)$\;
    $x_{t_i}^{(i)}\leftarrow \alpha_{t_i}x^{(i)}+\sigma_{t_i}\epsilon^{(i)}$ \;
    $V_\mathrm{IKL}(x^{(i)})\leftarrow \alpha_{t_i}\left(\nabla\log p_{t_i}(x_{t_i}^{(i)})-\nabla\log q_{t_i}(x_{t_i}^{(i)})\right)$ \tcp*{\cref{eq:dmd_wgf}}
    $V_\mathrm{sDMD}(x^{(i)})\leftarrow V_\mathrm{IKL}(x^{(i)})$\;
    \If(\tcp*[f]{\cref{eq:lambda}}){$\lambda_\phi>\lambda$}{
        $T_{q,p}\leftarrow \operatorname{Sinkhorn}(\{x^{(i)}\}_{i=1}^{M_q},\{y^{(j)}\}_{j=1}^{M_p},\beta,L)$ \tcp*{\cref{alg:sinkhorn}}
        $T_{q,q}\leftarrow \operatorname{Sinkhorn}(\{x^{(i)}\}_{i=1}^{M_q},\{x^{(i)}\}_{i=1}^{M_q},\beta,L)$ \;
        $V_{S_\beta}(x^{(i)})\leftarrow T_{q,p}(x^{(i)})-T_{q,q}(x^{(i)})$\;
        $V_\mathrm{sDMD}(x^{(i)})\leftarrow V_\mathrm{IKL}(x^{(i)})+w\,V_{S_\beta}(x^{(i)})$\;
    }
    $\mathcal{L}_\mathrm{sDMD}(\theta,\phi)\leftarrow \mathbb{E}_{z\sim p_\mathrm{ref}}\norm{f_\theta(z)-\mathrm{sg}\!\left(f_\theta(z)+V_\mathrm{sDMD}(f_\theta(z))\right)}^2
    +\mathcal{L}_\mathrm{adv}(\phi,\theta)$ \tcp*{\cref{eq:sdmd_gated}}
    Update $(\theta,\phi)$ with $\nabla_{\theta,\phi}\mathcal{L}_\mathrm{sDMD}$\;
    Update $\nabla \log q_t$ by denoising score matching on $\{x^{(i)}\}_{i=1}^{M_q}$\;
}
\end{algorithm}

\begin{algorithm}[H]
\small
\DontPrintSemicolon
\SetAlgoNlRelativeSize{-1}
\SetKwProg{Fn}{Function}{:}{end}
\caption{Sinkhorn barycentric projection}
\label{alg:sinkhorn}
\KwIn{Samples $\{x^{(i)}\}_{i=1}^{N}$ and $\{y^{(j)}\}_{j=1}^{M}$, regularization $\beta>0$, iterations $L\geq 1$.}
\KwOut{Barycentric projections $\{T(x^{(i)})\}_{i=1}^{N}$.}
\Fn{$\operatorname{Sinkhorn}(\{x^{(i)}\}_{i=1}^{N},\{y^{(j)}\}_{j=1}^{M},\beta,L)$}{
    $a\leftarrow \mathbf{1}_N/N$\;
    $b\leftarrow \mathbf{1}_M/M$\;
    $K\leftarrow \left[\exp\!\left(-\norm{x^{(i)}-y^{(j)}}^2/(2\beta)\right)\right]_{i,j}$\;
    $v\leftarrow \mathbf{1}_M$\;
    \For{$\ell=1,\dots,L$}{
        $u\leftarrow a\oslash(Kv)$\;
        $v\leftarrow b\oslash(K^\top u)$\;
    }
    $\Pi\leftarrow \operatorname{diag}(u)K\operatorname{diag}(v)$\;
    $T(x^{(i)})\leftarrow \dfrac{\sum_{j=1}^{M}\Pi_{ij}y^{(j)}}{\sum_{j=1}^{M}\Pi_{ij}}$,\quad $i=1,\dots,N$\;
    \Return{$\{T(x^{(i)})\}_{i=1}^{N}$}\;
}
\end{algorithm}

\newpage
\clearpage

\vspace{-2pt}\section{Derivations}\vspace{-2pt}
\label{app:derivations}

\vspace{-2pt}\subsection{Connection of DMD to WGF}\vspace{-2pt}
\label{app:dmd_gradient}

We derive the connection of DMD~\citep{yin2024:dmd} to WGF presented in \cref{sec:dmd_wgf}.
Starting from the IKL energy, we derive its WGF velocity and substitute it into the fixed-point relation to recover the DMD gradient.

\vspace{-2pt}\paragraph{IKL Energy.}
Let $q=(f_\theta)_\#p_\mathrm{ref}$ be the student distribution and $p$ the target distribution.
We choose the integral KL divergence~\citep{luo2023:diffinstruct} as the energy functional:
\begin{equation}
    \mathcal{F}_\mathrm{IKL}(q)=\E_t\!\left[\KL(q_t\,\|\,p_t)\right],
\end{equation}
where $t$ denotes the diffusion noise level and the perturbed marginals are:
\begin{equation}
    q_t(y)=\int k_t(y\mid x)\,dq(x),
    \qquad p_t(y)=\int k_t(y\mid x)\,dp(x),
    \qquad k_t(y\mid x)=\mathcal N(y;\alpha_t x,\sigma_t^2 I_d).
\end{equation}\vspace{-2pt}

We use continuous noise schedules on $t\in[0,1]$ satisfying:
\begin{equation}
    \alpha_0=\sigma_1=0,\qquad \alpha_1=\sigma_0=1,\qquad
    \dot\alpha_t>0,\quad\dot\sigma_t<0\quad(0<t<1).
    \label{eq:app_noise_schedule}
\end{equation}
Thus $0\leq\alpha_t,\sigma_t\leq1$, with $p_{t=0}=\mathcal N(0,I_d)$ and $p_{t=1}=p$ the clean target.
The expectations over $t$ use sampled noise levels in $(0,1)$, and the noised marginals have positive smooth densities.

\vspace{-2pt}\paragraph{IKL Gradient Flow.}
By \cref{eq:wgf}, the velocity is the negative spatial gradient of the first variation of $\mathcal F_\mathrm{IKL}$.
To compute this variation, consider a perturbation $q_h=q+h\chi$, where $\chi$ is a finite measure with zero total mass and $q_h$ remains in $\mathcal P_2(\R^d)$.
Writing $\chi_t(y)=\int k_t(y\mid x)\,d\chi(x)$, we have:
\begin{equation}
\begin{aligned}
    \left.\frac{d}{dh}\mathcal{F}_\mathrm{IKL}(q_h)\right|_{h=0}
    &=\E_t\!\left[\int\chi_t(y)\left(\log\frac{q_t(y)}{p_t(y)}+1\right)dy\right] \\
    &=\int\E_t\!\left[\int k_t(y\mid x)\left(\log\frac{q_t(y)}{p_t(y)}+1\right)dy\right]d\chi(x).
\end{aligned}
\end{equation}
Since $\chi$ has zero total mass, we can omit the additive constant and write:
\begin{equation}
    \frac{\delta\mathcal{F}_\mathrm{IKL}}{\delta q}(x)
    =\E_t\!\left[\int k_t(y\mid x)\log\frac{q_t(y)}{p_t(y)}\,dy\right].
\end{equation}
Using $\nabla_x k_t(y\mid x)=-\alpha_t\nabla_y k_t(y\mid x)$ and integrating by parts yields:
\begin{equation}
\begin{aligned}
    \nabla_x\frac{\delta\mathcal{F}_\mathrm{IKL}}{\delta q}(x)
    &=-\E_t\!\left[\alpha_t\int\nabla_y k_t(y\mid x)\log\frac{q_t(y)}{p_t(y)}\,dy\right] \\
    &=\E_t\!\left[\alpha_t\int k_t(y\mid x)\bigl[\nabla_y\log q_t(y)-\nabla_y\log p_t(y)\bigr]dy\right].
\end{aligned}
\end{equation}
Taking the negative gradient and writing the kernel integral gives:
\begin{equation}
\begin{aligned}
    V_\mathrm{IKL}(x)
    &=-\nabla_x\frac{\delta\mathcal F_\mathrm{IKL}}{\delta q}(x) \\
    &=\E_{t,\epsilon}\!\left[\alpha_t\bigl(\nabla\log p_t(x_t)-\nabla\log q_t(x_t)\bigr)\right],
\end{aligned}
    \label{eq:app_velocity}
\end{equation}
where $x_t=\alpha_t x+\sigma_t\epsilon$, $\epsilon\sim\mathcal N(0,I_d)$.
This recovers the IKL velocity in \cref{eq:dmd_wgf}.\vspace{-2pt}

\vspace{-2pt}\paragraph{Recovering the DMD Gradient.}
Substituting $V_\mathrm{IKL}$ into \cref{eq:fixed_point} gives:
\begin{equation}
    \mathcal L(\theta)=\E_{z\sim p_\mathrm{ref}}\norm{f_\theta(z)-\mathrm{sg}\!\left(f_\theta(z)+V_\mathrm{IKL}(f_\theta(z))\right)}^2.
\end{equation}
The stop-gradient holds the regression target fixed, and only $f_\theta(z)$ contributes to the gradient:
\begin{equation}
\begin{aligned}
    \nabla_\mathrm{IKL}(\theta)\coloneqq\nabla_\theta\mathcal L(\theta)
    &=-2\,\E_z\!\left[V_\mathrm{IKL}(f_\theta(z))\frac{\partial f_\theta(z)}{\partial\theta}\right] \\
    &=-2\,\E_{t,z,\epsilon}\!\left[\alpha_t\bigl(\nabla\log p_t(x_t)-\nabla\log q_t(x_t)\bigr)\frac{\partial f_\theta(z)}{\partial\theta}\right] \\
    &=2\,\nabla_\mathrm{DMD}(\theta),
\end{aligned}
    \label{eq:app_loss_grad}
\end{equation}
where $x_t=\alpha_t f_\theta(z)+\sigma_t\epsilon$.
Thus, the fixed-point relation under the IKL energy recovers the DMD gradient in \cref{eq:dmd} up to a constant factor, as stated in \cref{eq:ikl_gradient}.\vspace{-2pt}

\vspace{-2pt}\paragraph{Comparison with Concurrent and Previous Works.}
DMD~\citep{yin2024:dmd} uses diffusion-perturbed scores to approximate the clean reverse KL gradient, without an explicit connection to WGF.
Diff-Instruct~\citep{luo2023:diffinstruct} derives the IKL gradient for generator training without connecting it to WGF.
ProlificDreamer~\citep{wang2023:prolificdreamer} derives a WGF update for 3D particles through a renderer, without considering fixed-point regression.
While concurrent works~\citep{gretton2026:wgfdrifting,turan2026:scorematching} analyze kernel-smoothed KL and fixed-point training for drifting models, they do not explicitly study IKL averaged over diffusion noise levels or its connection to DMD.
We focus on diffusion distillation with pretrained teacher scores, recovering the widely used DMD update through fixed-point regression under the IKL energy.
This WGF perspective provides a common framework for deriving and combining velocity fields from complementary energy functionals to address the limitations of DMD in few-step image and video generation.

\vspace{-2pt}\subsection{IKL Failure and Exact OT Regularization}\vspace{-2pt}
\label{app:ot_regularization}

We first give an IKL failure case, then show how exact OT can restore contraction toward the target.

\vspace{-2pt}\subsubsection{Stationarity and Negative Curvature of IKL}\vspace{-2pt}
Take $p=\tfrac12\mathcal N(-m,s^2)+\tfrac12\mathcal N(m,s^2)$ with $m>s>0$ and $q_0=\delta_0$.
The perturbed densities are even, their score difference is odd, and the noise at $x=0$ is symmetric.
Consequently $V_\mathrm{IKL}(0)=0$, so $q_\tau=\delta_0$ is a stationary solution of the IKL continuity equation despite $q_\tau\ne p$.
This example rules out convergence from arbitrary initializations.
Note that it does not show failure from smooth positive densities.

Separately, IKL need not be convex along Wasserstein geodesics, even for a Gaussian target.
Take $p=\mathcal N(0,b^2)$ and $q_r=\mathcal N(0,r^2)$ with $b,r>0$.
Linear interpolation in $r$ is a Wasserstein geodesic, and at each noise level:
\begin{equation}
\begin{aligned}
    \KL(q_{r,t}\|p_t)
    &=\frac12\left[
        \frac{\sigma_t^2+\alpha_t^2r^2}{\sigma_t^2+\alpha_t^2b^2}
        -1-\log\frac{\sigma_t^2+\alpha_t^2r^2}{\sigma_t^2+\alpha_t^2b^2}
    \right],\\
    \lim_{r\to0^+}\frac{d^2}{dr^2}\KL(q_{r,t}\|p_t)
    &=-\frac{\alpha_t^4b^2}{\sigma_t^2(\alpha_t^2b^2+\sigma_t^2)}<0,
\end{aligned}
    \label{eq:ikl_gaussian_curvature}
\end{equation}
where $q_{r,t}$ is the diffusion-perturbed law of $q_r$.
For $t$ sampled from $(0,t_{\max}]$ with $t_{\max}<1$, the KL curvature remains negative after averaging over $t$ for sufficiently small $r>0$.
While negative curvature alone does not imply nonconvergence, it shows that IKL need not be convex, even for Gaussian targets.

\vspace{-2pt}\subsubsection{Exact OT Contracts Toward the Target}\vspace{-2pt}
With our cost convention, exact OT is $\tfrac12W_2^2(q,p)$.
For an absolutely continuous source $q$, let $T^0_{q,p}$ be the optimal transport map to $p$ and define $V_\mathrm{OT}(x)=T^0_{q,p}(x)-x$.
The superscript $0$ denotes zero entropic regularization and $T_{q,p}$ in the main text denotes the entropic barycentric map.
The squared-distance chain rule~\citep{ambrosio2008:gradient} gives:
\begin{equation}
    \frac{d}{d\tau}\frac12W_2^2(q_\tau,p)
    =-\int V_\mathrm{OT}(x)^\top V_\tau(x)\,dq_\tau(x),
    \label{eq:transport_distance_chain}
\end{equation}
where $V_\tau$ is the flow velocity and $V_\mathrm{OT}$ is evaluated at the current distribution $q_\tau$.
Throughout the convergence arguments, a flow is locally absolutely continuous in $W_2$, has a density at almost every time, and has square-integrable velocities.
Setting $V_\tau=V_\mathrm{OT}$ gives:
\begin{equation}
    \frac{d}{d\tau}\frac12W_2^2(q_\tau,p)=-W_2^2(q_\tau,p),
    \qquad W_2(q_\tau,p)=e^{-\tau}W_2(q_0,p).
    \label{eq:exact_ot_alone_contraction}
\end{equation}

\vspace{-2pt}\subsubsection{Convergence of IKL with Exact OT}\vspace{-2pt}

\begin{theorem}[Convergence of IKL with exact OT]
\label{thm:exact_ot_contraction}
Let $p\in\mathcal P_2(\mathbb R^d)$ and consider the energy:
\begin{equation}
    \widetilde{\mathcal F}(q)
    =\mathcal F_\mathrm{IKL}(q)+\frac{w}{2}W_2^2(q,p),
    \qquad w>0.
    \label{eq:exact_ot_energy}
\end{equation}

Let $q_\tau$ be a Wasserstein gradient flow of \cref{eq:exact_ot_energy}.
Define:
\begin{equation}
\begin{aligned}
    M_t&=\sup_y\lambda_{\max}\!\left(
        \operatorname{Cov}_p(X\mid\alpha_tX+\sigma_t\epsilon=y)\right),\\
    L&=\E_t\!\left[\frac{\alpha_t^4}{\sigma_t^4}M_t\right]<\infty,
    \qquad X\sim p,\quad\epsilon\sim\mathcal N(0,I_d),
\end{aligned}
    \label{eq:ikl_posterior_curvature}
\end{equation}
where $X$ and $\epsilon$ are independent, and $\lambda_{\max}$ denotes the largest eigenvalue of the covariance matrix.
Assume differentiation under the integrals and expectations and integration by parts are valid for the OT interpolations in \cref{eq:ikl_transport_path}.
Then the flow satisfies:
\begin{equation}
    W_2(q_\tau,p)\leq e^{-(w-L/2)\tau}W_2(q_0,p).
    \label{eq:exact_ot_contraction}
\end{equation}
\end{theorem}
When $w>L/2$, the right-hand side of \cref{eq:exact_ot_contraction} decays exponentially to zero, proving convergence of the combined IKL and OT flow to $p$.

\begin{proof}
We differentiate IKL along the OT interpolation from the current distribution $q$ to $p$:
\begin{equation}
\begin{aligned}
    x_u&=(1-u)x+uT^0_{q,p}(x)=x+uV_\mathrm{OT}(x),\\
    q_u&=(\mathrm{id}+uV_\mathrm{OT})_\#q,\qquad
    q_{u=0}=q,\quad q_{u=1}=p,\quad 0\leq u\leq1.
\end{aligned}
    \label{eq:ikl_transport_path}
\end{equation}
Here $q$ and its optimal map stay fixed as $u$ varies; $\tau$ is flow time and $t$ is the diffusion noise level.
Differentiating $q_{u,t}(y)=\int k_t(y\mid x_u)\,dq(x)$ gives:
\begin{equation}
\begin{aligned}
    \partial_u q_{u,t}(y)
    &=-\alpha_t\int V_\mathrm{OT}(x)^\top\nabla_y k_t(y\mid x_u)\,dq(x),\\
    \partial_u^2q_{u,t}(y)
    &=\alpha_t^2\int V_\mathrm{OT}(x)^\top
        \nabla_y^2k_t(y\mid x_u)V_\mathrm{OT}(x)\,dq(x).
\end{aligned}
    \label{eq:ikl_path_density_derivatives}
\end{equation}
Substituting \cref{eq:ikl_path_density_derivatives} into the second derivative of KL and integrating by parts yields:
\begin{equation}
\begin{aligned}
    \frac{d^2}{du^2}\KL(q_{u,t}\|p_t)
    &=\int\frac{(\partial_u q_{u,t})^2}{q_{u,t}}\,dy
      +\int(\partial_u^2q_{u,t})\log\frac{q_{u,t}}{p_t}\,dy\\
    &=\int\frac{(\partial_u q_{u,t})^2}{q_{u,t}}\,dy\\
    &\quad+\alpha_t^2\int\!\int k_t(y\mid x_u)V_\mathrm{OT}(x)^\top
        \nabla_y^2\log\frac{q_{u,t}(y)}{p_t(y)}V_\mathrm{OT}(x)\,dy\,dq(x).
\end{aligned}
    \label{eq:ikl_second_variation}
\end{equation}

To bound the negative curvature of IKL, differentiate $p_t(y)=\int k_t(y\mid x)\,dp(x)$:
\begin{equation}
    \nabla_y^2\log p_t(y)
    =-\frac{I_d}{\sigma_t^2}
      +\frac{\alpha_t^2}{\sigma_t^4}
        \operatorname{Cov}_p(X\mid\alpha_tX+\sigma_t\epsilon=y).
    \label{eq:ikl_posterior_hessian}
\end{equation}
Apply \cref{eq:ikl_posterior_hessian} also to $q_u$, subtract, and use positive semidefiniteness of covariance and \cref{eq:ikl_posterior_curvature}:
\begin{equation}
\begin{aligned}
    \nabla_y^2\log\frac{q_{u,t}(y)}{p_t(y)}
    &=\frac{\alpha_t^2}{\sigma_t^4}
        \operatorname{Cov}_{q_u}(X\mid\alpha_tX+\sigma_t\epsilon=y)\\
    &\quad-\frac{\alpha_t^2}{\sigma_t^4}
        \operatorname{Cov}_p(X\mid\alpha_tX+\sigma_t\epsilon=y)\\
    &\succeq-\frac{\alpha_t^2}{\sigma_t^4}M_tI_d.
\end{aligned}
    \label{eq:ikl_log_ratio_hessian_bound}
\end{equation}
Substitute \cref{eq:ikl_log_ratio_hessian_bound} into \cref{eq:ikl_second_variation}, drop the nonnegative first term, and use $\int k_t(y\mid x_u)\,dy=1$:
\begin{equation}
\begin{aligned}
    \frac{d^2}{du^2}\mathcal F_\mathrm{IKL}(q_u)
    &=\E_t\!\left[\frac{d^2}{du^2}\KL(q_{u,t}\|p_t)\right]\\
    &\geq-\E_t\!\left[\frac{\alpha_t^4}{\sigma_t^4}M_t\right]
        \int\|V_\mathrm{OT}(x)\|^2\,dq(x)\\
    &=-LW_2^2(q,p).
\end{aligned}
    \label{eq:ikl_curvature_lower_bound}
\end{equation}

To prove convergence, we must control how much the IKL velocity can oppose transport toward $p$.
Indeed, substituting $V_\tau=V_\mathrm{IKL}+wV_\mathrm{OT}$ into \cref{eq:transport_distance_chain} gives:
\begin{equation}
    \frac{d}{d\tau}\frac12W_2^2(q_\tau,p)
    =-\int V_\mathrm{OT}(x)^\top V_\mathrm{IKL}(x)\,dq_\tau(x)
      -wW_2^2(q_\tau,p).
    \label{eq:combined_ot_distance_chain}
\end{equation}
The OT term decreases the distance, while the IKL term can increase it if the velocity inner product is negative.
We use the curvature bound to lower-bound this inner product at an arbitrary current distribution $q$.

First, \cref{eq:ikl_curvature_lower_bound} implies that $\mathcal F_\mathrm{IKL}(q_u)+\tfrac L2u^2W_2^2(q,p)$ is convex in $u$.
Therefore, using $q_{u=0}=q$, $q_{u=1}=p$, and $\mathcal F_\mathrm{IKL}(p)=0$ gives:
\begin{equation}
    \mathcal F_\mathrm{IKL}(q_u)
    \leq(1-u)\mathcal F_\mathrm{IKL}(q)
       +\frac L2u(1-u)W_2^2(q,p).
    \label{eq:ikl_target_curvature}
\end{equation}
Subtracting $\mathcal F_\mathrm{IKL}(q)$, dividing by $u>0$, and taking $u\to0^+$ bounds the initial energy increase along the OT direction:
\begin{equation}
    \left.\frac{d}{du}\mathcal F_\mathrm{IKL}(q_u)\right|_{u=0}
    \leq-\mathcal F_\mathrm{IKL}(q)+\frac L2W_2^2(q,p).
    \label{eq:ikl_initial_slope_bound}
\end{equation}

Next, we express this initial slope as the negative inner product of the two velocities.
The interpolation satisfies
$\left.\partial_u q_u\right|_{u=0}=-\nabla_x\cdot(qV_\mathrm{OT})$.
The first-variation chain rule and integration by parts give:
\begin{equation}
\begin{aligned}
    \left.\frac{d}{du}\mathcal F_\mathrm{IKL}(q_u)\right|_{u=0}
    &= \int\frac{\delta\mathcal F_\mathrm{IKL}}{\delta q}(x)
        \left.\partial_u q_u(x)\right|_{u=0}\,dx\\
    &=-\int\frac{\delta\mathcal F_\mathrm{IKL}}{\delta q}(x)
        \nabla_x\cdot\bigl(q(x)V_\mathrm{OT}(x)\bigr)\,dx\\
    &=\int\left[\nabla_x\frac{\delta\mathcal F_\mathrm{IKL}}{\delta q}(x)\right]^\top
        V_\mathrm{OT}(x)\,dq(x)\\
    &=-\int V_\mathrm{IKL}(x)^\top V_\mathrm{OT}(x)\,dq(x).
\end{aligned}
    \label{eq:ikl_path_first_derivative}
\end{equation}
Combining \cref{eq:ikl_initial_slope_bound,eq:ikl_path_first_derivative} yields the required alignment bound:
\begin{equation}
    \int V_\mathrm{IKL}(x)^\top V_\mathrm{OT}(x)\,dq(x)
    \geq\mathcal F_\mathrm{IKL}(q)-\frac L2W_2^2(q,p).
    \label{eq:ikl_transport_alignment}
\end{equation}
Since $\mathcal F_\mathrm{IKL}(q)\geq0$, the inner product cannot fall below $-\tfrac L2W_2^2(q,p)$.

Finally, apply this bound at $q=q_\tau$ in \cref{eq:combined_ot_distance_chain}:
\begin{equation}
\begin{aligned}
    \frac{d}{d\tau}\frac12W_2^2(q_\tau,p)
    &\leq-\mathcal F_\mathrm{IKL}(q_\tau)
        -(w-L/2)W_2^2(q_\tau,p)\\
    &\leq-(w-L/2)W_2^2(q_\tau,p).
\end{aligned}
    \label{eq:exact_ot_distance_decay}
\end{equation}
Thus $w>L/2$ ensures that OT dominates any opposing IKL contribution.
Multiplying by the integrating factor gives
\begin{equation*}
    \frac{d}{d\tau}\left[e^{2(w-L/2)\tau}W_2^2(q_\tau,p)\right]\leq0.
\end{equation*}
Integrating from $0$ to $\tau$ and taking square roots yields
\begin{equation*}
    W_2(q_\tau,p)\leq e^{-(w-L/2)\tau}W_2(q_0,p),
\end{equation*}
which proves \cref{eq:exact_ot_contraction}.
\end{proof}

\begin{corollary}[A sufficient weight for Gaussian-mixture targets]
\label{cor:mixture_ot_contraction}
Under the flow and calculus assumptions of \cref{thm:exact_ot_contraction}, let $p=\sum_j\pi_j\mathcal N(\mu_j,s^2I_d)$ have finitely many components, $s>0$, and positive weights summing to one.
Define:
\begin{equation*}
    \Delta=\max_{i,j}\|\mu_i-\mu_j\|,\qquad
    v_t=\alpha_t^2s^2+\sigma_t^2.
\end{equation*}
Then $w>L_\mathrm{mix}/2$ suffices for \cref{eq:exact_ot_contraction} with $L$ replaced by:
\begin{equation}
    L_\mathrm{mix}=\E_t\!\left[
        \frac{\alpha_t^4s^2}{\sigma_t^2v_t}
        +\frac{\alpha_t^4\Delta^2}{4v_t^2}
    \right],
    \label{eq:ikl_mixture_curvature}
\end{equation}
which is finite when the noise level is bounded away from zero.
\end{corollary}

\begin{proof}
To compute the target posterior covariance, let $J$ be the component index. Gaussian conditioning gives:
\begin{equation}
\begin{aligned}
    \E[X\mid\alpha_tX+\sigma_t\epsilon=y,J=j]
    &=\frac{\alpha_t s^2}{v_t}y+\frac{\sigma_t^2}{v_t}\mu_j,\\
    \operatorname{Cov}(X\mid\alpha_tX+\sigma_t\epsilon=y,J=j)
    &=\frac{s^2\sigma_t^2}{v_t}I_d.
\end{aligned}
    \label{eq:ikl_mixture_component_posterior}
\end{equation}
Substituting \cref{eq:ikl_mixture_component_posterior} into the law of total covariance gives:
\begin{equation}
\begin{aligned}
    &\operatorname{Cov}(X\mid\alpha_tX+\sigma_t\epsilon=y)\\
    &\quad=\frac{s^2\sigma_t^2}{v_t}I_d
      +\frac{\sigma_t^4}{v_t^2}
        \operatorname{Cov}(\mu_J\mid\alpha_tX+\sigma_t\epsilon=y).
\end{aligned}
    \label{eq:ikl_mixture_posterior_covariance}
\end{equation}

To bound $L$, note that in any unit direction, projected means span an interval of length at most $\Delta$, so their variance is at most $\Delta^2/4$.
Using this in \cref{eq:ikl_mixture_posterior_covariance} and then \cref{eq:ikl_posterior_curvature} gives:
\begin{equation*}
\begin{aligned}
    M_t&\leq\frac{s^2\sigma_t^2}{v_t}
        +\frac{\sigma_t^4\Delta^2}{4v_t^2},\\
    L&\leq\E_t\!\left[
        \frac{\alpha_t^4s^2}{\sigma_t^2v_t}
        +\frac{\alpha_t^4\Delta^2}{4v_t^2}\right]
      =L_\mathrm{mix} <\infty.
\end{aligned}
\end{equation*}
Then apply \cref{thm:exact_ot_contraction} with $w>L_\mathrm{mix}/2\geq L/2$.
\end{proof}

\vspace{-2pt}\subsection{Identifiability of \Ours{}}\vspace{-2pt}
\label{app:identifiability}
We establish identifiability of the velocity induced by \cref{eq:sdmd_energy}, \ie~$V_\text{sDMD-U}(x)=0$ for every $x\in\mathbb R^d$ implies $p=q$.
Let $p,q$ have compact support, and fix $w,\beta>0$.
Assume $\alpha_t,\sigma_t>0$ and sampled noise levels bounded away from zero.

\vspace{-2pt}\paragraph{IKL Identifiability.}
Since noise averaging preserves convexity and Gaussian smoothing is injective, the identifiability argument for kernel-smoothed scores~\citep{turan2026:scorematching} extends to IKL:
\begin{equation*}
    V_\mathrm{IKL}\equiv0
    \quad\Longleftrightarrow\quad \mathcal F_\mathrm{IKL}(q)=0
    \quad\Longleftrightarrow\quad q=p.
\end{equation*}\vspace{-2pt}

\vspace{-2pt}\paragraph{Sinkhorn Identifiability.}
Following \citep{he2026:sinkhorndrifting}, the exact Sinkhorn field satisfies:
\begin{equation*}
    V_{S_\beta}\equiv0
    \quad\Longleftrightarrow\quad S_\beta(q,p)=0
    \quad\Longleftrightarrow\quad q=p.
\end{equation*}\vspace{-2pt}

\vspace{-2pt}\paragraph{\Ours{} Identifiability.}
Since noising is linear and KL is convex, $\mathcal F_\mathrm{IKL}$ is convex in $q$.
$S_\beta$ is also convex in $q$ for the quadratic cost~\citep{feydy2019:interpolating}.
Hence $\mathcal F_\text{sDMD-U}=\mathcal F_\mathrm{IKL}+wS_\beta$ is convex under mixtures of distributions.
For $q_h=(1-h)q+hp=q+h(p-q)$ with $0<h\leq1$, convexity gives:
\begin{equation*}
    \mathcal F_\text{sDMD-U}(q_h)
    \leq(1-h)\mathcal F_\text{sDMD-U}(q)+h\mathcal F_\text{sDMD-U}(p).
\end{equation*}
Rearranging and taking $h\to0^+$, the definition of the first variation yields:
\begin{equation*}
\begin{aligned}
    \mathcal F_\text{sDMD-U}(p)-\mathcal F_\text{sDMD-U}(q)
    &\geq\lim_{h\to0^+}\frac{\mathcal F_\text{sDMD-U}(q_h)-\mathcal F_\text{sDMD-U}(q)}{h} \\
    &=\int\frac{\delta\mathcal F_\text{sDMD-U}}{\delta q}(x)\bigl[p(x)-q(x)\bigr]\,dx.
\end{aligned}
\end{equation*}
If $V_\text{sDMD-U}\equiv0$, the first variation is spatially constant, and the integral vanishes as $\int [p(x)-q(x)]\,dx=1-1=0$.
Since both divergences are nonnegative and vanish at $q=p$, we obtain:
\begin{equation*}
    0=\mathcal F_\text{sDMD-U}(p)\geq\mathcal F_\text{sDMD-U}(q)\geq0.
\end{equation*}
Since $w>0$, both $\mathcal F_\mathrm{IKL}(q)$ and $S_\beta(q,p)$ must vanish, implying $q=p$.
Conversely, $q=p$ makes both velocity terms zero, establishing:
\begin{equation*}
    V_\text{sDMD-U}\equiv0
    \quad\Longleftrightarrow\quad q=p.
\end{equation*}

\vspace{-2pt}\subsection{Convergence with Sinkhorn Regularization}\vspace{-2pt}
\label{app:sinkhorn_proxy}

In this section, we extend \cref{app:ot_regularization} to Sinkhorn flows \cref{eq:sdmd_energy}.
We first analyze how Sinkhorn regularization affects convergence (\cref{thm:sinkhorn_relative_contraction,thm:sinkhorn_error_bound}), then specialize the analysis to Gaussian distributions (\cref{prop:gaussian_sinkhorn_convergence}).

\vspace{-2pt}\subsubsection{Convergence of IKL with Sinkhorn}\vspace{-2pt}
On compact supports, $S_\beta(q,p)\to\tfrac12W_2^2(q,p)$ as $\beta\to0^+$~\citep{genevay2019:sample}.
To compare velocities, write $\pi^\beta_{q,p}$ for the entropic plan and $T^\beta_{q,p}$ for its barycentric map.
Since $V_{S_\beta}-V_\mathrm{OT}=(T^\beta_{q,p}-T^0_{q,p})-(T^\beta_{q,q}-\mathrm{id})$, the triangle and Jensen inequalities gives:
\begin{equation}
\begin{aligned}
    &\left(\int\|V_{S_\beta}(x)-V_\mathrm{OT}(x)\|^2\,dq(x)\right)^{1/2}\\
    &\quad\leq\left(\int\|y-T^0_{q,p}(x)\|^2\,d\pi^\beta_{q,p}(x,y)\right)^{1/2}
       +\sqrt{2\mathrm{OT}_\beta(q,q)}.
\end{aligned}
    \label{eq:sinkhorn_velocity_approximation}
\end{equation}
For fixed $p,q$ the Sinkhorn velocity converges to the OT velocity as $\beta\to0^+$.
This supports using Sinkhorn regularization to approximate the OT guidance in \cref{app:ot_regularization} at small $\beta$.

When the Sinkhorn velocity remains sufficiently close to the exact OT velocity along the flow, the following theorem establishes convergence of IKL and Sinkhorn flow to the target distribution.
\begin{theorem}[Convergence under relative velocity control]
\label{thm:sinkhorn_relative_contraction}
Let $p\in\mathcal P_2(\mathbb R^d)$, $w,\beta>0$, and $q_\tau$ be a regular Wasserstein gradient flow of $\mathcal F_\mathrm{IKL}+wS_\beta(\cdot,p)$.
Assume \cref{eq:ikl_posterior_curvature} with $L<\infty$ and the assumptions of \cref{thm:exact_ot_contraction}.
If a fixed $0\leq\eta<1$ satisfies:
\begin{equation}
    \left(\int\|V_{S_\beta}(x)-V_\mathrm{OT}(x)\|^2\,dq_\tau(x)\right)^{1/2}
    \leq\eta W_2(q_\tau,p)
    \quad\text{along the flow},
    \label{eq:sinkhorn_relative_approximation}
\end{equation}
then:
\begin{equation}
    W_2(q_\tau,p)\leq e^{-[w(1-\eta)-L/2]\tau}W_2(q_0,p).
    \label{eq:sinkhorn_conditional_contraction}
\end{equation}
In particular, $w>L/[2(1-\eta)]$ gives exponential convergence to $p$.
\end{theorem}

\begin{proof}
At $q=q_\tau$, set $e_\beta=V_{S_\beta}-V_\mathrm{OT}$.
Applying Cauchy--Schwarz inequality and the identity $\int\|V_\mathrm{OT}(x)\|^2\,dq(x)=W_2^2(q,p)$ gives:
\begin{align}
    \int V_\mathrm{OT}(x)^\top V_{S_\beta}(x)\,dq(x)
    &=W_2^2(q,p)+\int V_\mathrm{OT}(x)^\top e_\beta(x)\,dq(x)\notag\\
    &\geq W_2^2(q,p)-W_2(q,p)
        \left(\int\|e_\beta(x)\|^2\,dq(x)\right)^{1/2}\notag\\
    &\geq(1-\eta)W_2^2(q,p). \quad\text{(by \cref{eq:sinkhorn_relative_approximation})}
    \label{eq:sinkhorn_transport_alignment}
\end{align}

Using \cref{eq:transport_distance_chain,eq:ikl_transport_alignment,eq:sinkhorn_transport_alignment}:
\begin{align}
    \frac{d}{d\tau}\frac12W_2^2(q_\tau,p)
    &=-\int V_\mathrm{OT}(x)^\top V_\mathrm{IKL}(x)\,dq_\tau(x)
      -w\int V_\mathrm{OT}(x)^\top V_{S_\beta}(x)\,dq_\tau(x)\notag\\
    &\leq-\mathcal F_\mathrm{IKL}(q_\tau)
        -[w(1-\eta)-L/2]W_2^2(q_\tau,p)\notag\\
    &\leq-[w(1-\eta)-L/2]W_2^2(q_\tau,p).
    \label{eq:sinkhorn_relative_distance_decay}
\end{align}

With $a_\eta=w(1-\eta)-L/2$, the integrating factor gives:
\begin{equation*}
\begin{aligned}
    \frac{d}{d\tau}\left[e^{2a_\eta\tau}W_2^2(q_\tau,p)\right]&\leq0,\\
    W_2^2(q_\tau,p)&\leq e^{-2a_\eta\tau}W_2^2(q_0,p).
\end{aligned}
\end{equation*}
Taking square roots gives \cref{eq:sinkhorn_conditional_contraction}.
\end{proof}

\label{app:sinkhorn_general}

\begin{theorem}[Distance bound under general transport error]
\label{thm:sinkhorn_error_bound}
Let $p\in\mathcal P_2(\mathbb R^d)$ and $q_\tau$ be a Wasserstein gradient flow with velocity:
\begin{equation}
    V_\tau=V_\mathrm{IKL}
        +w\bigl[V_{S_{\beta(\tau)}}+\zeta_\tau\bigr],
    \label{eq:sinkhorn_inexact_flow}
\end{equation}
where $w>0$, $\beta(\tau)>0$, and $\zeta_\tau$ is transport-estimation or solver error.
Assume the regularity and calculus conditions of \cref{thm:exact_ot_contraction} and the IKL curvature bound \cref{eq:ikl_curvature_lower_bound}, with a uniform constant $L<\infty$ along the flow.
Let $b_\tau$ be a nonnegative, locally integrable error bound satisfying:
\begin{equation}
    \left(\int\|V_{S_{\beta(\tau)}}(x)-V_\mathrm{OT}(x)+\zeta_\tau(x)\|^2\,dq_\tau(x)\right)^{1/2}
    \leq b_\tau.
    \label{eq:sinkhorn_total_error}
\end{equation}
If $w>L/2$, then:
\begin{equation}
    W_2(q_\tau,p)
    \leq e^{-(w-L/2)\tau}W_2(q_0,p)
        +w\int_0^\tau e^{-(w-L/2)(\tau-s)}b_s\,ds.
    \label{eq:sinkhorn_robust_distance}
\end{equation}
\end{theorem}

\begin{proof}
The Cauchy--Schwarz argument in \cref{eq:sinkhorn_transport_alignment}, with \cref{eq:sinkhorn_total_error}, gives:
\begin{equation}
    \int V_\mathrm{OT}(x)^\top
        [V_{S_{\beta(\tau)}}(x)+\zeta_\tau(x)]\,dq_\tau(x)
    \geq W_2^2(q_\tau,p)-b_\tau W_2(q_\tau,p).
    \label{eq:sinkhorn_inexact_transport_alignment}
\end{equation}
Using \cref{eq:transport_distance_chain,eq:ikl_transport_alignment,eq:sinkhorn_inexact_transport_alignment}:
\begin{align}
    \frac{d}{d\tau}\frac12W_2^2(q_\tau,p)
    &\leq-\mathcal F_\mathrm{IKL}(q_\tau)
        -(w-L/2)W_2^2(q_\tau,p)+wb_\tau W_2(q_\tau,p)\notag\\
    &\leq-(w-L/2)W_2^2(q_\tau,p)+wb_\tau W_2(q_\tau,p).
    \label{eq:sinkhorn_inexact_distance_decay}
\end{align}

We divide \cref{eq:sinkhorn_inexact_distance_decay} by $W_2(q_\tau,p)$ where the distance is positive.
When the distance is zero, its derivative is zero almost everywhere, and the inequality still holds:
\begin{equation}
    \frac{d}{d\tau}W_2(q_\tau,p)
    \leq-(w-L/2)W_2(q_\tau,p)+wb_\tau.
    \label{eq:sinkhorn_inexact_distance_inequality}
\end{equation}
Multiplying by $e^{(w-L/2)\tau}$ and integrating proves \cref{eq:sinkhorn_robust_distance}.
\end{proof}

\begin{corollary}[Distance bound under bounded transport error]
\label{cor:sinkhorn_bounded_error}
Under the assumptions of \cref{thm:sinkhorn_error_bound}, suppose $b_\tau\leq b$ uniformly for some $b\geq0$.
Then:
\begin{align}
    W_2(q_\tau,p)&\leq e^{-(w-L/2)\tau}W_2(q_0,p)
        +\frac{wb}{w-L/2}\bigl(1-e^{-(w-L/2)\tau}\bigr),\notag\\
    \limsup_{\tau\to\infty}W_2(q_\tau,p)&\leq\frac{wb}{w-L/2}.
    \label{eq:sinkhorn_accuracy_bound}
\end{align}
\end{corollary}

\begin{proof}
Substitute $b_s\leq b$ into \cref{eq:sinkhorn_robust_distance} and integrate the exponential.
Taking the limit superior gives the second claim.
\end{proof}

At fixed $\beta$, uniform bounds on approximation, estimation, and solver errors give $b$.
The radius $wb/(w-L/2)$ approaches $b$ as $w\to\infty$ for fixed $b$.
Note that this does not mean that the flow cannot converge exactly to $p$.

\vspace{-2pt}\subsubsection{Gaussian IKL--Sinkhorn Convergence}\vspace{-2pt}
\label{app:sinkhorn_gaussian}

For Gaussian targets and initializations, we verify the velocity bound in \cref{eq:sinkhorn_relative_approximation}.
This gives exact convergence at fixed, sufficiently small $\beta>0$.

\begin{proposition}[Gaussian IKL--Sinkhorn convergence]
\label{prop:gaussian_sinkhorn_convergence}
Let $p=\mathcal N(m_p,B)$ and $q_0=\mathcal N(m_0,A_0)$ satisfy
$0<\lambda_- I_d\preceq A_0,B\preceq\lambda_+ I_d$.
Under the schedule in \cref{eq:app_noise_schedule}, assume:
\begin{equation}
    L=\E_t\!\left[
        \alpha_t^2\left(\frac{1}{\sigma_t^2}
        -\frac{1}{\alpha_t^2\lambda_++\sigma_t^2}\right)
    \right]<\infty.
    \label{eq:gaussian_ikl_curvature_bound}
\end{equation}
For $\beta>0$ and $w\geq0$, the flow of $\mathcal F_\mathrm{IKL}+wS_\beta(\cdot,p)$ from $q_0$ admits a global Gaussian solution $q_\tau=\mathcal N(m_\tau,A_\tau)$ with $\lambda_-I_d\preceq A_\tau\preceq\lambda_+I_d$.
Set $\eta=\beta^2\lambda_+^{3/2}/(8\lambda_-^{7/2})$.
If $\eta<1$ and $w>L/[2(1-\eta)]$, then:
\begin{equation}
    W_2(q_\tau,p)
    \leq e^{-[w(1-\eta)-L/2]\tau}W_2(q_0,p).
    \label{eq:gaussian_sinkhorn_convergence}
\end{equation}
\end{proposition}

\begin{proof}
For $q_\tau=\mathcal N(m_\tau,A_\tau)$, set $Z_\tau=A_\tau^{1/2}BA_\tau^{1/2}$ and define the symmetric matrices:
\begin{align*}
    C_{\mathrm{IKL},\tau}
    &=\E_t\!\left[\alpha_t^2\left\{
        (\alpha_t^2A_\tau+\sigma_t^2I_d)^{-1}
        -(\alpha_t^2B+\sigma_t^2I_d)^{-1}
    \right\}\right],\\
    C_{\beta,\tau}
    &=A_\tau^{-1/2}\left[
        (Z_\tau+\tfrac{\beta^2}{4}I_d)^{1/2}
        -(A_\tau^2+\tfrac{\beta^2}{4}I_d)^{1/2}
    \right]A_\tau^{-1/2}.
\end{align*}
Gaussian scores in \cref{eq:app_velocity} and the Gaussian maps of \citep{pooladian2022:debiaser} give:
\begin{align}
    V_\mathrm{IKL}(x)
    &=-\E_t[\alpha_t^2(\alpha_t^2B+\sigma_t^2I_d)^{-1}](m_\tau-m_p)
        +C_{\mathrm{IKL},\tau}(x-m_\tau),\notag\\
    V_{S_\beta}(x)&=m_p-m_\tau+C_{\beta,\tau}(x-m_\tau).
    \label{eq:gaussian_flow_velocities}
\end{align}

Affine velocities preserve Gaussian distributions.
Therefore, using $\E_{q_\tau}[X-m_\tau]=0$ and $\E_{q_\tau}[(X-m_\tau)(X-m_\tau)^\top]=A_\tau$, the mean and covariance evolve according to:
\begin{align}
    \dot m_\tau
    &=-\left(\E_t[\alpha_t^2(\alpha_t^2B+\sigma_t^2I_d)^{-1}]+wI_d\right)(m_\tau-m_p),\notag\\
    \dot A_\tau
    &=(C_{\mathrm{IKL},\tau}+wC_{\beta,\tau})A_\tau
        +A_\tau(C_{\mathrm{IKL},\tau}+wC_{\beta,\tau}).
    \label{eq:gaussian_flow_moments}
\end{align}

We show that the covariance remains bounded above and away from zero.
For a unit vector $u$ with $A_\tau u=\lambda_-u$, the variance is at its lower limit.
Using $B\succeq\lambda_-I_d$ and $Z_\tau\succeq\lambda_-A_\tau$ gives:
\begin{align}
    u^\top C_{\mathrm{IKL},\tau}u
    &=\E_t\!\left[\alpha_t^2\left(
        \frac{1}{\alpha_t^2\lambda_-+\sigma_t^2}
        -u^\top(\alpha_t^2B+\sigma_t^2I_d)^{-1}u
    \right)\right]\geq0,
    \label{eq:gaussian_ikl_boundary}\\
    u^\top C_{\beta,\tau}u
    &=\frac{u^\top(Z_\tau+\tfrac{\beta^2}{4}I_d)^{1/2}u
        -\sqrt{\lambda_-^2+\beta^2/4}}{\lambda_-}\geq0.
    \label{eq:gaussian_sinkhorn_boundary}
\end{align}
Thus $u^\top\dot A_\tau u=2\lambda_-u^\top(C_{\mathrm{IKL},\tau}+wC_{\beta,\tau})u\geq0$, and the variance cannot decrease at its lower limit.
At the upper limit $A_\tau u=\lambda_+u$, the inequalities reverse: $u^\top\dot A_\tau u\leq0$, and the variance cannot increase.
Hence $\lambda_-I_d\preceq A_\tau\preceq\lambda_+I_d$ for all $\tau$.
These bounds keep the covariance and its inverse bounded.

We next bound the relative velocity error without assuming that $A_\tau$ and $B$ commute.
The Gaussian OT map gives:
\begin{equation}
    V_\mathrm{OT}(x)=m_p-m_\tau
        +A_\tau^{-1/2}(Z_\tau^{1/2}-A_\tau)A_\tau^{-1/2}(x-m_\tau).
    \label{eq:gaussian_transport_velocities}
\end{equation}

Set $g_\beta(z)=\sqrt{z+\beta^2/4}-\sqrt z$.
Since $\lambda_-^2I_d\preceq Z_\tau,A_\tau^2\preceq\lambda_+^2I_d$, we use the scalar bound:
\begin{equation}
    |g_\beta'(z)|
    =\frac{\beta^2}{8\sqrt z\sqrt{z+\beta^2/4}
        (\sqrt z+\sqrt{z+\beta^2/4})}
    \leq\frac{\beta^2}{16\lambda_-^3},
    \qquad z\in[\lambda_-^2,\lambda_+^2].
    \label{eq:gaussian_scalar_lipschitz}
\end{equation}

Let $M,N$ be symmetric matrices with eigenvalues in $[\lambda_-^2,\lambda_+^2]$. Write their orthonormal eigenvectors as $u_i,v_j$, with corresponding eigenvalues $\mu_i,\nu_j$. Then \cref{eq:gaussian_scalar_lipschitz} implies:
\begin{align*}
    \|g_\beta(M)-g_\beta(N)\|_{\mathrm F}^2
    &=\sum_{i,j}|g_\beta(\mu_i)-g_\beta(\nu_j)|^2|u_i^\top v_j|^2\\
    &\leq\left(\frac{\beta^2}{16\lambda_-^3}\right)^2
        \sum_{i,j}|\mu_i-\nu_j|^2|u_i^\top v_j|^2\\
    &=\left(\frac{\beta^2}{16\lambda_-^3}\right)^2\|M-N\|_{\mathrm F}^2.
\end{align*}
Apply this with $M=Z_\tau,N=A_\tau^2$ and use
$Z_\tau-A_\tau^2=Z_\tau^{1/2}(Z_\tau^{1/2}-A_\tau)+(Z_\tau^{1/2}-A_\tau)A_\tau$:
\begin{align}
    \|g_\beta(Z_\tau)-g_\beta(A_\tau^2)\|_{\mathrm F}
    &\leq\frac{\beta^2}{16\lambda_-^3}\|Z_\tau-A_\tau^2\|_{\mathrm F}\notag\\
    &\leq\frac{\beta^2\lambda_+}{8\lambda_-^3}
        \|Z_\tau^{1/2}-A_\tau\|_{\mathrm F}.
    \label{eq:gaussian_matrix_difference_bound}
\end{align}

Let $e_\beta=V_{S_\beta}-V_\mathrm{OT}$.
The mean terms cancel. Substituting $C_{\beta,\tau}$ and using $(A_\tau^2)^{1/2}=A_\tau$ gives:
\begin{align*}
    e_\beta(x)
    &=\left[C_{\beta,\tau}
        -A_\tau^{-1/2}(Z_\tau^{1/2}-A_\tau)A_\tau^{-1/2}\right](x-m_\tau)\\
    &=A_\tau^{-1/2}\Big[(Z_\tau+\tfrac{\beta^2}{4}I_d)^{1/2}-Z_\tau^{1/2}\\
    &\qquad\qquad-(A_\tau^2+\tfrac{\beta^2}{4}I_d)^{1/2}+A_\tau\Big]
        A_\tau^{-1/2}(x-m_\tau)\\
    &=A_\tau^{-1/2}[g_\beta(Z_\tau)-g_\beta(A_\tau^2)]
        A_\tau^{-1/2}(x-m_\tau).
\end{align*}

Substitute $X=m_\tau+A_\tau^{1/2}\xi$ with $\xi\sim\mathcal N(0,I_d)$.
Using $\E\|K\xi\|^2=\|K\|_{\mathrm F}^2$ for any matrix $K$ gives:
\begin{equation}
    \left(\int\|e_\beta(x)\|^2\,dq_\tau(x)\right)^{1/2}
    =\|A_\tau^{-1/2}[g_\beta(Z_\tau)-g_\beta(A_\tau^2)]\|_{\mathrm F}.
    \label{eq:gaussian_velocity_error_norm}
\end{equation}

For the transport distance, evaluate the exact OT velocity at the same $X$:
\begin{equation*}
    V_\mathrm{OT}(X)=m_p-m_\tau
        +A_\tau^{-1/2}(Z_\tau^{1/2}-A_\tau)\xi.
\end{equation*}
The mean cross term vanishes because $\E[\xi]=0$.
Applying the same identity to the remaining random term gives:
\begin{align}
    W_2^2(q_\tau,p)
    &=\E\|V_\mathrm{OT}(X)\|^2\notag\\
    &=\|m_\tau-m_p\|^2
        +\|A_\tau^{-1/2}(Z_\tau^{1/2}-A_\tau)\|_{\mathrm F}^2.
    \label{eq:gaussian_transport_distance}
\end{align}

Inserting $A_\tau^{1/2}A_\tau^{-1/2}=I_d$ gives:
\begin{align}
    \|Z_\tau^{1/2}-A_\tau\|_{\mathrm F}
    &\leq\|A_\tau^{1/2}\|_{\mathrm{op}}
        \|A_\tau^{-1/2}(Z_\tau^{1/2}-A_\tau)\|_{\mathrm F}\notag\\
    &\leq\lambda_+^{1/2}
        \|A_\tau^{-1/2}(Z_\tau^{1/2}-A_\tau)\|_{\mathrm F}.
    \label{eq:gaussian_covariance_norm_bound}
\end{align}

Using $\|A_\tau^{-1/2}\|_{\mathrm{op}}\leq\lambda_-^{-1/2}$ in the first inequality yields:
\begin{align}
    \left(\int\|e_\beta(x)\|^2\,dq_\tau(x)\right)^{1/2}
    &\leq\frac{\beta^2\lambda_+}{8\lambda_-^{7/2}}
        \|Z_\tau^{1/2}-A_\tau\|_{\mathrm F}
        &&\text{by \cref{eq:gaussian_velocity_error_norm,eq:gaussian_matrix_difference_bound}}\notag\\
    &\leq\frac{\beta^2\lambda_+^{3/2}}{8\lambda_-^{7/2}}
        \|A_\tau^{-1/2}(Z_\tau^{1/2}-A_\tau)\|_{\mathrm F}
        &&\text{by \cref{eq:gaussian_covariance_norm_bound}}\notag\\
    &\leq\eta W_2(q_\tau,p)
        &&\text{by \cref{eq:gaussian_transport_distance}},
    \label{eq:gaussian_sinkhorn_relative_error}
\end{align}
where $\eta=\beta^2\lambda_+^{3/2}/(8\lambda_-^{7/2})$.

Finally, Gaussian conditioning and $B\preceq\lambda_+I_d$ give:
\begin{align}
    \operatorname{Cov}_p(X\mid\alpha_tX+\sigma_t\epsilon=y)
    &=(B^{-1}+\alpha_t^2\sigma_t^{-2}I_d)^{-1}\notag\\
    &\preceq\frac{\lambda_+\sigma_t^2}{\alpha_t^2\lambda_++\sigma_t^2}I_d.
    \label{eq:gaussian_posterior_curvature}
\end{align}

Thus $M_t$ in \cref{eq:ikl_posterior_curvature} satisfies:
\begin{equation*}
    \E_t\!\left[\frac{\alpha_t^4}{\sigma_t^4}M_t\right]
    \leq\E_t\!\left[
        \frac{\alpha_t^4\lambda_+}{\sigma_t^2(\alpha_t^2\lambda_++\sigma_t^2)}
    \right]=L<\infty.
\end{equation*}
Applying \cref{thm:sinkhorn_relative_contraction} with this curvature bound $L$ and the uniform relative error $\eta$ proves \cref{eq:gaussian_sinkhorn_convergence}.
\end{proof}

\vspace{-2pt}\subsection{Analysis of IKL and Sinkhorn Velocity}\vspace{-2pt}
\label{app:vel_analysis}

\cref{sec:motivation} discusses two limitations of DMD: unreliable guidance for outlier samples and mode collapse.
Outlier behavior depends on multiple factors and is difficult to characterize in general, and we present empirical observations in \cref{fig:mot_disjoint,fig:mot_sinkhorn}.

In this section, we present analysis on mode coverage behavior using the toy example of \citep{gretton2026:wgfdrifting}.
We compare IKL velocity of DMD with \Ours{} and show how adding Sinkhorn preserves a correction toward an underrepresented mode when IKL guidance vanishes.
We assume exact scores and transport plans.
The diffusion coefficients $\alpha_t,\sigma_t$ control the noise level, while $\beta>0$ independently controls Sinkhorn regularization.

\vspace{-2pt}\paragraph{Two-Mode Setup.}
Consider the following two-mode distributions~\citep{gretton2026:wgfdrifting}:
\begin{equation}
    p=a\delta_{-D}+(1-a)\delta_D,
    \qquad q=b\delta_{-D}+(1-b)\delta_D,
    \qquad D>0,\quad 0<b<a<1.
    \label{eq:toy_measures}
\end{equation}
The student has excess mass at $D$ and insufficient mass at $-D$.
We evaluate each velocity at $x=D$, where a negative velocity points toward the underrepresented mode.\vspace{-2pt}

\vspace{-2pt}\paragraph{IKL Velocity.}
For simplicity, we analyze the IKL update for a single draw of $(t,\epsilon)$.
A similar analysis applies to the expected update.
The stochastic IKL update corresponding to \cref{eq:app_velocity} is:
\begin{equation}
    V_\mathrm{IKL}(x;t,\epsilon)
    =\alpha_t\bigl[\nabla\log p_t(x_t)-\nabla\log q_t(x_t)\bigr],
    \qquad x_t=\alpha_t x+\sigma_t\epsilon.
    \label{eq:toy_ikl_sample}
\end{equation}
Using the Gaussian diffusion kernel from \cref{app:dmd_gradient}, the perturbed distributions are:
\begin{equation}
\begin{aligned}
    p_t(y)&=a\,k_t(y\mid-D)+(1-a)\,k_t(y\mid D),\\
    q_t(y)&=b\,k_t(y\mid-D)+(1-b)\,k_t(y\mid D).
\end{aligned}
\end{equation}
Differentiating the two mixtures and subtracting their scores gives:
\begin{equation}
    \nabla\log p_t(y)-\nabla\log q_t(y)
    =-\frac{2\alpha_tD(a-b)}{\sigma_t^2}
    \frac{k_t(y\mid-D)k_t(y\mid D)}{p_t(y)q_t(y)}.
    \label{eq:toy_score_difference}
\end{equation}\vspace{-2pt}

At $x=D$, we have $x_t=\alpha_tD+\sigma_t\epsilon$.
To simplify the velocity, define the ratio of the left and right Gaussian kernels at this noisy query:
\begin{equation}
    \xi_t(\epsilon)
    \coloneqq\frac{k_t(x_t\mid-D)}{k_t(x_t\mid D)}
    =\exp\!\left[-\frac{2\alpha_t^2D^2}{\sigma_t^2}
    -\frac{2\alpha_tD}{\sigma_t}\epsilon\right].
    \label{eq:toy_kernel_ratio}
\end{equation}
Substituting this ratio into the velocity in \cref{eq:toy_ikl_sample} gives:
\begin{equation}
    V_\mathrm{IKL}(D;t,\epsilon)
    =-\frac{2\alpha_t^2D(a-b)}{\sigma_t^2}
    \frac{\xi_t(\epsilon)}
    {[1-a+a\xi_t(\epsilon)][1-b+b\xi_t(\epsilon)]}.
    \label{eq:toy_ikl_exact}
\end{equation}

\vspace{-2pt}\paragraph{Sinkhorn Velocity.}
Let $m_\beta(a,b)\coloneqq\pi_{q,p}(D,-D)$ be the mass transported from $D$ to $-D$.
A transport plan has student modes as rows and target modes as columns, both in the order $(-D,D)$.
Its row sums must be $(b,1-b)$ and column sums $(a,1-a)$, so any feasible plan has the form:
\begin{equation}
    \pi(m)=
    \begin{pmatrix}
        a-m & b-a+m\\
        m & 1-b-m
    \end{pmatrix},
    \qquad a-b\leq m\leq\min(a,1-b).
    \label{eq:toy_transport_plan}
\end{equation}
For $c(x,y)=\frac12\norm{x-y}^2$, staying at the same mode costs zero and crossing between modes costs $2D^2$.
Thus, \cref{eq:eot} reduces to minimizing over the feasible interval above:
\begin{equation}
    m_\beta(a,b)=\arg\min_m
    \left\{2D^2(b-a+2m)+\beta\KL\bigl(\pi(m)\,\|\,q\otimes p\bigr)\right\}.
    \label{eq:toy_transport_objective}
\end{equation}
Differentiating the objective with respect to $m$ and setting the derivative to zero gives the following condition for the optimal mass $m=m_\beta(a,b)$:
\begin{equation}
\begin{aligned}
    0&=4D^2+\beta\log\frac{m(b-a+m)}{(a-m)(1-b-m)},\\
    m(b-a+m)&=e^{-4D^2/\beta}(a-m)(1-b-m).
\end{aligned}
    \label{eq:toy_plan_optimality}
\end{equation}
At $m=m_\beta(a,b)$, dividing the destination-weighted masses by the source mass $1-b$ gives:
\begin{equation}
\begin{aligned}
    T_{q,p}(D)
    &=\frac{(-D)m+D(1-b-m)}{1-b}
    =D-\frac{2D}{1-b}m_\beta(a,b),\\
    T_{q,q}(D)
    &=D-\frac{2D}{1-b}m_\beta(b,b).
\end{aligned}
\end{equation}
Here $m_\beta(b,b)$ is the transported mass under self-transport $q\to q$.
Subtracting the two averages gives the Sinkhorn velocity:
\begin{equation}
    V_{S_\beta}(D)
    =T_{q,p}(D)-T_{q,q}(D)
    =-\frac{2D}{1-b}\bigl[m_\beta(a,b)-m_\beta(b,b)\bigr],
    \label{eq:toy_sinkhorn_exact}
\end{equation}
which recovers the unregularized transport velocity $-2D(a-b)/(1-b)$ of \cite{gretton2026:wgfdrifting} as $\beta \to 0$.\vspace{-2pt}

\vspace{-2pt}\paragraph{\Ours{} Velocity Analysis.}
We examine two cases where a student sample at $D$ receives diminishing IKL guidance toward the underrepresented target mode at $-D$: (i) approaching the clean diffusion timestep ($t\to1^-$, $\sigma_t/\alpha_t\to0$) at fixed $D$, and (ii) increasing mode separation ($D\to\infty$) at fixed $t$.
In both cases, with active Sinkhorn term, the velocity presents as:
\begin{equation}
    V_\mathrm{sDMD}(D;t,\epsilon)
    =V_\mathrm{IKL}(D;t,\epsilon)+wV_{S_\beta}(D).
    \label{eq:toy_sdmd_velocity}
\end{equation}\vspace{-2pt}

\emph{(i) Low diffusion noise.}
First, let the diffusion timestep approach the clean endpoint, with $D>0$ and $\epsilon$ fixed.
As $\sigma_t/\alpha_t\to0$ with $\alpha_t,\sigma_t>0$, the exponential decay of $\xi_t(\epsilon)$ dominates in \cref{eq:toy_ikl_exact}, and the IKL velocity vanishes:
\begin{equation}
    V_\mathrm{IKL}(D;t,\epsilon)\longrightarrow0.
    \label{eq:toy_ikl_annealing_limit}
\end{equation}
On the other hand, for fixed $w,\beta>0$, the Sinkhorn correction is independent of diffusion noise.
Substituting this limit into \cref{eq:toy_sdmd_velocity} gives:
\begin{equation}
    V_\mathrm{sDMD}(D;t,\epsilon)
    =V_\mathrm{IKL}(D;t,\epsilon)+wV_{S_\beta}(D)
    \longrightarrow wV_{S_\beta}(D).
    \label{eq:toy_annealing_limit}
\end{equation}
Hence, the Sinkhorn term in \Ours{} does not vanish and continues to guide the sample toward the underrepresented mode at $-D$.

\emph{(ii) Large mode separation.} We can see a similar result when the mode separation becomes large.
As $D\to\infty$, $\xi_t(\epsilon)$ decays exponentially in $D^2$, causing the IKL velocity in \cref{eq:toy_ikl_exact} to vanish.
Meanwhile, \cref{eq:toy_plan_optimality} gives $m_\beta(a,b)\to a-b$ and $m_\beta(b,b)\to0$, yielding:
\begin{equation}
    V_\mathrm{IKL}(D;t,\epsilon)\longrightarrow0,
    \qquad
    \frac{V_\mathrm{sDMD}(D;t,\epsilon)}{2D}
    \longrightarrow-w\frac{a-b}{1-b}.
    \label{eq:toy_distance_limit}
\end{equation}
As mode separation grows, IKL guidance vanishes, while the Sinkhorn term in \Ours{} continues to guide the sample toward $-D$, with magnitude growing proportionally to $D$.
These results motivate augmenting IKL with Sinkhorn in \Ours{} to retain guidance when the IKL correction becomes weak.

\vspace{-2pt}\paragraph{Improving DMD with Finite Sinkhorn Batches.}
These properties make Sinkhorn alone a natural candidate for distillation.
Indeed, concurrent works demonstrate that Sinkhorn can train generators from scratch~\citep{han2026:wflow,he2026:sinkhorndrifting}.
However, our analysis assumes that $p$ and $q$ are known exactly, while training computes Sinkhorn from finite batches.
Finite batches introduce estimation errors in the entropic transport cost and the barycentric maps used to compute the velocity~\citep{genevay2019:sample,pooladian2021:entropic}.\vspace{-2pt}

Therefore, small transport batches can yield inaccurate Sinkhorn guidance.
In our ablation, Sinkhorn alone underperforms IKL (\cref{tab:ablation_energy}).
Nevertheless, combining Sinkhorn with IKL improves generation quality even with relatively small transport batches (\cref{tab:ablation_batch}).
We interpret this improvement in terms of their complementary roles: despite its own estimation error, Sinkhorn can provide more informative guidance than IKL for outlier samples, while IKL captures fine details in regions well supported by the teacher (\cref{fig:mot_sinkhorn}).

\vspace{-2pt}\section{Additional Related Works}\vspace{-2pt}

\vspace{-2pt}\paragraph{Training Few-Step Generators from Scratch.}
Recent methods train one- or few-step generators directly from data~\citep{song2023:consistency,frans2025:shortcut,zhou2025:inductive,boffi2025:flowmaps}, with often using JVPs during training~\citep{lu2025:scm,geng2026:meanflow}.
Further advances improve stability and sample quality through revised objectives and training strategies~\citep{song2024:icm,geng2026:imf,lu2026:one,zhang2026:alphaflow}, and reduce computation costs through finite-difference approximations~\citep{wang2026:TiM}.
On the other hand, recent methods use mini-batch particle interactions to align generated and target distributions during training~\citep{han2026:wflow,deng2026:drifting}. 
Our work distills few-step image and video generators from pretrained multi-step diffusion models widely used in real-world applications~\citep{wu2025qwenimage,wan2025,flux2024}.\vspace{-2pt}

\vspace{-2pt}\paragraph{Wasserstein Gradient Flow for Generative Modeling.}\vspace{-2pt}

WGF evolves probability distributions by steepest descent of an energy functional under the 2-Wasserstein metric~\citep{ambrosio2008:gradient}.
The Jordan--Kinderlehrer--Otto (JKO) scheme~\citep{jordan1998:jko} approximates this flow by minimizing the energy functional at each step, with a squared 2-Wasserstein penalty that keeps the updated distribution close to the distribution from the preceding step.
Neural implementations~\citep{mokrov2021:large,fan2021:variational} optimize a transport map at each JKO step which incur quadratic training cost in the number of JKO steps.
More efficient formulations have been proposed~\citep{choi2024:scalable,bonet2021:efficient}, but their evaluation does not extend to the scale and dimensionality of modern generative models.
In this work, we apply the WGF framework to diffusion distillation, training few-step image and video generators from large pretrained diffusion teachers.

\vspace{-2pt}\paragraph{Distribution Matching Distillation.}
The high sample quality and scalability of DMD have led to broad adoption in high-resolution image synthesis~\citep{yin2024:dmd2}, hybrid consistency training for image and video generation~\citep{zheng2026:rcm}, and autoregressive video generation~\citep{yin2025:causvid,huang2026:self-forcing,yang2025:longlive}.
Recent advances further improve its training and sampling: SenseFlow~\citep{ge2025:senseflow} introduces fake-score alignment, while Decoupled DMD~\citep{liu2026:decoupled} uses separate noise schedules for CFG augmentation and distribution matching.
Other methods improve few-step distillation by matching intermediate distributions along the denoising trajectory~\citep{luo2025:tdm,fan2026:phased}.
Additionally, recent works combine DMD-based and trajectory-based training objectives for few-step generation~\citep{cheng2026:twinflow,nie2026:tmd,gu2026:anyflow}.\vspace{-2pt}

Meanwhile, AMD~\citep{bai2026:amd} shares our concern about unreliable teacher guidance, but uses reward scores as a diagnostic proxy to adapt distillation updates.
We derive the DMD update from WGF under the IKL energy, opening a design space of energy functionals for improving distillation.
\Ours{} builds on this formulation by adding complementary Sinkhorn transport to improve sample quality and diversity.
Additionally, this transport correction may complement the above DMD training strategies which we leave to future work.

\vspace{-2pt}\section{Implementation Details}\vspace{-2pt}
\label{sec:impl_detail}

This section describes the training setup and configuration for each application.
\Cref{tab:implementation} summarizes the corresponding hyperparameters.

In all applications, we generate teacher samples using the default guidance scale and number of sampling steps for each model.
For DMD2~\citep{yin2024:dmd2} and \Ours{}, we perform one generator update for every four discriminator and fake-score updates.
We use one H200 node for ImageNet, FLUX, and Wan 1.3B, two nodes for Qwen-Image, and four nodes for Wan 14B.

\begin{table}[t]\vspace{-3pt}
    \centering
    \scriptsize
    \setlength{\tabcolsep}{3pt}
    \renewcommand{\arraystretch}{1.12}
    \begin{tabularx}{\textwidth}{@{}lYYYY@{}}
        \toprule
         & ImageNet-256 & FLUX.1 & Qwen-Image & Wan2.1 \\
         & \cref{sec:exp_image} & \cref{sec:exp_flux} & \cref{sec:exp_t2i} & \cref{sec:exp_video} \\
        \midrule
        \rowcolor{gray!10}
        \multicolumn{5}{l}{\textbf{Model architecture}} \\
        Backbone & SiT-XL/2 & FLUX.1-dev & Qwen-Image & 1.3B / 14B \\
        Resolution & $256\times256$ & $512\times512$ & $1024\times1024$ & $832\times480$ \\
        Latent resolution & $32\times32\times4$ & $64\times64\times16$ & $128\times128\times16$ & $104\times60$ \\
        Trainable parameters & Full & Full & LoRA & LoRA \\
        Student / score LoRA rank & --- & --- & 256 / 64 & 256 / 64 \\
        \midrule
        \rowcolor{gray!10}
        \multicolumn{5}{l}{\textbf{Optimization}} \\
        Batch size & 256 & 128 & 128 & 48 / 32 \\
        Training iterations & 100K & 5K & 3K & 1K \\
        Student / score learning rate & $3\times10^{-5}$ & $10^{-5}$ & $5\times10^{-5}$ & $10^{-5}$ \\
        Discriminator learning rate & $2\times10^{-6}$ & $10^{-5}$ & $10^{-5}$ & $10^{-5}$ \\
        Student / score weight decay & 0 & 0.01 & 0.01 & 0.01 \\
        Discriminator weight decay & 0.01 & 0.01 & 0.01 & 0.01 \\
        \midrule
        \rowcolor{gray!10}
        \multicolumn{5}{l}{\textbf{Distillation}} \\
        Student NFE & 1 & 4 & 2 / 4 & 4 \\
        Teacher guidance scale & 1.5 & 3.5 & 4 & 5 \\
        Adversarial loss weight & 0.003 & 0.03 & 0.03 & 0.03 \\
        \midrule
        \rowcolor{gray!10}
        \multicolumn{5}{l}{\textbf{Sinkhorn transport}} \\
        Samples per condition $M_q=M_p$ & 8 & 8 & 8 & 6 / 8 \\
        Self-transport $T_{q,q}$ & Independent batch & Diagonal mask & Diagonal mask & Diagonal mask \\
        Regularization strength $\beta$ & 0.05 & 0.05 & 1 & 1 / 0.5 \\
        Sinkhorn iterations $L$ & 40 & 10 & 100 & 50 \\
        Sinkhorn weight $w$ & 0.5 & 0.5 & 2 & 0.5 \\
        Transport representation & MAE & DINOv2 & Latent & Latent \\
        \midrule
        \rowcolor{gray!10}
        \multicolumn{5}{l}{\textbf{Gating schedule}} \\
        Gate threshold $\lambda$ & 0.50 & 0.55 & 0.55 & 0.55 \\
        Gate warmup iterations & 1,000 & 1,000 & 1,000 & 50 \\
        \bottomrule
    \end{tabularx}
    \caption{
        \textbf{Implementation Configurations.}
        Hyperparameters used in \Ours{}.
    }
    \label{tab:implementation}
\vspace{-3pt}\end{table}

\vspace{-2pt}\paragraph{ImageNet-256.}
We distill the pretrained SiT-XL/2 teacher~\citep{ma2024:sit} (675M) into a one-step, class-conditional generator at $256\times256$ resolution.
Training uses ImageNet-1K images~\citep{deng2009:imagenet} encoded into $32\times32\times4$ latents with a pretrained VAE~\citep{rombach2022:ldm}.
The data loader groups the batch on each GPU by class, and each transport problem uses 8 clean student samples and 8 clean target samples from the same class.
We compute Sinkhorn transport using features from a frozen MAE encoder, following the latent feature extraction protocol of \citet{deng2026:drifting}.
We use two-batch sampling to estimate the self-transport Sinkhorn term.

\vspace{-2pt}\paragraph{FLUX.1.}
We distill FLUX.1-dev~\citep{flux2024} (12B) into a four-step generator at $512\times512$ resolution using $64\times64\times16$ latents.
The training set contains $8{,}000$ teacher-generated images from $1{,}000$ GenEval prompts~\citep{ghosh2023:geneval}, with eight images per prompt.
Each GPU processes two prompts per batch, with $M_q=M_p=8$ samples per prompt, and transport is computed independently for each prompt.
For feature transport, we use a frozen DINOv2 ViT-B/14 backbone~\citep{oquab2023:dinov2} with inputs at $224\times224$ resolution and extract activations from blocks 3, 6, 9, and 12 to construct a multiscale representation.

\vspace{-2pt}\paragraph{Qwen-Image.}
We distill Qwen-Image~\citep{wu2025qwenimage} (20B) into two- and four-step generators at $1024\times1024$ resolution with $128\times128\times16$ latents.
To reduce memory use, we train LoRA~\citep{hu2021:lora} adapters with rank 256 for the student and rank 64 for the fake score network.
Teacher images are generated from the HPSv2 benchmark set~\citep{wu2023:hpsv2}.
Following \citet{deng2026:drifting}, we mask the diagonal in self-transport to exclude matches between a sample and itself.

\vspace{-2pt}\paragraph{Wan2.1.}
We distill Wan2.1~\citep{wan2025} (1.3B and 14B) into four-step generators producing 81-frame videos at 16 frames per second and $832\times480$ using the latents which have a spatial resolution of $104\times60$.
Training uses teacher video latents generated from the prompts in MovieGenVideoBench~\citep{polyak2024:movie}.

For Wan 14B, the memory constraints limit the per-GPU batch size.
Therefore, we adopt cross-node sample gathering to pool samples conditioned on the same prompt across GPUs for Sinkhorn transport computation.
We use the four-step sampling schedule $(0.999,0.937,0.833,0.624,0)$ provided in the FastGen framework~\citep{fastgen2026}.

\clearpage
\newpage 

\vspace{-2pt}\section{Additional Experimental Results}\vspace{-2pt}
\label{sec:additional_results}

\begin{experimentinset}{0.42\textwidth}{\centering
\renewcommand{\arraystretch}{1.15}
\setlength{\tabcolsep}{3pt}
\scriptsize
\begin{tabular}{l c c}
    \toprule
    Method & NFE & FID $\downarrow$ \\
    \midrule
    SiT\reportedmark~\tabcite{ma2024:sit} & $250\times2$ & 2.06 \\
    SiT + REPA\reportedmark~\tabcite{yu2024:repa} & $250\times2$ & 1.42 \\
    \midrule
    DMD2~\tabcite{yin2024:dmd2} & 1 & 3.14 \\
    sCM~\tabcite{lu2025:scm} & 1 & 5.11 \\
    rCM~\tabcite{zheng2026:rcm} & 1 & 3.51 \\
    MeanFlow\reportedmark~\tabcite{geng2026:meanflow} & 1 & 3.43 \\
    \rowcolor{gray!10}
    {\Oursbf{}} (Ours) & 1 & \textbf{2.83} \\
    \midrule
    REPA + $\pi$-Flow\reportedmark~\tabcite{chen2026:piflow} & 1 & 2.85 \\
    \rowcolor{gray!10}
    REPA + {\Oursbf{}} (Ours) & 1 & \textbf{2.62} \\
    \bottomrule
\end{tabular}
\normalsize
\caption{
    \textbf{ImageNet-256 One-Step Distillation.}
    The first two rows show the teacher results.
    \reportedmark{} denotes reported metrics in previous works.
}
\label{tab:imagenet}
\vspace{-3pt}
}{%
    \vspace{-2pt}\subsection{Class-Conditional Image Diffusion Distillation}\vspace{-2pt}
    \label{sec:exp_image}}
We evaluate one-step class-conditional generation on ImageNet-256, using SiT-XL/2~\citep{ma2024:sit}.
We also distill from a SiT-XL/2 variant pretrained with representation alignment (REPA)~\citep{yu2024:repa}, which accelerates teacher pretraining. The pretraining and distillation costs are summarized in \cref{tab:cost}.

The student uses the same backbone as the teacher and generates samples with a single network function evaluation (NFE).
Training uses real ImageNet-256 images encoded with the Stable Diffusion VAE~\citep{rombach2022:ldm}, with Sinkhorn transport computed between samples of the same class.
We evaluate FID~\citep{heusel2017:fid} on 50,000 generated images and compare against trajectory-based baselines~\citep{lu2025:scm,geng2026:meanflow,zheng2026:rcm} and the distribution-matching baseline~\citep{yin2024:dmd2}.

\vspace{-2pt}\paragraph{Results.}
As shown in \cref{tab:imagenet}, \Ours{} outperforms the one-step baselines, including DMD2~\citep{yin2024:dmd2}, supporting the benefit of incorporating Sinkhorn transport into distribution matching distillation.
Additionally, distilling from the SiT-XL/2+REPA~\citep{yu2024:repa} further improves generation quality.
We compare the computational cost of distillation with that of training a one-step model from scratch in \cref{tab:cost}.
\end{experimentinset}\vspace{-2pt}

\begin{experimentinset}{0.50\textwidth}{\centering
\renewcommand{\arraystretch}{1.15}
\setlength{\tabcolsep}{3pt}
\scriptsize
\begin{tabular}{l c c c c}
    \toprule
    Method & NFE & FID $\downarrow$ & CMMD $\downarrow$ & CLIP $\uparrow$ \\
    \midrule
    Teacher & $50 \times 2$ & 11.10 & 0.19 & 27.31 \\
    \midrule
    DMD2~\tabcite{yin2024:dmd2} & 4 & 16.67 & 0.22 & 27.14 \\
    sCM~\tabcite{lu2025:scm} & 4 & 14.20 & 0.25 & 26.17 \\
    \rowcolor{gray!10}
    {\Oursbf{}} (Ours) & 4 & \textbf{13.78} & \textbf{0.19} & \textbf{27.46} \\
    \bottomrule
\end{tabular}
\normalsize
\caption{
    \textbf{FLUX Distillation.}
    Four-step text-to-image generation at $512 \times 512$ resolution with FLUX.1 12B~\citep{flux2024}.
}
\label{tab:flux}
\vspace{-1pt}
}{%
    \vspace{-2pt}\subsection{FLUX}\vspace{-2pt}
    \label{sec:exp_flux}}
We distill the FLUX.1 12B~\citep{flux2024} teacher into a four-step student at $512 \times 512$ resolution.
Training uses $8,000$ teacher-generated images obtained from $1,000$ prompts in the GenEval~\citep{ghosh2023:geneval}, with eight images per prompt.
Sample quality is assessed using FID~\citep{heusel2017:fid}, CMMD~\citep{jayasumana2024:cmmd}, and CLIP score~\citep{radford2021:clip}, computed on the same generated samples, with $30,000$ images from MJHQ~\citep{li2024:mjhq} as the reference set.
For the baselines, we use the FastGen~\citep{fastgen2026} framework and compare against DMD2~\citep{yin2024:dmd2} and sCM~\citep{lu2025:scm}, as training instability prevented us from reproducing rCM~\citep{zheng2026:rcm} in our experiments.

\vspace{-2pt}\paragraph{Results.}
On FLUX, \Ours{} outperforms the few-step baselines across all metrics and achieves CMMD and CLIP scores comparable to those of the teacher (\cref{tab:flux}).
The qualitative comparison on FLUX.1 in \cref{sec:additional_qualitative} shows that \Ours{} preserves fine details while generating more diverse samples than DMD2~\citep{yin2024:dmd2}, consistent with the quantitative diversity results in \cref{sec:sample_diversity}.
\end{experimentinset}\vspace{-2pt}

\begin{experimentinset}{0.46\textwidth}{    \centering
    \captionsetup[subtable]{justification=centering,singlelinecheck=false}
    \begin{subtable}[b]{0.48\linewidth}
        \centering
        \renewcommand{\arraystretch}{1.15}
        \setlength{\tabcolsep}{5pt}
        \footnotesize
        \begin{tabular}{c c c}
            \toprule
            $M_q$ & $M_p$ & FID $\downarrow$ \\
            \midrule
            1 & 8 & 3.23 \\
            4 & 8 & 2.98 \\
            \rowcolor{gray!10}
            8 & 8 & 2.83 \\
            16 & 8 & 2.84 \\
            \bottomrule
        \end{tabular}
        \caption{Student sweep\\($M_p=8$).}
        \label{tab:ablation_batch_q}
    \end{subtable}
    \hfill
    \begin{subtable}[b]{0.48\linewidth}
        \centering
        \renewcommand{\arraystretch}{1.15}
        \setlength{\tabcolsep}{5pt}
        \footnotesize
        \begin{tabular}{c c c}
            \toprule
            $M_q$ & $M_p$ & FID $\downarrow$ \\
            \midrule
            8 & 1 & 3.97 \\
            8 & 4 & 3.02 \\
            \rowcolor{gray!10}
            8 & 8 & 2.83 \\
            8 & 16 & 2.85 \\
            \bottomrule
        \end{tabular}
        \caption{Teacher sweep\\($M_q=8$).}
        \label{tab:ablation_batch_p}
    \end{subtable}
    \caption{
        \textbf{Ablation Study of Sinkhorn Batch Size.}
        \Ours{} achieves robust performance with relatively small batch sizes ($M_q, M_p \geq 8$).
    }
    \label{tab:ablation_batch}
\vspace{-1pt}
}{%
    \vspace{-2pt}\subsection{Sinkhorn Batch Size}\vspace{-2pt}}
As discussed in \cref{subsec:complementarity}, while Sinkhorn transport provides global guidance toward the teacher distribution, in practice it is computed using a finite number of samples per condition, which may introduce errors into the transport velocity.
To analyze the effect of this finite-sample approximation, we show ablation results on ImageNet-256~\citep{deng2009:imagenet} in \cref{tab:ablation_batch_q} and \cref{tab:ablation_batch_p}, which sweep the student sample count $M_q$ with $M_p=8$ and the teacher sample count $M_p$ with $M_q=8$, respectively.
In both cases, using too few samples degrades generation quality, with a more pronounced degradation when fewer teacher samples ($M_p$) are used.
However, \Ours{} recovers robust performances even with relatively low number of samples from each distribution ($M_q \geq 8, M_p \geq 8$).
These results suggest that Sinkhorn velocities estimated from relatively few samples can still provide more informative guidance than IKL velocities for out-of-distribution samples, thereby improving generation quality.
\end{experimentinset}

\vspace{-2pt}\subsection{Training Cost}\vspace{-2pt}

\begin{table}[h]\vspace{-3pt}
    \centering
    \begin{threeparttable}
    \renewcommand{\arraystretch}{1.15}
    \setlength{\tabcolsep}{8pt}
    \scriptsize
        \begin{tabular*}{\linewidth}{@{\extracolsep{\fill}}l c c c c c}
        \toprule
        Method & $\#$Params & Iter. & NFE & GPU Hours & FID $\downarrow$ \\
        \midrule
        SiT~\citep{ma2024:sit} & 675M & 7M & $250\times2$ & 2,290 & 2.06 \\
        SiT + REPA~\citep{yu2024:repa} & 675M / 87M & 4M & $250\times2$ & 1,220 & 1.42\\
        \midrule
        W-Flow\reportedmark~\citep{han2026:wflow} & 675M / 49M & 200K & 1 & 4,639 & 1.29 \\
        DMD2~\citep{yin2024:dmd2} & 675M & 100K & 1 & 114 & 3.14 \\
        \midrule
        {\Oursbf{}} (Diag. Mask) & 675M / 49M & 100K & 1 & 118 & 2.91 \\
        {\Oursbf{}} (Two Batch) & 675M / 49M & 100K & 1 & 145 & 2.83 \\
        REPA + {\Oursbf{}} (Two Batch) & 675M / 49M & 100K & 1 & 145 & 2.62 \\
        \bottomrule
    \end{tabular*}
    \normalsize
    \caption{
        \textbf{Training and Sampling Cost.}
        Comparison of teacher pretraining~\citep{ma2024:sit,yu2024:repa}, training a one-step generator from scratch~\citep{han2026:wflow}, and distillation.
        GPU hours for distillation exclude teacher pretraining, whose cost is listed separately in the first two rows.
        \reportedmark{} denotes reported metrics in previous works.
    }
    \label{tab:cost}
    \end{threeparttable}
\vspace{-3pt}\end{table}

\Cref{tab:cost} compares the training and sampling costs of \Ours{} with DMD2~\citep{yin2024:dmd2} and training a one-step generator from scratch~\citep{han2026:wflow}.
All experiments in our cost analysis are conducted on NVIDIA H200 GPUs.
For \Ours{}, we consider two choices of computing the self-transport map $T_{q,q}$ in \cref{eq:sinkhorn_velocity}: diagonal masking~\citep{deng2026:drifting} and two-batch~\citep{han2026:wflow}.
Diagonal masking reuses a single student batch while preventing each sample from matching itself, whereas two-batch estimation computes transport between two independent student batches.
With diagonal masking, \Ours{} improves generation quality over DMD2 while increasing training cost by only around 3.5\%.
Using two batches further improves FID at additional training cost, consistent with the observations of \citet{han2026:wflow}.

\vspace{-2pt}\paragraph{Distillation and Training from Scratch.}
Recent methods achieve high generation quality by training one-step generators from scratch~\citep{han2026:wflow,deng2026:drifting,geng2026:meanflow}.
In our analysis, we observe that teacher pretraining followed by distillation still requires fewer GPU hours than training a model from scratch~\citep{han2026:wflow} (\cref{tab:cost}).
Additionally, advances in representation learning further accelerate convergence and improve generation quality of pretrained models~\citep{yu2024:repa,zheng2026:rae}, supporting for pretraining followed by distillation as an efficient practice.\vspace{-2pt}

\clearpage
\newpage 

\vspace{-2pt}\section{Limitations and Future Work}\vspace{-2pt}
\label{sec:limitations}

For conditional generation, \Ours{} requires multiple teacher samples for each conditioning input to estimate Sinkhorn guidance.
While our ablations show that relatively few samples suffice to provide effective guidance for outlier samples, generating these samples still adds computational costs.
Additionally, although our gated Sinkhorn scheduler improves performance, it uses discriminator outputs as a proxy for identifying samples with relatively weak teacher support.
Exploring more reliable criteria for identifying samples that benefit from Sinkhorn guidance remains an interesting direction for future work.
Lastly, we compute transport in the clean latent space for video diffusion models to reduce computational overhead.
Exploring video feature spaces that provide a favorable balance between sample quality and computational cost remains future work.

\FloatBarrier
\section{Additional Qualitative Results}\vspace{-2pt}
\label{sec:additional_qualitative}

Extending the comparisons in \cref{fig:qual_image,fig:qual_video}, we present additional text-to-image results with FLUX.1~\citep{flux2024} and Qwen-Image 20B~\citep{wu2025qwenimage}, along with text-to-video results with Wan2.1-14B~\citep{wan2025}.
For each prompt, we show samples generated with different random seeds to illustrate the fidelity and diversity of \Ours{}.

\clearpage
\providecommand{\QualRowLabel}[1]{\adjustbox{valign=m}{\rotatebox[origin=c]{90}{\scriptsize #1}}}
\begin{figure}[p]\vspace{-3pt}
    \centering
    \captionsetup{font=normalsize}
    \caption{
        \textbf{Additional Qualitative Comparison on FLUX.1.}
        \Ours{} more closely matches the diversity of the teacher than DMD2 while preserving visual quality.
    }
    \label{fig:flux_qual_image}
    \setlength{\tabcolsep}{1.5pt}
    \renewcommand{\arraystretch}{1.05}
    \scriptsize
    \begin{adjustbox}{max totalsize={\textwidth}{0.92\textheight}}
    \begin{tabular}{c cccc}
        \QualRowLabel{Teacher} &
        \adjustbox{valign=m}{\includegraphics[width=0.23\textwidth]{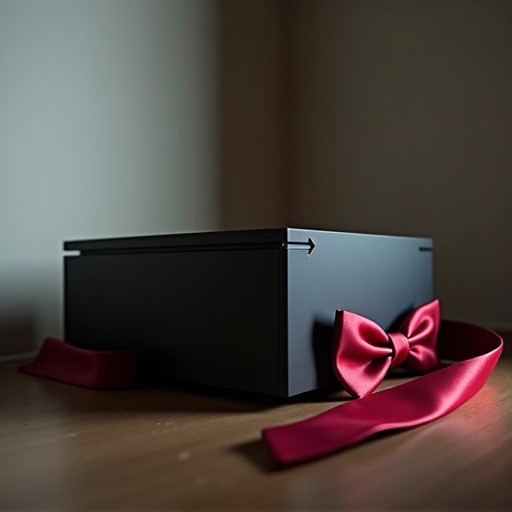}} & \adjustbox{valign=m}{\includegraphics[width=0.23\textwidth]{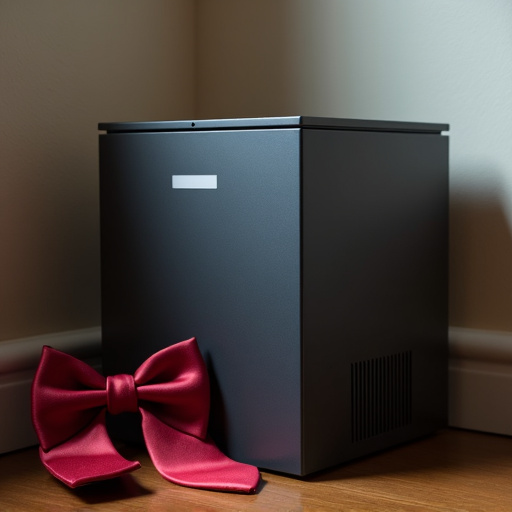}} & \adjustbox{valign=m}{\includegraphics[width=0.23\textwidth]{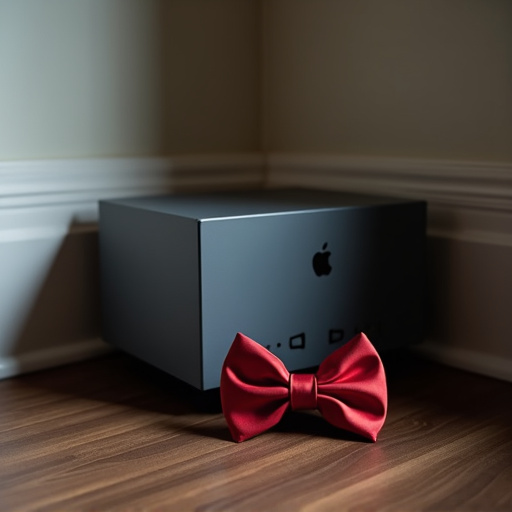}} & \adjustbox{valign=m}{\includegraphics[width=0.23\textwidth]{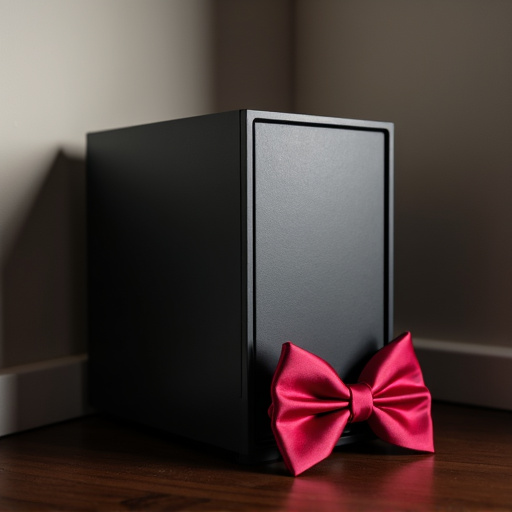}} \\
        \addlinespace[2pt]
        \QualRowLabel{\makecell{DMD2 \\ {\tiny\citep{yin2024:dmd2}}}} &
        \adjustbox{valign=m}{\includegraphics[width=0.23\textwidth]{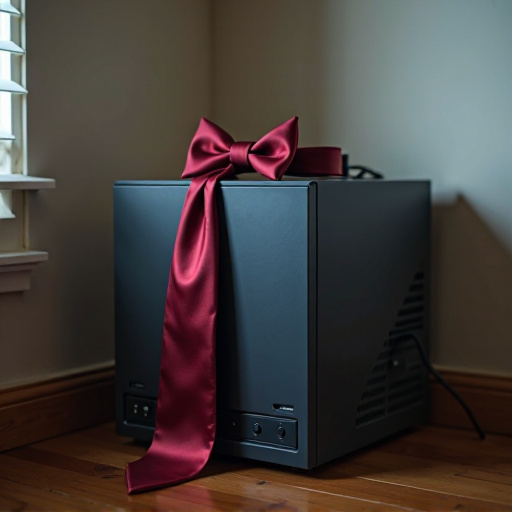}} & \adjustbox{valign=m}{\includegraphics[width=0.23\textwidth]{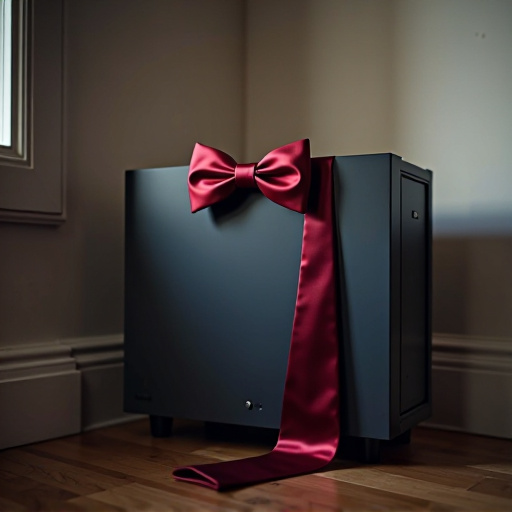}} & \adjustbox{valign=m}{\includegraphics[width=0.23\textwidth]{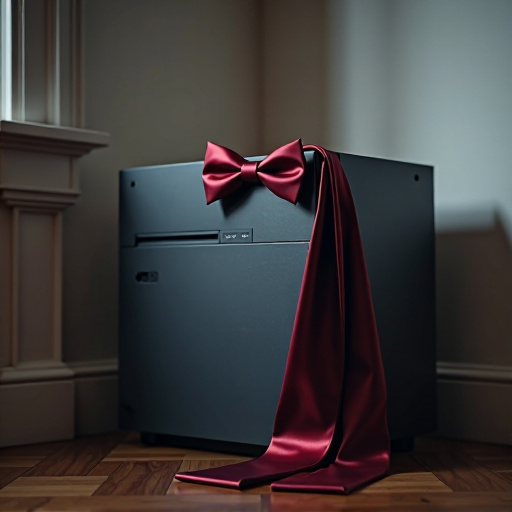}} & \adjustbox{valign=m}{\includegraphics[width=0.23\textwidth]{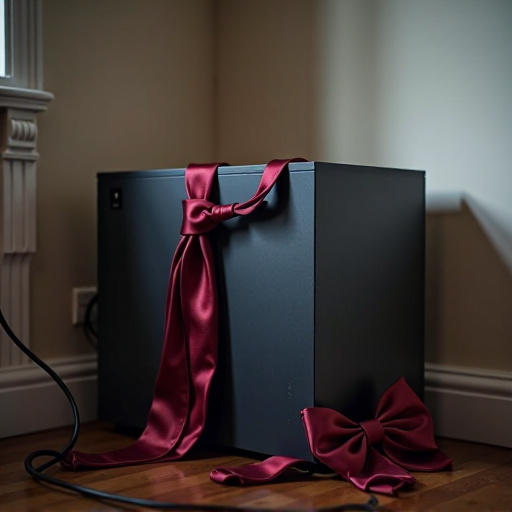}} \\
        \addlinespace[2pt]
        \QualRowLabel{\makecell{{\Oursbf{}} \\ {(Ours)}}} &
        \adjustbox{valign=m}{\includegraphics[width=0.23\textwidth]{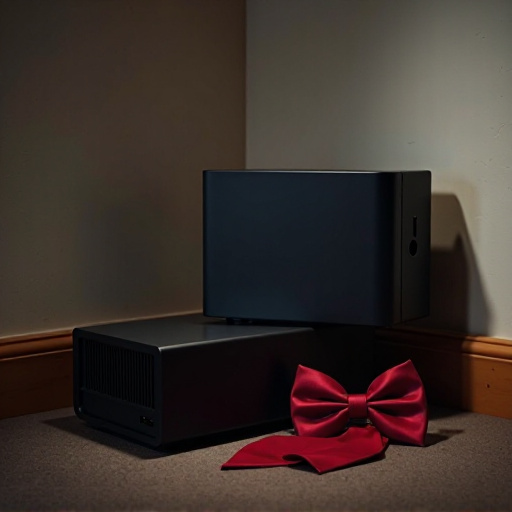}} & \adjustbox{valign=m}{\includegraphics[width=0.23\textwidth]{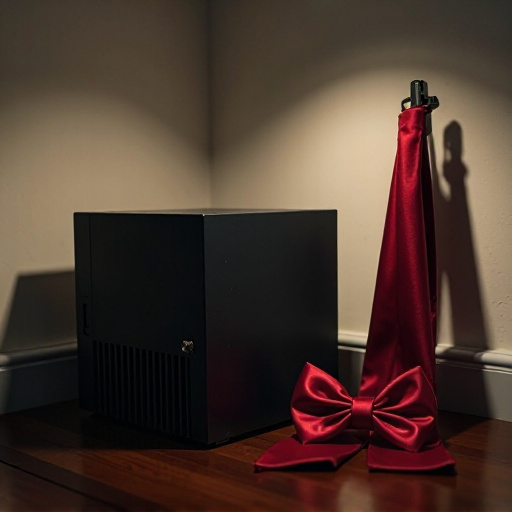}} & \adjustbox{valign=m}{\includegraphics[width=0.23\textwidth]{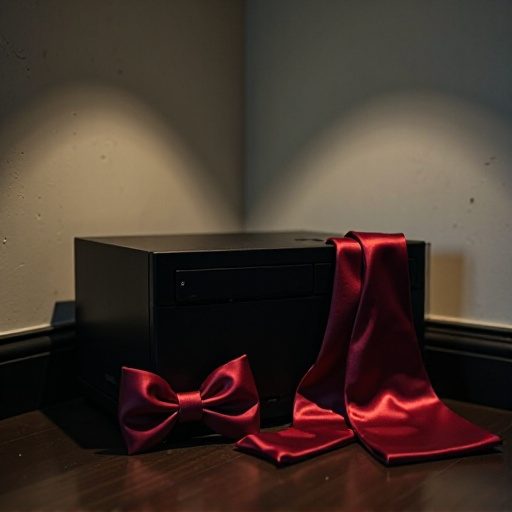}} & \adjustbox{valign=m}{\includegraphics[width=0.23\textwidth]{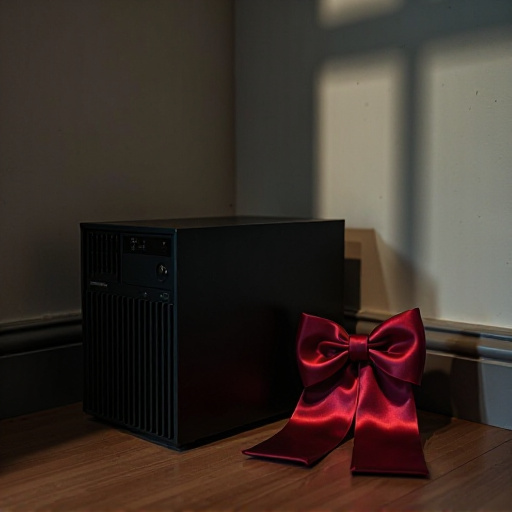}} \\
        \addlinespace[4pt]
        \multicolumn{5}{c}{\parbox{0.94\textwidth}{\centering\tiny\itshape ``In the corner of a dimly lit room, a rectangular, charcoal-gray computer box exhibits its sharp edges as it rests beside an elegantly draped, cherry-red silk bow tie. The contrast between the electronic's matte finish and the tie's glossy sheen is striking. The computer appears dormant, while the bow tie seems almost ready to be picked up and worn to an event. Behind these objects, the ambient lighting casts soft shadows against the plain, light-colored wall.''}} \\[9pt]
        \addlinespace[8pt]
        \QualRowLabel{Teacher} &
        \adjustbox{valign=m}{\includegraphics[width=0.23\textwidth]{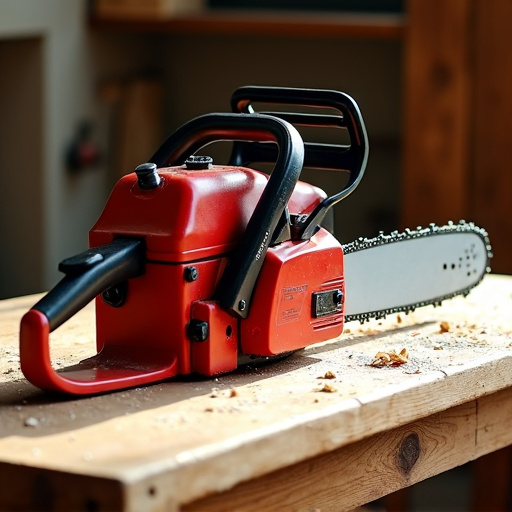}} & \adjustbox{valign=m}{\includegraphics[width=0.23\textwidth]{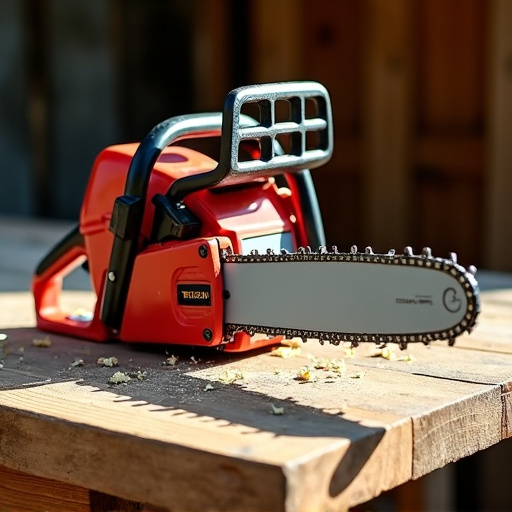}} & \adjustbox{valign=m}{\includegraphics[width=0.23\textwidth]{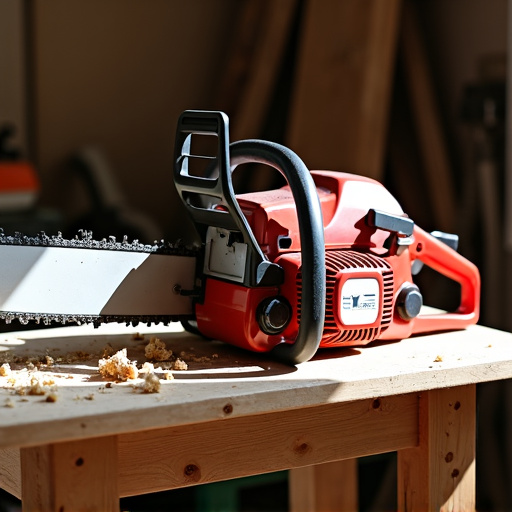}} & \adjustbox{valign=m}{\includegraphics[width=0.23\textwidth]{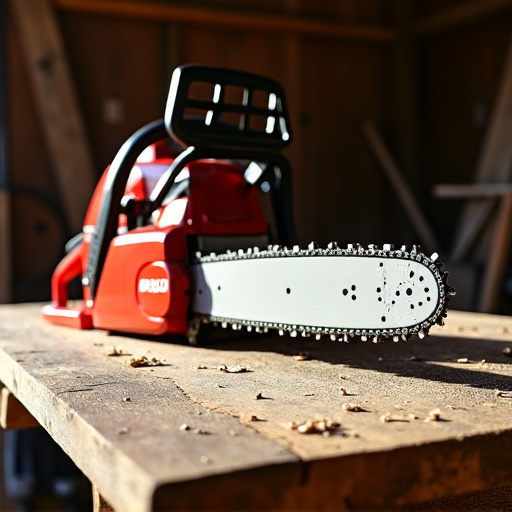}} \\
        \addlinespace[2pt]
        \QualRowLabel{\makecell{DMD2 \\ {\tiny\citep{yin2024:dmd2}}}} &
        \adjustbox{valign=m}{\includegraphics[width=0.23\textwidth]{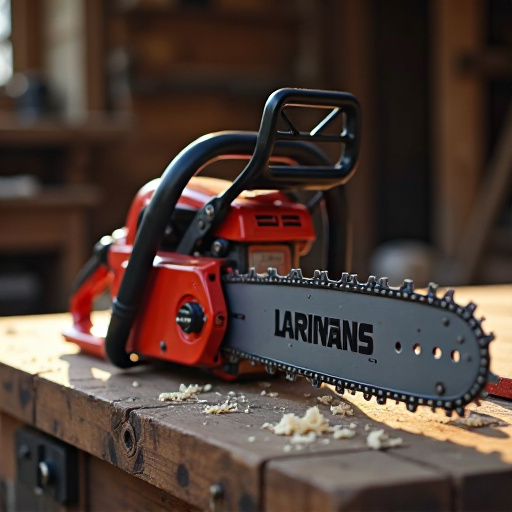}} & \adjustbox{valign=m}{\includegraphics[width=0.23\textwidth]{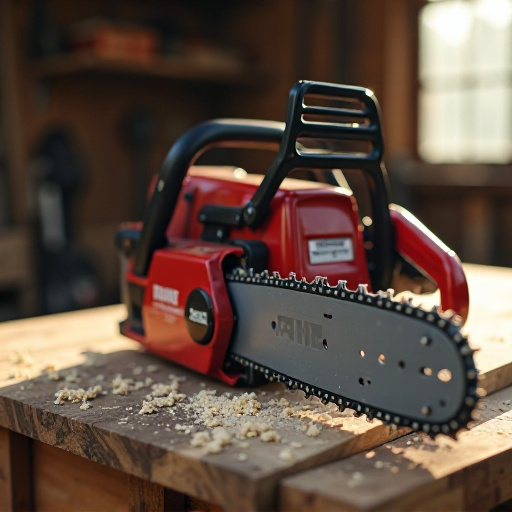}} & \adjustbox{valign=m}{\includegraphics[width=0.23\textwidth]{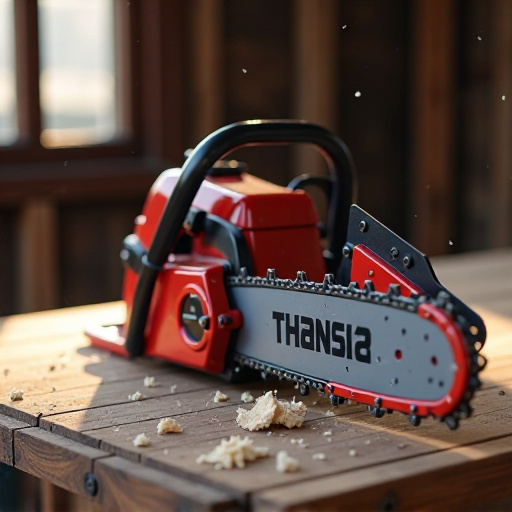}} & \adjustbox{valign=m}{\includegraphics[width=0.23\textwidth]{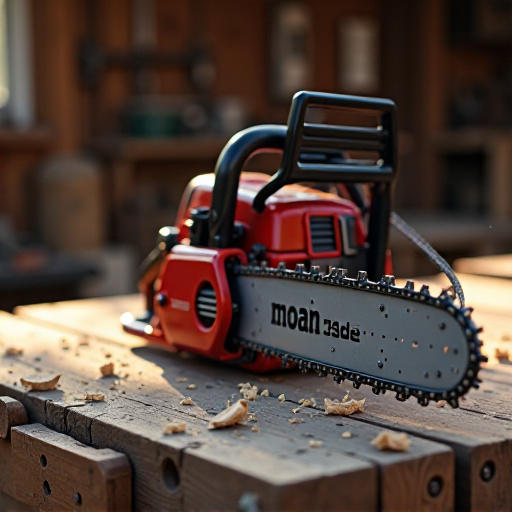}} \\
        \addlinespace[2pt]
        \QualRowLabel{\makecell{{\Oursbf{}} \\ {(Ours)}}} &
        \adjustbox{valign=m}{\includegraphics[width=0.23\textwidth]{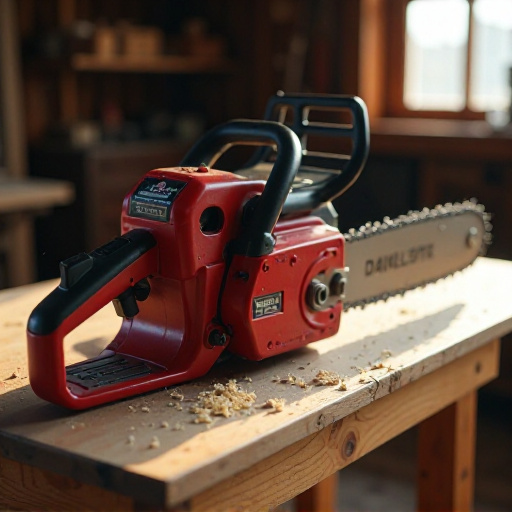}} & \adjustbox{valign=m}{\includegraphics[width=0.23\textwidth]{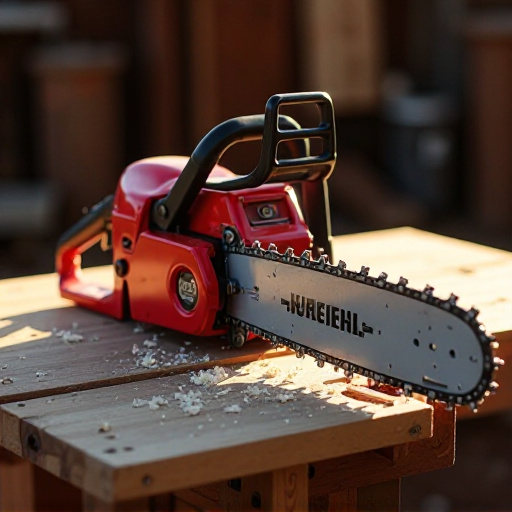}} & \adjustbox{valign=m}{\includegraphics[width=0.23\textwidth]{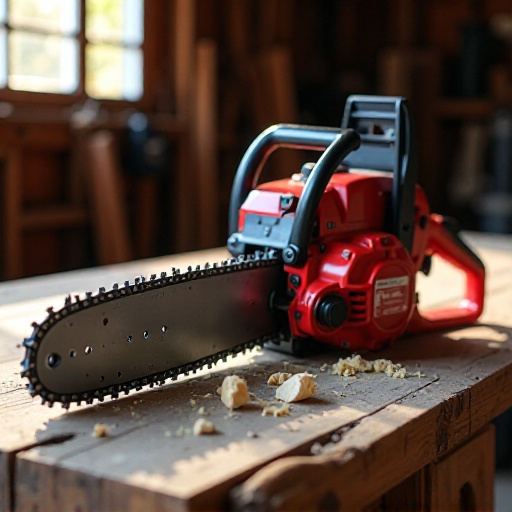}} & \adjustbox{valign=m}{\includegraphics[width=0.23\textwidth]{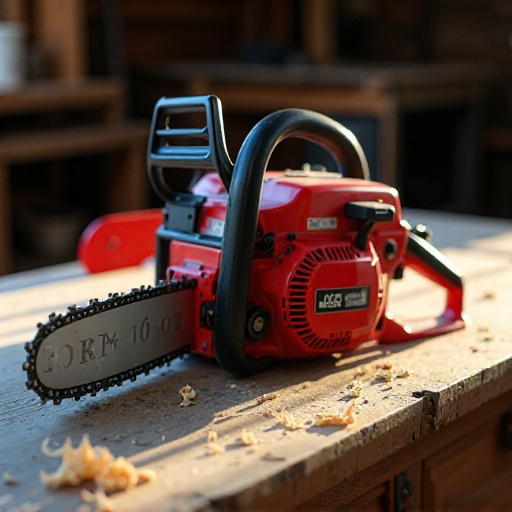}} \\
        \addlinespace[4pt]
        \multicolumn{5}{c}{\parbox{0.94\textwidth}{\centering\tiny\itshape ``A hefty, red-painted chainsaw rests prominently on the surface of a sturdy, solid oak table. The metal components of the chainsaw catch the soft glow of the early morning sunlight, causing a shimmering effect. The wood's rich, deep grain stands in contrast to the boldness of the chainsaw, and there are wood shavings scattered nearby, hinting at recent activity.''}} \\[9pt]
    \end{tabular}
    \end{adjustbox}
\vspace{-3pt}\end{figure}

\begin{figure}[p]\vspace{-3pt}
    \centering
    \captionsetup{font=normalsize}
    \caption{
        \textbf{Additional Qualitative Comparison on FLUX.1 (Continued).}
        \Ours{} more closely matches the diversity of the teacher than DMD2 while preserving visual quality.
    }
    \label{fig:flux_qual_image_2}
    \setlength{\tabcolsep}{1.5pt}
    \renewcommand{\arraystretch}{1.05}
    \scriptsize
    \begin{adjustbox}{max totalsize={\textwidth}{0.92\textheight}}
    \begin{tabular}{c cccc}
        \QualRowLabel{Teacher} &
        \adjustbox{valign=m}{\includegraphics[width=0.23\textwidth]{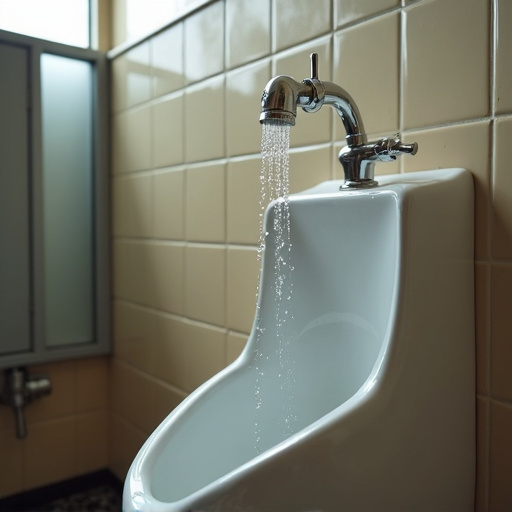}} & \adjustbox{valign=m}{\includegraphics[width=0.23\textwidth]{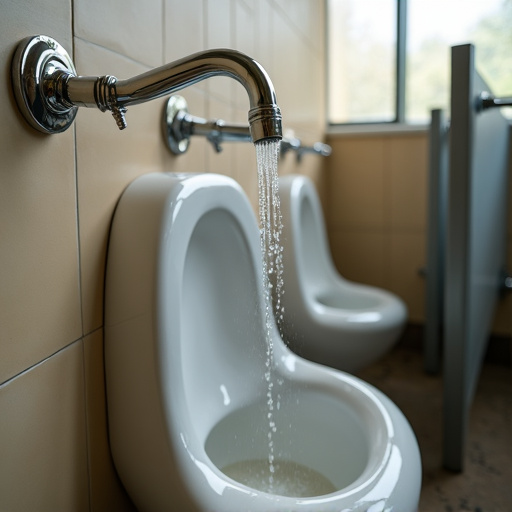}} & \adjustbox{valign=m}{\includegraphics[width=0.23\textwidth]{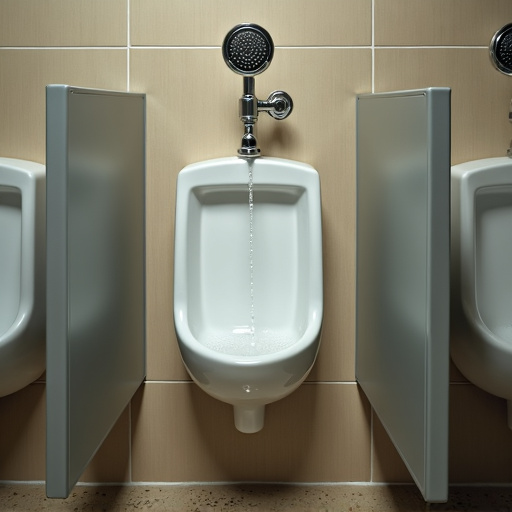}} & \adjustbox{valign=m}{\includegraphics[width=0.23\textwidth]{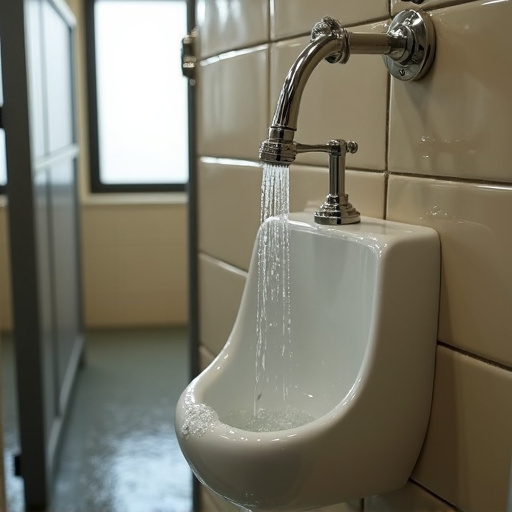}} \\
        \addlinespace[2pt]
        \QualRowLabel{\makecell{DMD2 \\ {\tiny\citep{yin2024:dmd2}}}} &
        \adjustbox{valign=m}{\includegraphics[width=0.23\textwidth]{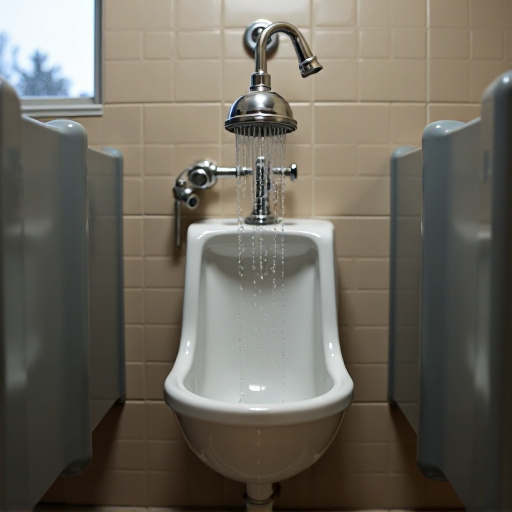}} & \adjustbox{valign=m}{\includegraphics[width=0.23\textwidth]{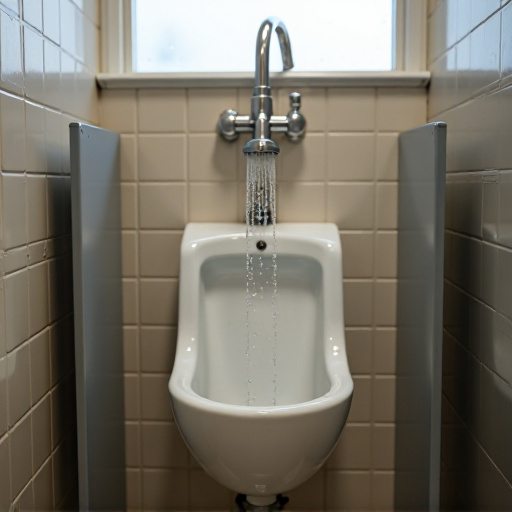}} & \adjustbox{valign=m}{\includegraphics[width=0.23\textwidth]{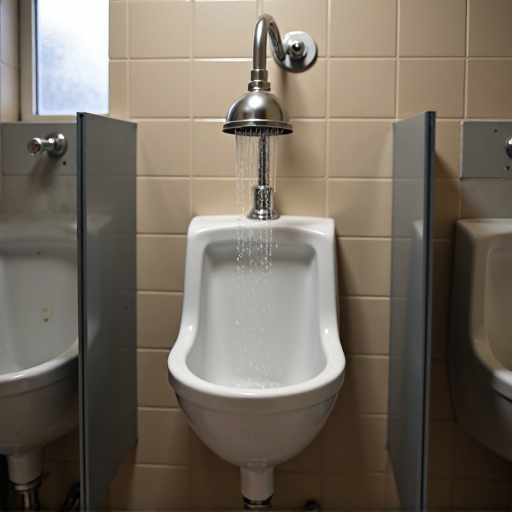}} & \adjustbox{valign=m}{\includegraphics[width=0.23\textwidth]{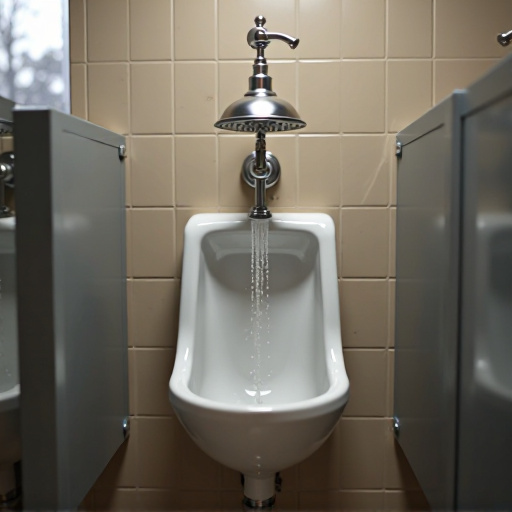}} \\
        \addlinespace[2pt]
        \QualRowLabel{\makecell{{\Oursbf{}} \\ {(Ours)}}} &
        \adjustbox{valign=m}{\includegraphics[width=0.23\textwidth]{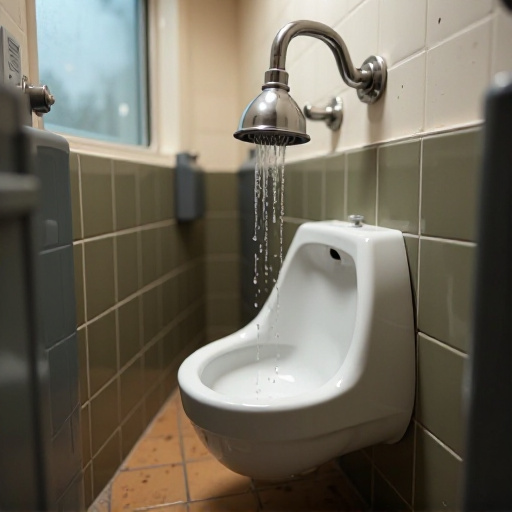}} & \adjustbox{valign=m}{\includegraphics[width=0.23\textwidth]{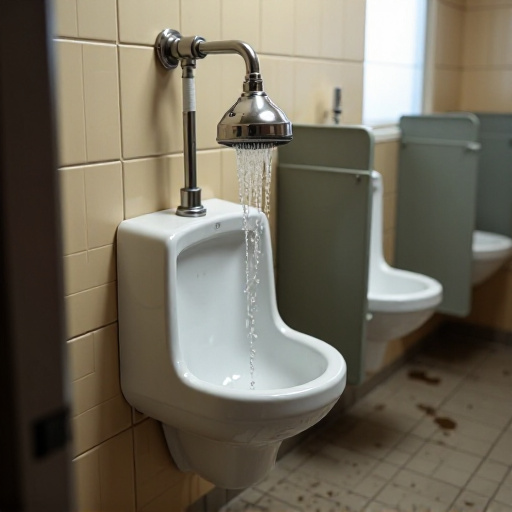}} & \adjustbox{valign=m}{\includegraphics[width=0.23\textwidth]{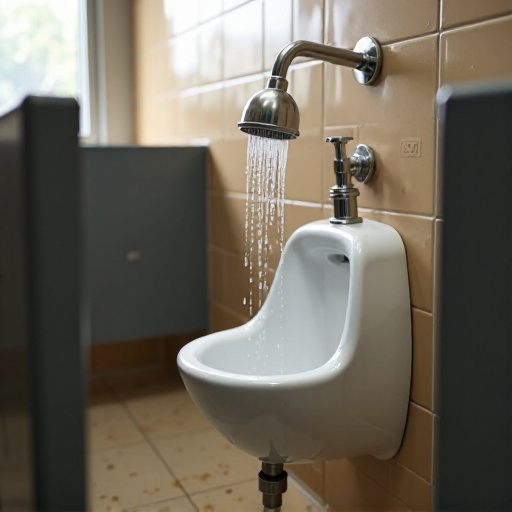}} & \adjustbox{valign=m}{\includegraphics[width=0.23\textwidth]{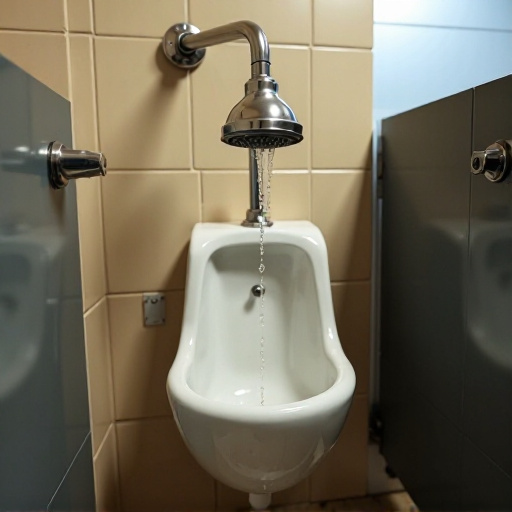}} \\
        \addlinespace[4pt]
        \multicolumn{5}{c}{\parbox{0.94\textwidth}{\centering\tiny\itshape ``A gleaming stainless steel showerhead mounted on a beige tiled wall, with beads of water steadily dripping from its nozzle onto the glossy white porcelain urinal below. The urinal is flanked by grey dividers, ensuring privacy, and the floor beneath is speckled with small, wet splashes reflecting the dim early morning light filtering through a frosted window just out of view. The metallic fixtures of the showerhead are complemented by the chrome handles and plumbing visible under the urinal, creating a cohesive and functional restroom vignette.''}} \\[9pt]
        \addlinespace[8pt]
        \QualRowLabel{Teacher} &
        \adjustbox{valign=m}{\includegraphics[width=0.23\textwidth]{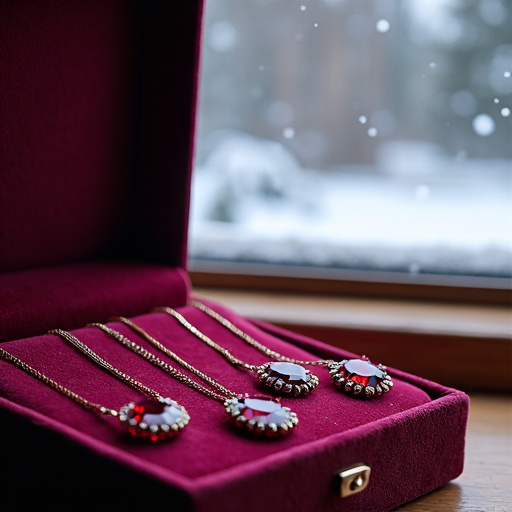}} & \adjustbox{valign=m}{\includegraphics[width=0.23\textwidth]{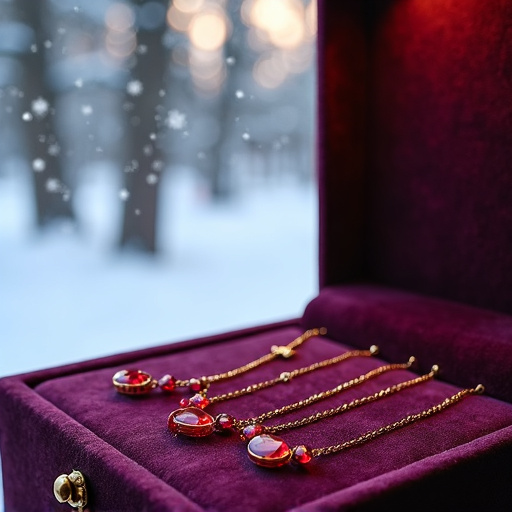}} & \adjustbox{valign=m}{\includegraphics[width=0.23\textwidth]{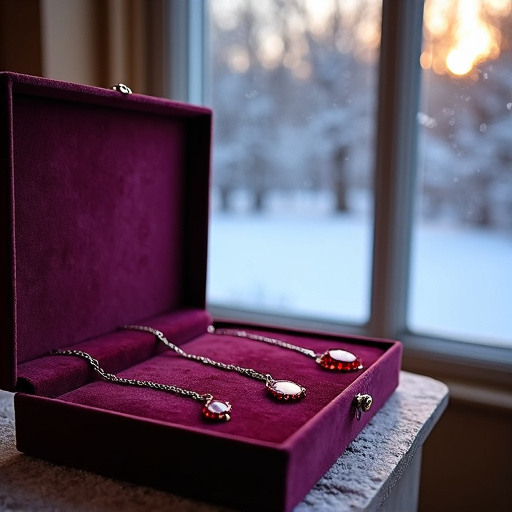}} & \adjustbox{valign=m}{\includegraphics[width=0.23\textwidth]{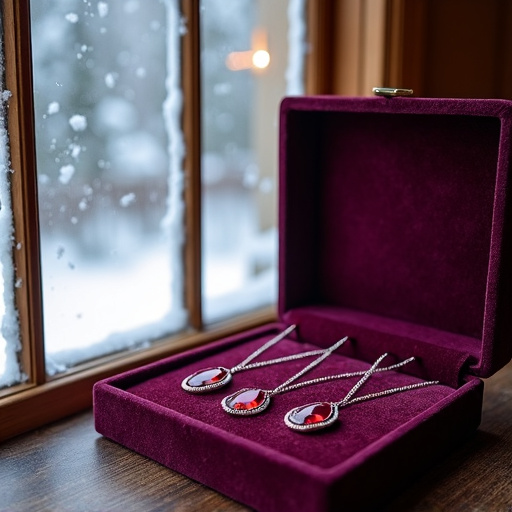}} \\
        \addlinespace[2pt]
        \QualRowLabel{\makecell{DMD2 \\ {\tiny\citep{yin2024:dmd2}}}} &
        \adjustbox{valign=m}{\includegraphics[width=0.23\textwidth]{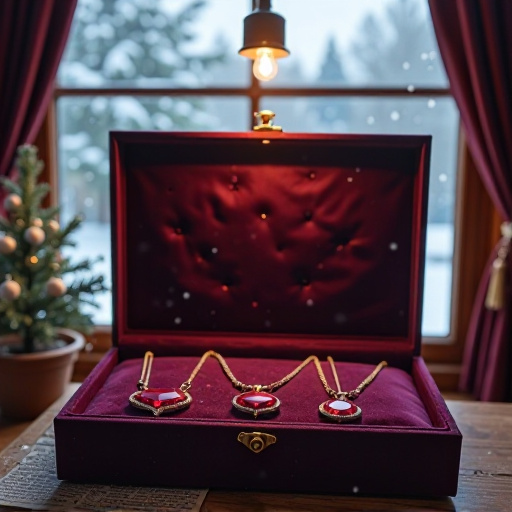}} & \adjustbox{valign=m}{\includegraphics[width=0.23\textwidth]{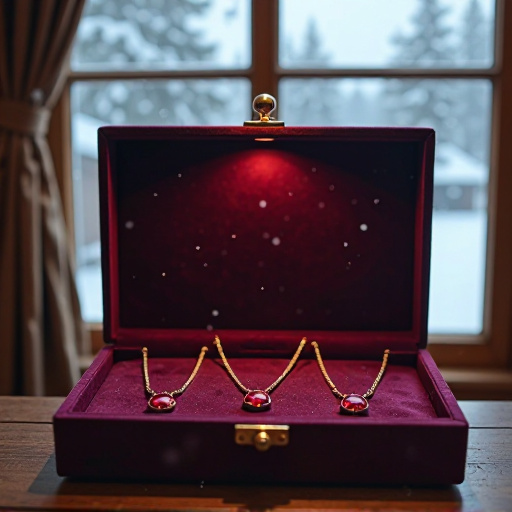}} & \adjustbox{valign=m}{\includegraphics[width=0.23\textwidth]{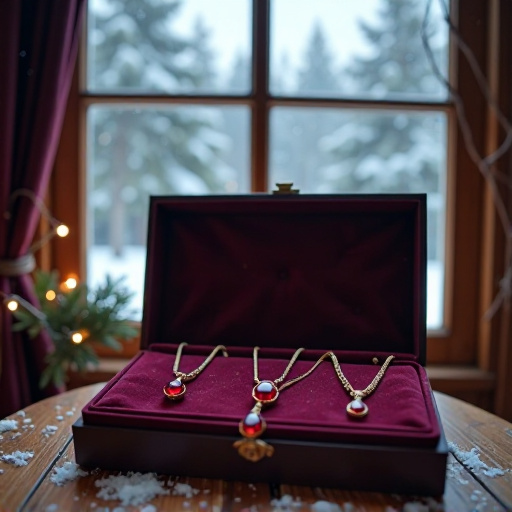}} & \adjustbox{valign=m}{\includegraphics[width=0.23\textwidth]{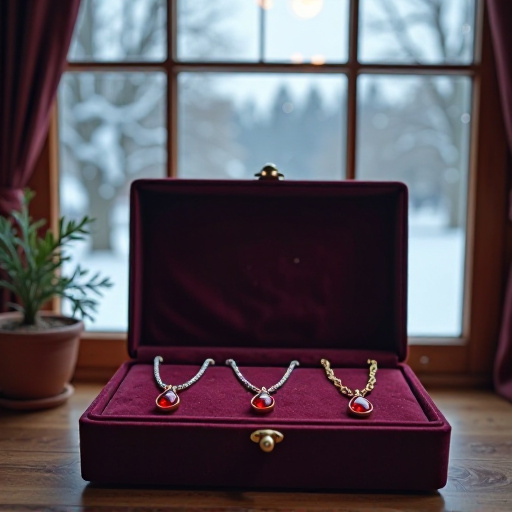}} \\
        \addlinespace[2pt]
        \QualRowLabel{\makecell{{\Oursbf{}} \\ {(Ours)}}} &
        \adjustbox{valign=m}{\includegraphics[width=0.23\textwidth]{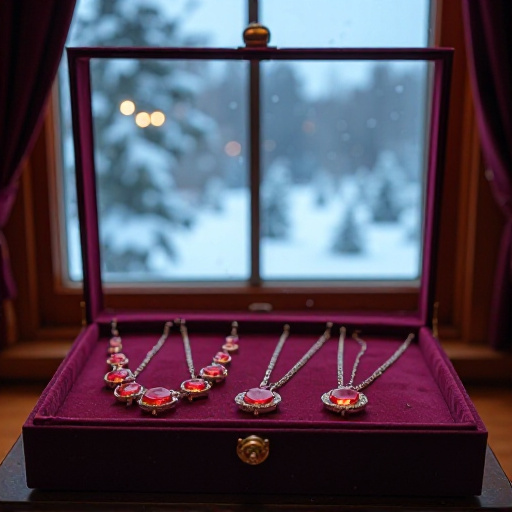}} & \adjustbox{valign=m}{\includegraphics[width=0.23\textwidth]{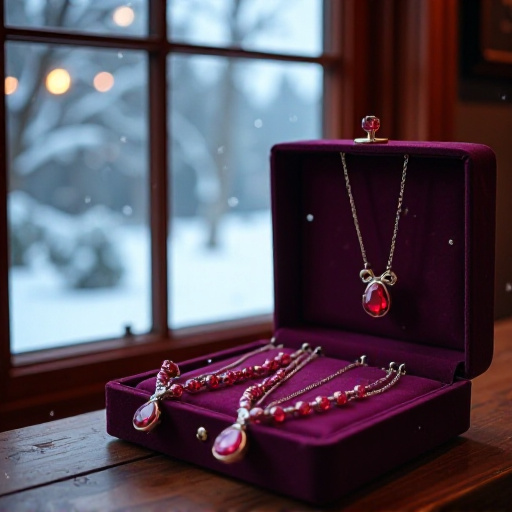}} & \adjustbox{valign=m}{\includegraphics[width=0.23\textwidth]{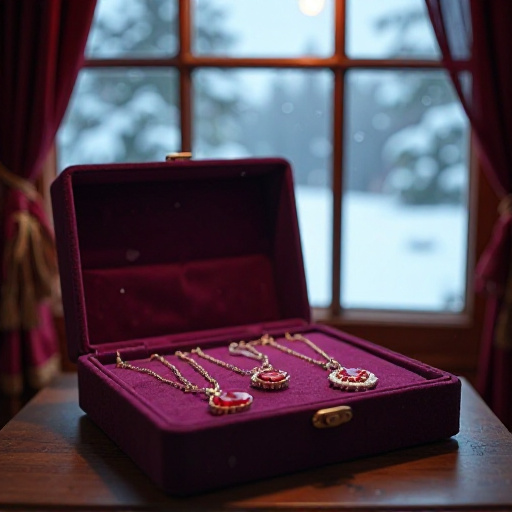}} & \adjustbox{valign=m}{\includegraphics[width=0.23\textwidth]{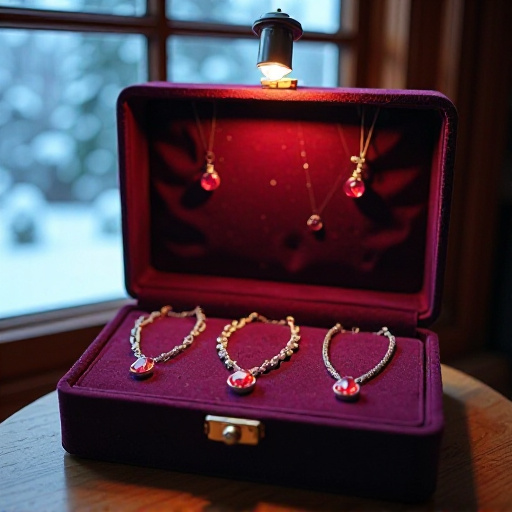}} \\
        \addlinespace[4pt]
        \multicolumn{5}{c}{\parbox{0.94\textwidth}{\centering\tiny\itshape ``Inside a warm room with a large window showcasing a picturesque winter landscape, three gleaming ruby red necklaces are elegantly laid out on the plush surface of a deep purple velvet jewelry box. The gentle glow from the overhead light accentuates the rich color and intricate design of the necklaces. Just beyond the glass pane, snowflakes can be seen gently falling to coat the ground outside in a blanket of white.''}} \\[9pt]
    \end{tabular}
    \end{adjustbox}
\vspace{-3pt}\end{figure}

\begin{figure}[p]\vspace{-3pt}
    \centering
    \captionsetup{font=normalsize}
    \caption{
        \textbf{Additional Qualitative Comparison on FLUX.1 (Continued).}
        \Ours{} more closely matches the diversity of the teacher than DMD2 while preserving visual quality.
    }
    \label{fig:flux_qual_image_3}
    \setlength{\tabcolsep}{1.5pt}
    \renewcommand{\arraystretch}{1.05}
    \scriptsize
    \begin{adjustbox}{max totalsize={\textwidth}{0.92\textheight}}
    \begin{tabular}{c cccc}
        \QualRowLabel{Teacher} &
        \adjustbox{valign=m}{\includegraphics[width=0.23\textwidth]{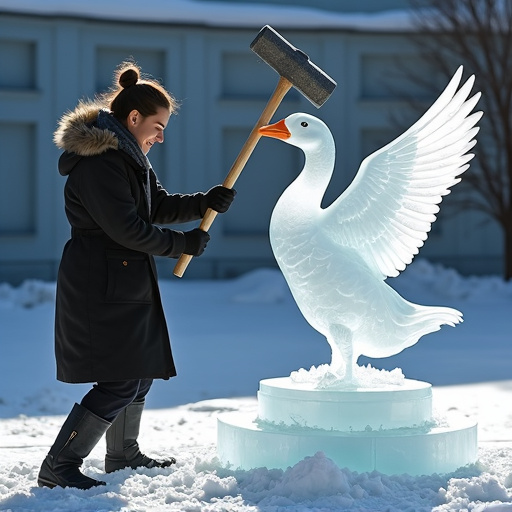}} & \adjustbox{valign=m}{\includegraphics[width=0.23\textwidth]{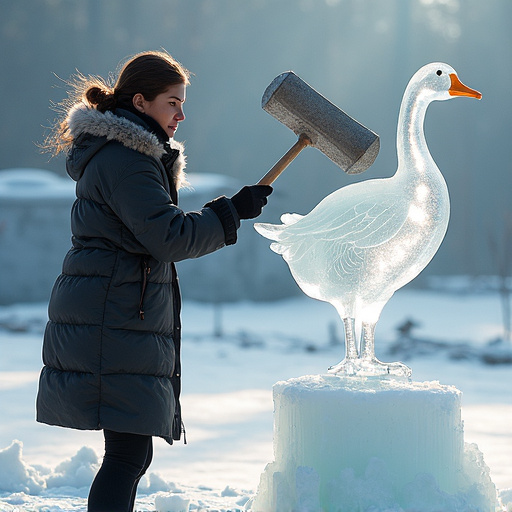}} & \adjustbox{valign=m}{\includegraphics[width=0.23\textwidth]{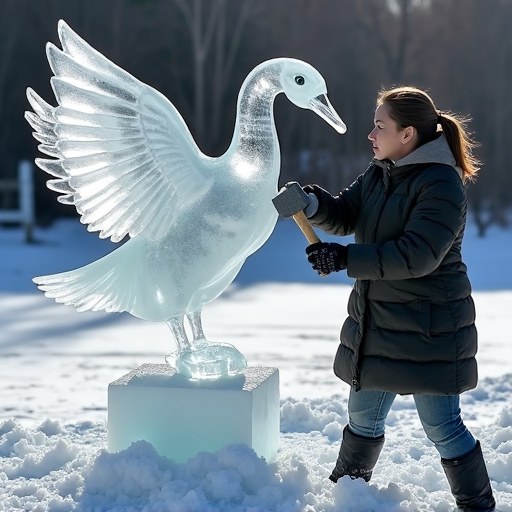}} & \adjustbox{valign=m}{\includegraphics[width=0.23\textwidth]{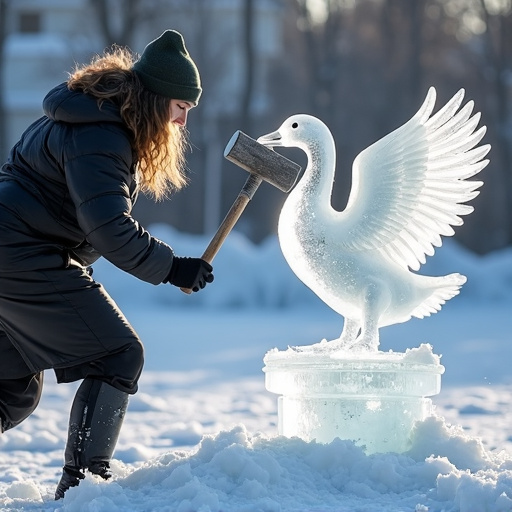}} \\
        \addlinespace[2pt]
        \QualRowLabel{\makecell{DMD2 \\ {\tiny\citep{yin2024:dmd2}}}} &
        \adjustbox{valign=m}{\includegraphics[width=0.23\textwidth]{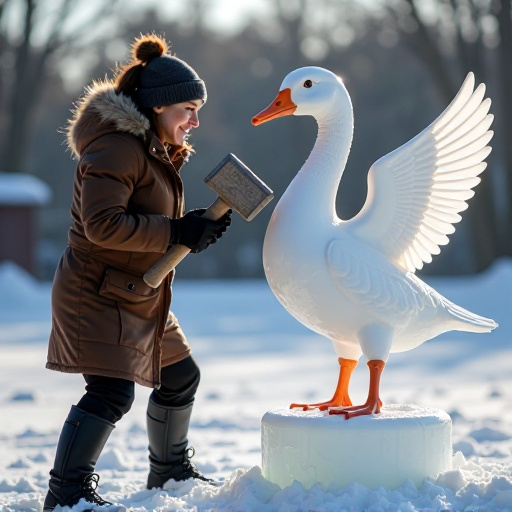}} & \adjustbox{valign=m}{\includegraphics[width=0.23\textwidth]{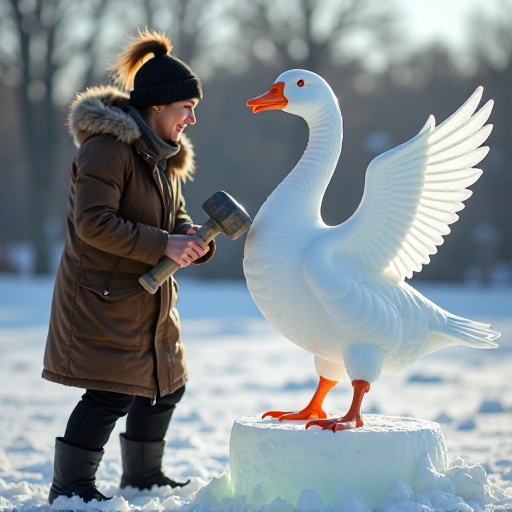}} & \adjustbox{valign=m}{\includegraphics[width=0.23\textwidth]{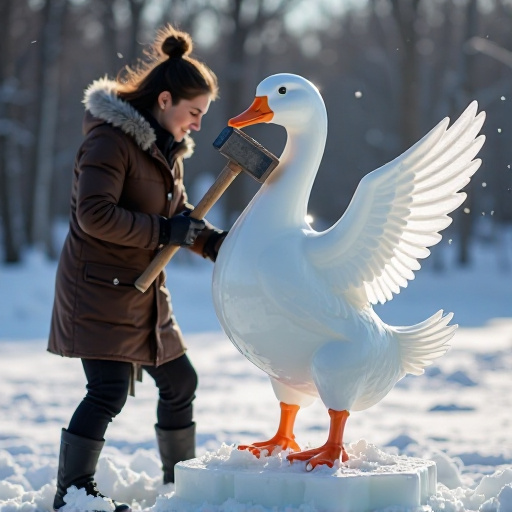}} & \adjustbox{valign=m}{\includegraphics[width=0.23\textwidth]{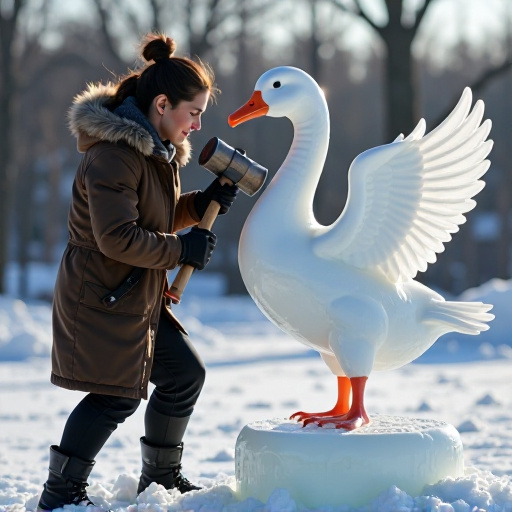}} \\
        \addlinespace[2pt]
        \QualRowLabel{\makecell{{\Oursbf{}} \\ {(Ours)}}} &
        \adjustbox{valign=m}{\includegraphics[width=0.23\textwidth]{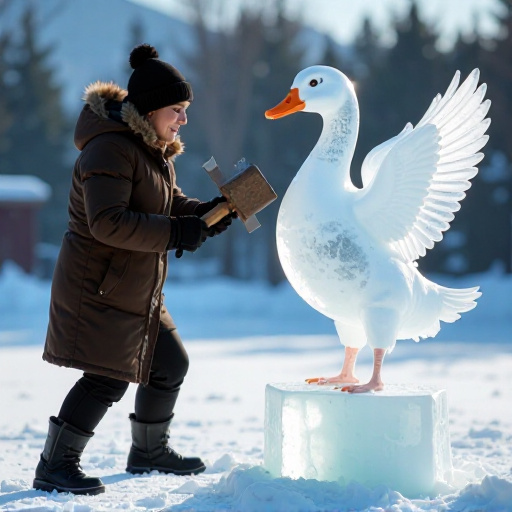}} & \adjustbox{valign=m}{\includegraphics[width=0.23\textwidth]{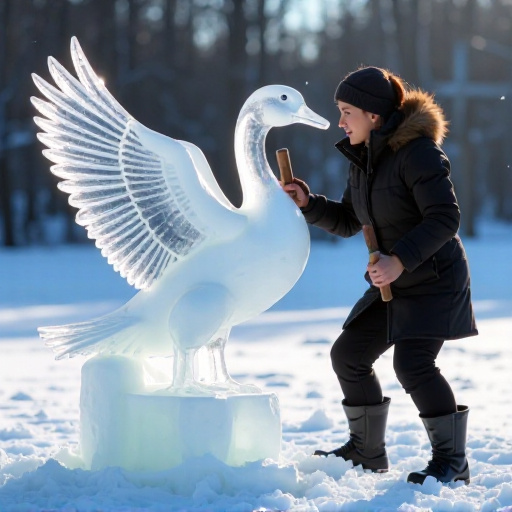}} & \adjustbox{valign=m}{\includegraphics[width=0.23\textwidth]{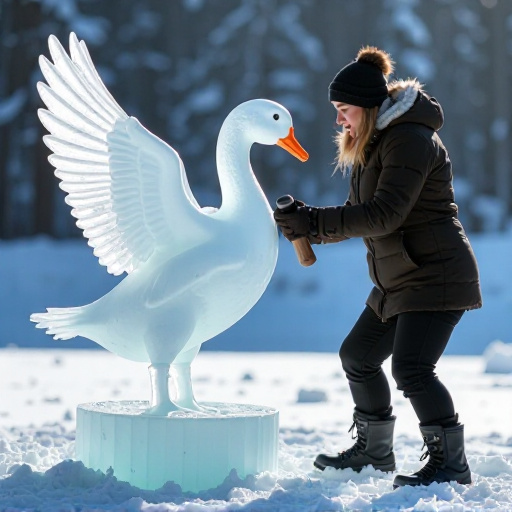}} & \adjustbox{valign=m}{\includegraphics[width=0.23\textwidth]{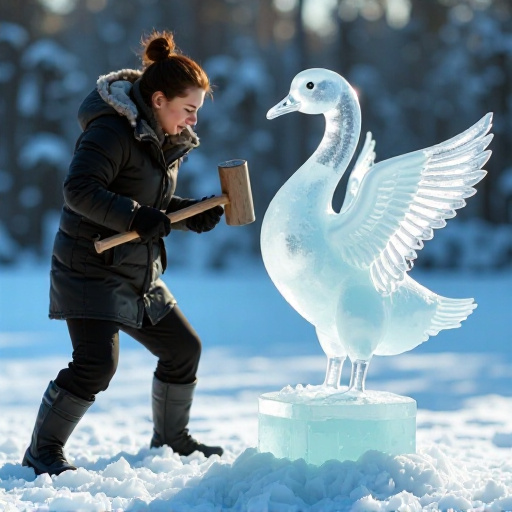}} \\
        \addlinespace[4pt]
        \multicolumn{5}{c}{\parbox{0.94\textwidth}{\centering\tiny\itshape ``a focused woman wielding a heavy sledgehammer, poised to strike an intricately carved ice sculpture of a goose. The sculpture glistens in the light, showcasing its detailed wings and feathers, standing on a pedestal of snow. Around her, shards of ice are scattered across the ground, evidence of her previous strikes.''}} \\[9pt]
        \addlinespace[8pt]
        \QualRowLabel{Teacher} &
        \adjustbox{valign=m}{\includegraphics[width=0.23\textwidth]{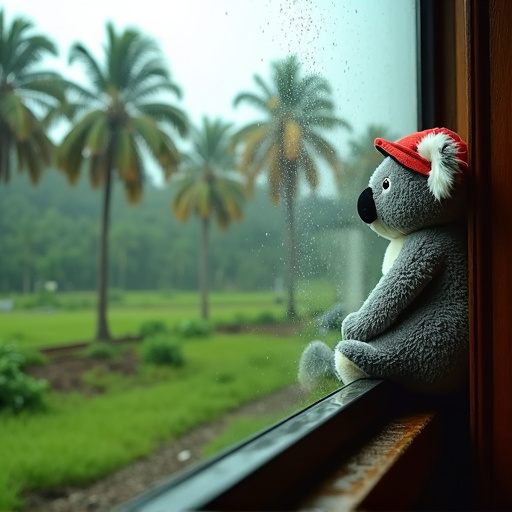}} & \adjustbox{valign=m}{\includegraphics[width=0.23\textwidth]{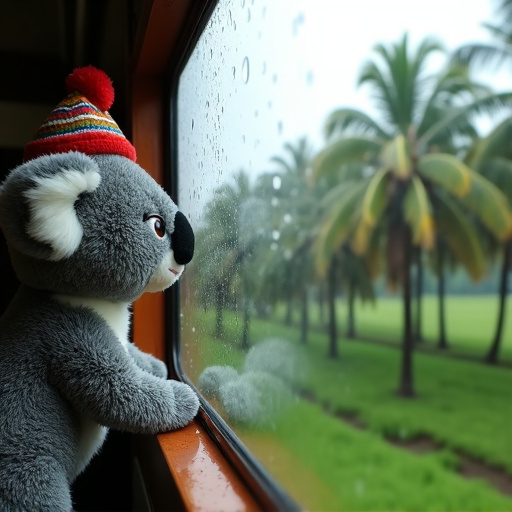}} & \adjustbox{valign=m}{\includegraphics[width=0.23\textwidth]{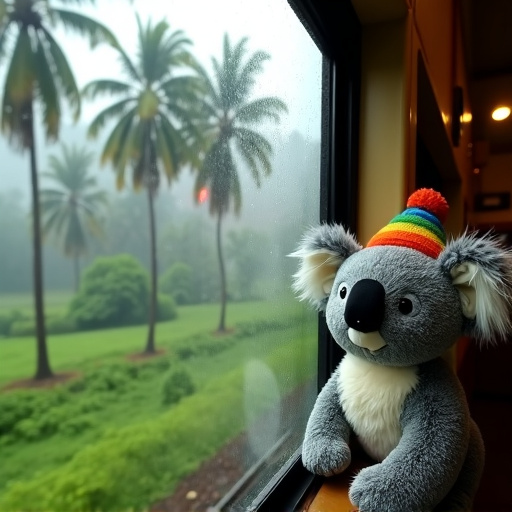}} & \adjustbox{valign=m}{\includegraphics[width=0.23\textwidth]{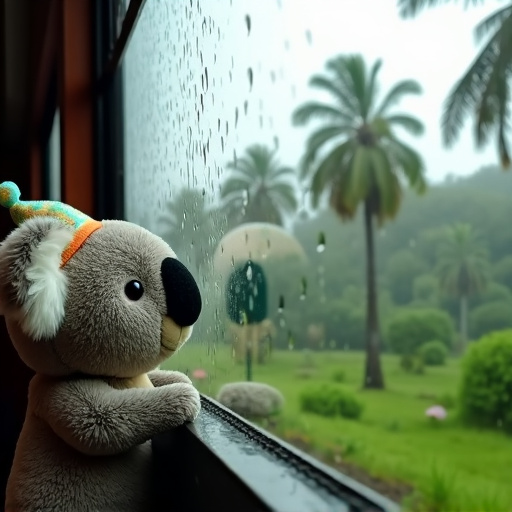}} \\
        \addlinespace[2pt]
        \QualRowLabel{\makecell{DMD2 \\ {\tiny\citep{yin2024:dmd2}}}} &
        \adjustbox{valign=m}{\includegraphics[width=0.23\textwidth]{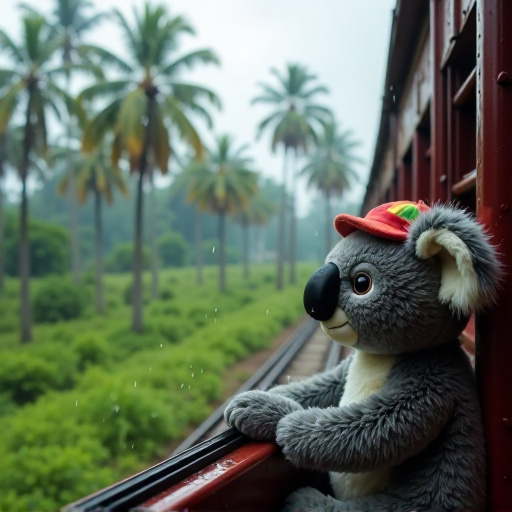}} & \adjustbox{valign=m}{\includegraphics[width=0.23\textwidth]{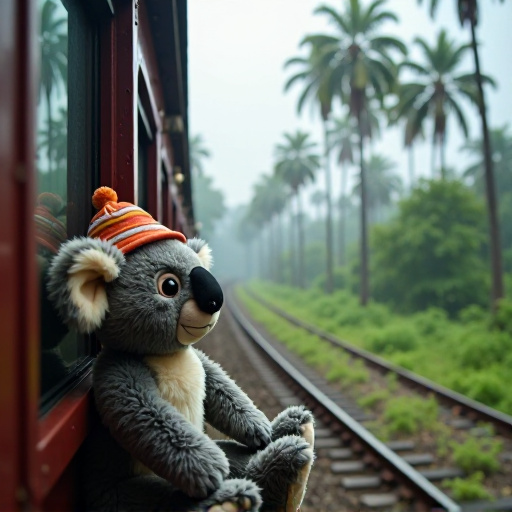}} & \adjustbox{valign=m}{\includegraphics[width=0.23\textwidth]{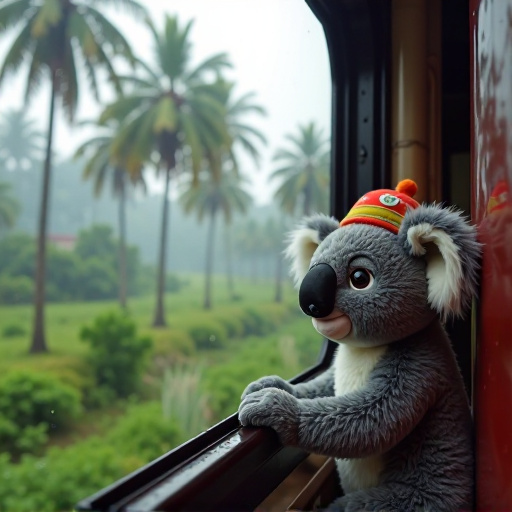}} & \adjustbox{valign=m}{\includegraphics[width=0.23\textwidth]{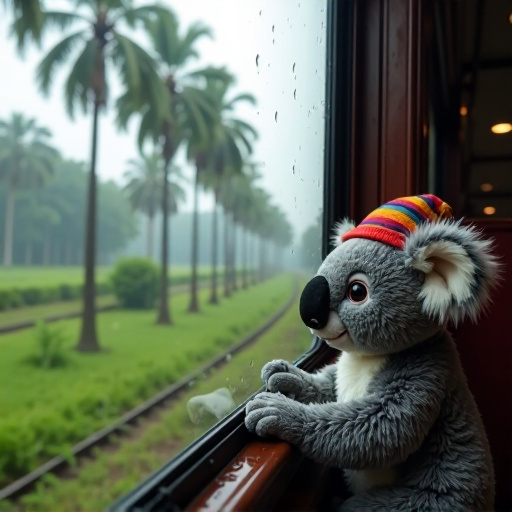}} \\
        \addlinespace[2pt]
        \QualRowLabel{\makecell{{\Oursbf{}} \\ {(Ours)}}} &
        \adjustbox{valign=m}{\includegraphics[width=0.23\textwidth]{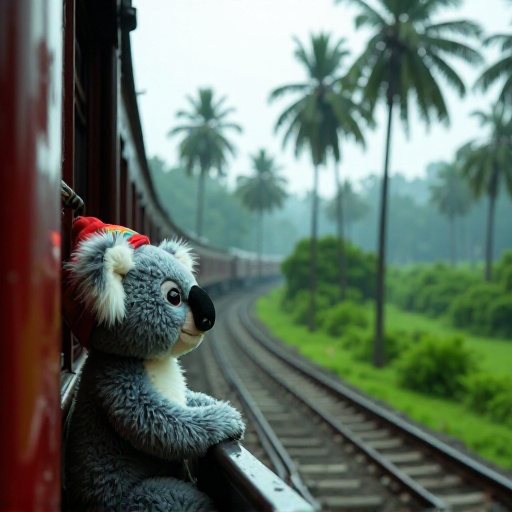}} & \adjustbox{valign=m}{\includegraphics[width=0.23\textwidth]{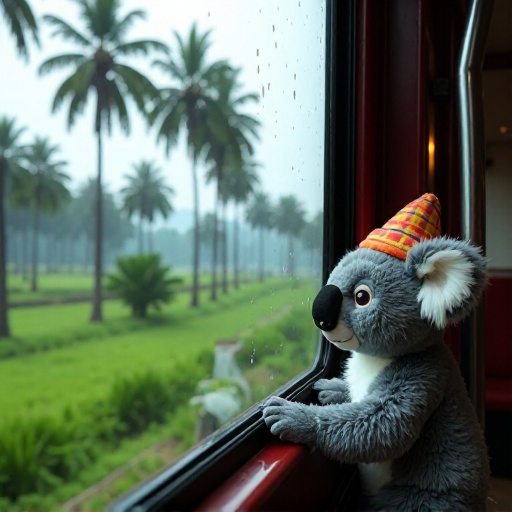}} & \adjustbox{valign=m}{\includegraphics[width=0.23\textwidth]{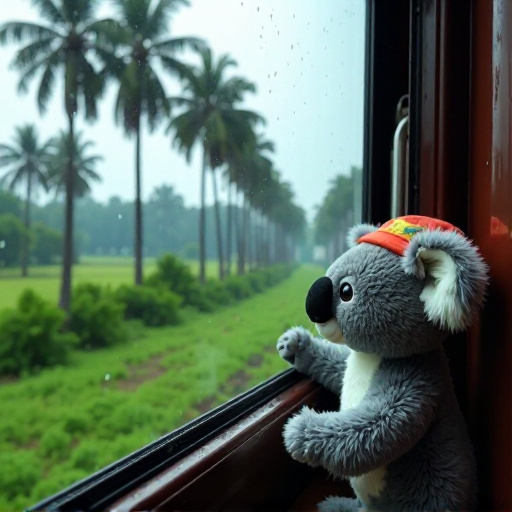}} & \adjustbox{valign=m}{\includegraphics[width=0.23\textwidth]{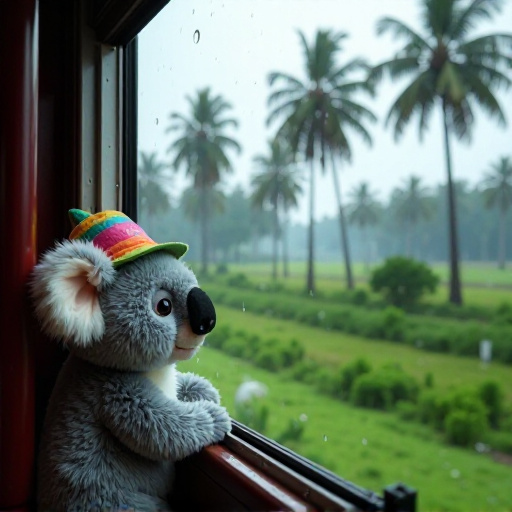}} \\
        \addlinespace[4pt]
        \multicolumn{5}{c}{\parbox{0.94\textwidth}{\centering\tiny\itshape ``A scenic train journey unfolds during the monsoon season in Kerala, with raindrops rhythmically tapping against the windowpane. Inside one of the train's carriages, a plush koala bear toy, donning a small, colorful hat, is propped up against the glass, peering out at the lush landscape. Outside, numerous tall coconut trees sway gently in the rain, their green fronds glistening with moisture, creating a verdant backdrop as the train meanders through the countryside.''}} \\[9pt]
    \end{tabular}
    \end{adjustbox}
\vspace{-3pt}\end{figure}

\begin{figure}[p]\vspace{-3pt}
    \centering
    \captionsetup{font=normalsize}
    \caption{
        \textbf{Additional Qualitative Comparison on FLUX.1 (Continued).}
        \Ours{} more closely matches the diversity of the teacher than DMD2 while preserving visual quality.
    }
    \label{fig:flux_qual_image_4}
    \setlength{\tabcolsep}{1.5pt}
    \renewcommand{\arraystretch}{1.05}
    \scriptsize
    \begin{adjustbox}{max totalsize={\textwidth}{0.92\textheight}}
    \begin{tabular}{c cccc}
        \QualRowLabel{Teacher} &
        \adjustbox{valign=m}{\includegraphics[width=0.23\textwidth]{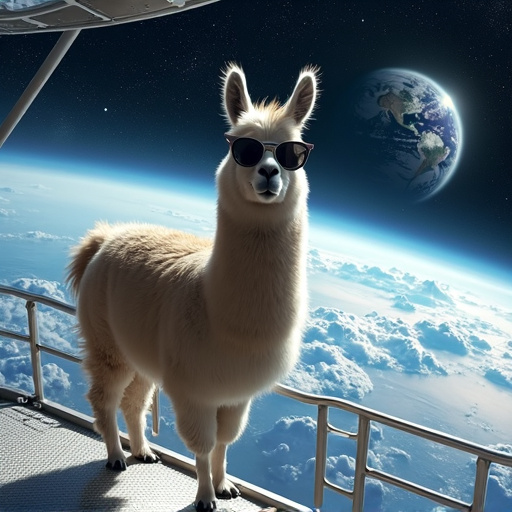}} & \adjustbox{valign=m}{\includegraphics[width=0.23\textwidth]{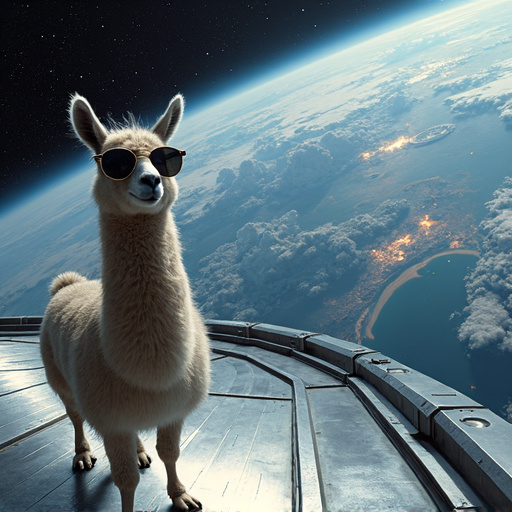}} & \adjustbox{valign=m}{\includegraphics[width=0.23\textwidth]{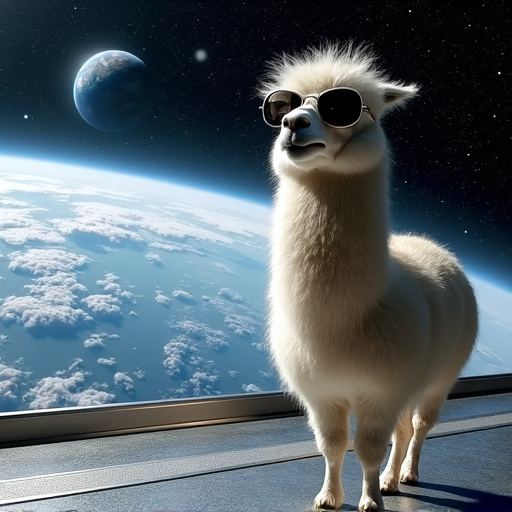}} & \adjustbox{valign=m}{\includegraphics[width=0.23\textwidth]{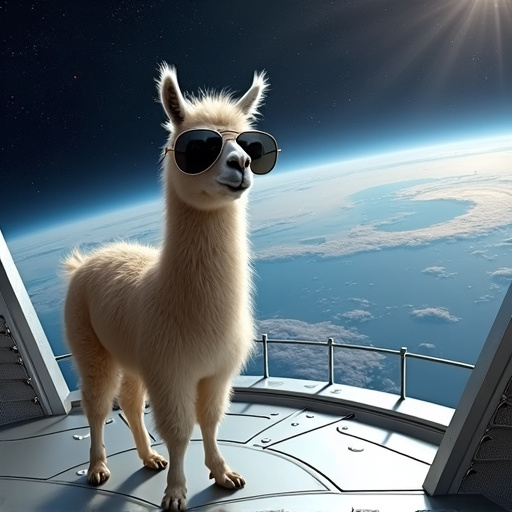}} \\
        \addlinespace[2pt]
        \QualRowLabel{\makecell{DMD2 \\ {\tiny\citep{yin2024:dmd2}}}} &
        \adjustbox{valign=m}{\includegraphics[width=0.23\textwidth]{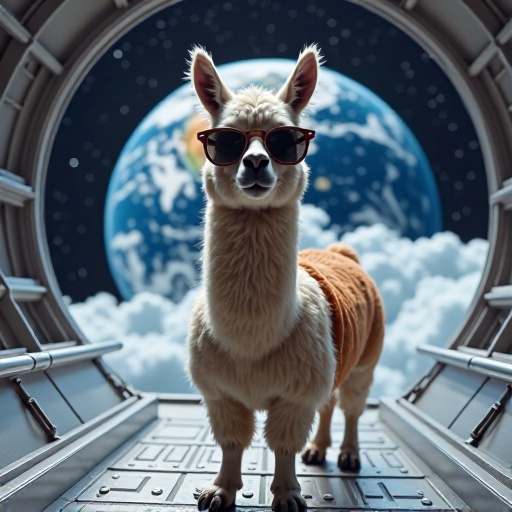}} & \adjustbox{valign=m}{\includegraphics[width=0.23\textwidth]{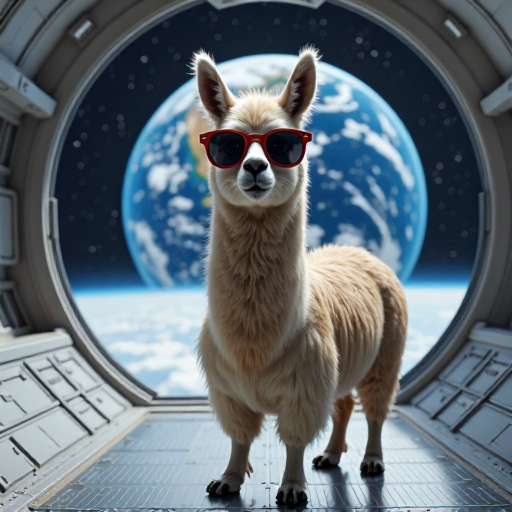}} & \adjustbox{valign=m}{\includegraphics[width=0.23\textwidth]{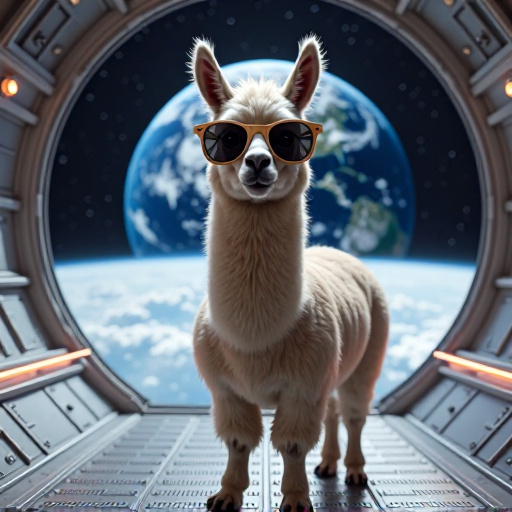}} & \adjustbox{valign=m}{\includegraphics[width=0.23\textwidth]{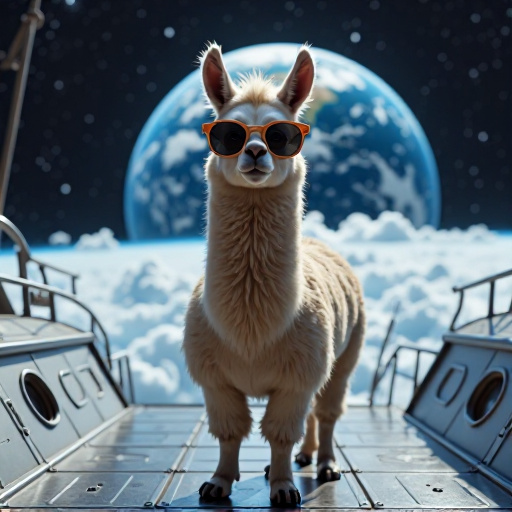}} \\
        \addlinespace[2pt]
        \QualRowLabel{\makecell{{\Oursbf{}} \\ {(Ours)}}} &
        \adjustbox{valign=m}{\includegraphics[width=0.23\textwidth]{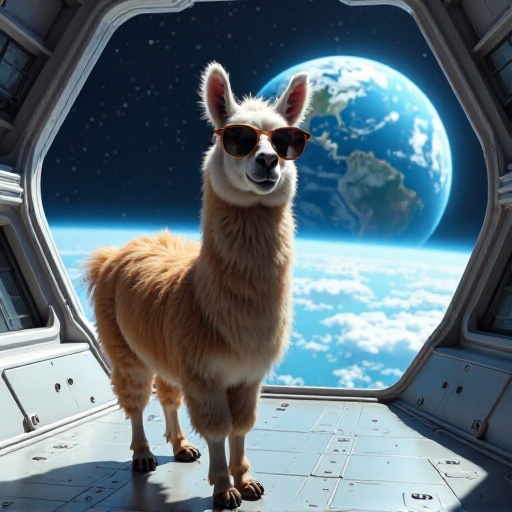}} & \adjustbox{valign=m}{\includegraphics[width=0.23\textwidth]{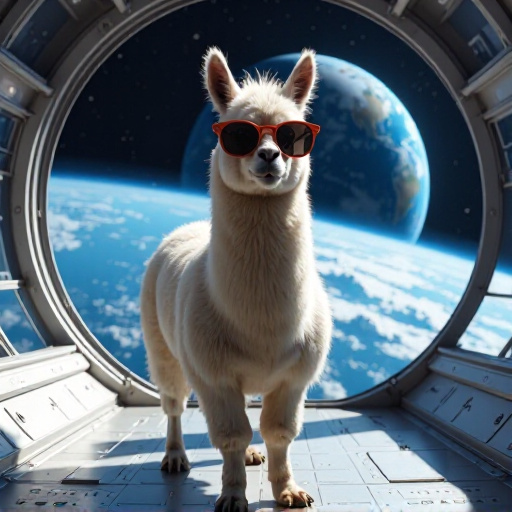}} & \adjustbox{valign=m}{\includegraphics[width=0.23\textwidth]{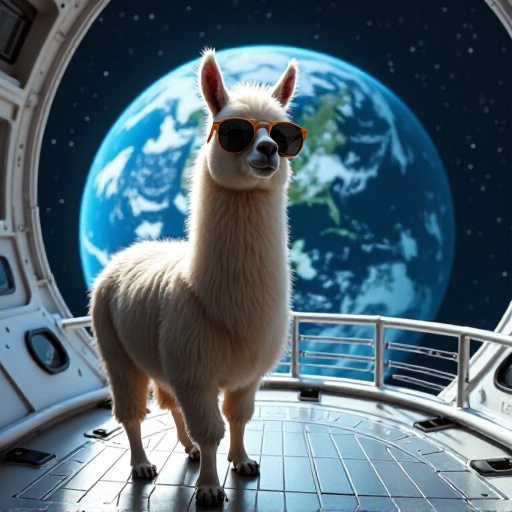}} & \adjustbox{valign=m}{\includegraphics[width=0.23\textwidth]{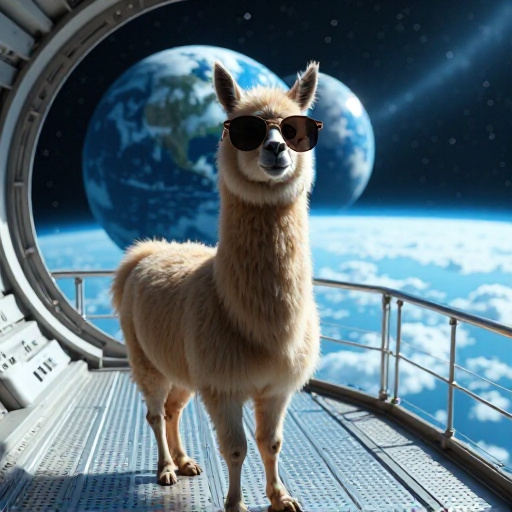}} \\
        \addlinespace[4pt]
        \multicolumn{5}{c}{\parbox{0.94\textwidth}{\centering\tiny\itshape ``A whimsical scene where a llama, adorned with a pair of oversized, round sunglasses, stands confidently on the metallic deck of a spacecraft. The deck beneath the llama's hooves gleams with a polished silver finish, reflecting the starry cosmos that surrounds the vessel. In the vast backdrop, the Earth looms large, a swirl of blue oceans and white clouds, providing a stunning contrast to the spaceship's sleek, futuristic design.''}} \\[9pt]
        \addlinespace[8pt]
        \QualRowLabel{Teacher} &
        \adjustbox{valign=m}{\includegraphics[width=0.23\textwidth]{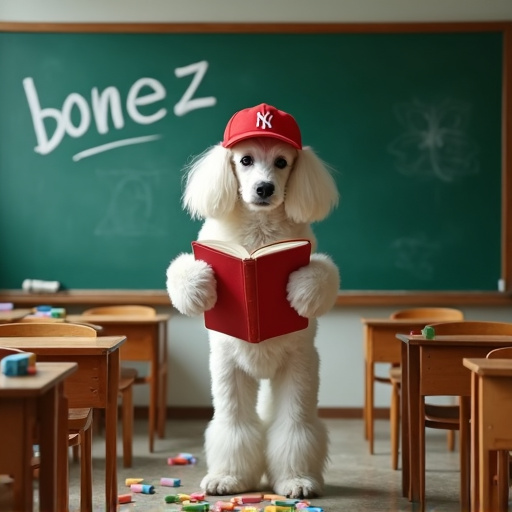}} & \adjustbox{valign=m}{\includegraphics[width=0.23\textwidth]{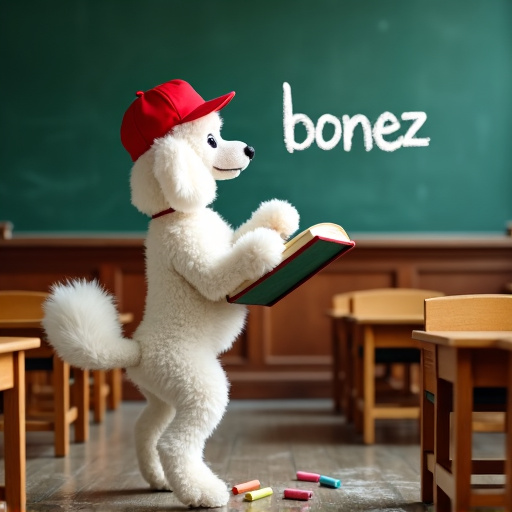}} & \adjustbox{valign=m}{\includegraphics[width=0.23\textwidth]{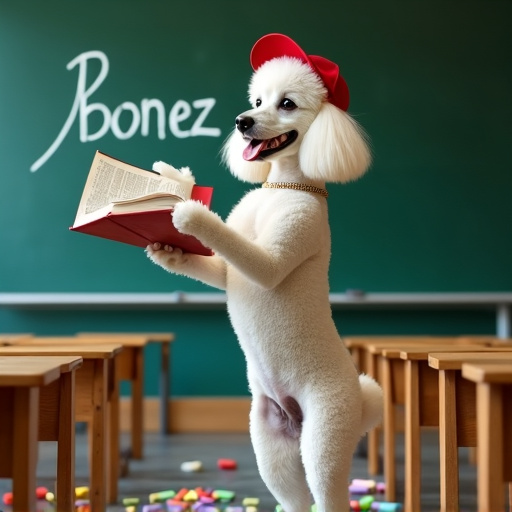}} & \adjustbox{valign=m}{\includegraphics[width=0.23\textwidth]{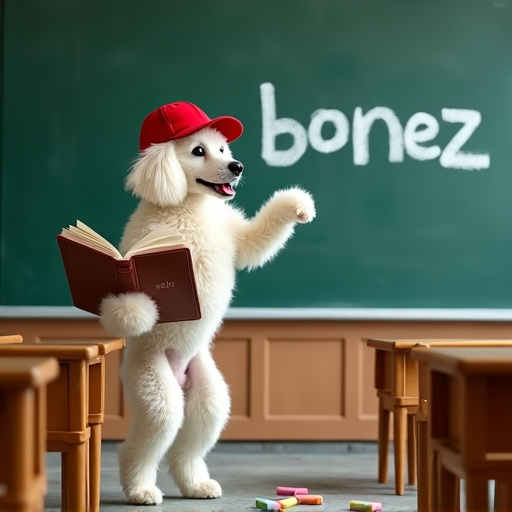}} \\
        \addlinespace[2pt]
        \QualRowLabel{\makecell{DMD2 \\ {\tiny\citep{yin2024:dmd2}}}} &
        \adjustbox{valign=m}{\includegraphics[width=0.23\textwidth]{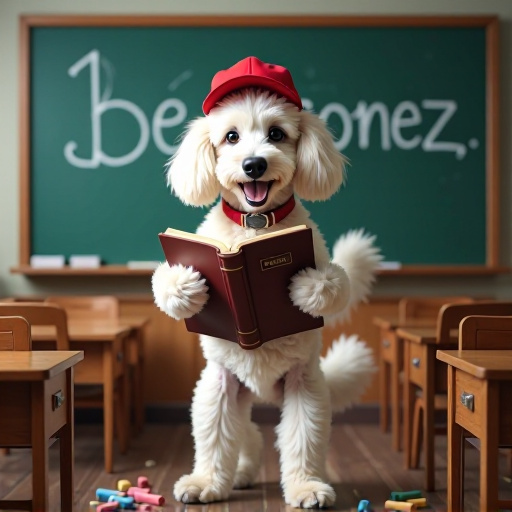}} & \adjustbox{valign=m}{\includegraphics[width=0.23\textwidth]{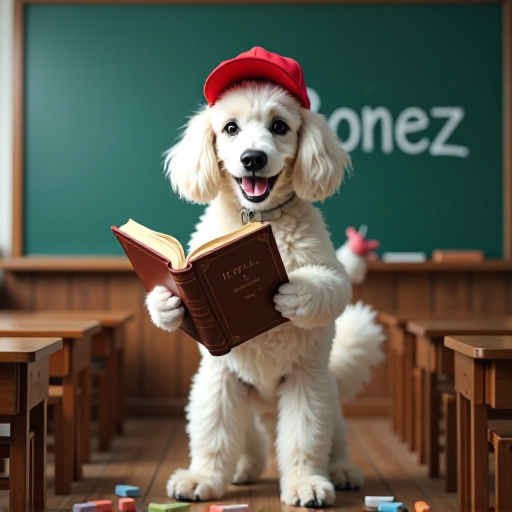}} & \adjustbox{valign=m}{\includegraphics[width=0.23\textwidth]{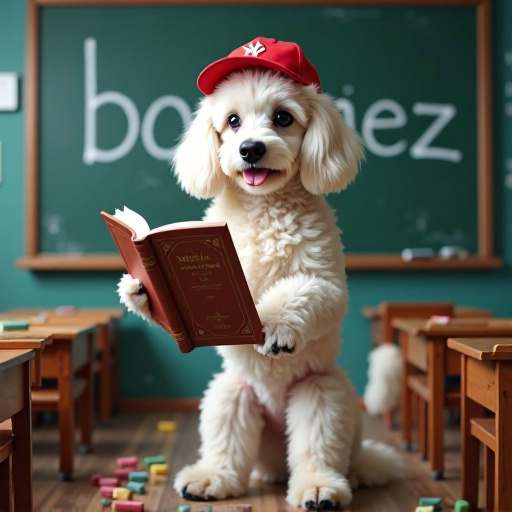}} & \adjustbox{valign=m}{\includegraphics[width=0.23\textwidth]{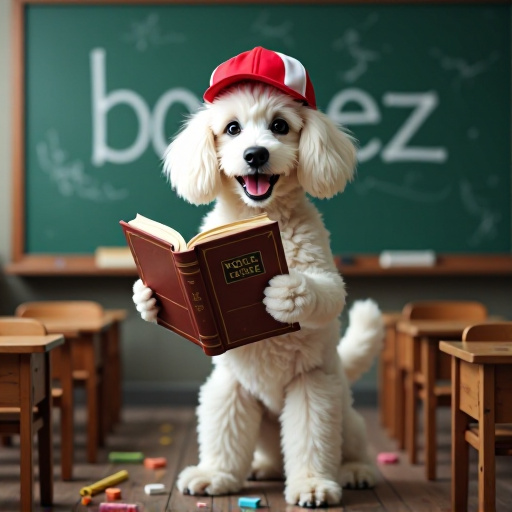}} \\
        \addlinespace[2pt]
        \QualRowLabel{\makecell{{\Oursbf{}} \\ {(Ours)}}} &
        \adjustbox{valign=m}{\includegraphics[width=0.23\textwidth]{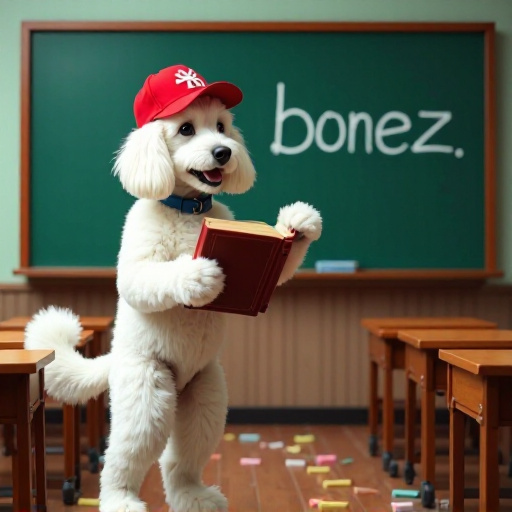}} & \adjustbox{valign=m}{\includegraphics[width=0.23\textwidth]{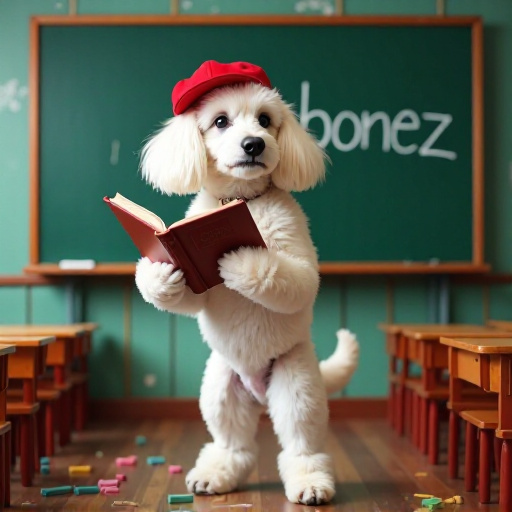}} & \adjustbox{valign=m}{\includegraphics[width=0.23\textwidth]{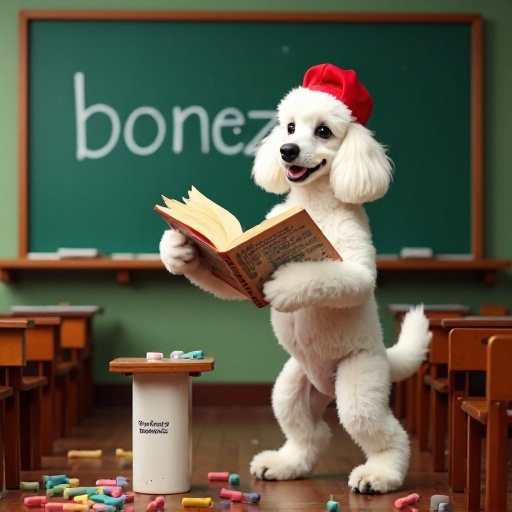}} & \adjustbox{valign=m}{\includegraphics[width=0.23\textwidth]{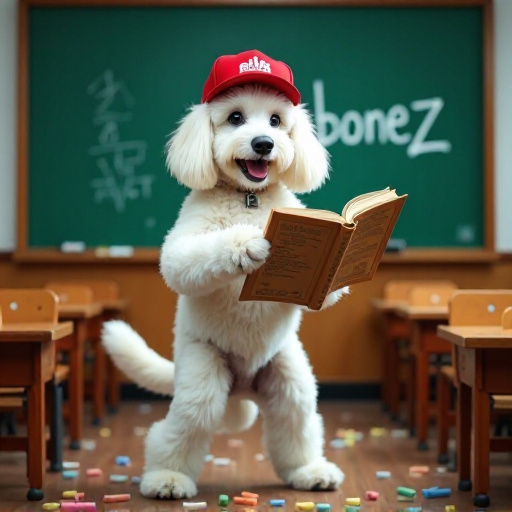}} \\
        \addlinespace[4pt]
        \multicolumn{5}{c}{\parbox{0.94\textwidth}{\centering\tiny\itshape ``A fluffy white poodle, adorned with a red baseball cap, stands on its hind legs in front of a green chalkboard. In one paw, it holds a large, open dictionary, while the other paw is used to scrawl the word ``bonez'' in white chalk. The floor beneath the poodle is scattered with colorful pieces of chalk, and the room is filled with small wooden desks and chairs, suggesting a classroom setting.''}} \\[9pt]
    \end{tabular}
    \end{adjustbox}
\vspace{-3pt}\end{figure}

\providecommand{\QualRowLabel}[1]{\adjustbox{valign=m}{\rotatebox[origin=c]{90}{\scriptsize #1}}}
\begin{figure}[p]\vspace{-3pt}
    \centering
    \captionsetup{font=normalsize}
    \caption{
        \textbf{Additional Qualitative Comparison on Qwen-Image 20B.}
        \Ours{} more closely matches the diversity of the teacher than DMD-based Qwen-Lightning while preserving visual quality.
        \checkpointmark{} denotes an officially released checkpoint.
    }
    \label{fig:qwen_qual_image}
    \setlength{\tabcolsep}{1.5pt}
    \renewcommand{\arraystretch}{1.05}
    \scriptsize
    \begin{adjustbox}{max totalsize={\textwidth}{0.92\textheight}}
    \begin{tabular}{c cccc}
        \QualRowLabel{Teacher} &
        \adjustbox{valign=m}{\includegraphics[width=0.23\textwidth]{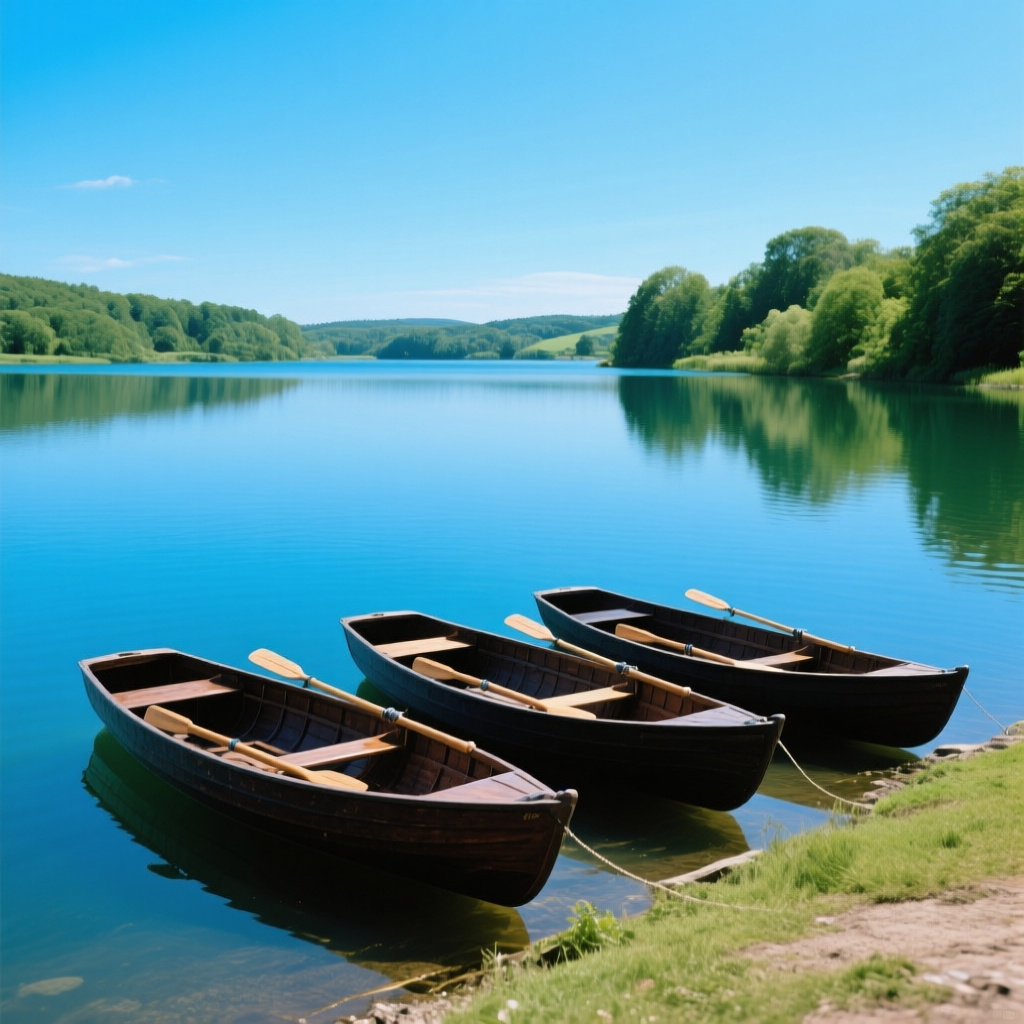}} & \adjustbox{valign=m}{\includegraphics[width=0.23\textwidth]{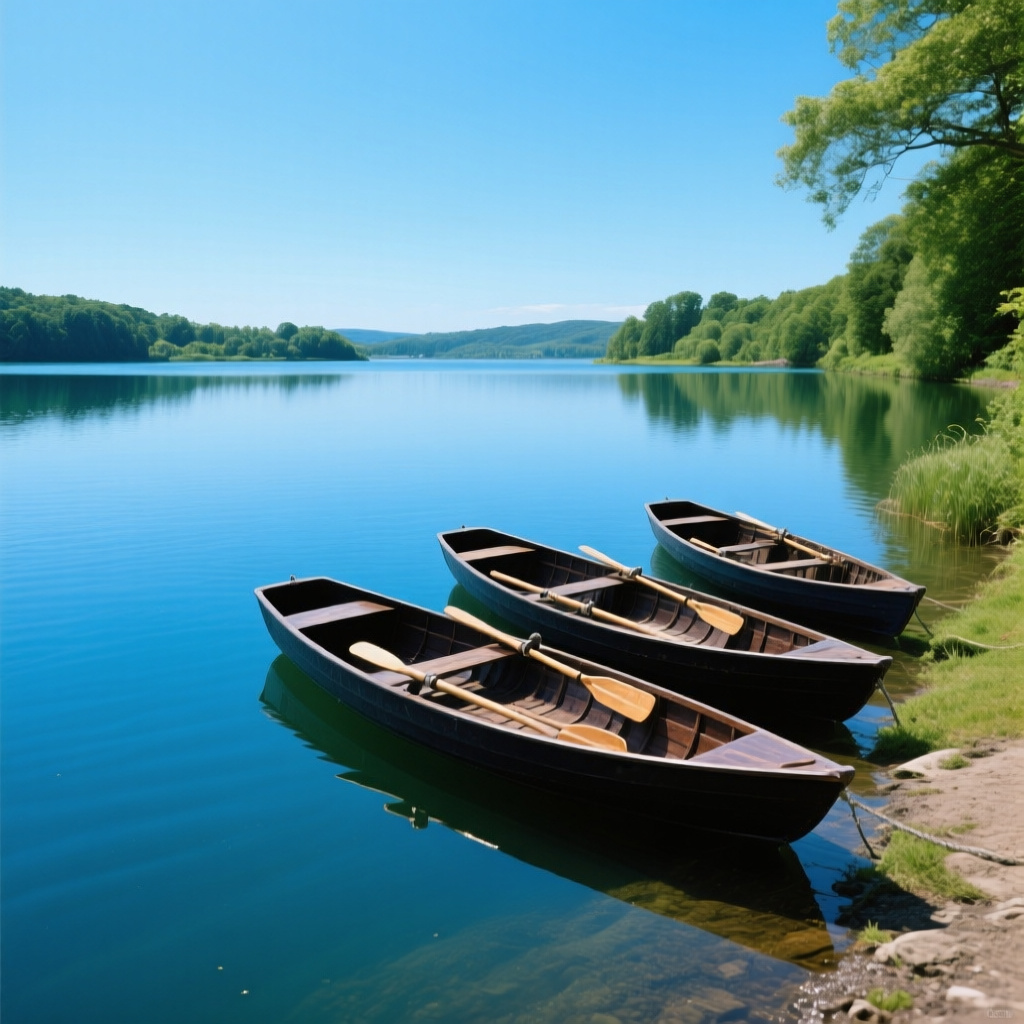}} & \adjustbox{valign=m}{\includegraphics[width=0.23\textwidth]{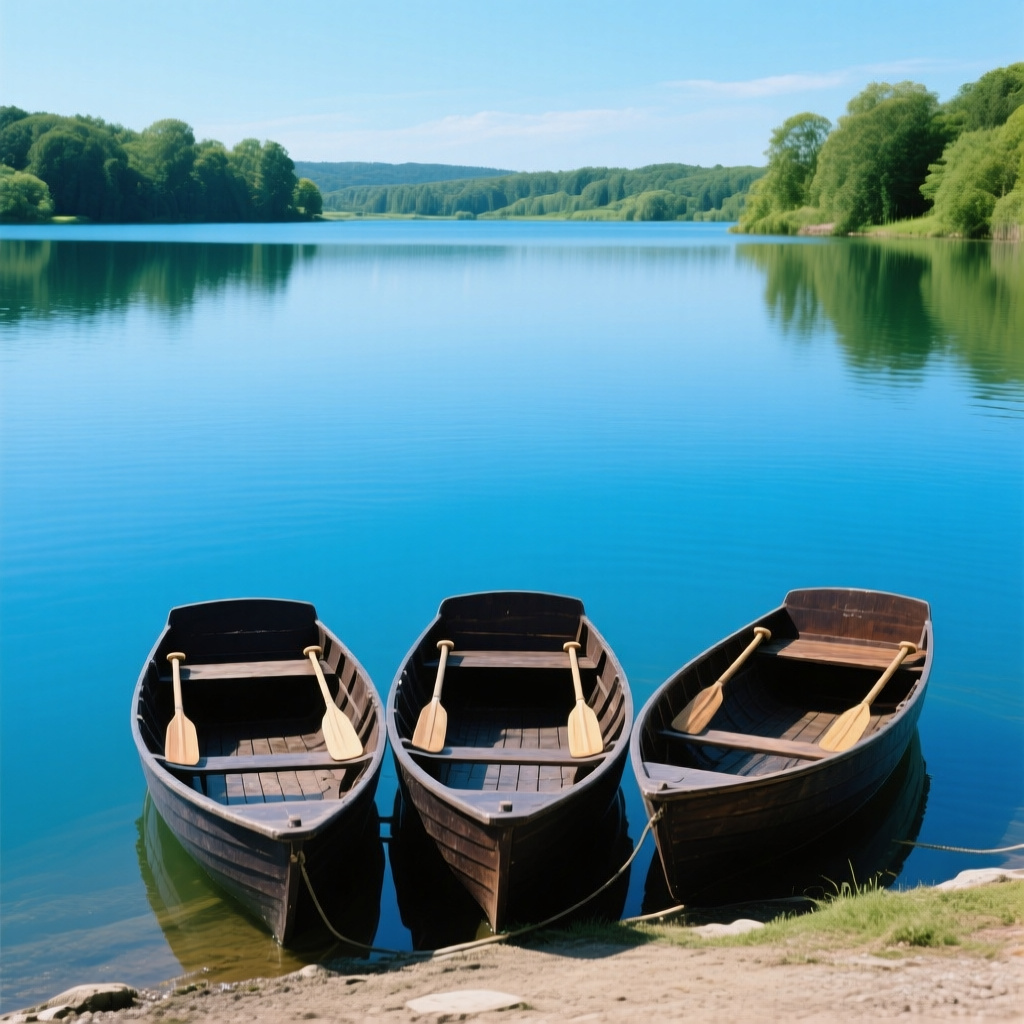}} & \adjustbox{valign=m}{\includegraphics[width=0.23\textwidth]{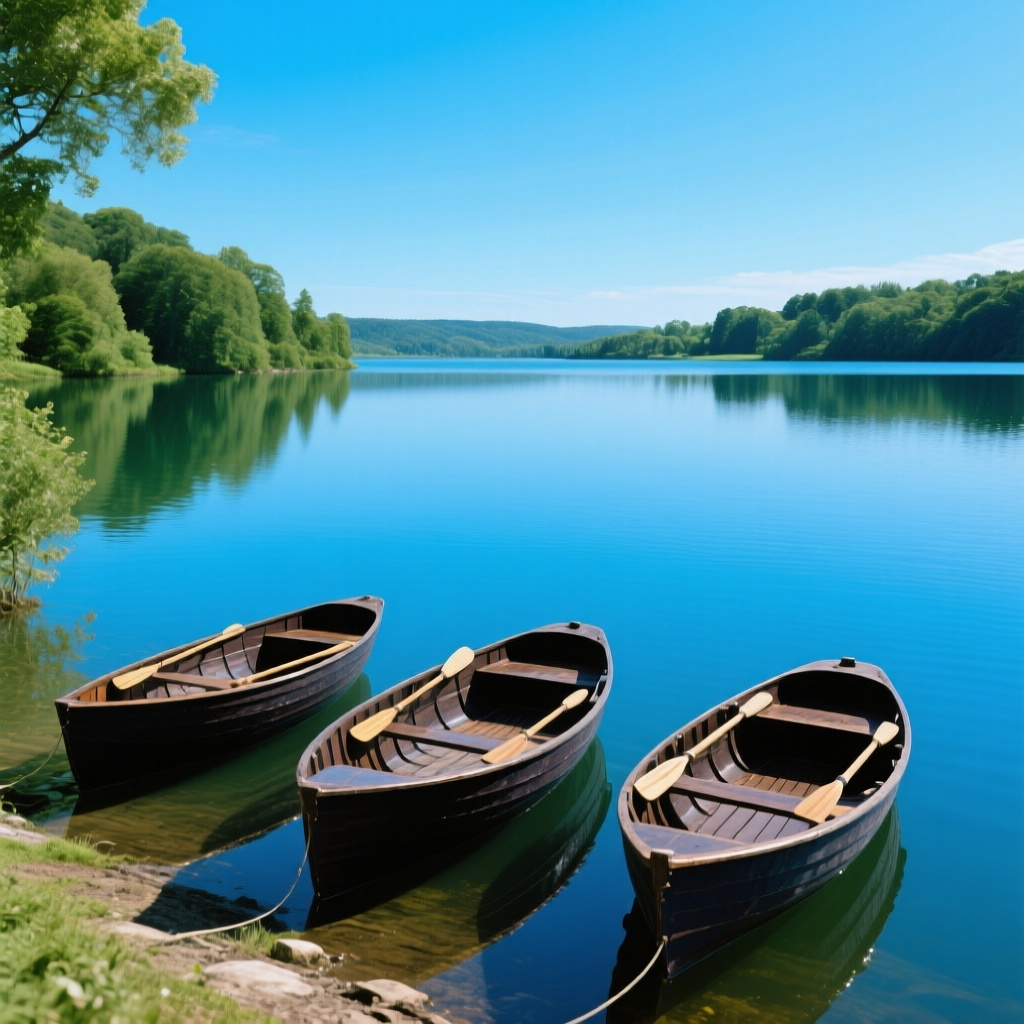}} \\
        \addlinespace[2pt]
        \QualRowLabel{\makecell{Qwen-Lightning\checkpointmark \\ {\tiny\citep{fan2025:qwenlightning}}}} &
        \adjustbox{valign=m}{\includegraphics[width=0.23\textwidth]{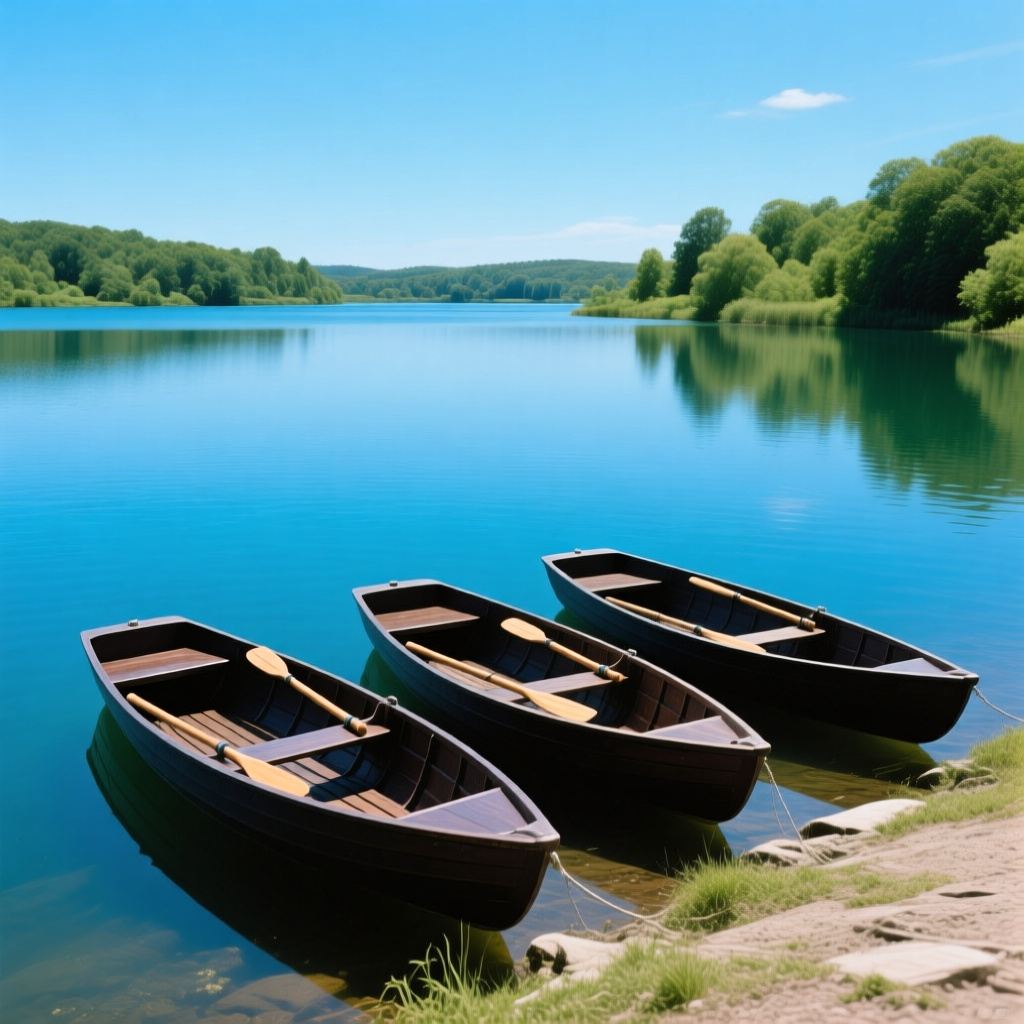}} & \adjustbox{valign=m}{\includegraphics[width=0.23\textwidth]{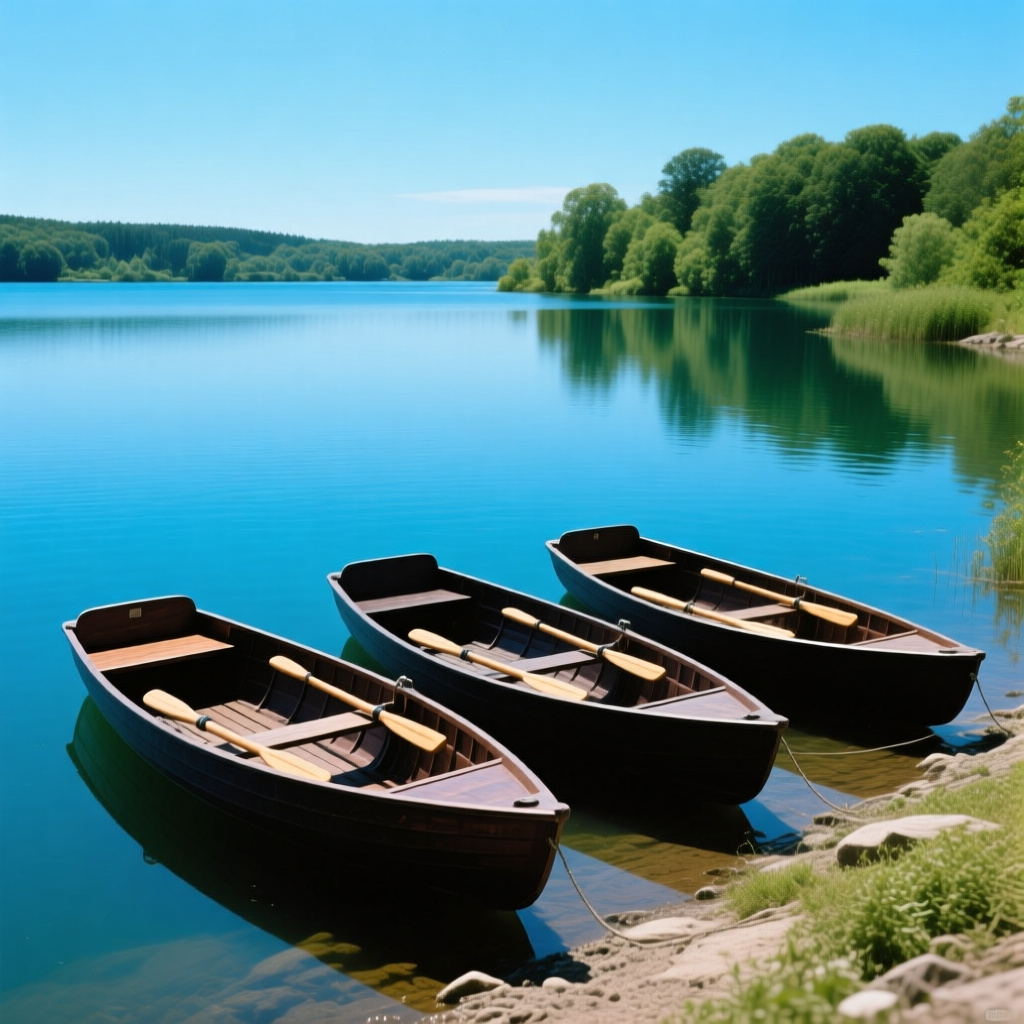}} & \adjustbox{valign=m}{\includegraphics[width=0.23\textwidth]{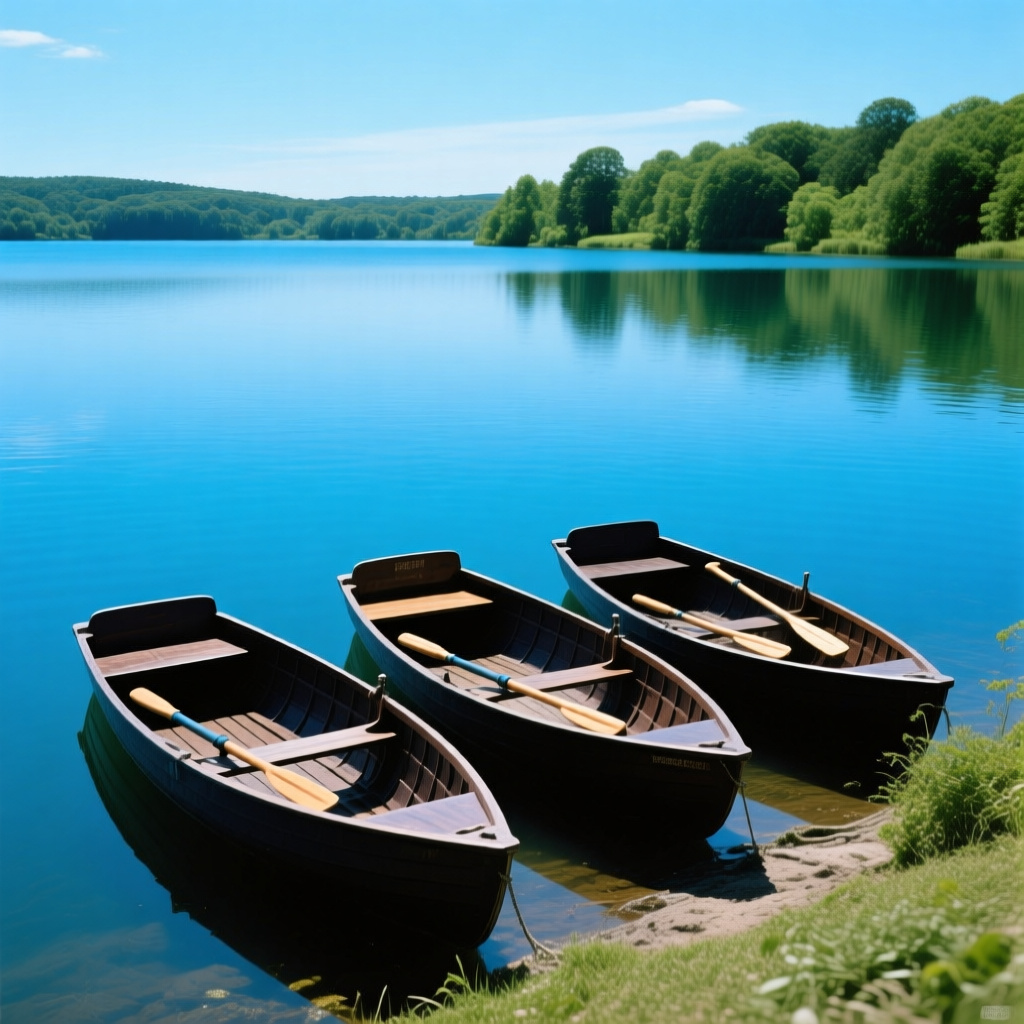}} & \adjustbox{valign=m}{\includegraphics[width=0.23\textwidth]{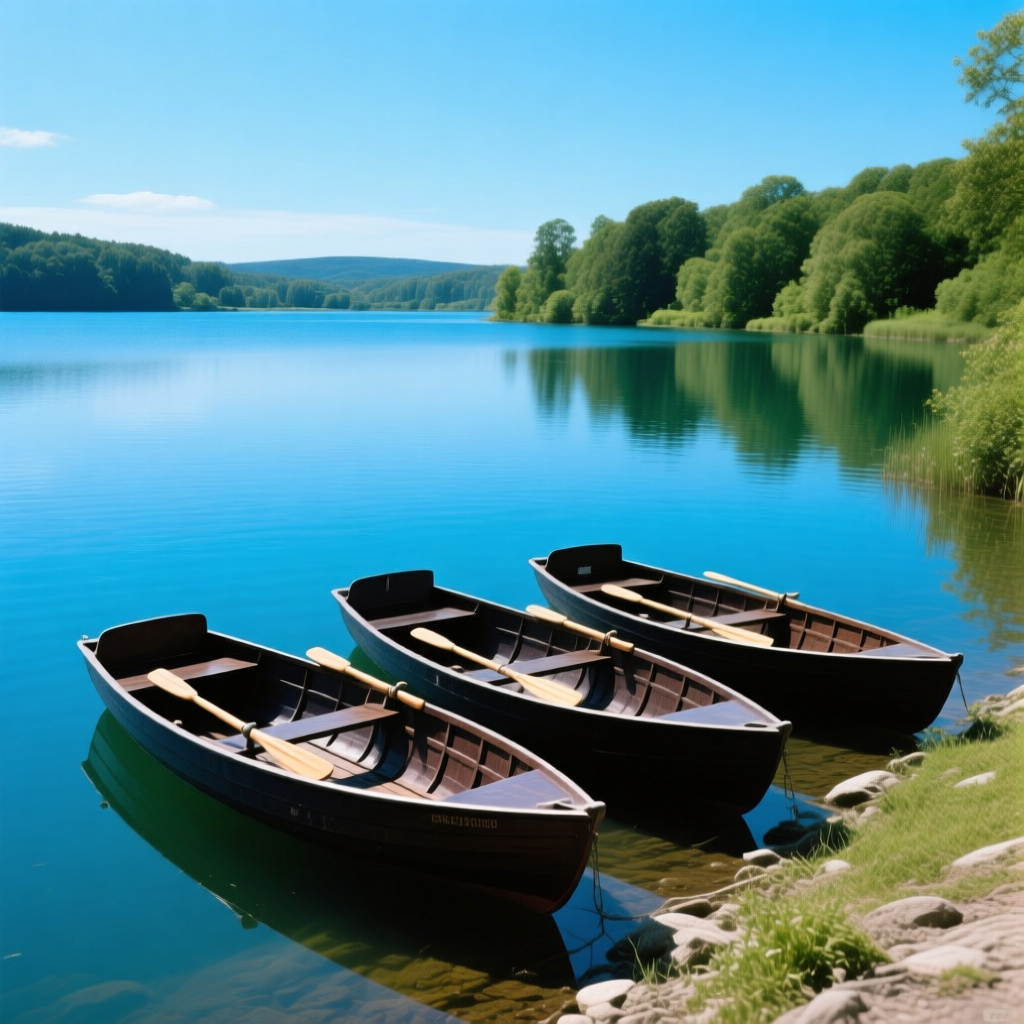}} \\
        \addlinespace[2pt]
        \QualRowLabel{\makecell{{\Oursbf{}} \\ {(Ours)}}} &
        \adjustbox{valign=m}{\includegraphics[width=0.23\textwidth]{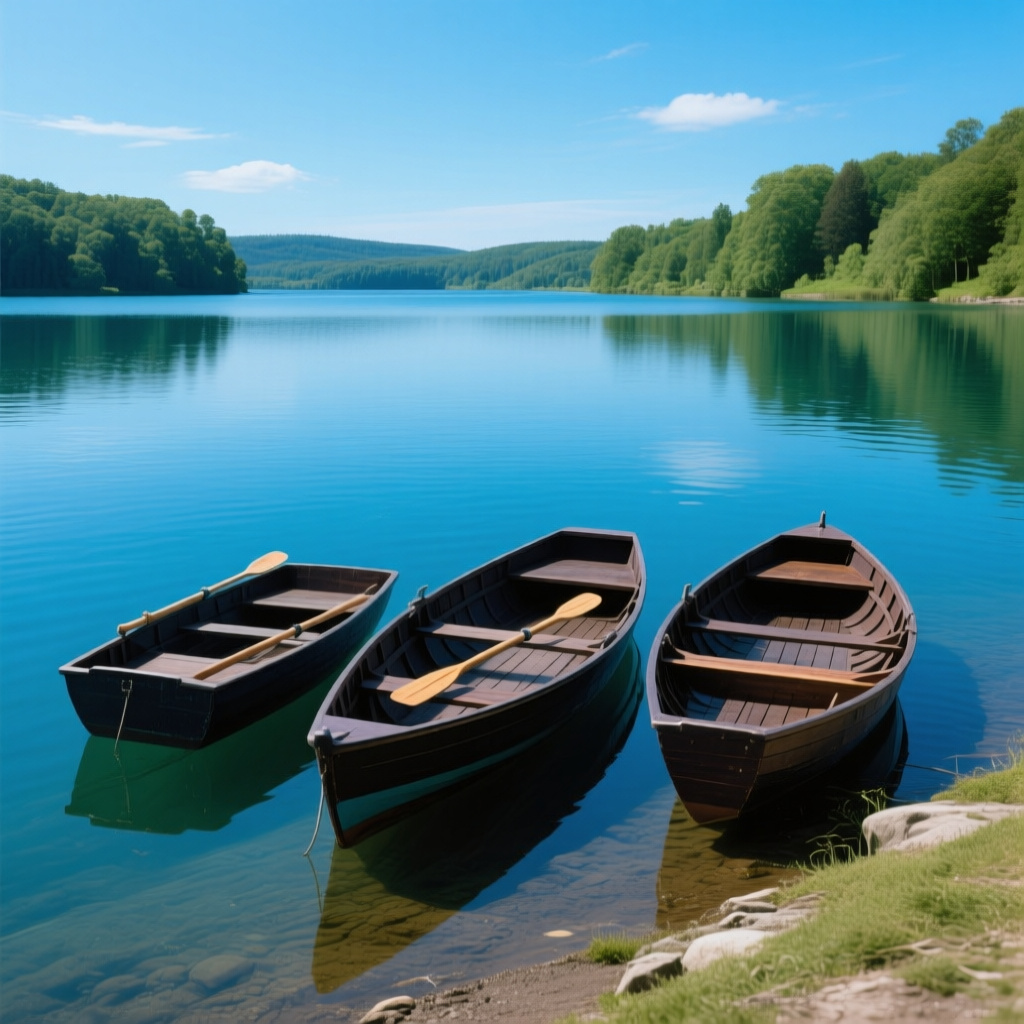}} & \adjustbox{valign=m}{\includegraphics[width=0.23\textwidth]{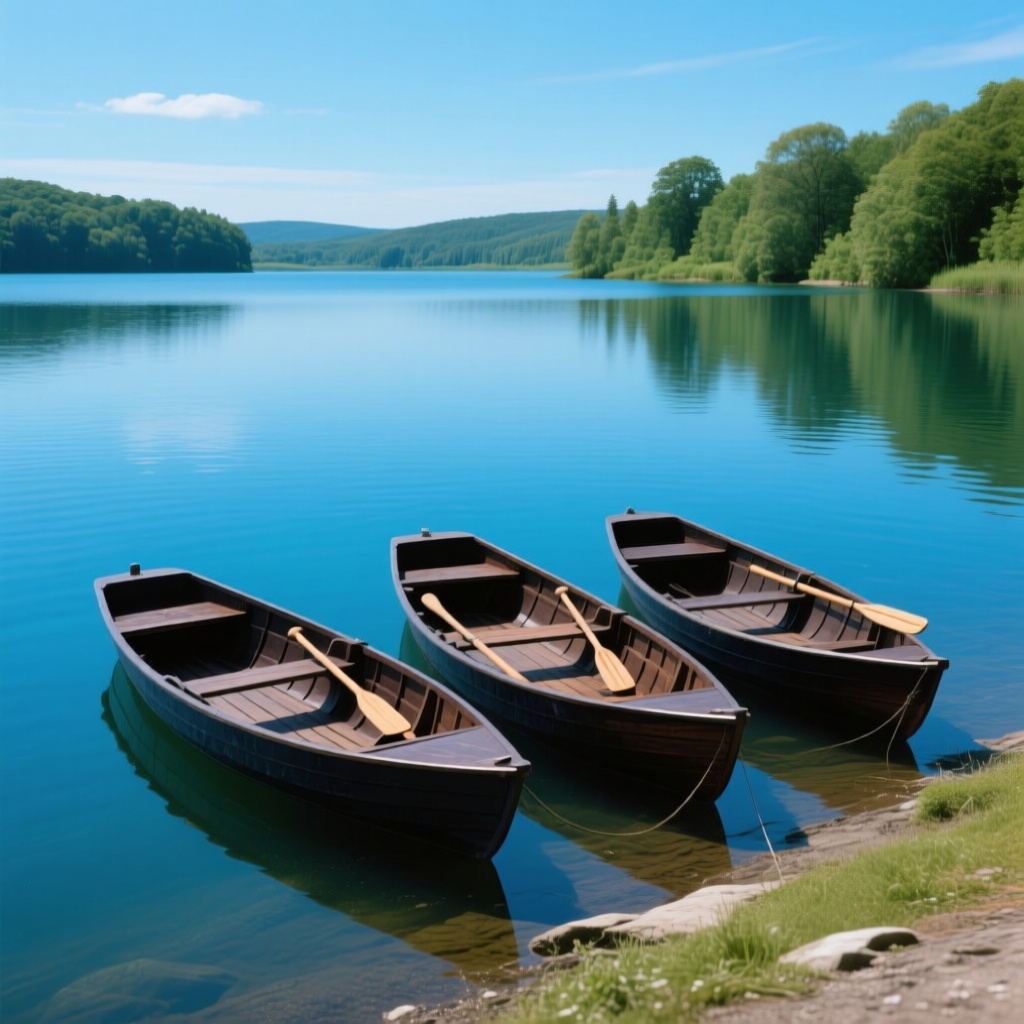}} & \adjustbox{valign=m}{\includegraphics[width=0.23\textwidth]{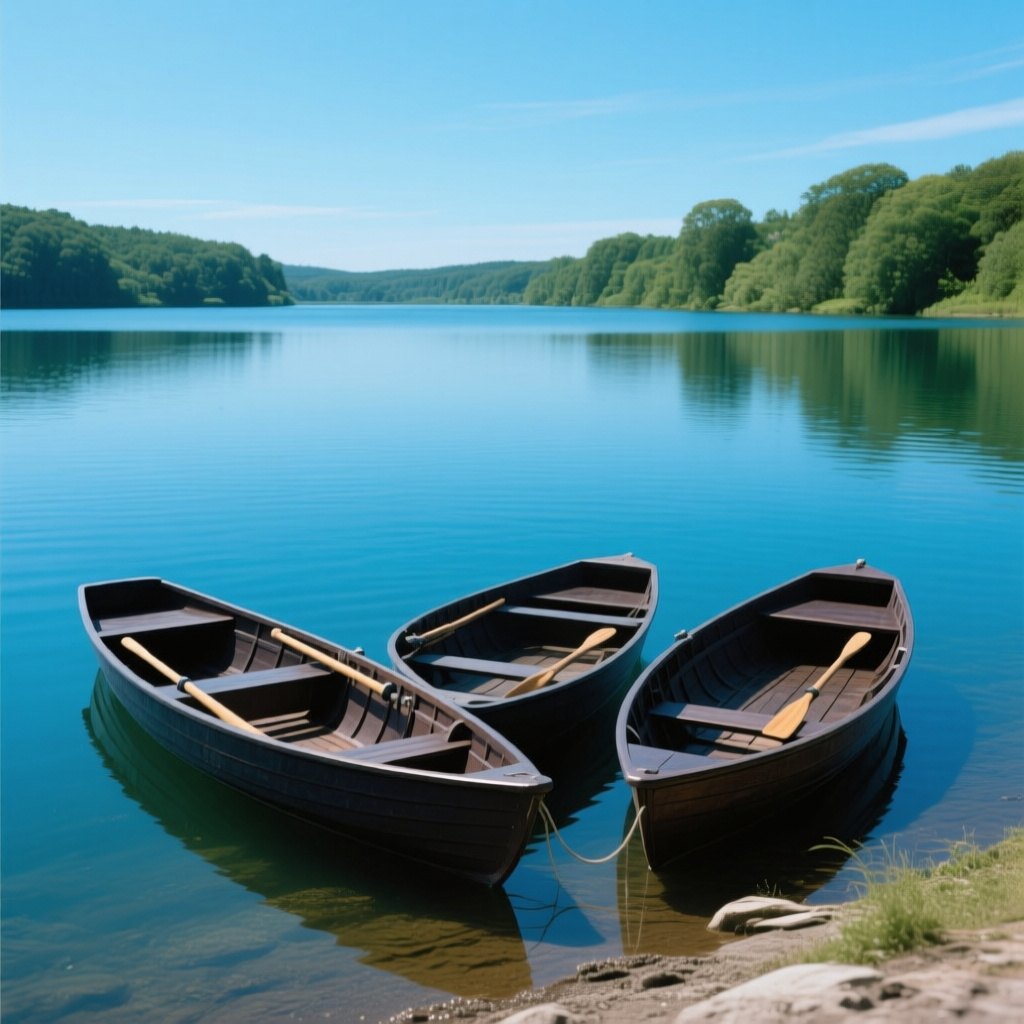}} & \adjustbox{valign=m}{\includegraphics[width=0.23\textwidth]{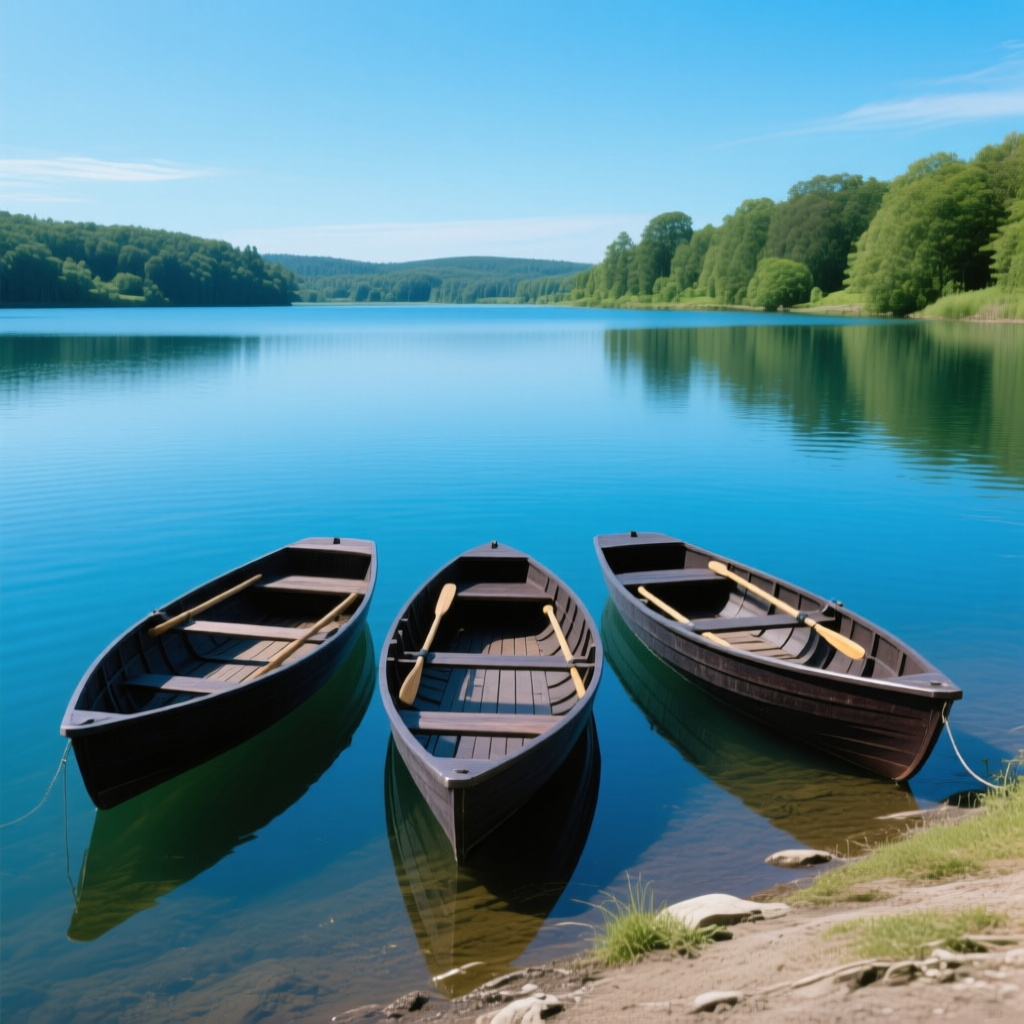}} \\
        \addlinespace[4pt]
        \multicolumn{5}{c}{\parbox{0.94\textwidth}{\centering\tiny\itshape ``Three sleek, dark wooden boats are resting along the banks of a tranquil, azure blue lake. The smooth surface of the water reflects the clear sky above and the lush greenery surrounding the lake's edge. Positioned close to each other, the boats' oars are tucked neatly inside, hinting at a recent or upcoming journey across the calm waters.''}} \\[9pt]
        \QualRowLabel{Teacher} &
        \adjustbox{valign=m}{\includegraphics[width=0.23\textwidth]{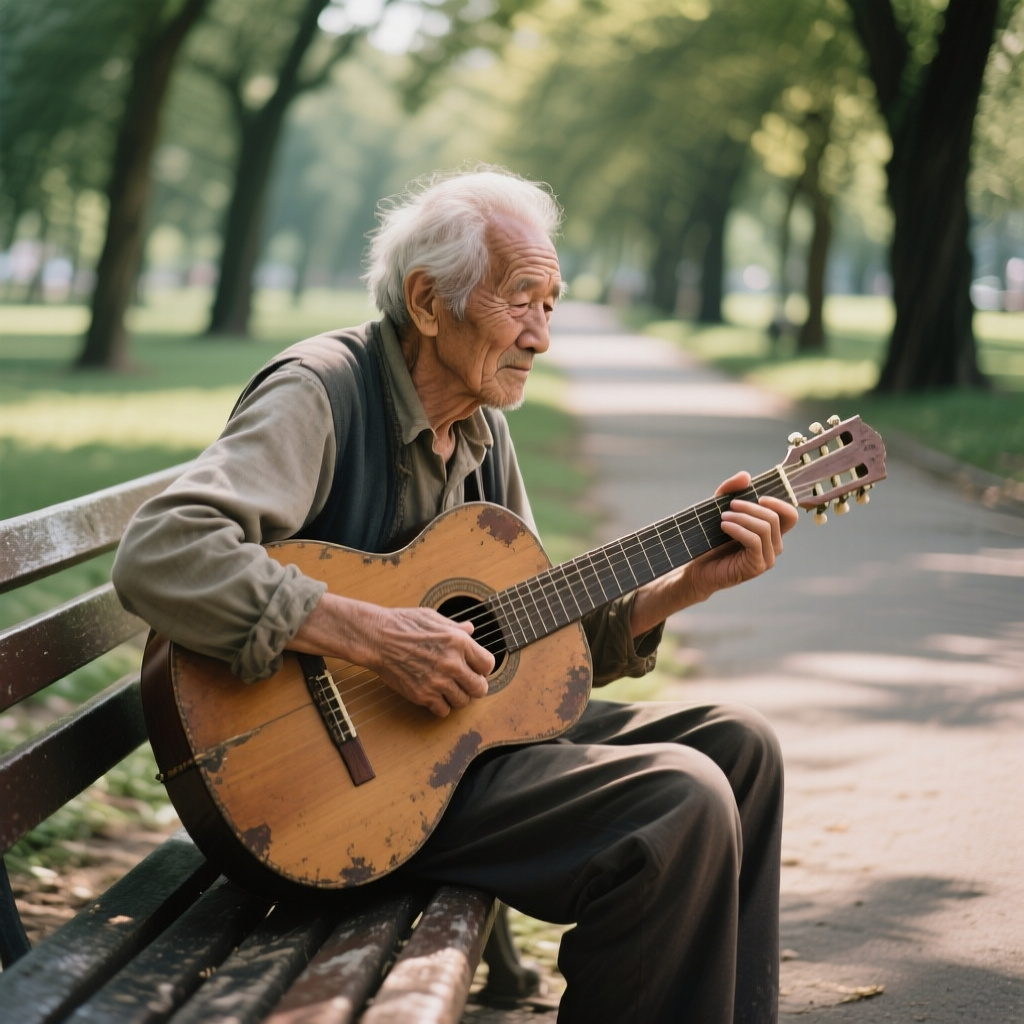}} & \adjustbox{valign=m}{\includegraphics[width=0.23\textwidth]{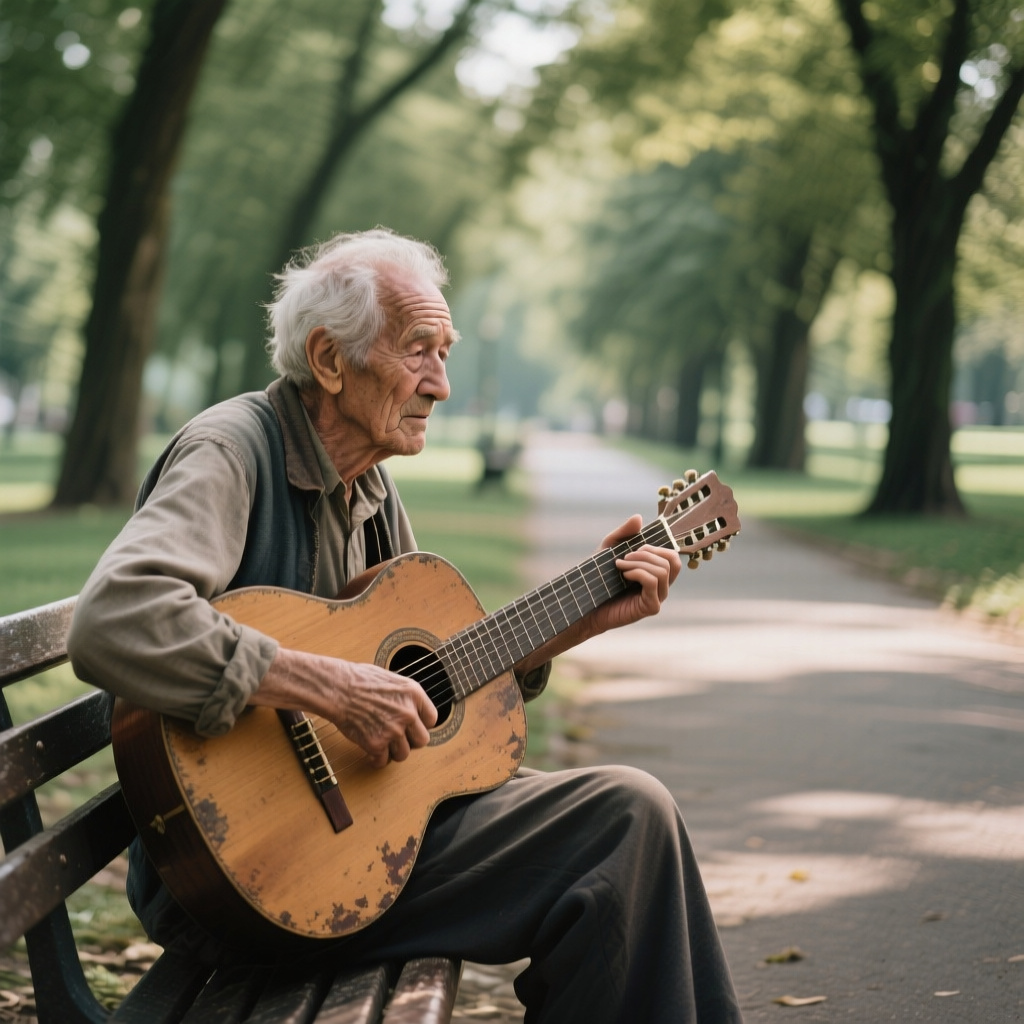}} & \adjustbox{valign=m}{\includegraphics[width=0.23\textwidth]{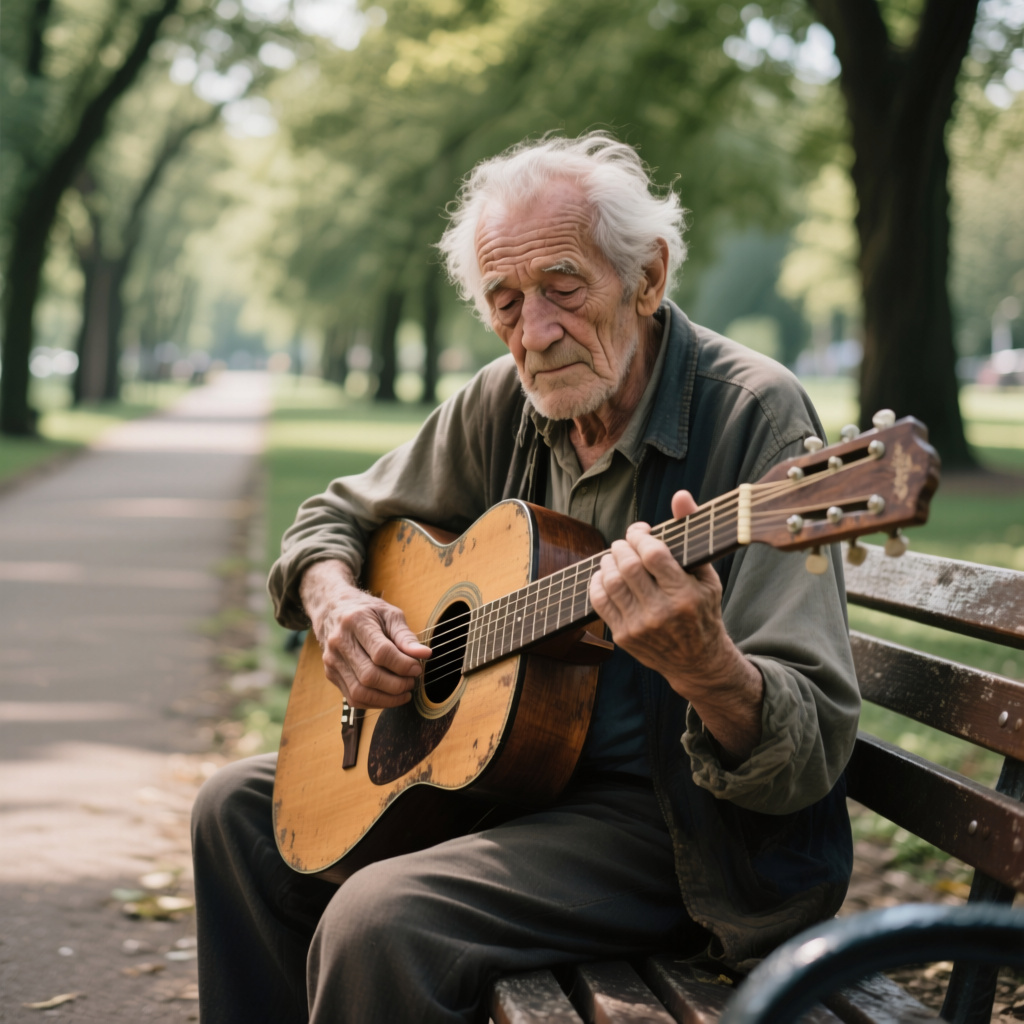}} & \adjustbox{valign=m}{\includegraphics[width=0.23\textwidth]{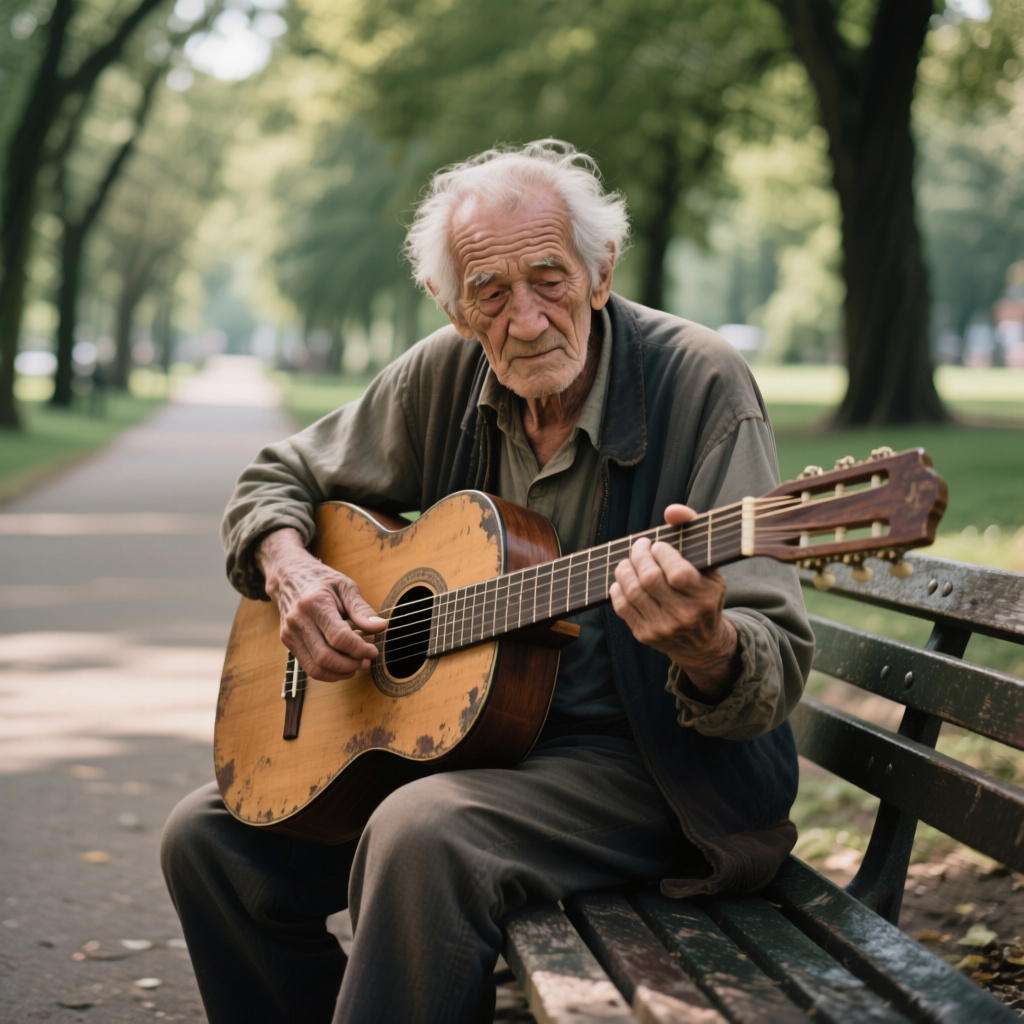}} \\
        \addlinespace[2pt]
        \QualRowLabel{\makecell{Qwen-Lightning\checkpointmark \\ {\tiny\citep{fan2025:qwenlightning}}}} &
        \adjustbox{valign=m}{\includegraphics[width=0.23\textwidth]{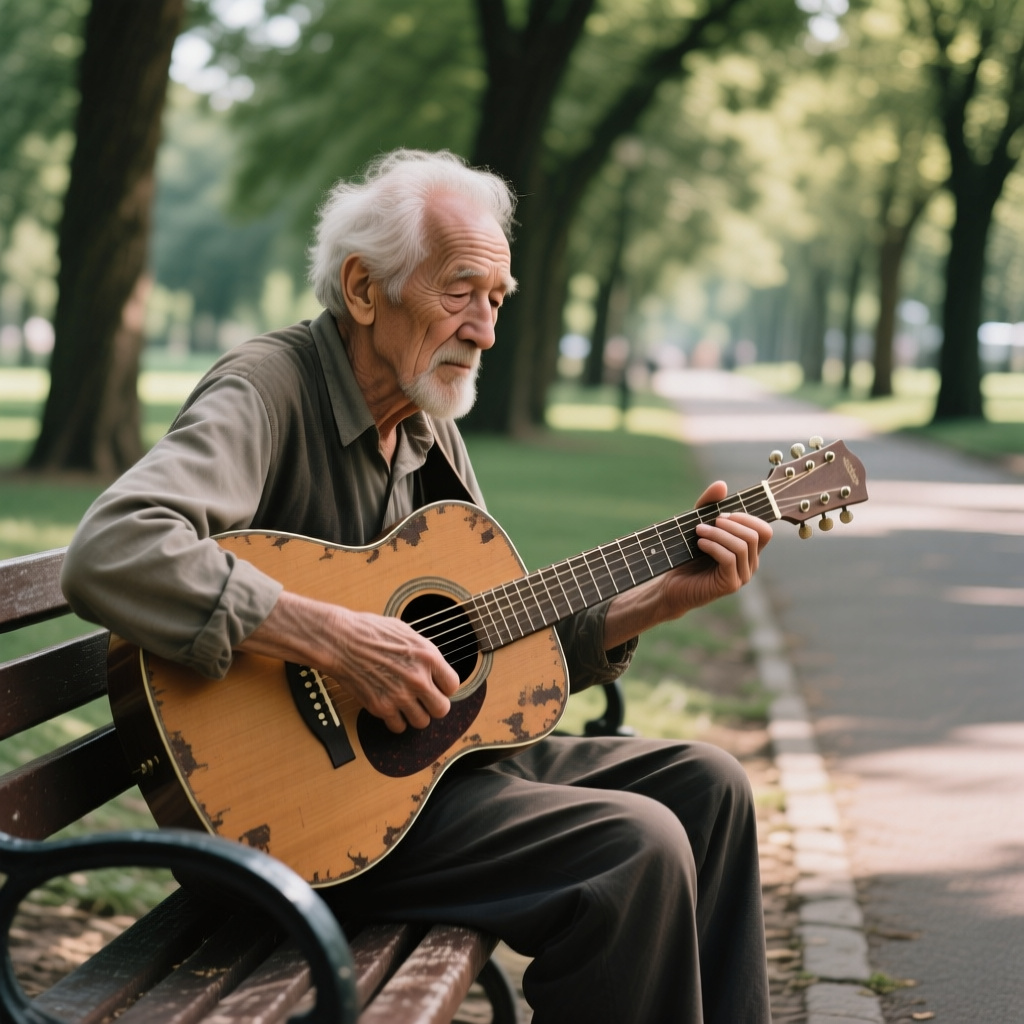}} & \adjustbox{valign=m}{\includegraphics[width=0.23\textwidth]{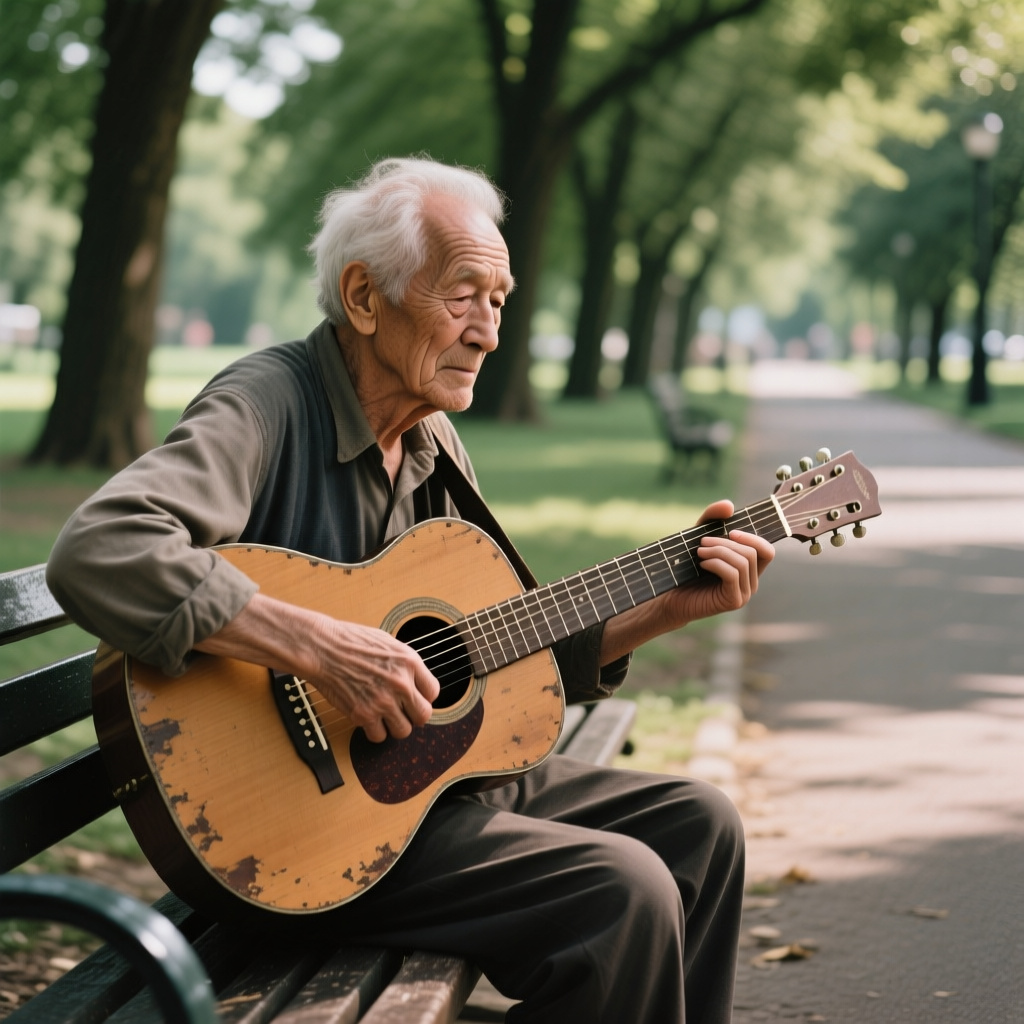}} & \adjustbox{valign=m}{\includegraphics[width=0.23\textwidth]{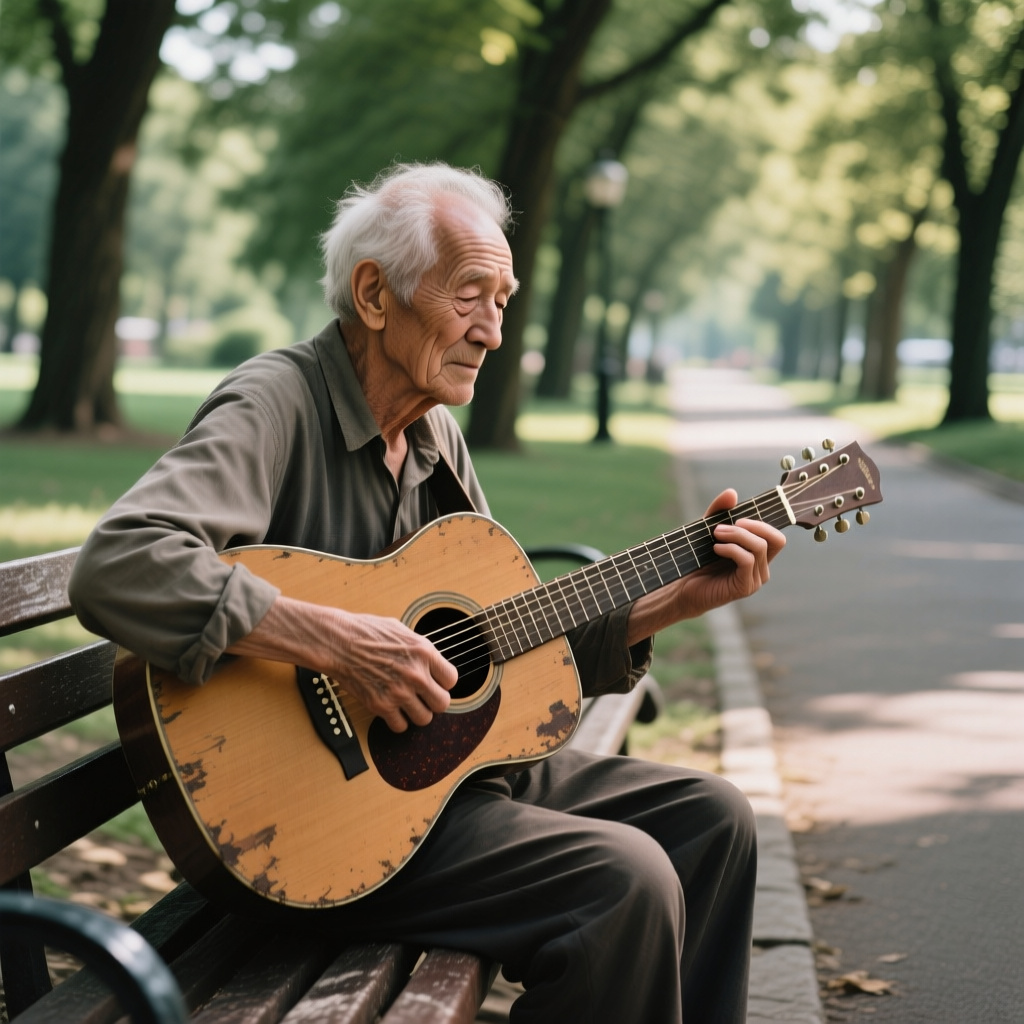}} & \adjustbox{valign=m}{\includegraphics[width=0.23\textwidth]{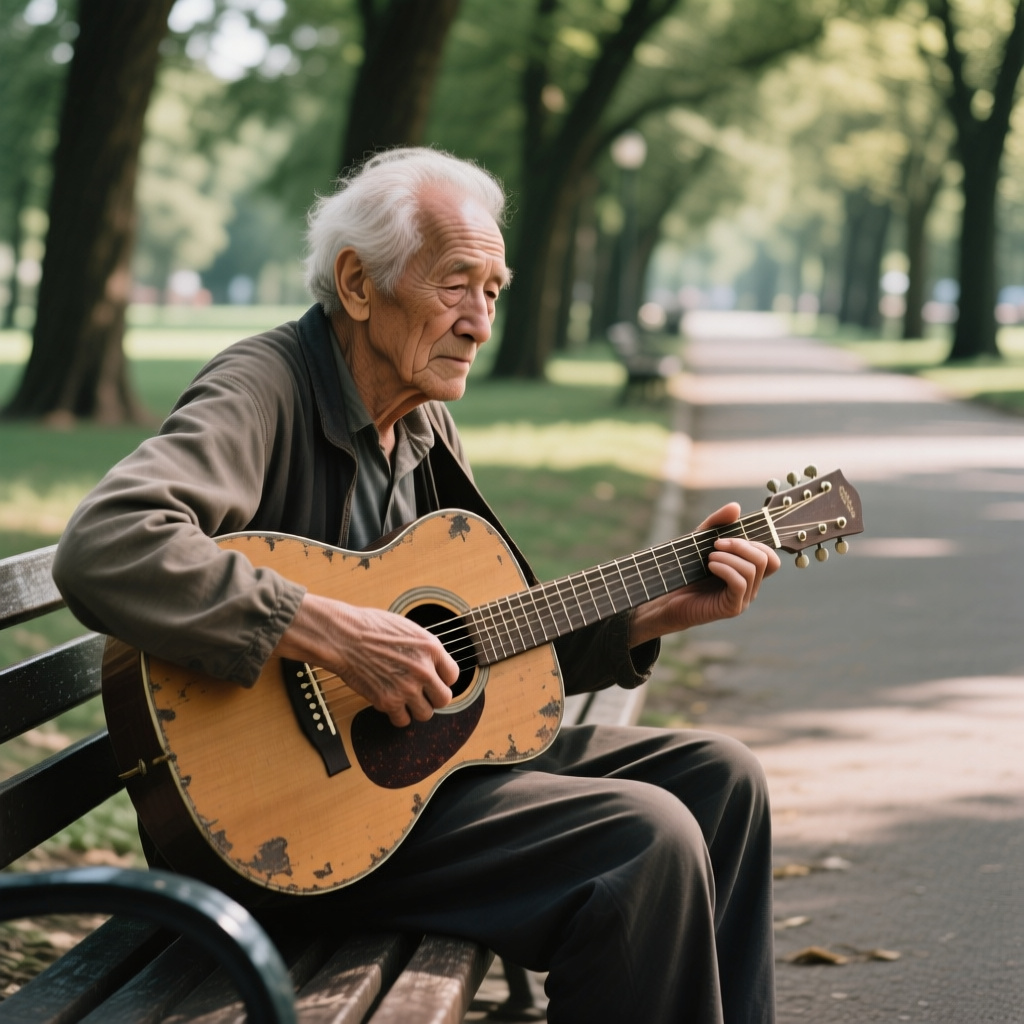}} \\
        \addlinespace[2pt]
        \QualRowLabel{\makecell{{\Oursbf{}} \\ {(Ours)}}} &
        \adjustbox{valign=m}{\includegraphics[width=0.23\textwidth]{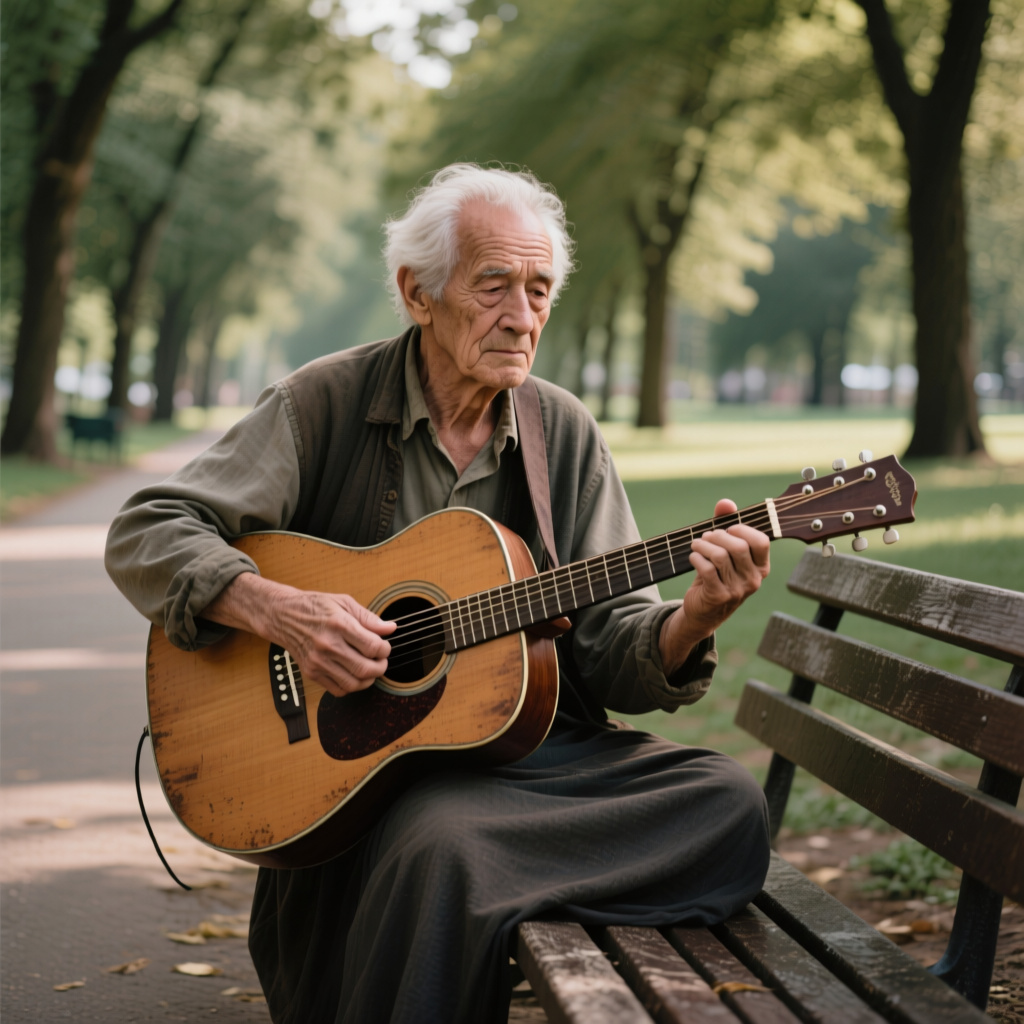}} & \adjustbox{valign=m}{\includegraphics[width=0.23\textwidth]{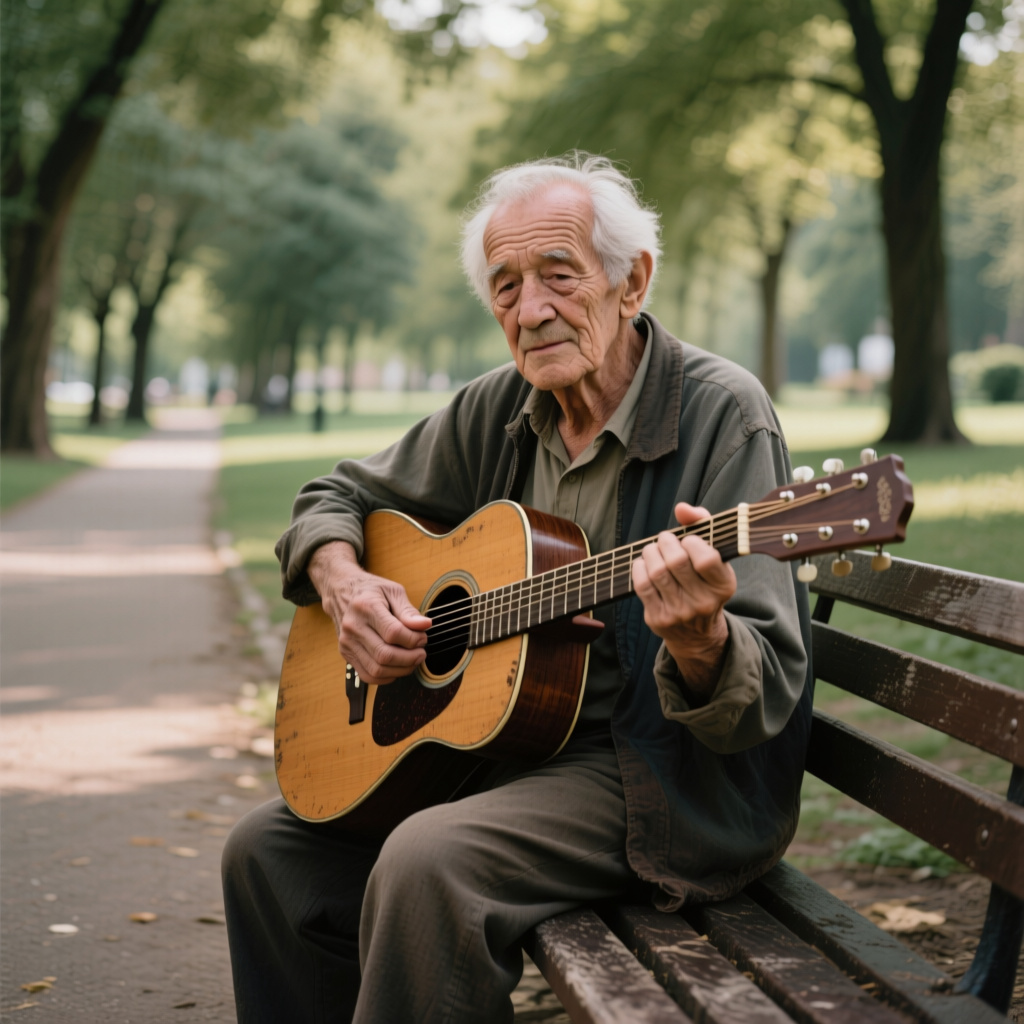}} & \adjustbox{valign=m}{\includegraphics[width=0.23\textwidth]{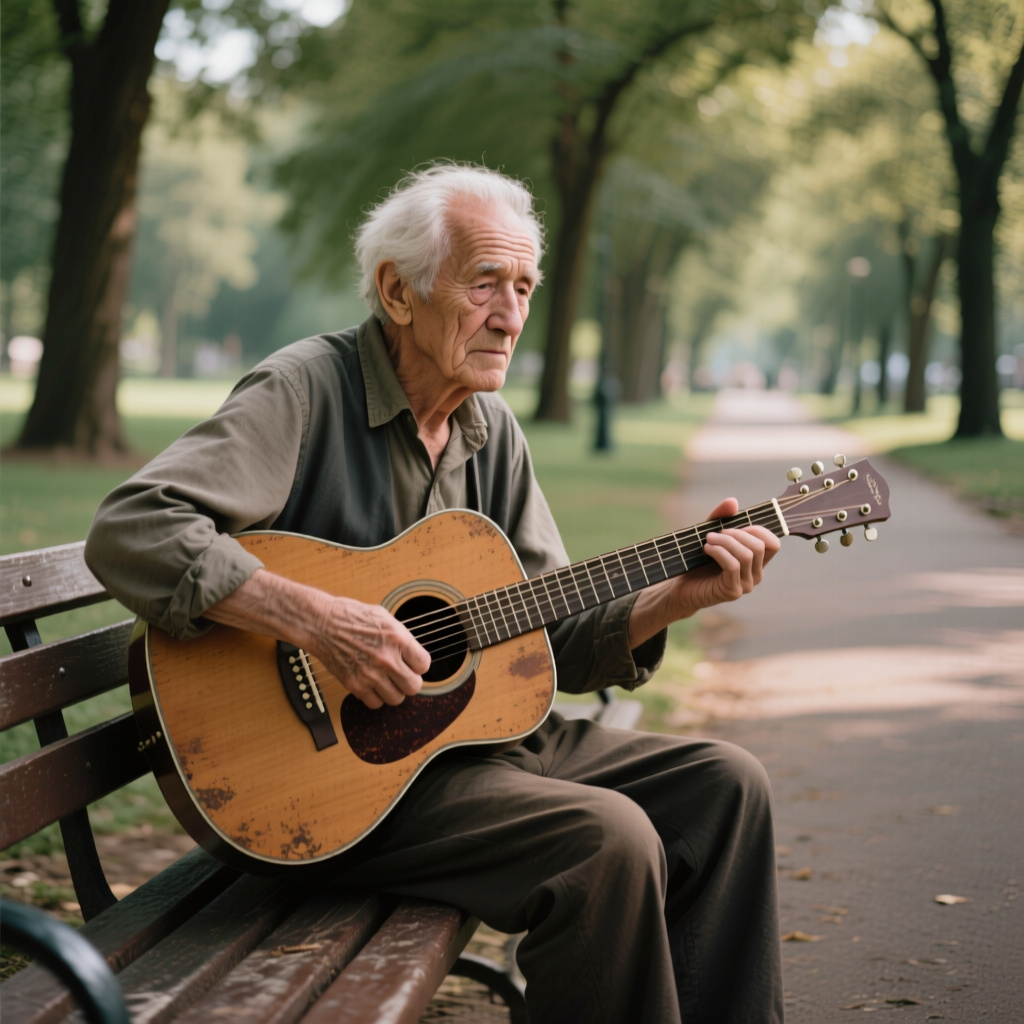}} & \adjustbox{valign=m}{\includegraphics[width=0.23\textwidth]{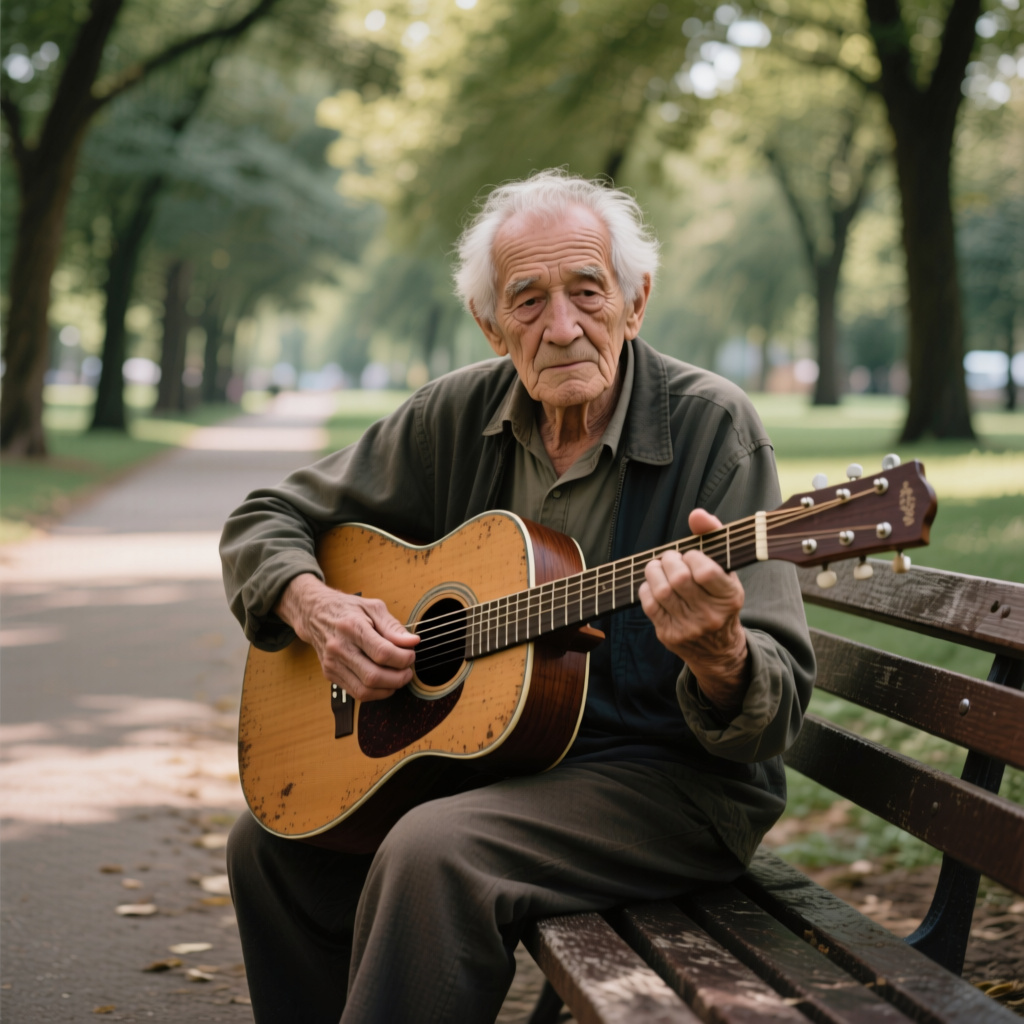}} \\
        \addlinespace[4pt]
        \multicolumn{5}{c}{\parbox{0.94\textwidth}{\centering\tiny\itshape ``An elderly gentleman with a weathered face sits on a park bench, cradling a well-worn wooden guitar in his arms. His fingers are poised over the strings, ready to play a tune. The bench is situated along a path lined with trees that provide a canopy of shade overhead.''}} \\[9pt]
    \end{tabular}
    \end{adjustbox}
\vspace{-3pt}\end{figure}

\begin{figure}[p]\vspace{-3pt}
    \centering
    \captionsetup{font=normalsize}
    \caption{
        \textbf{Additional Qualitative Comparison on Qwen-Image 20B (Continued).}
        \Ours{} more closely matches the diversity of the teacher than DMD-based Qwen-Lightning while preserving visual quality.
        \checkpointmark{} denotes an officially released checkpoint.
    }
    \label{fig:qwen_qual_image_2}
    \setlength{\tabcolsep}{1.5pt}
    \renewcommand{\arraystretch}{1.05}
    \scriptsize
    \begin{adjustbox}{max totalsize={\textwidth}{0.92\textheight}}
    \begin{tabular}{c cccc}
        \QualRowLabel{Teacher} &
        \adjustbox{valign=m}{\includegraphics[width=0.23\textwidth]{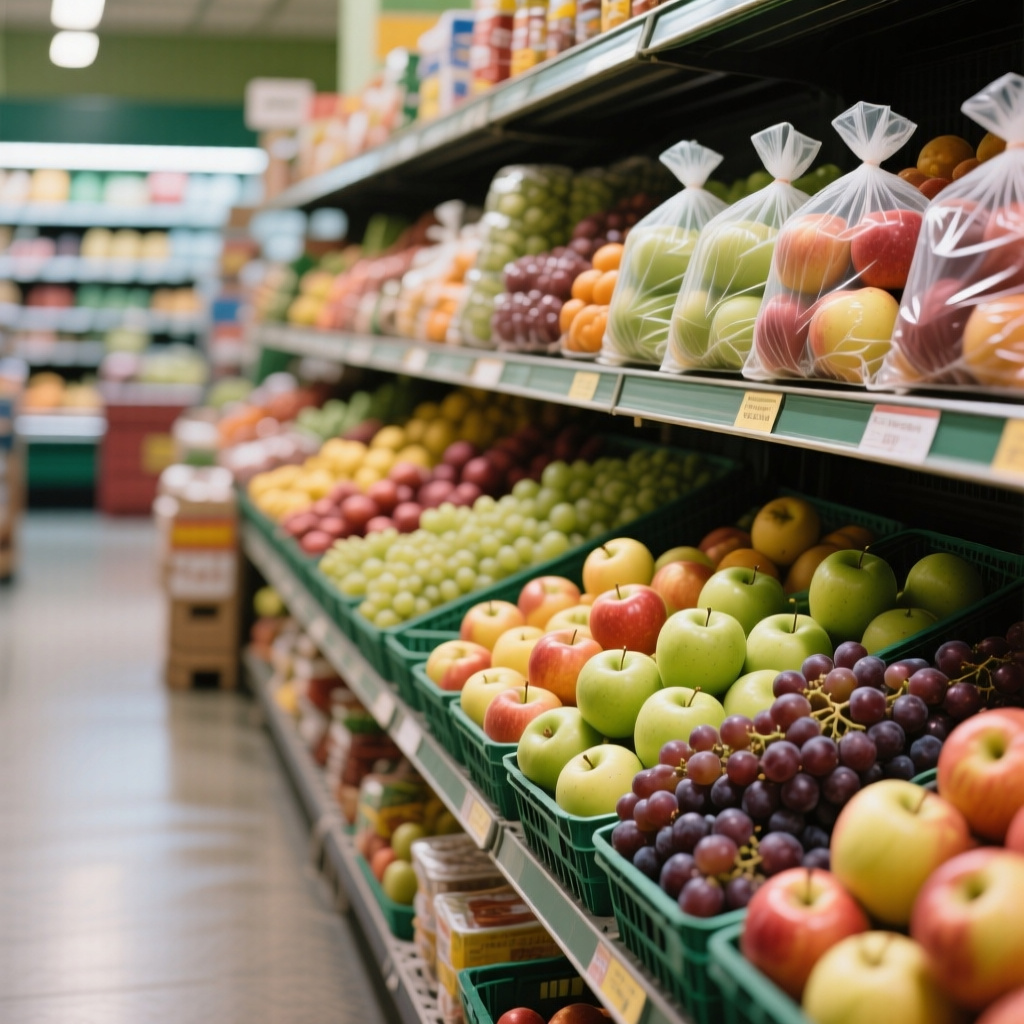}} & \adjustbox{valign=m}{\includegraphics[width=0.23\textwidth]{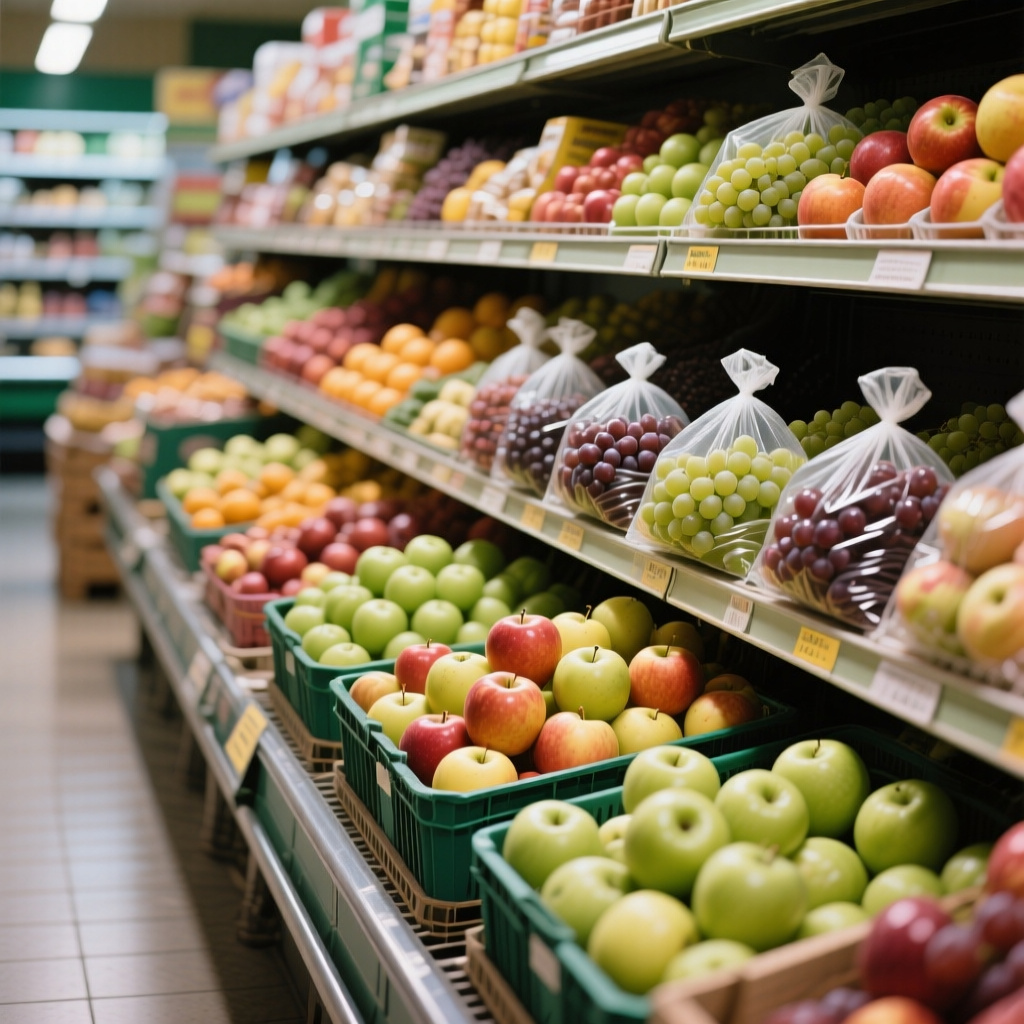}} & \adjustbox{valign=m}{\includegraphics[width=0.23\textwidth]{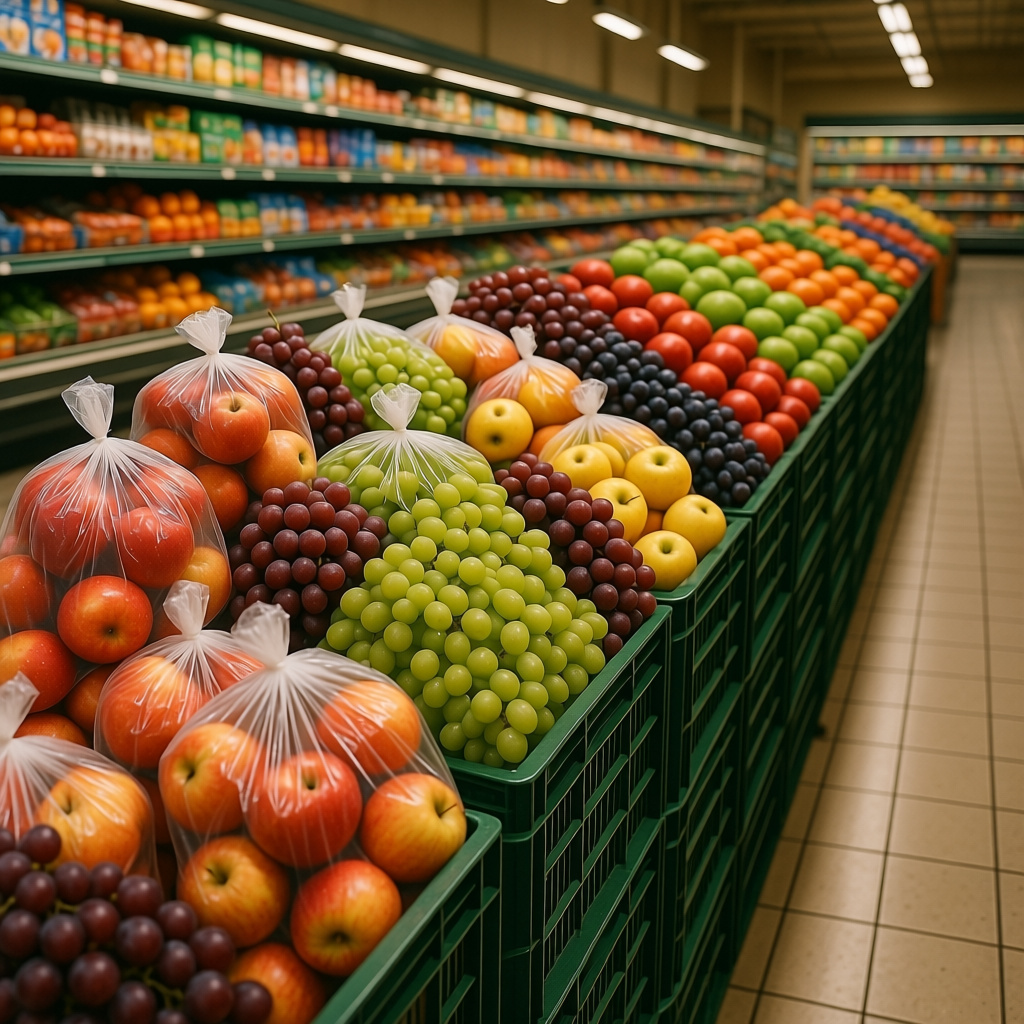}} & \adjustbox{valign=m}{\includegraphics[width=0.23\textwidth]{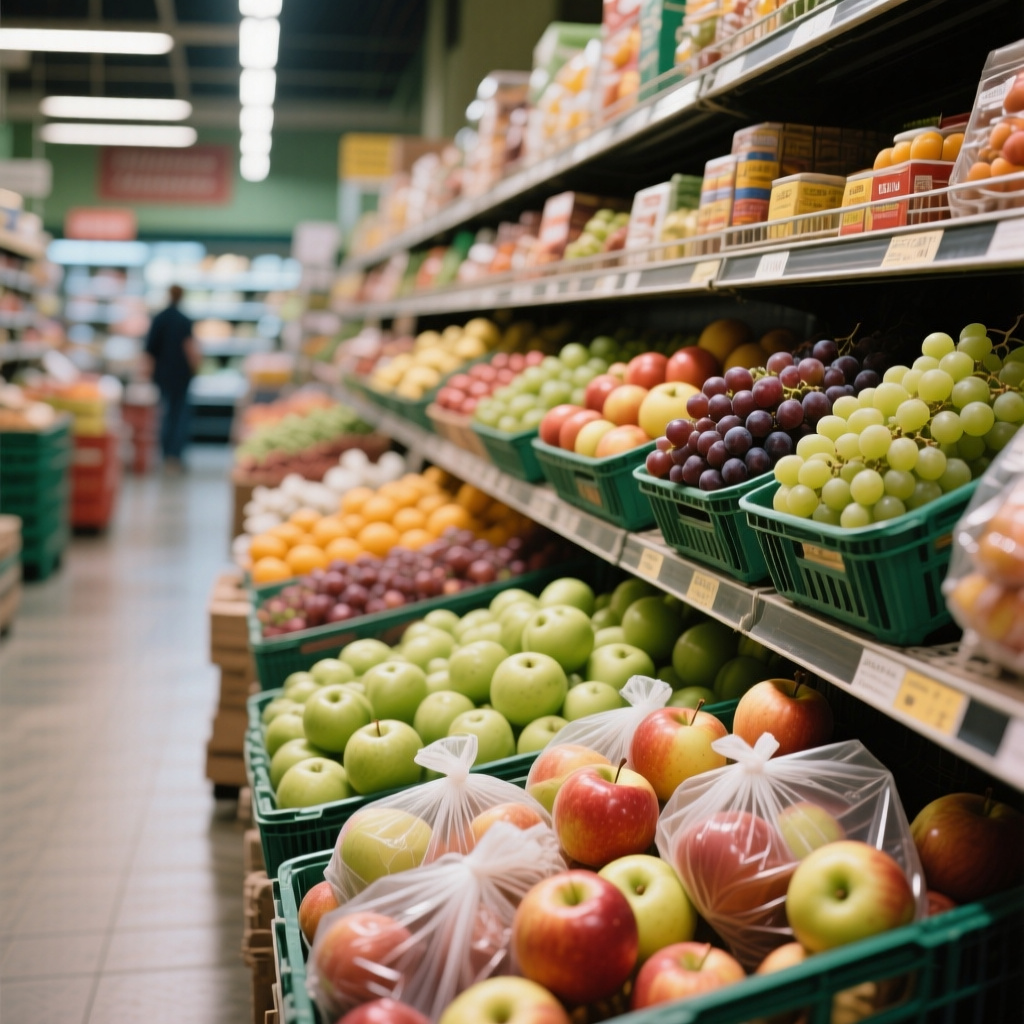}} \\
        \addlinespace[2pt]
        \QualRowLabel{\makecell{Qwen-Lightning\checkpointmark \\ {\tiny\citep{fan2025:qwenlightning}}}} &
        \adjustbox{valign=m}{\includegraphics[width=0.23\textwidth]{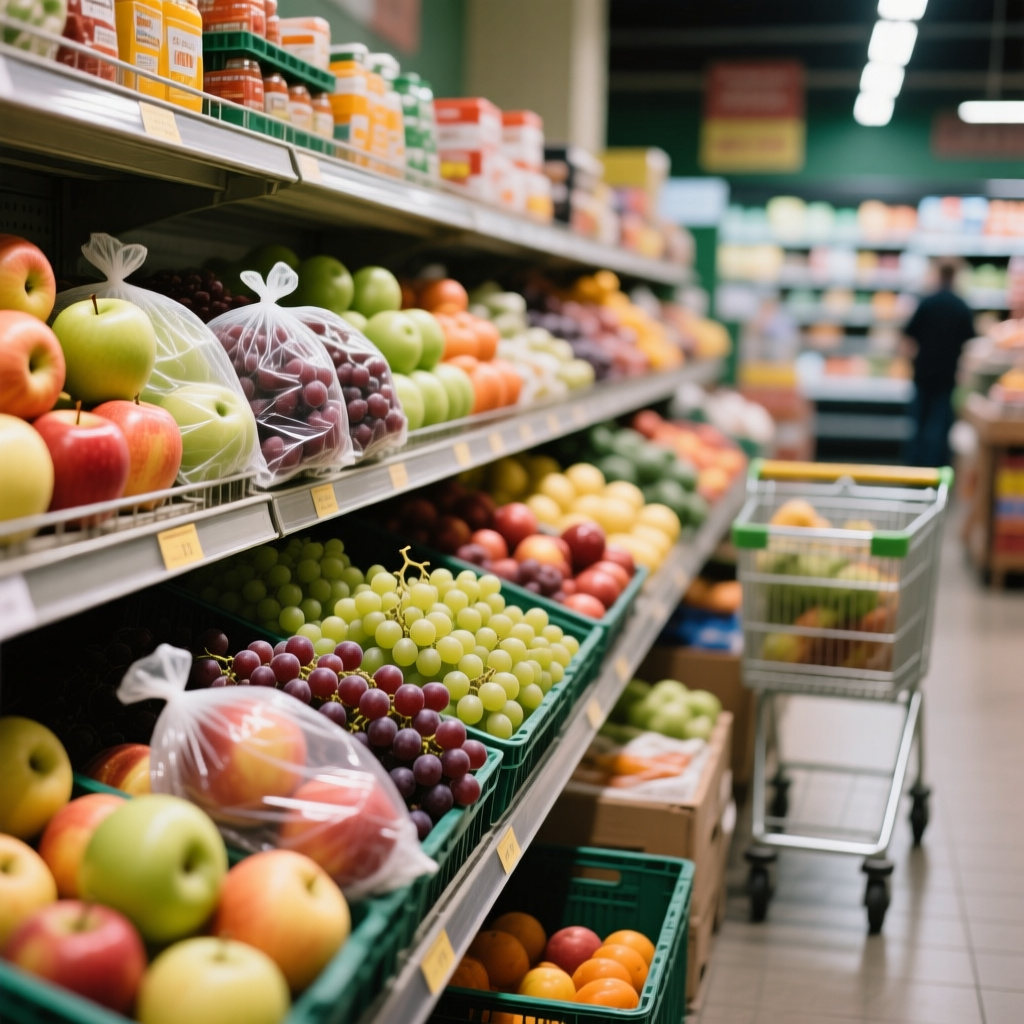}} & \adjustbox{valign=m}{\includegraphics[width=0.23\textwidth]{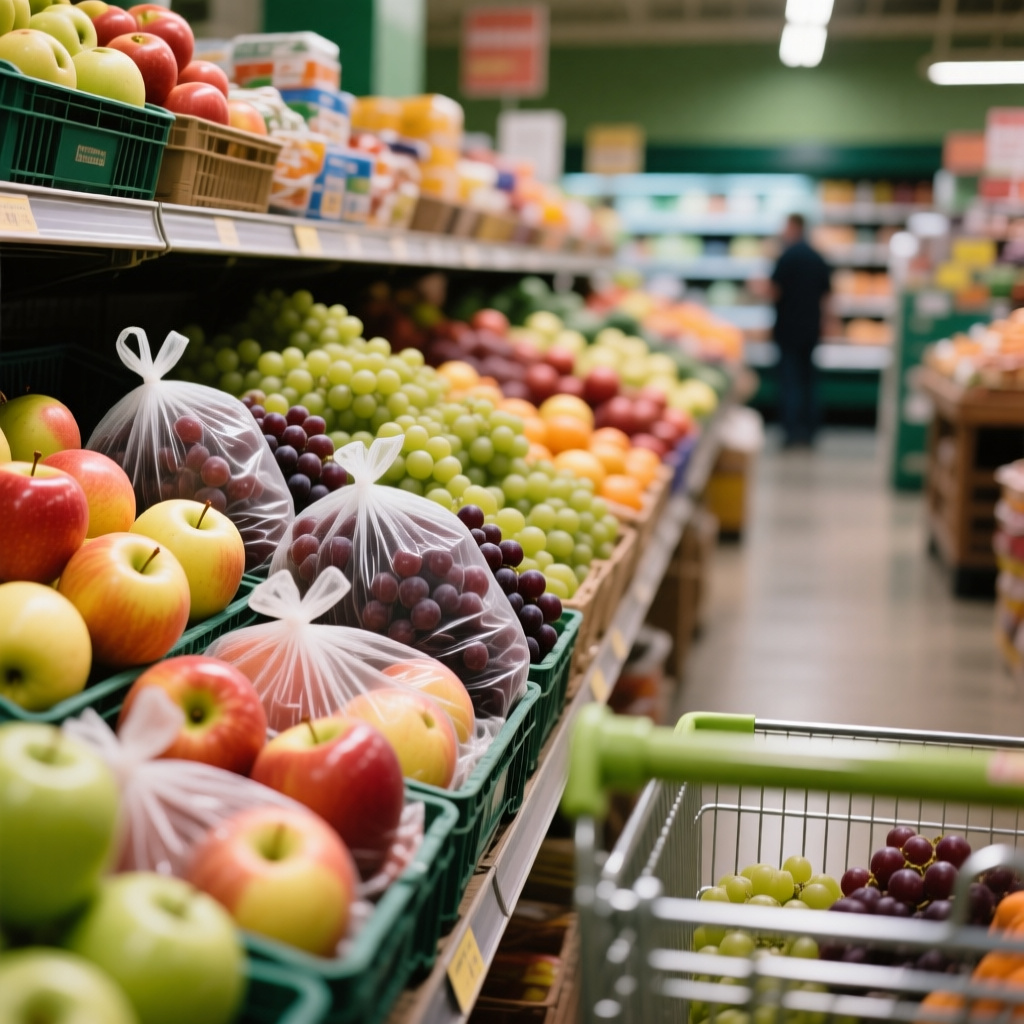}} & \adjustbox{valign=m}{\includegraphics[width=0.23\textwidth]{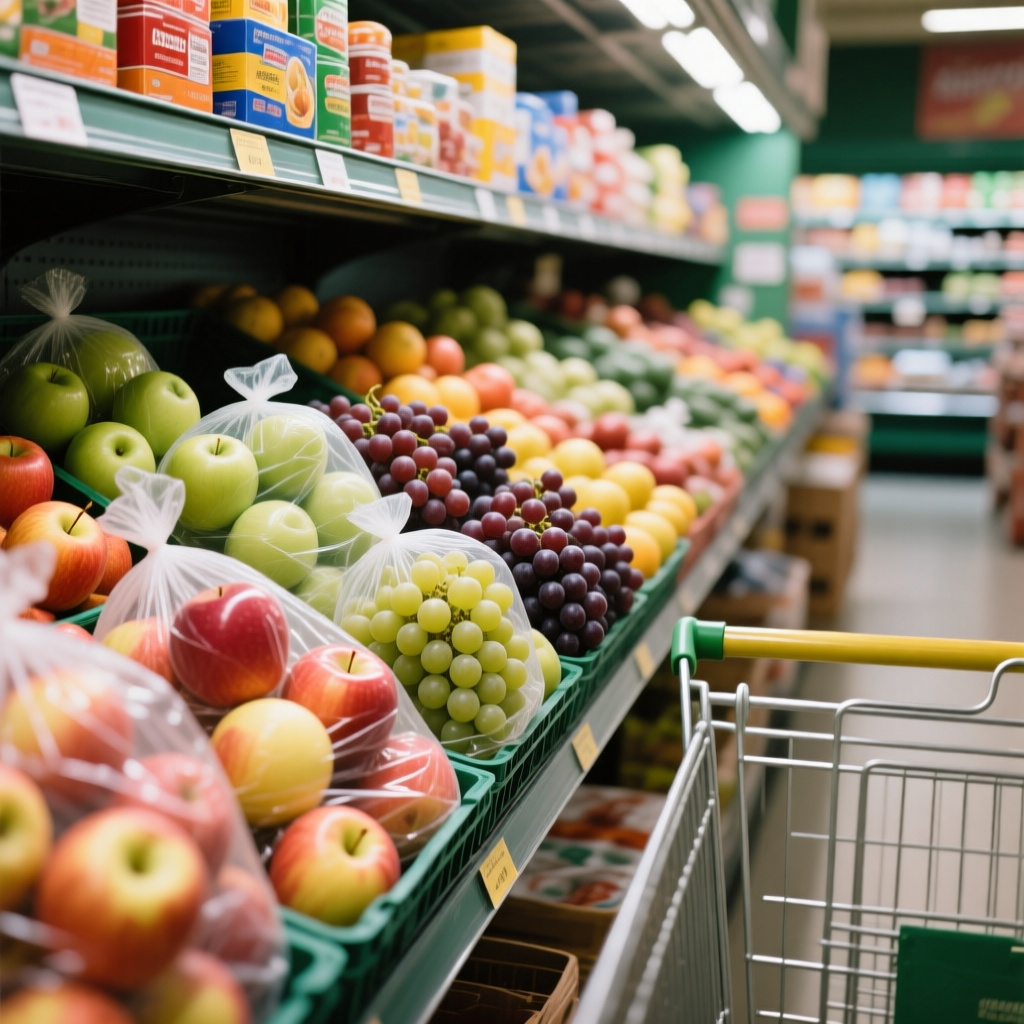}} & \adjustbox{valign=m}{\includegraphics[width=0.23\textwidth]{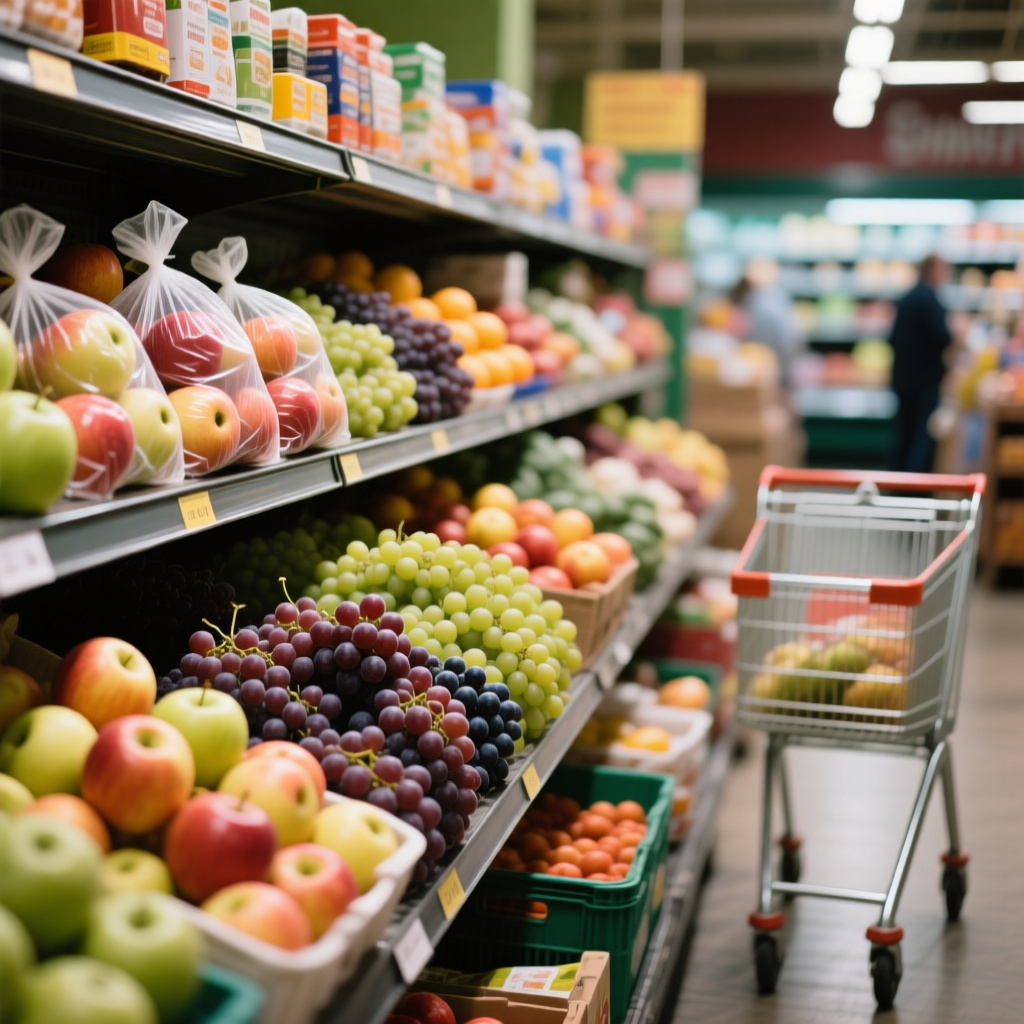}} \\
        \addlinespace[2pt]
        \QualRowLabel{\makecell{{\Oursbf{}} \\ {(Ours)}}} &
        \adjustbox{valign=m}{\includegraphics[width=0.23\textwidth]{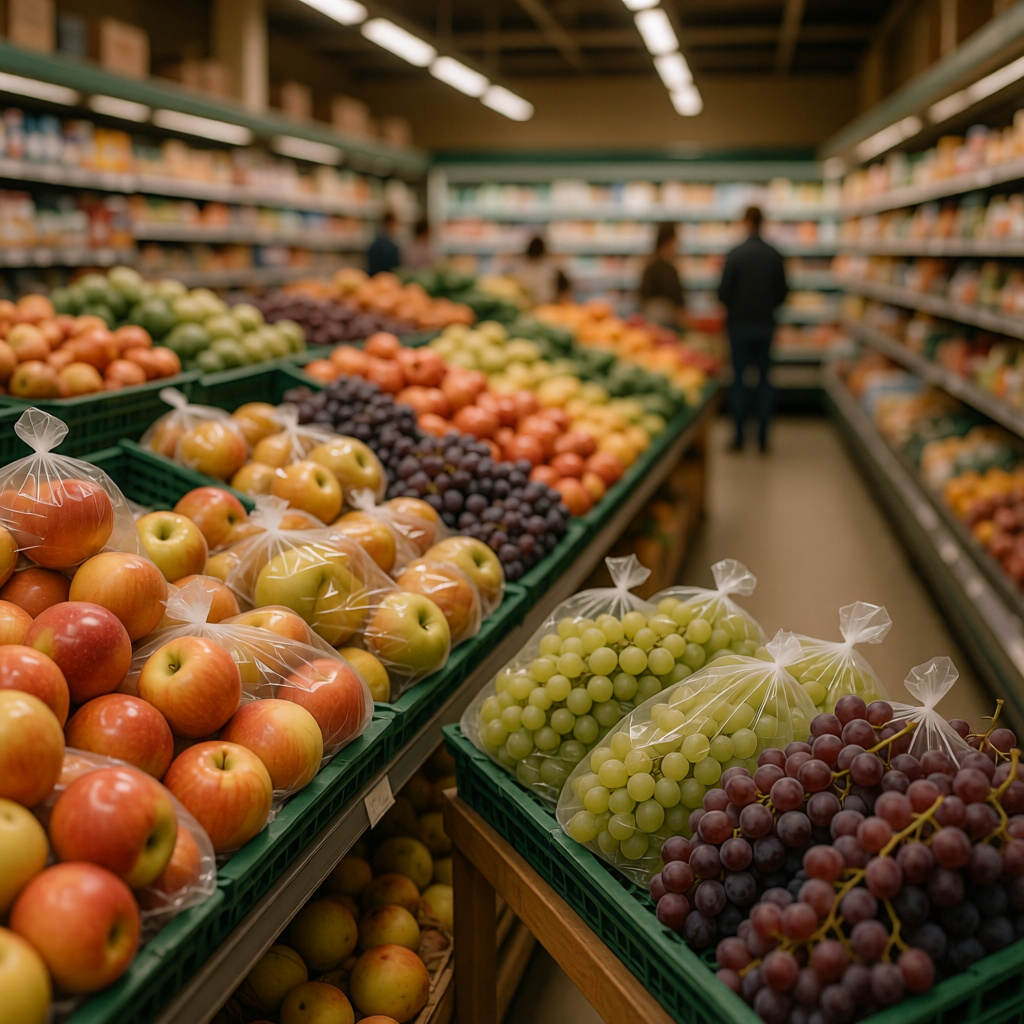}} & \adjustbox{valign=m}{\includegraphics[width=0.23\textwidth]{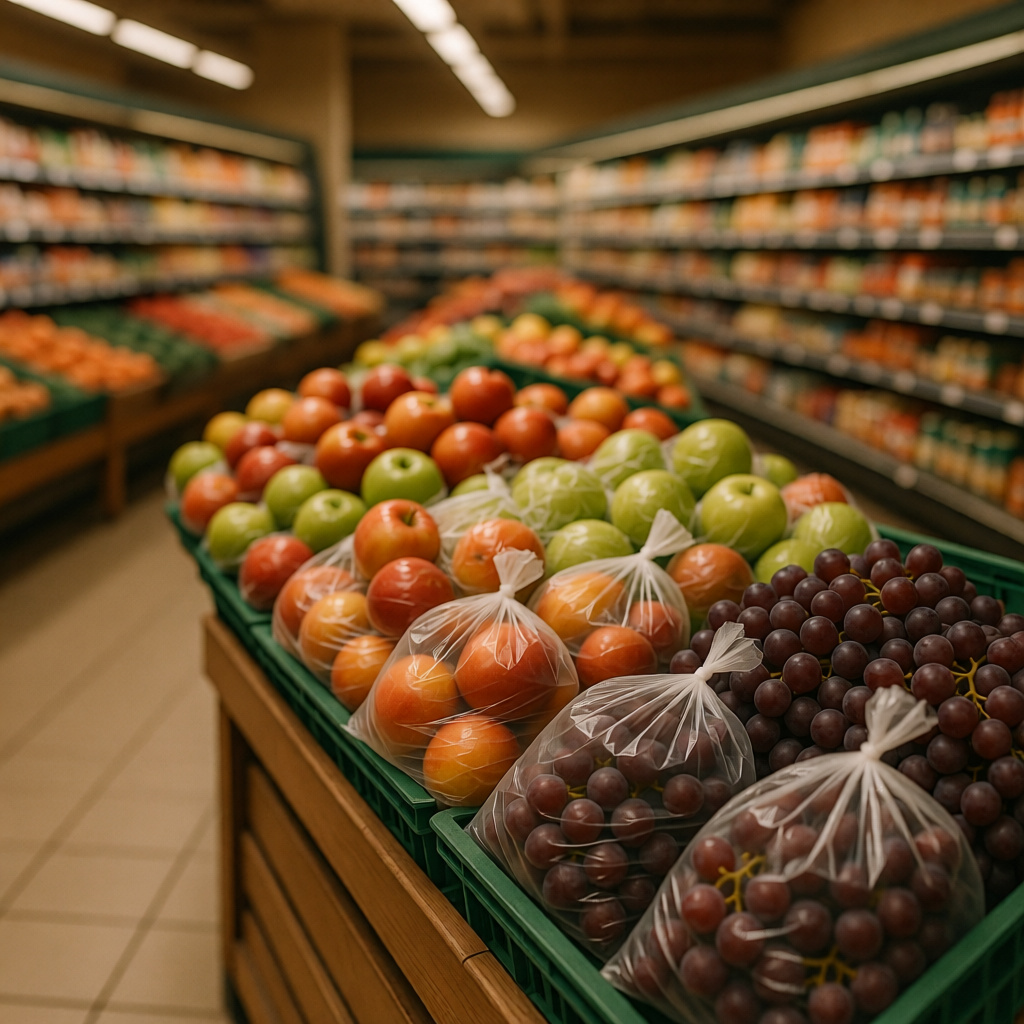}} & \adjustbox{valign=m}{\includegraphics[width=0.23\textwidth]{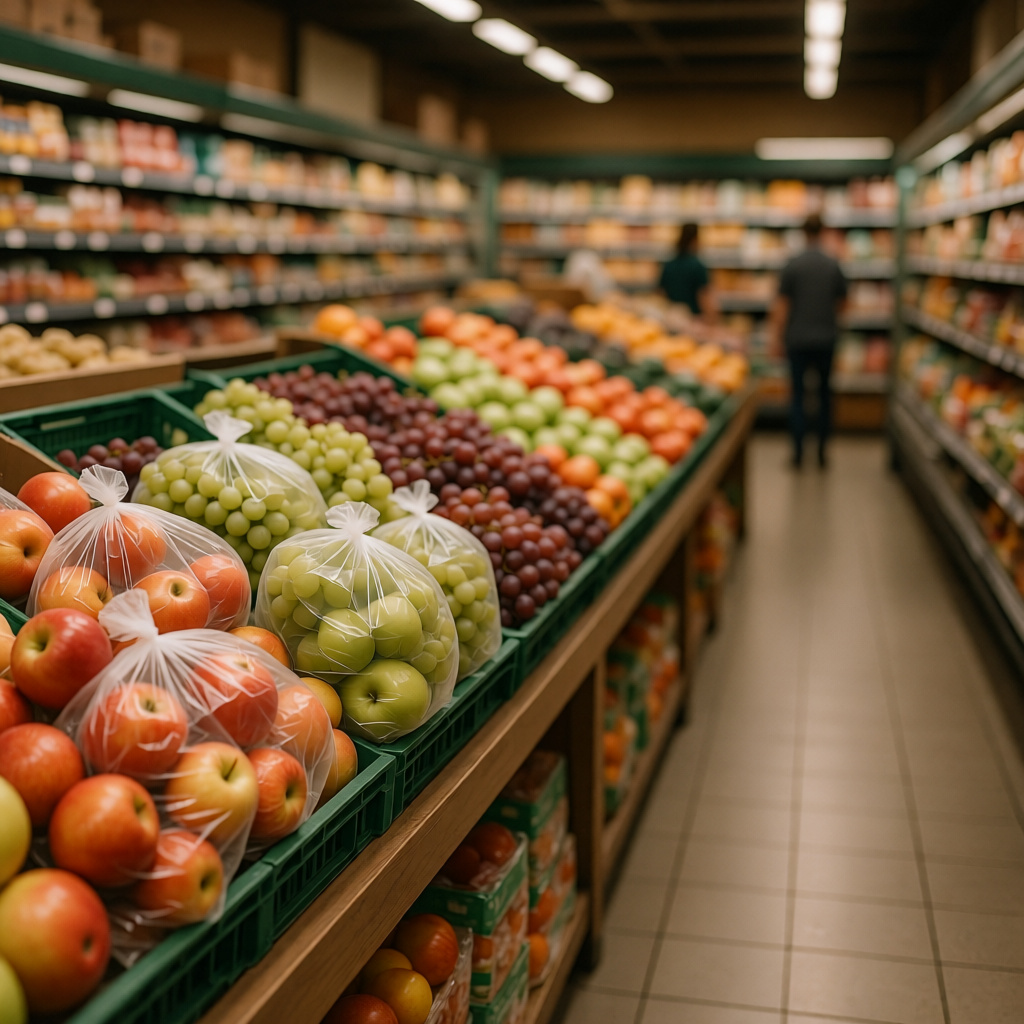}} & \adjustbox{valign=m}{\includegraphics[width=0.23\textwidth]{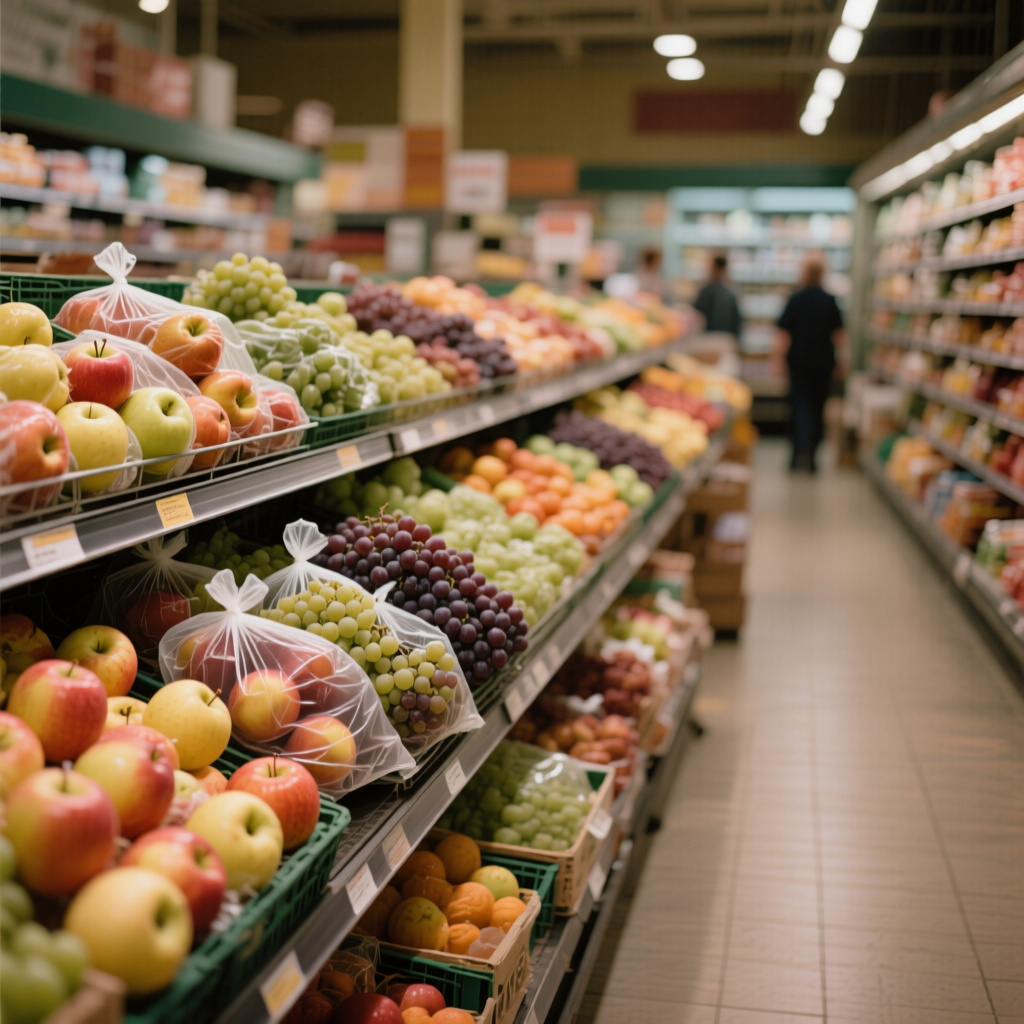}} \\
        \addlinespace[4pt]
        \multicolumn{5}{c}{\parbox{0.94\textwidth}{\centering\tiny\itshape ``Rows of fresh produce line the interior of a bustling grocery store. Bags filled with crisp apples and plump grapes are neatly arranged on the shelves, inviting shoppers to add them to their carts. The fruits' vibrant colors stand out against the backdrop of other grocery items in the background.''}} \\[9pt]
        \QualRowLabel{Teacher} &
        \adjustbox{valign=m}{\includegraphics[width=0.23\textwidth]{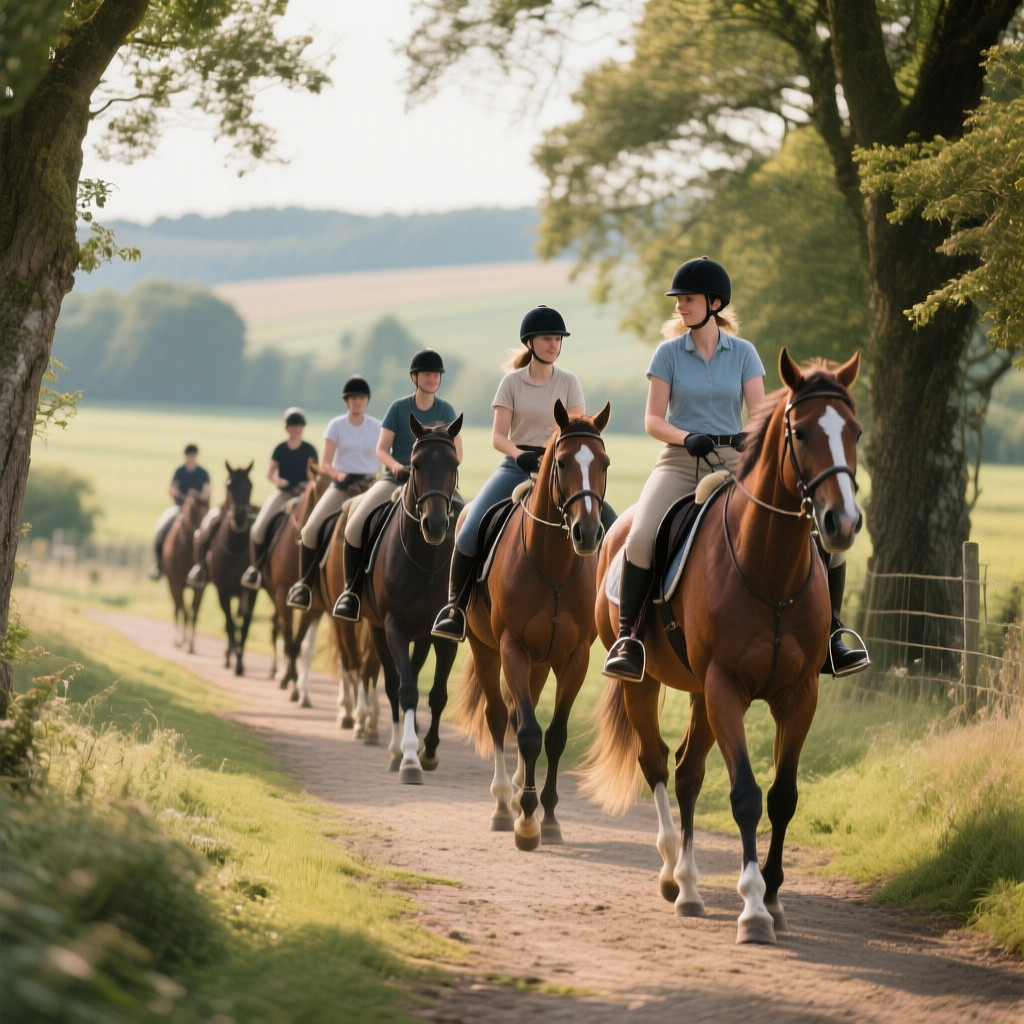}} & \adjustbox{valign=m}{\includegraphics[width=0.23\textwidth]{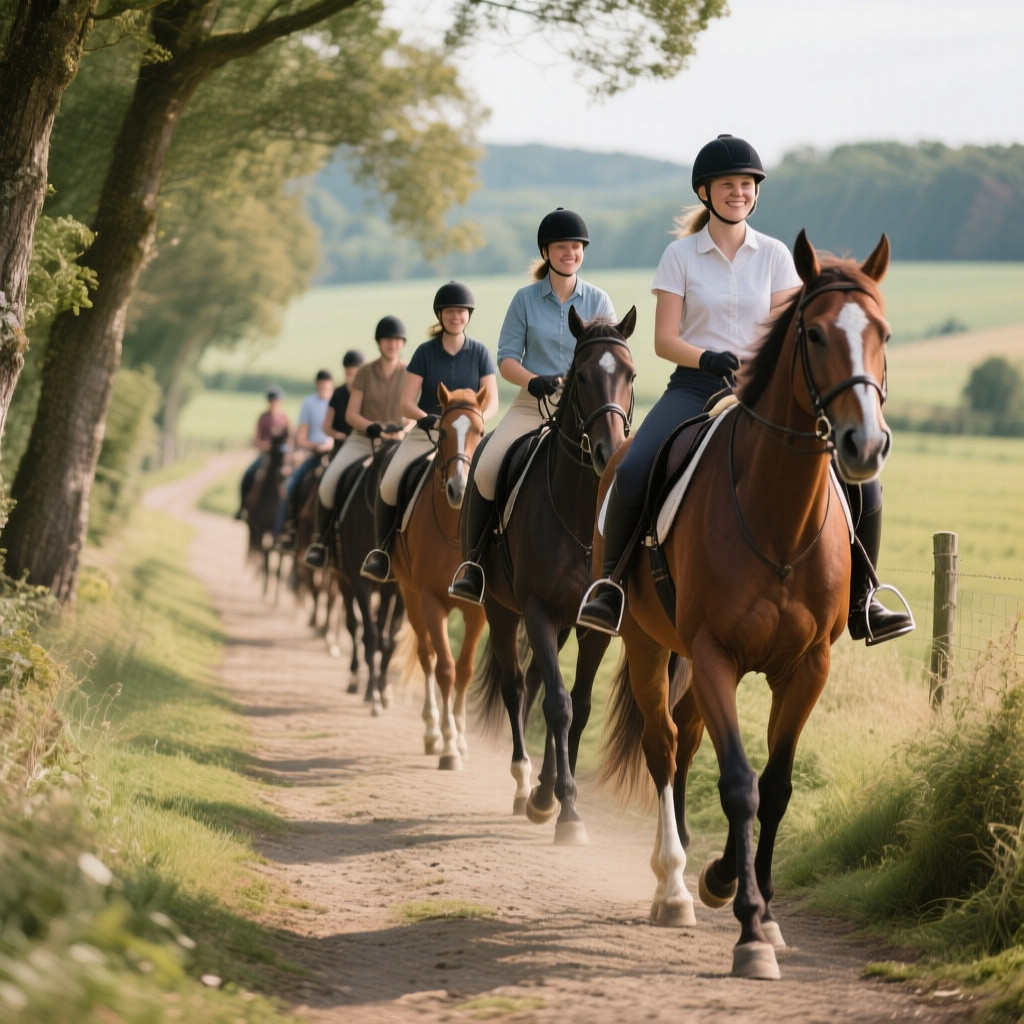}} & \adjustbox{valign=m}{\includegraphics[width=0.23\textwidth]{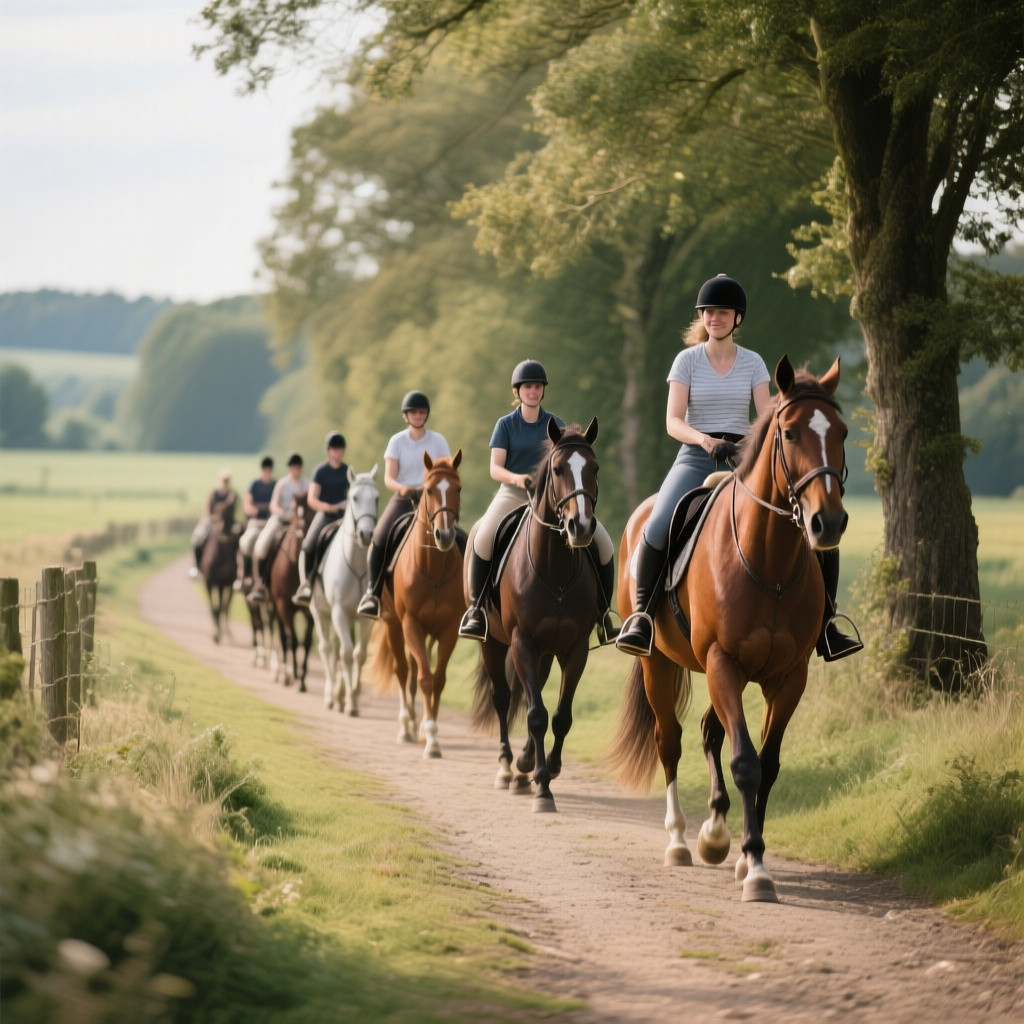}} & \adjustbox{valign=m}{\includegraphics[width=0.23\textwidth]{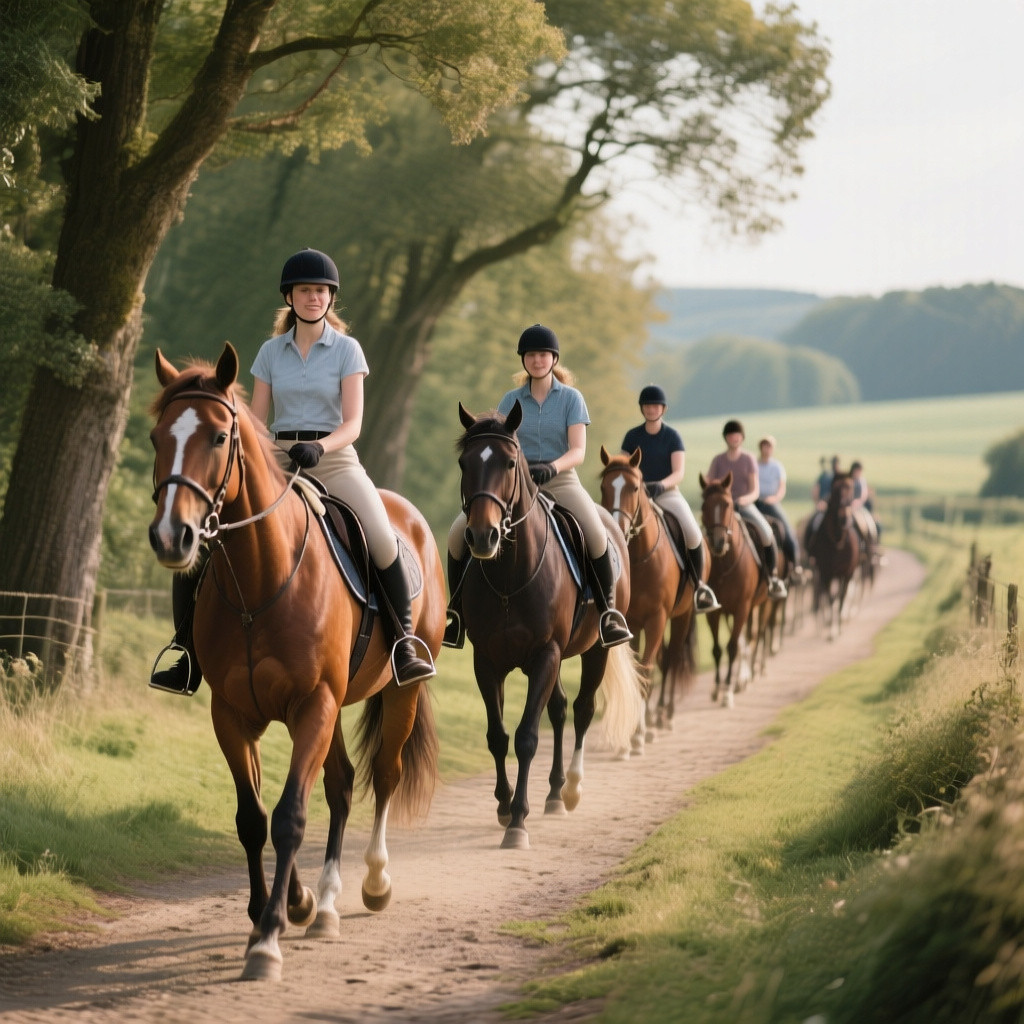}} \\
        \addlinespace[2pt]
        \QualRowLabel{\makecell{Qwen-Lightning\checkpointmark \\ {\tiny\citep{fan2025:qwenlightning}}}} &
        \adjustbox{valign=m}{\includegraphics[width=0.23\textwidth]{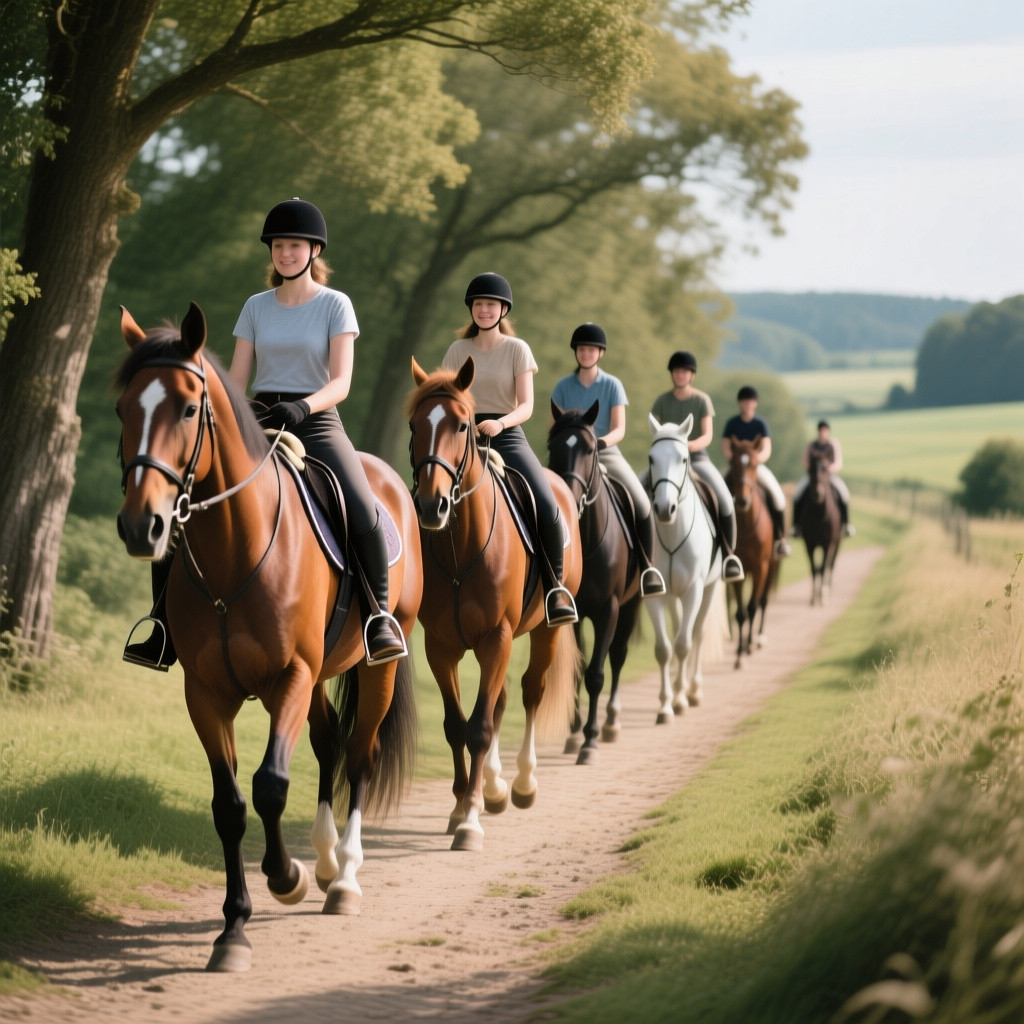}} & \adjustbox{valign=m}{\includegraphics[width=0.23\textwidth]{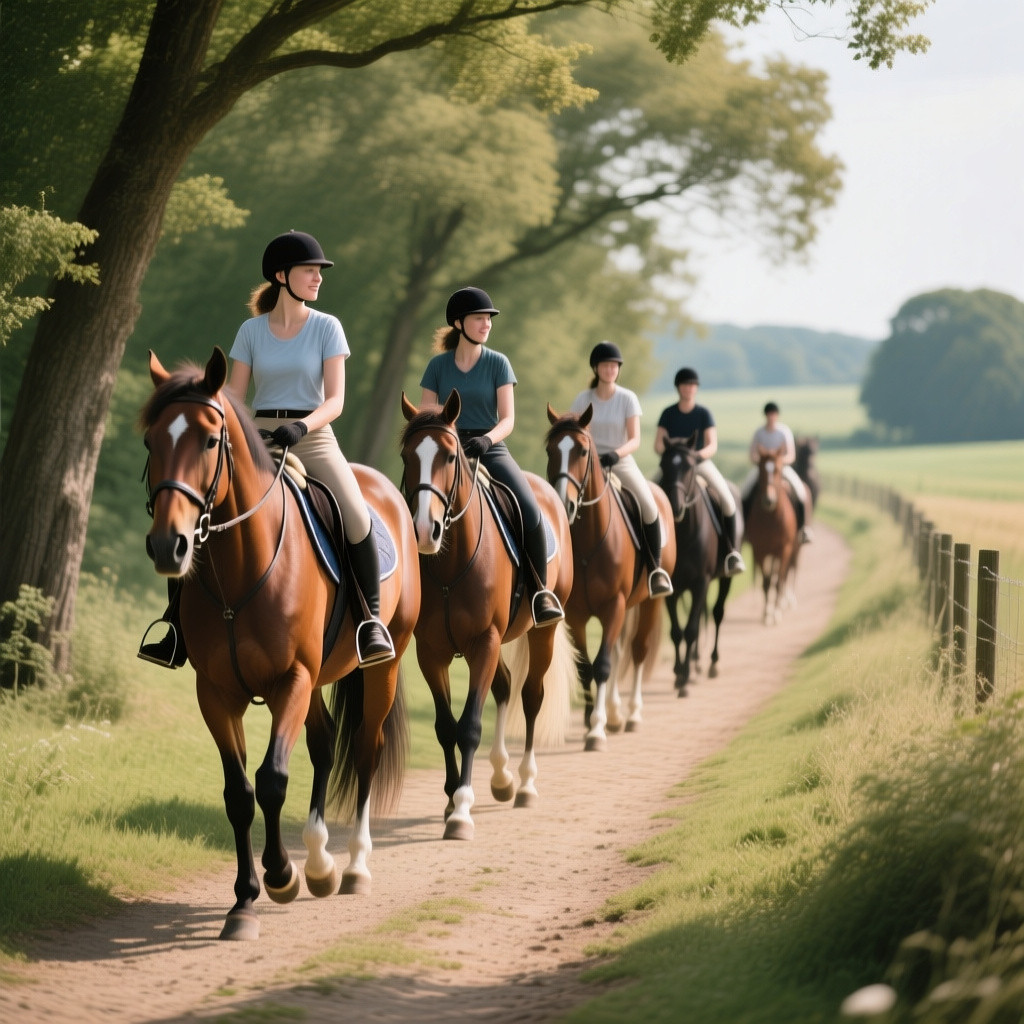}} & \adjustbox{valign=m}{\includegraphics[width=0.23\textwidth]{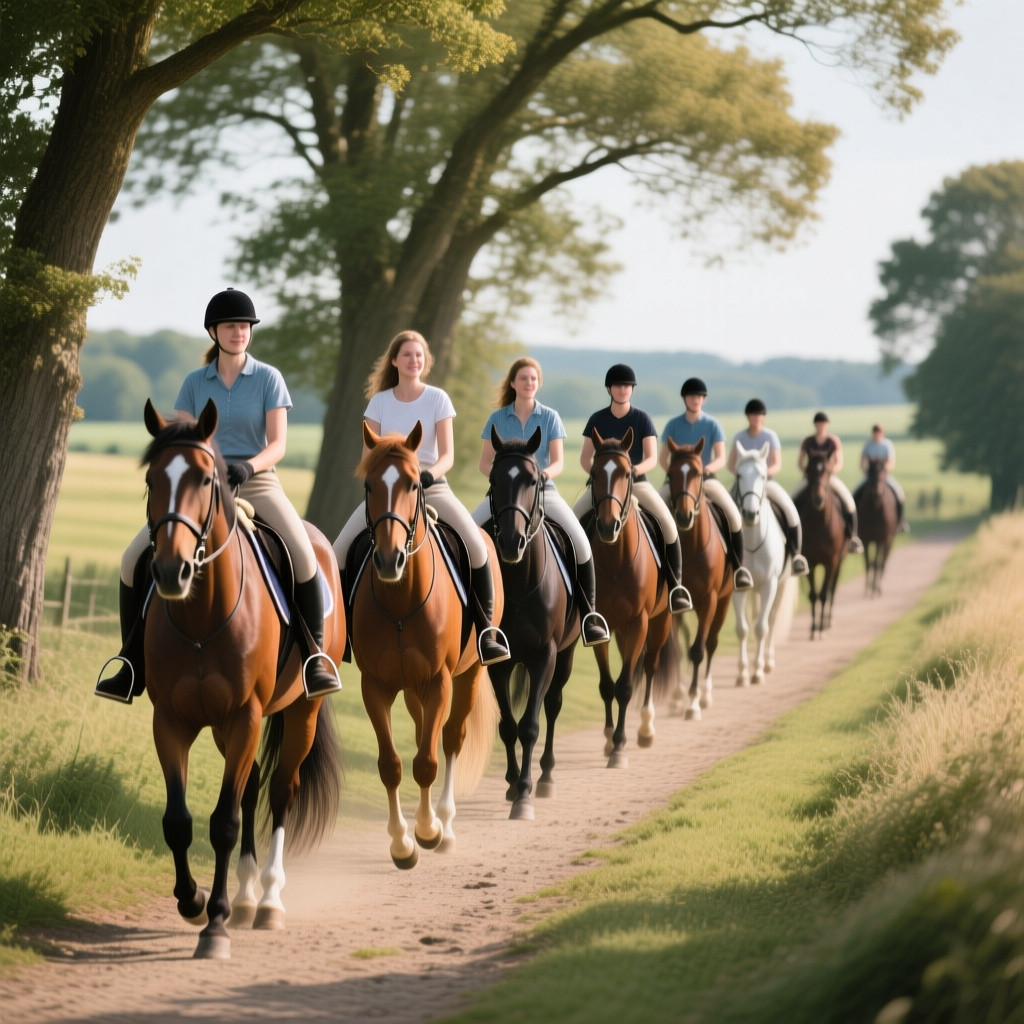}} & \adjustbox{valign=m}{\includegraphics[width=0.23\textwidth]{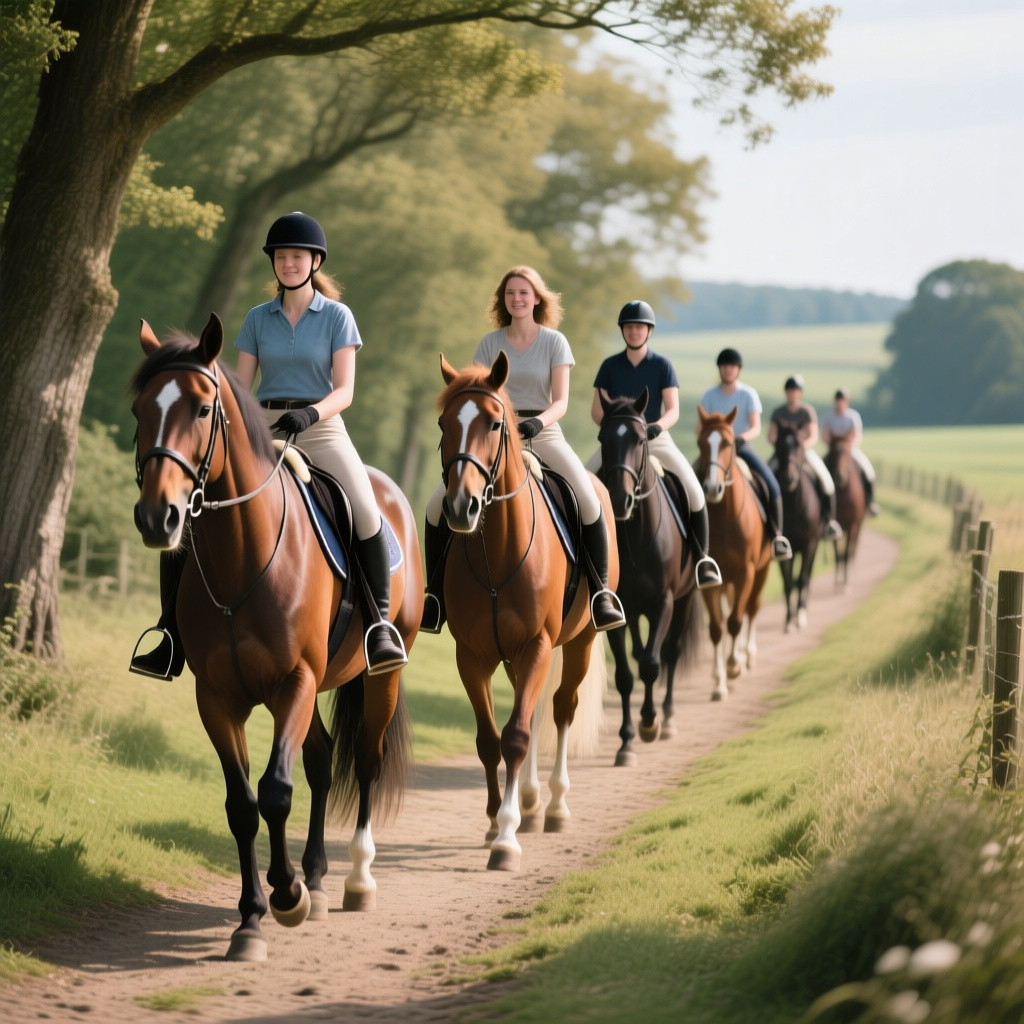}} \\
        \addlinespace[2pt]
        \QualRowLabel{\makecell{{\Oursbf{}} \\ {(Ours)}}} &
        \adjustbox{valign=m}{\includegraphics[width=0.23\textwidth]{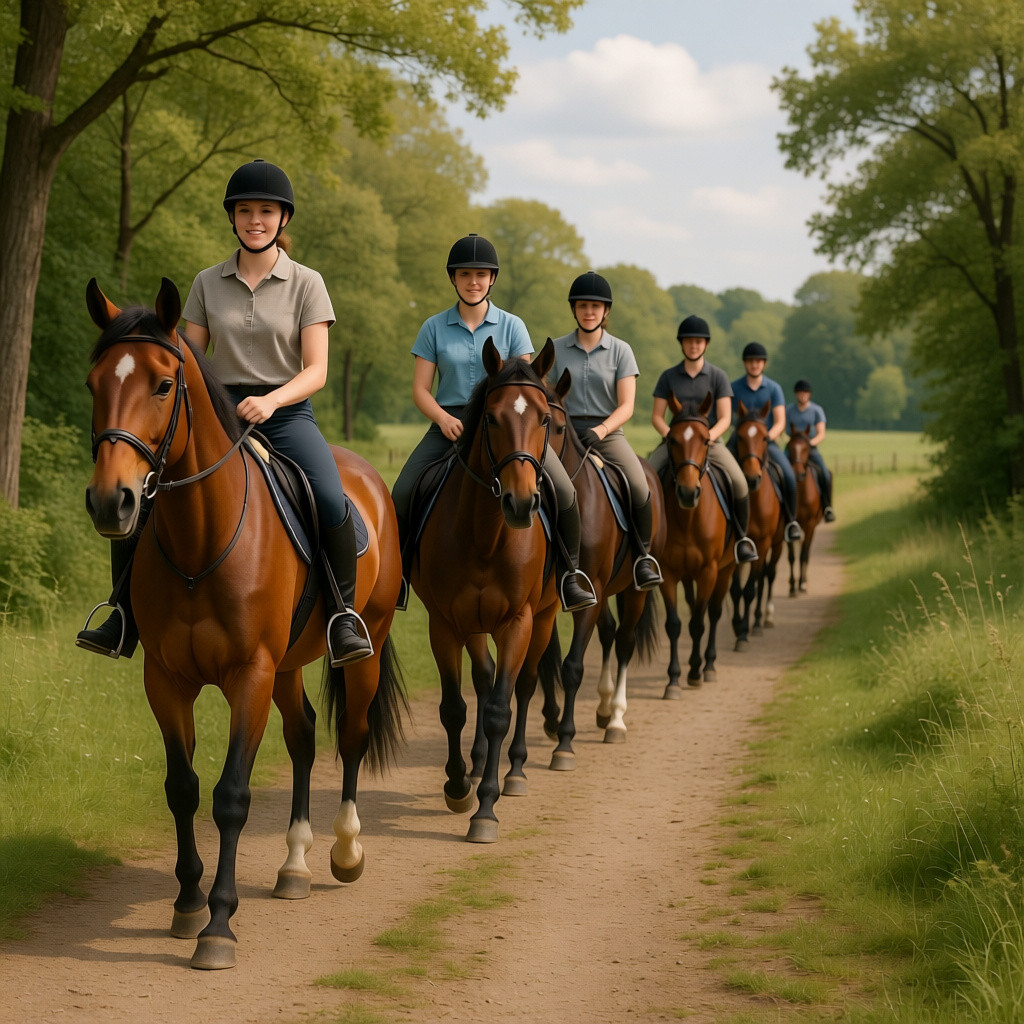}} & \adjustbox{valign=m}{\includegraphics[width=0.23\textwidth]{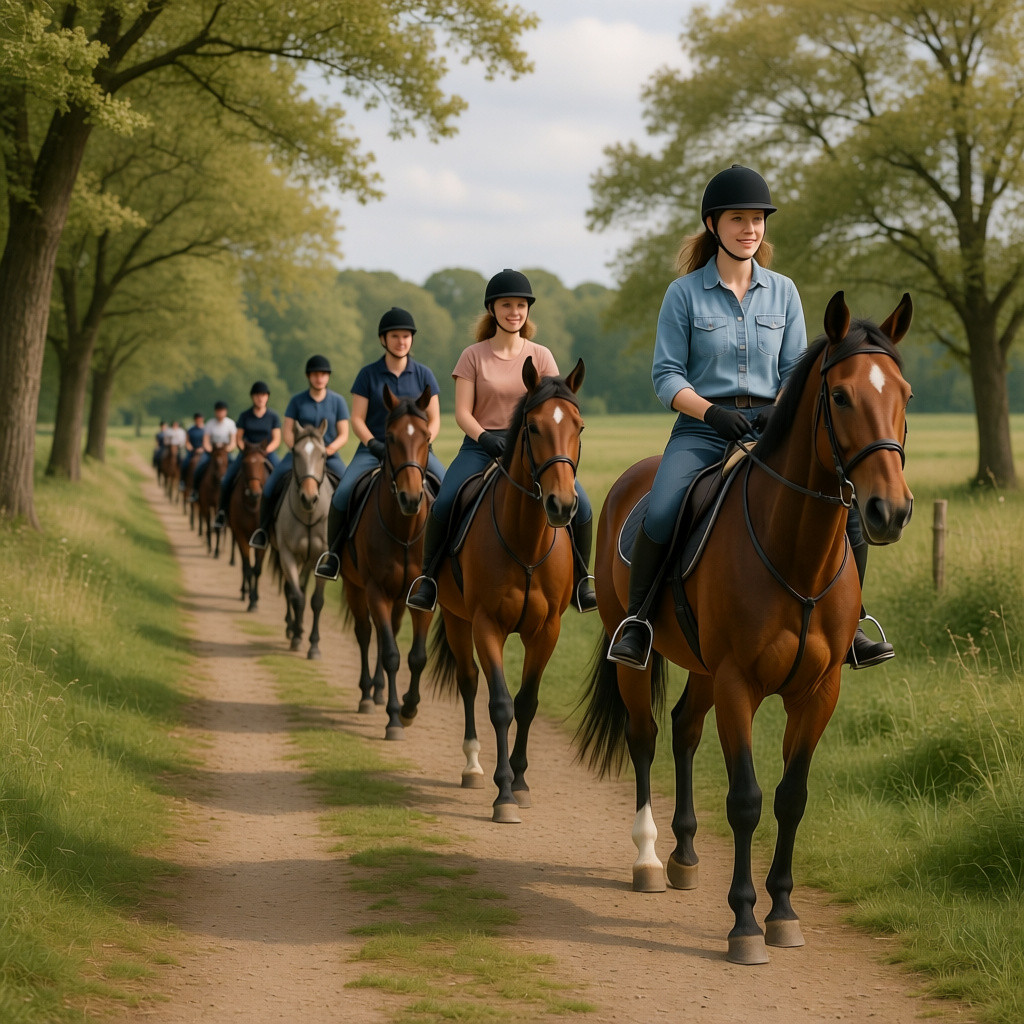}} & \adjustbox{valign=m}{\includegraphics[width=0.23\textwidth]{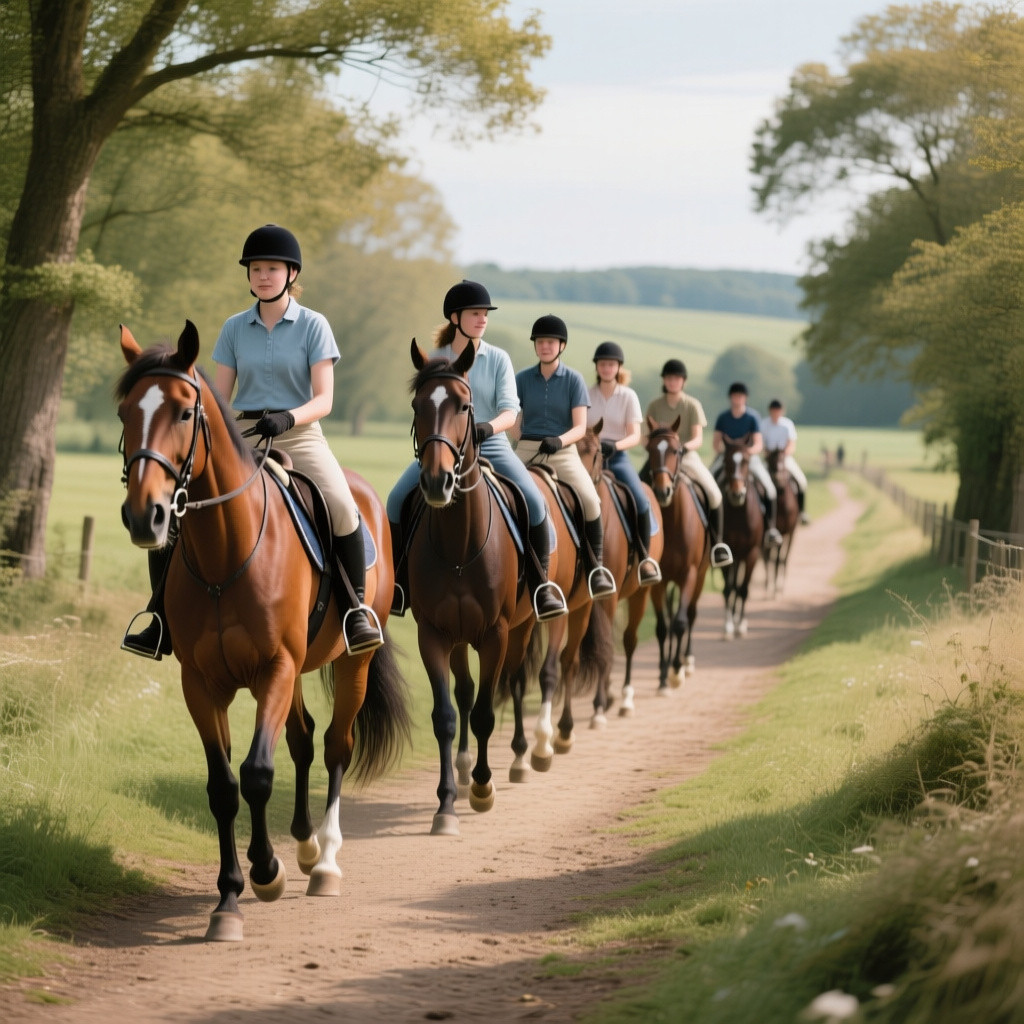}} & \adjustbox{valign=m}{\includegraphics[width=0.23\textwidth]{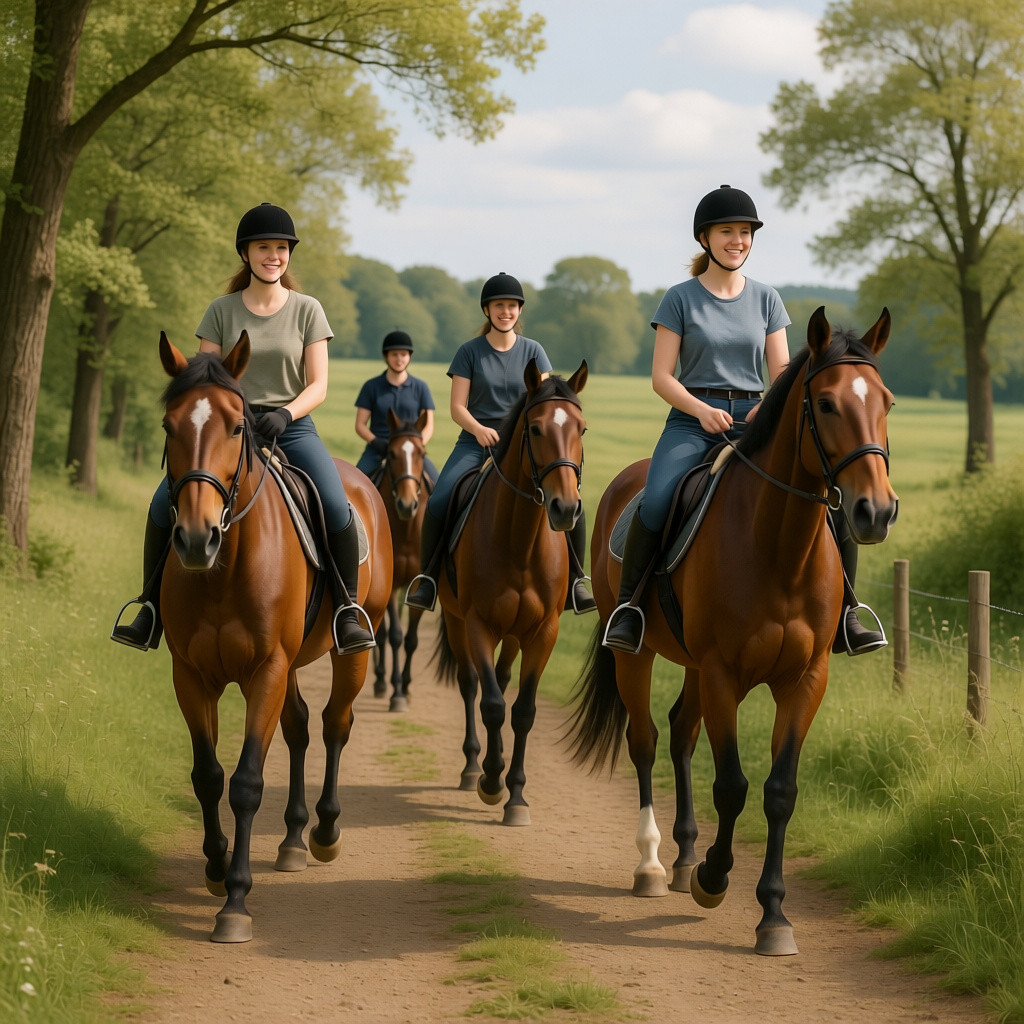}} \\
        \addlinespace[4pt]
        \multicolumn{5}{c}{\parbox{0.94\textwidth}{\centering\tiny\itshape ``A scenic trail where a group of riders are mounted on their horses, moving in a line. The horses appear well-groomed, and the riders are dressed in casual riding attire. The path they are on is bordered by trees, and the group seems to be enjoying a leisurely ride in the countryside.''}} \\[9pt]
    \end{tabular}
    \end{adjustbox}
\vspace{-3pt}\end{figure}

\begin{figure}[p]\vspace{-3pt}
    \centering
    \captionsetup{font=normalsize}
    \caption{
        \textbf{Additional Qualitative Comparison on Qwen-Image 20B (Continued).}
        \Ours{} more closely matches the diversity of the teacher than DMD-based Qwen-Lightning while preserving visual quality.
        \checkpointmark{} denotes an officially released checkpoint.
    }
    \label{fig:qwen_qual_image_3}
    \setlength{\tabcolsep}{1.5pt}
    \renewcommand{\arraystretch}{1.05}
    \scriptsize
    \begin{adjustbox}{max totalsize={\textwidth}{0.92\textheight}}
    \begin{tabular}{c cccc}
        \QualRowLabel{Teacher} &
        \adjustbox{valign=m}{\includegraphics[width=0.23\textwidth]{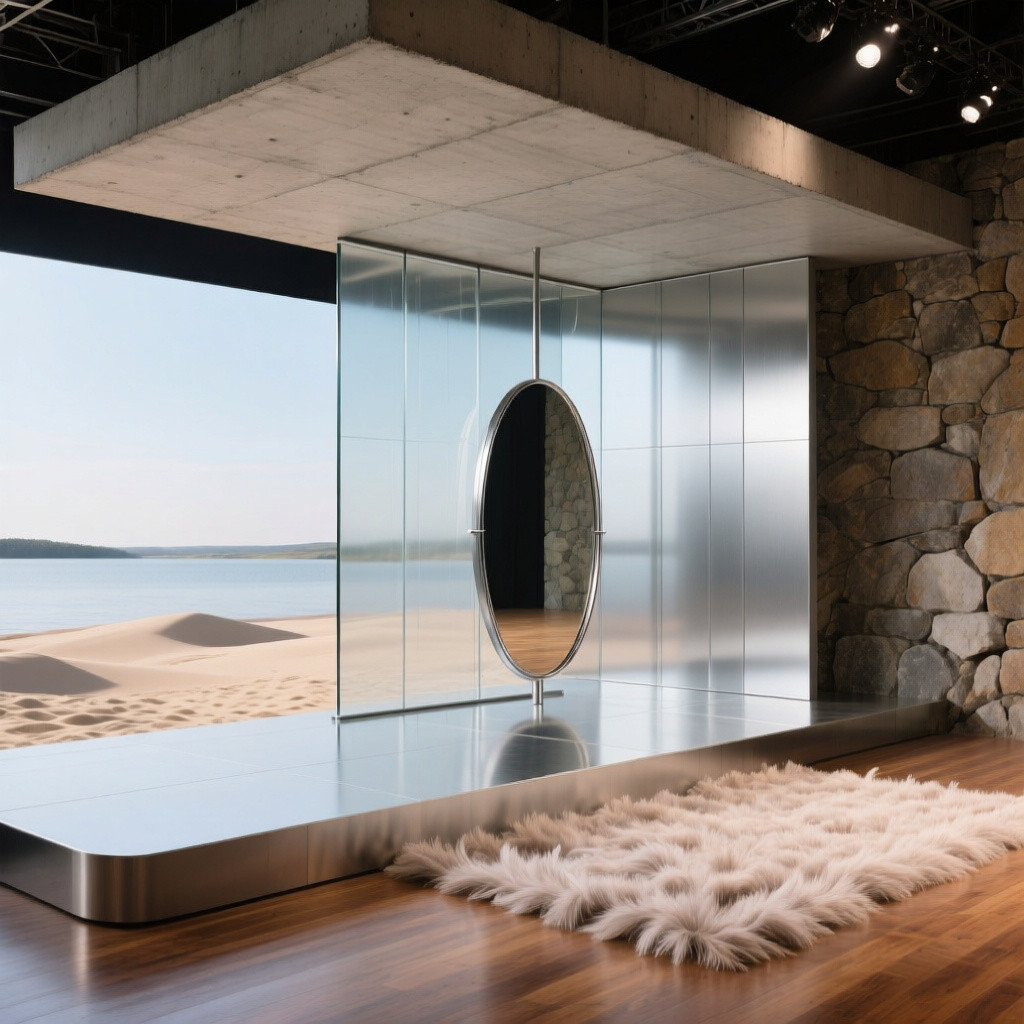}} & \adjustbox{valign=m}{\includegraphics[width=0.23\textwidth]{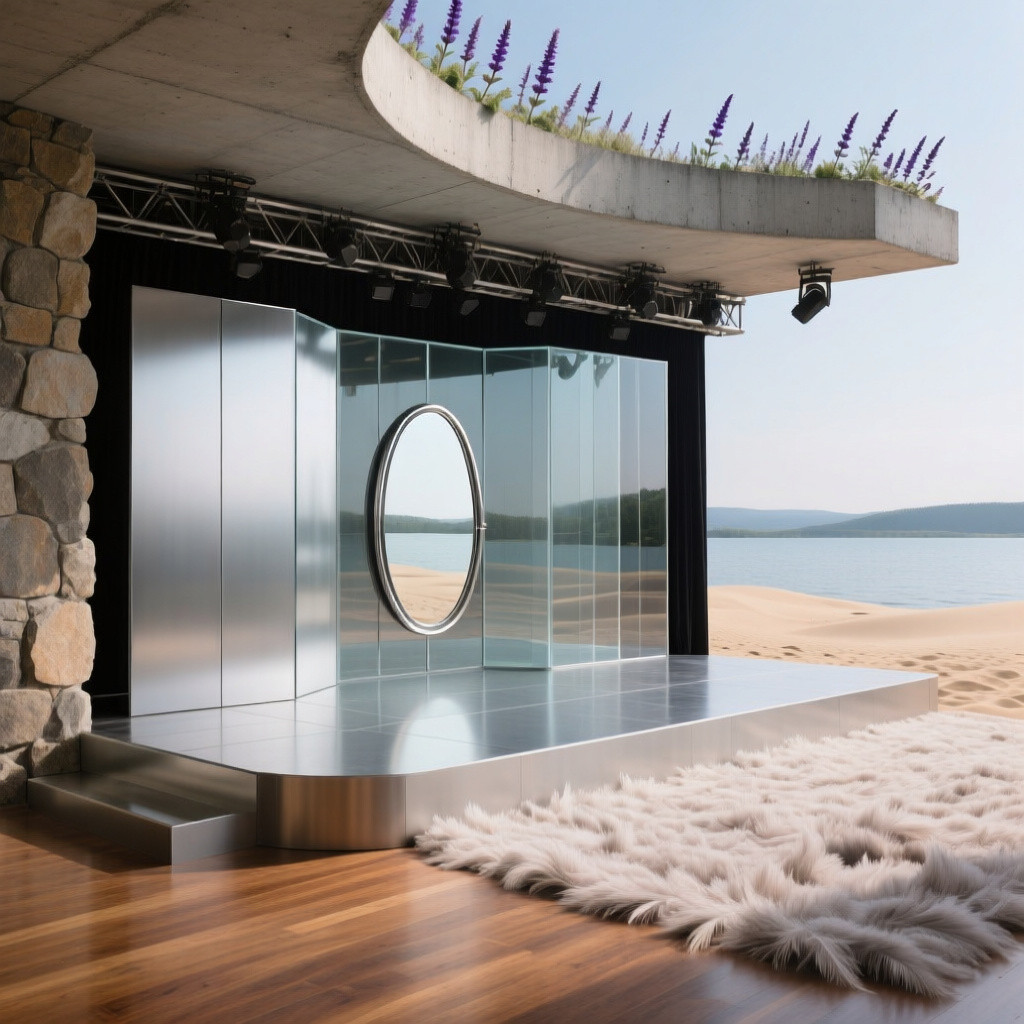}} & \adjustbox{valign=m}{\includegraphics[width=0.23\textwidth]{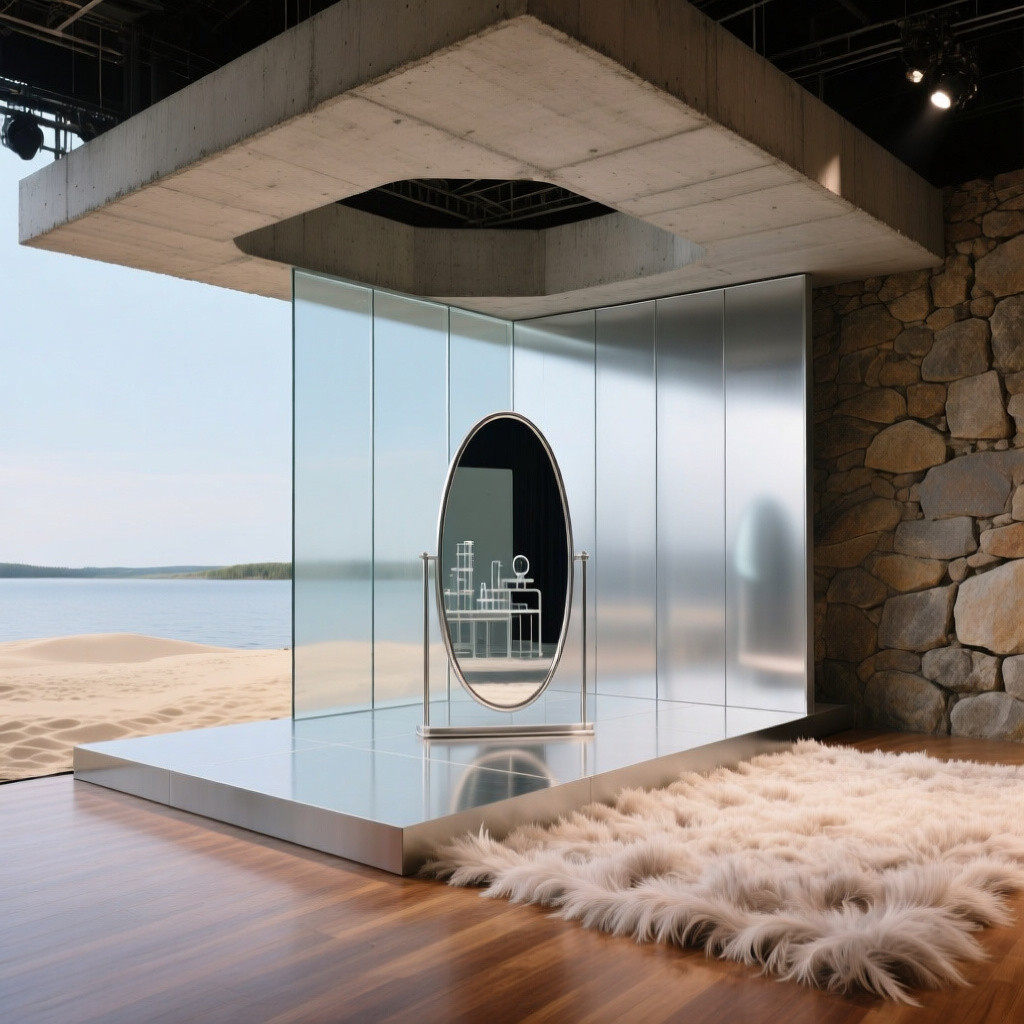}} & \adjustbox{valign=m}{\includegraphics[width=0.23\textwidth]{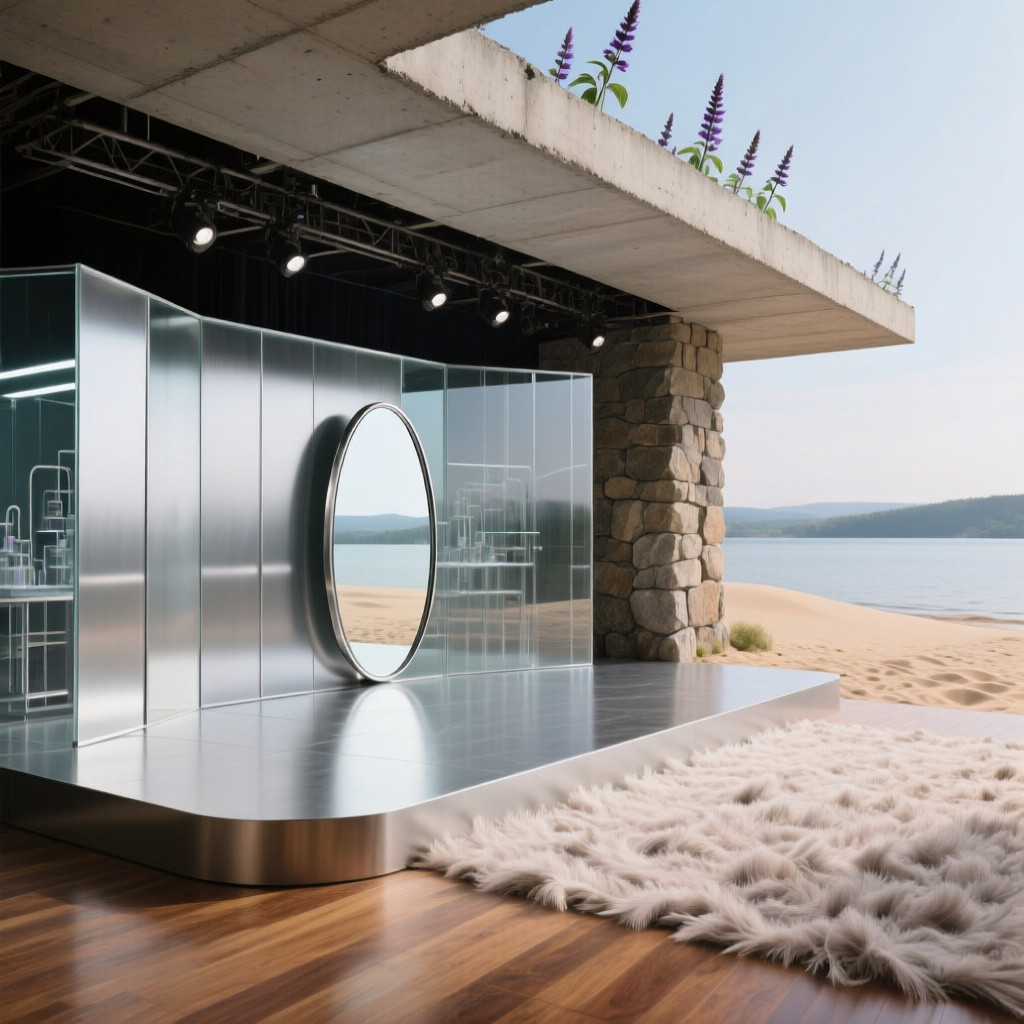}} \\
        \addlinespace[2pt]
        \QualRowLabel{\makecell{Qwen-Lightning\checkpointmark \\ {\tiny\citep{fan2025:qwenlightning}}}} &
        \adjustbox{valign=m}{\includegraphics[width=0.23\textwidth]{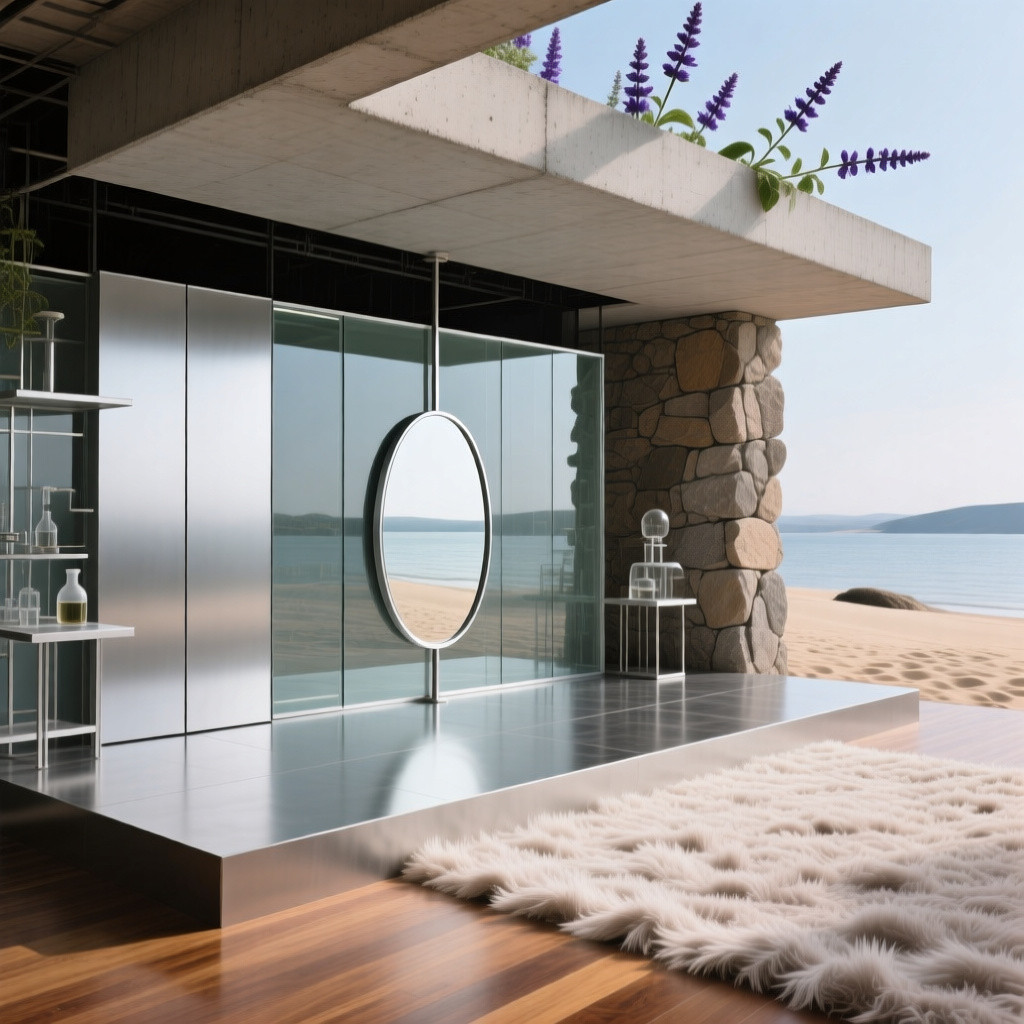}} & \adjustbox{valign=m}{\includegraphics[width=0.23\textwidth]{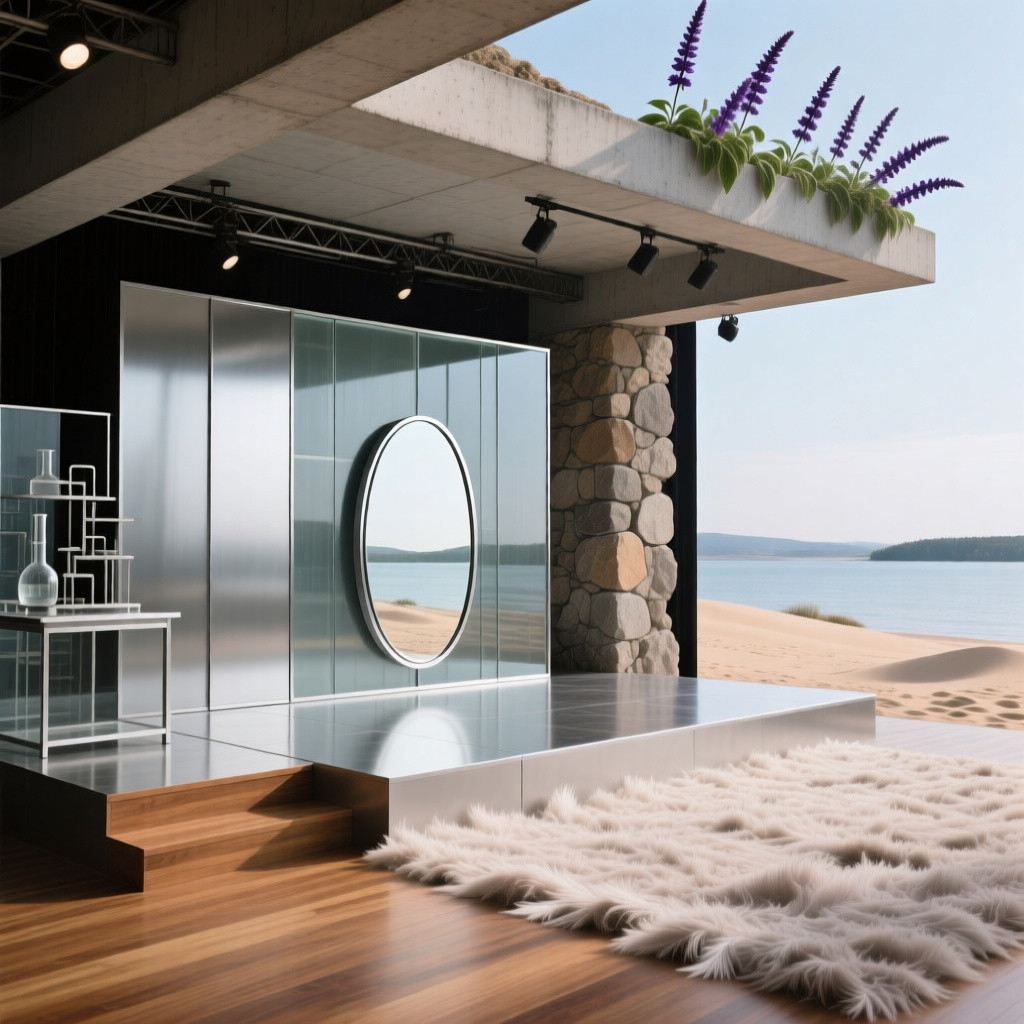}} & \adjustbox{valign=m}{\includegraphics[width=0.23\textwidth]{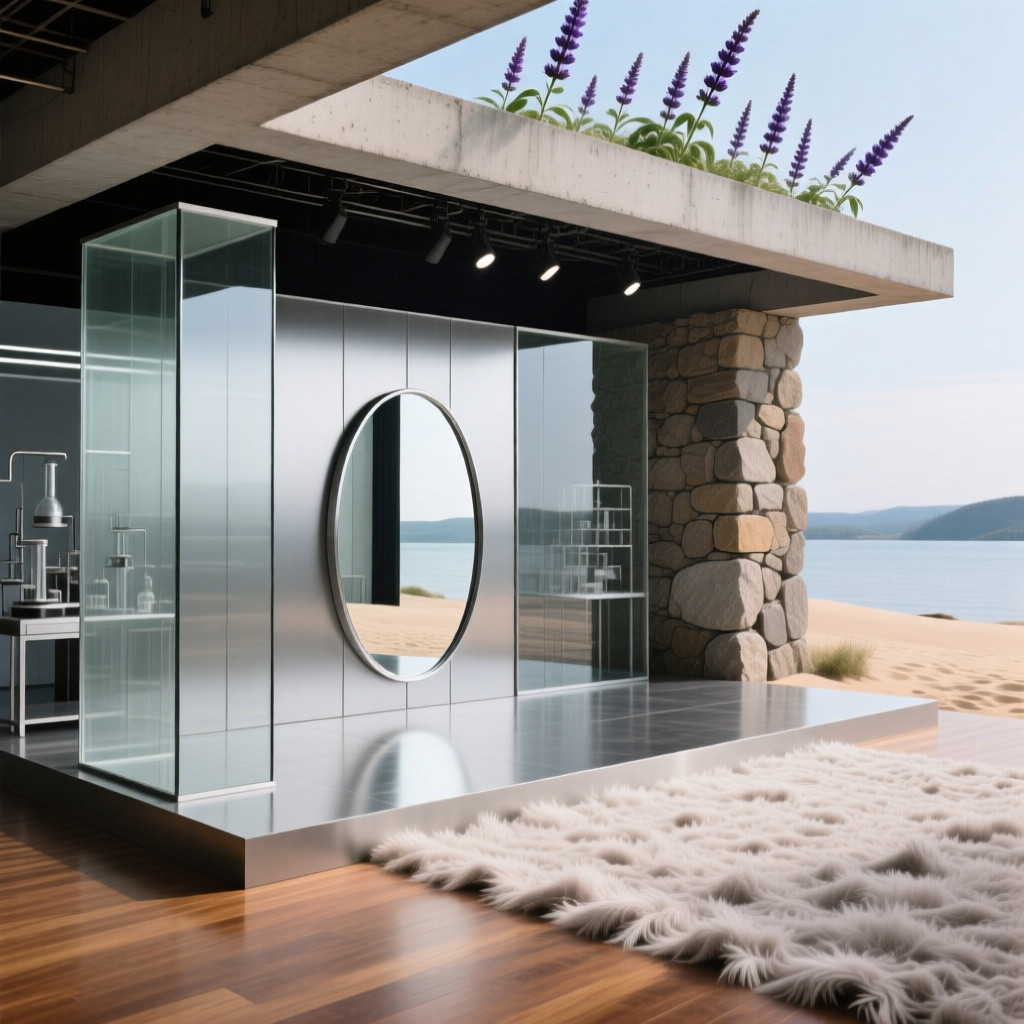}} & \adjustbox{valign=m}{\includegraphics[width=0.23\textwidth]{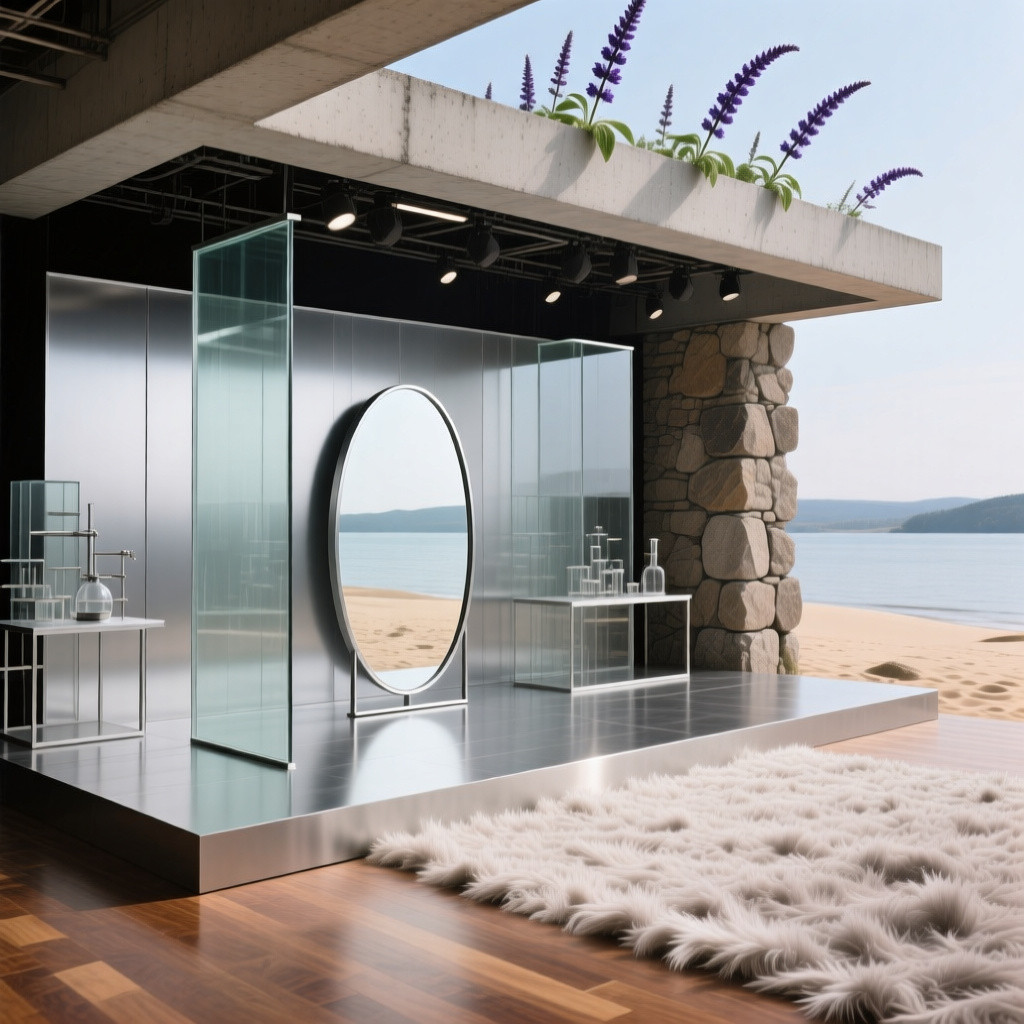}} \\
        \addlinespace[2pt]
        \QualRowLabel{\makecell{{\Oursbf{}} \\ {(Ours)}}} &
        \adjustbox{valign=m}{\includegraphics[width=0.23\textwidth]{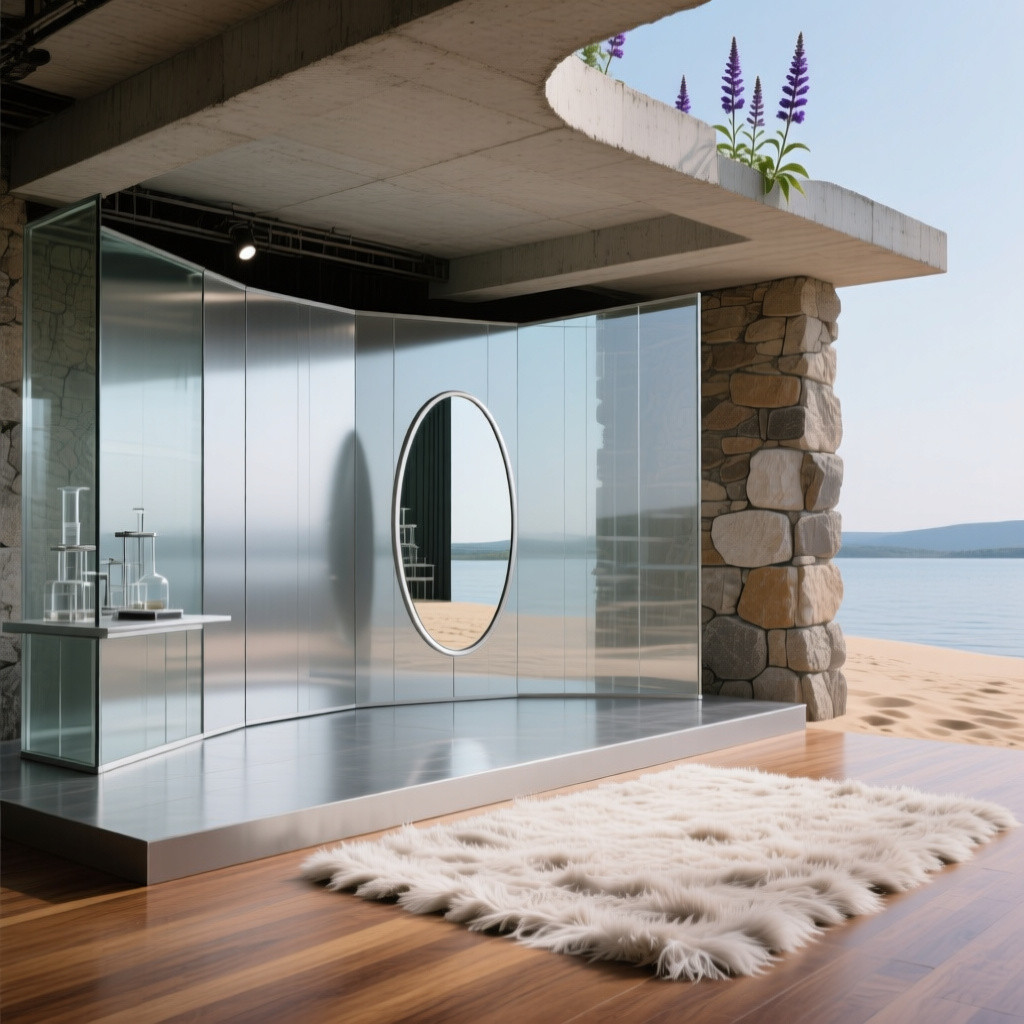}} & \adjustbox{valign=m}{\includegraphics[width=0.23\textwidth]{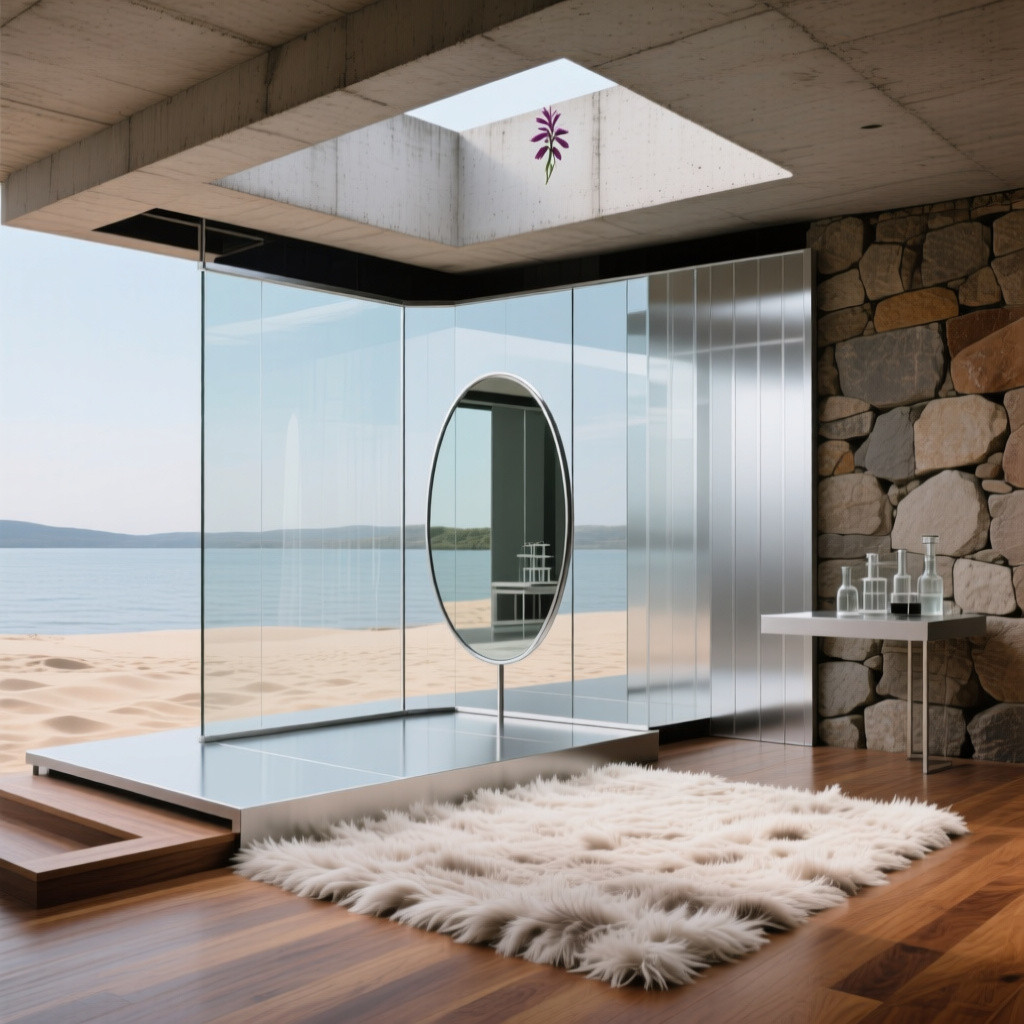}} & \adjustbox{valign=m}{\includegraphics[width=0.23\textwidth]{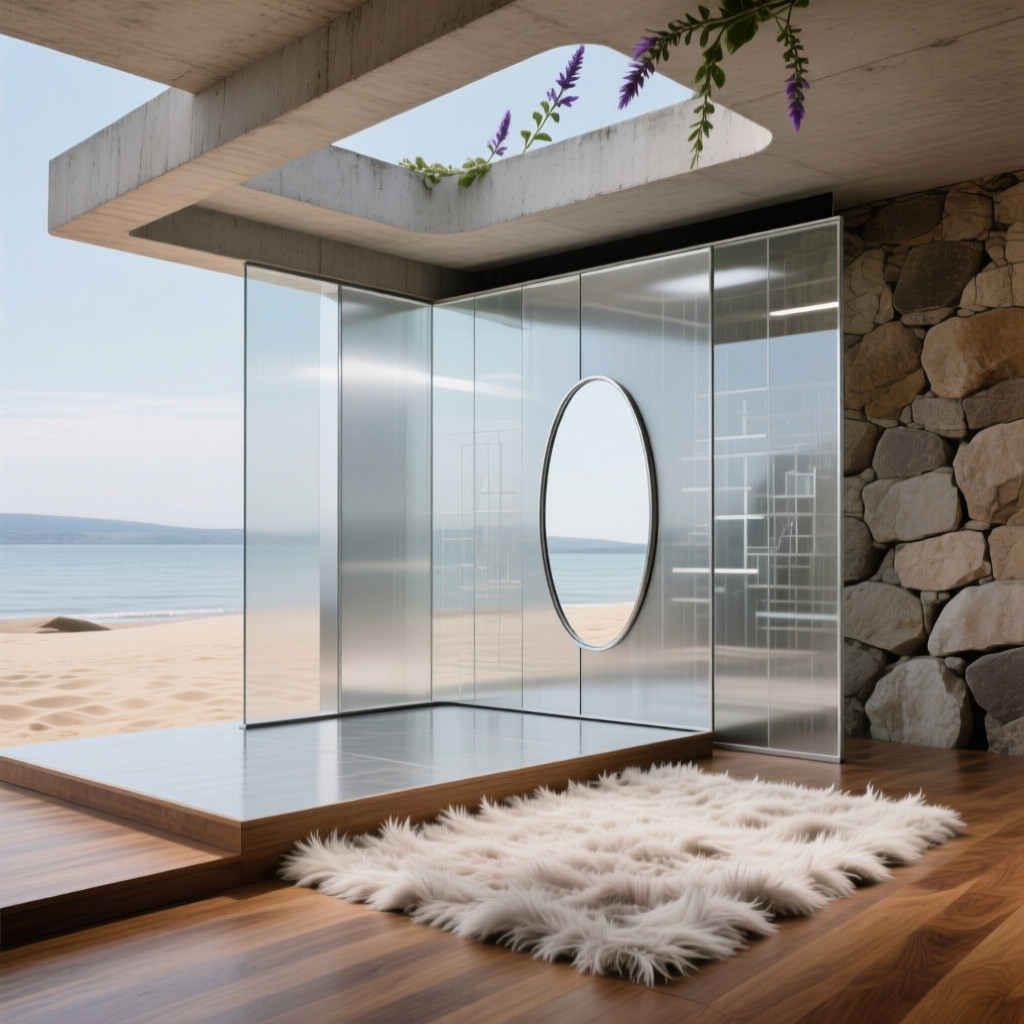}} & \adjustbox{valign=m}{\includegraphics[width=0.23\textwidth]{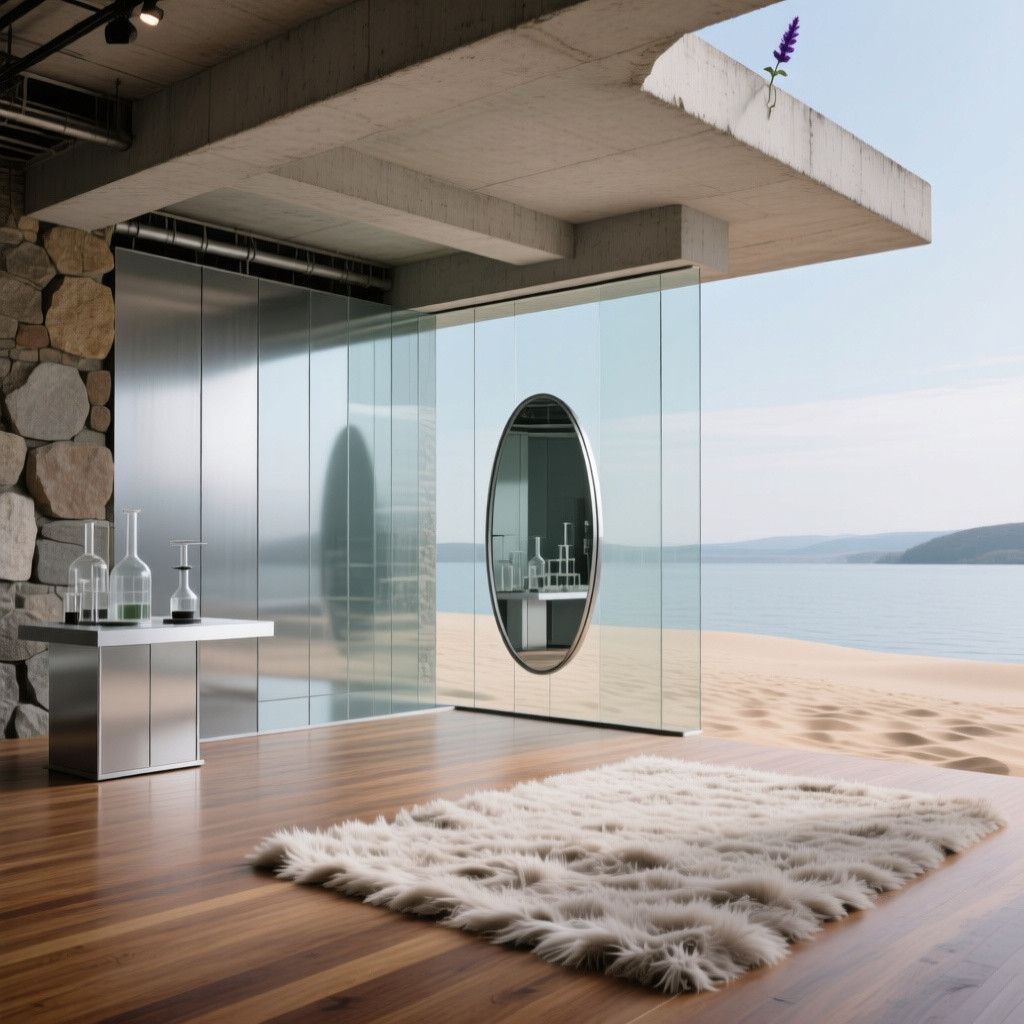}} \\
        \addlinespace[4pt]
        \multicolumn{5}{c}{\parbox{0.94\textwidth}{\centering\tiny\itshape ``A stage designed with a fusion of natural and industrial elements, featuring sleek silver glass panels reminiscent of a high-tech laboratory, situated alongside a soft, feathered carpet. The flooring transitions into an area of polished wood leading to a sturdy stone wall, adding an earthy touch to the synthetic ambiance. Overhead, raw concrete merges into the form of a salvia patens roof, juxtaposed with a tranquil sand-laden backdrop that suggests the gentle shores of a lake. In the center, an oval mirror with a slim steel frame reflects the composed space, including the transparent screen that adds a layer of modernity to this eclectic setting.''}} \\[9pt]
        \QualRowLabel{Teacher} &
        \adjustbox{valign=m}{\includegraphics[width=0.23\textwidth]{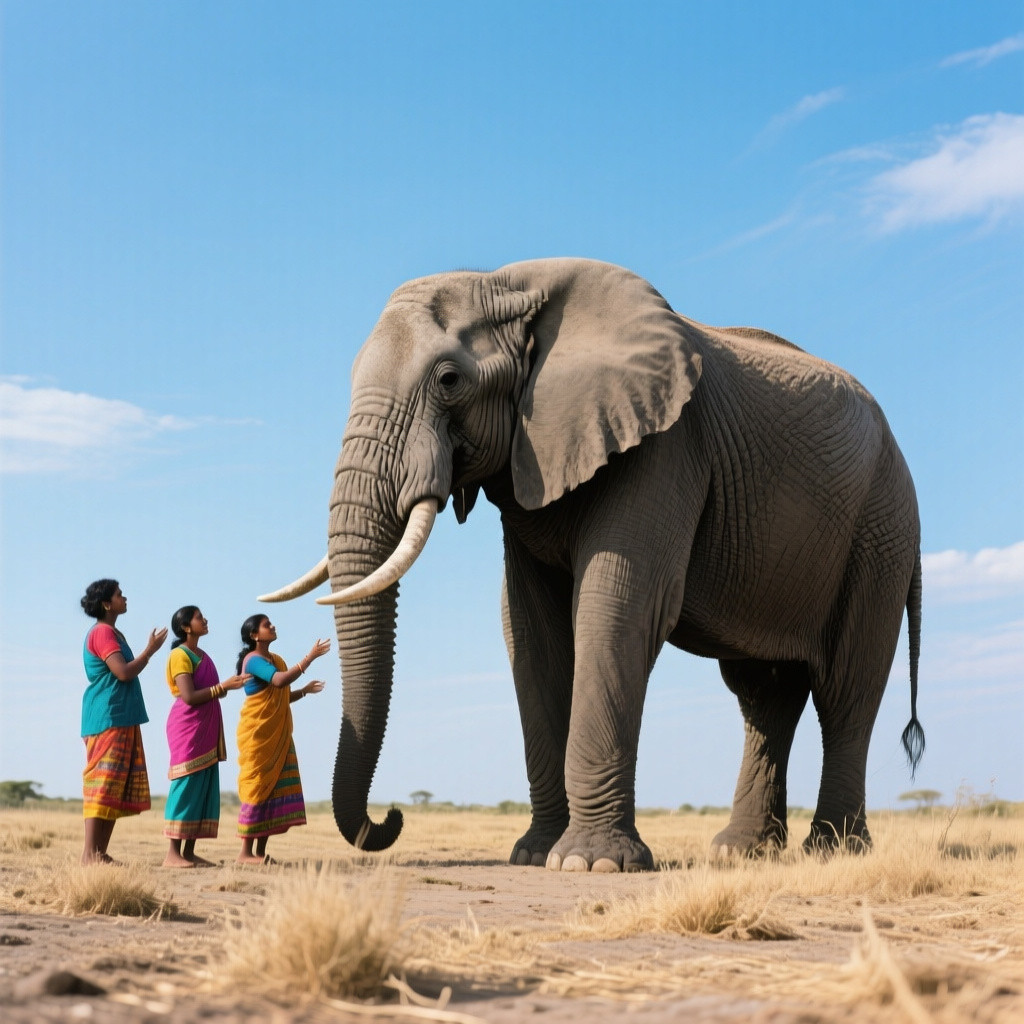}} & \adjustbox{valign=m}{\includegraphics[width=0.23\textwidth]{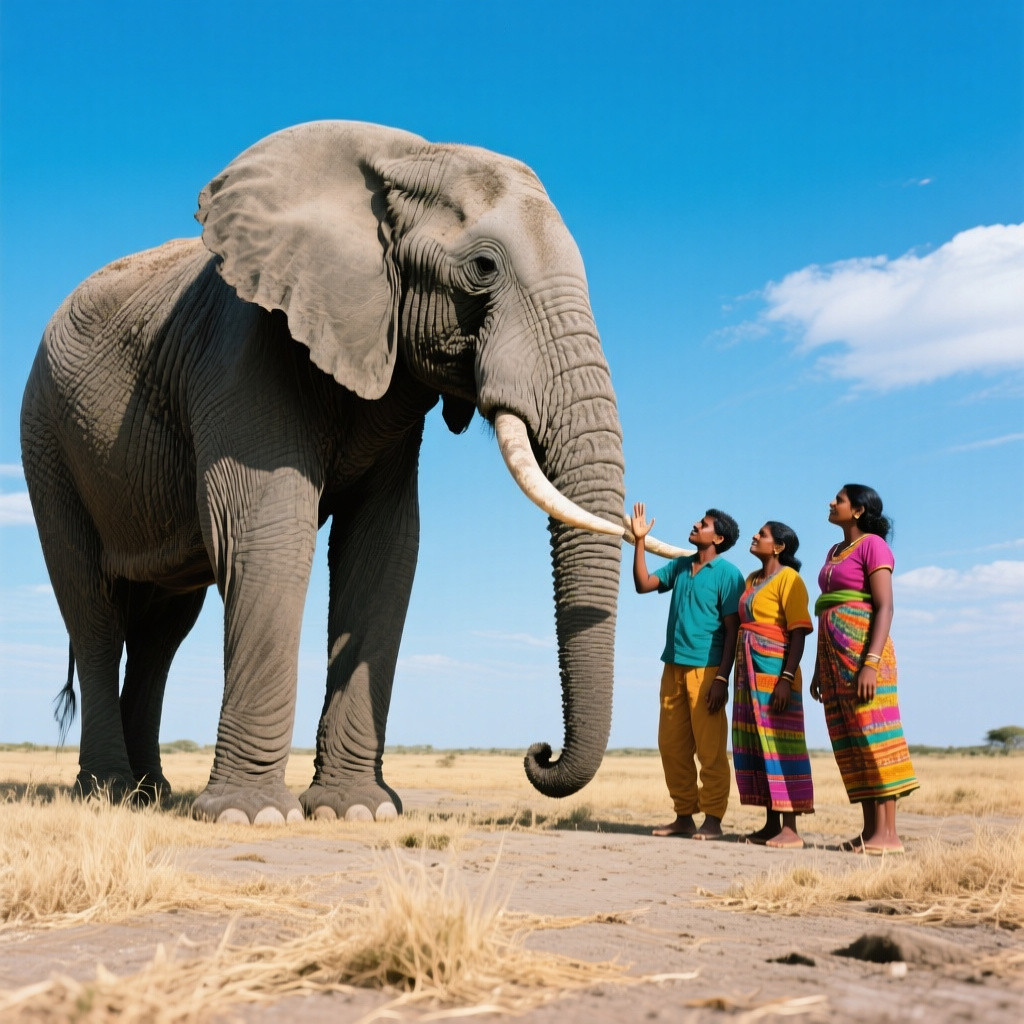}} & \adjustbox{valign=m}{\includegraphics[width=0.23\textwidth]{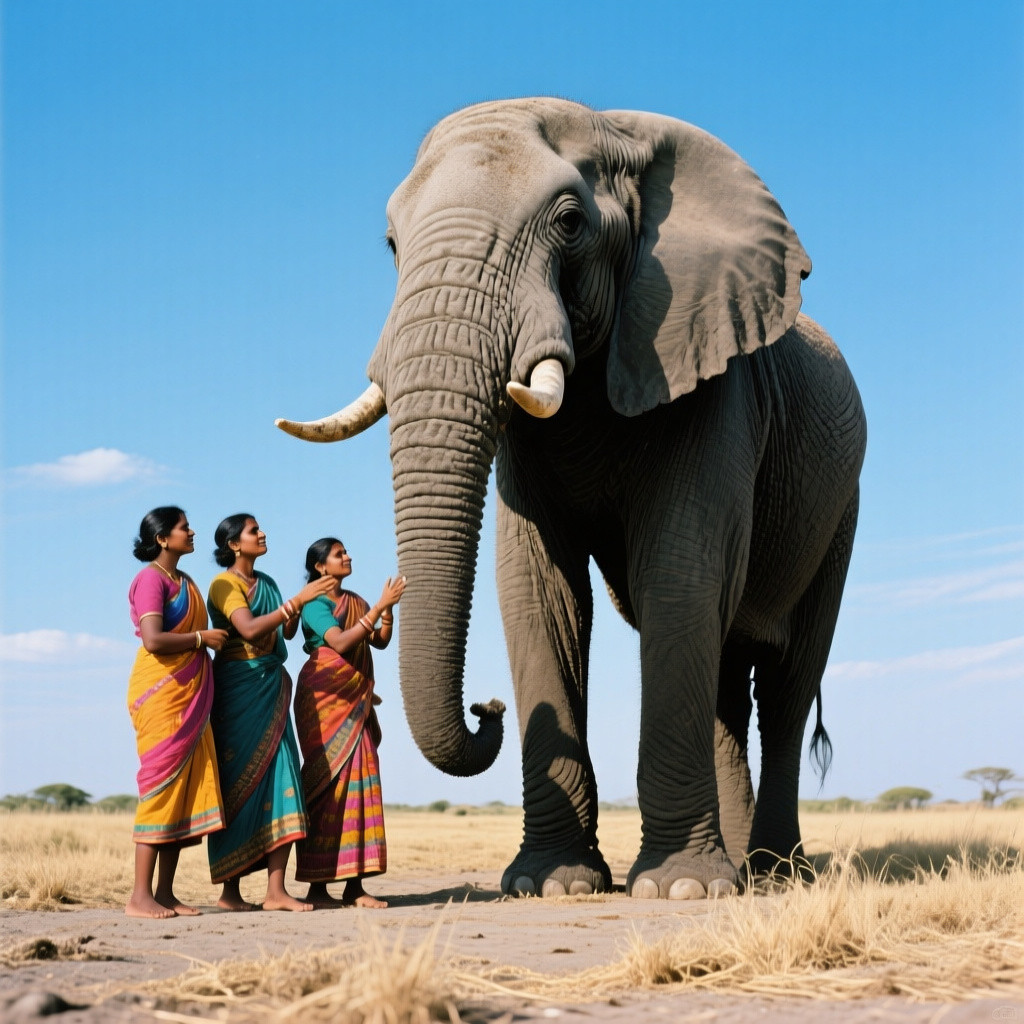}} & \adjustbox{valign=m}{\includegraphics[width=0.23\textwidth]{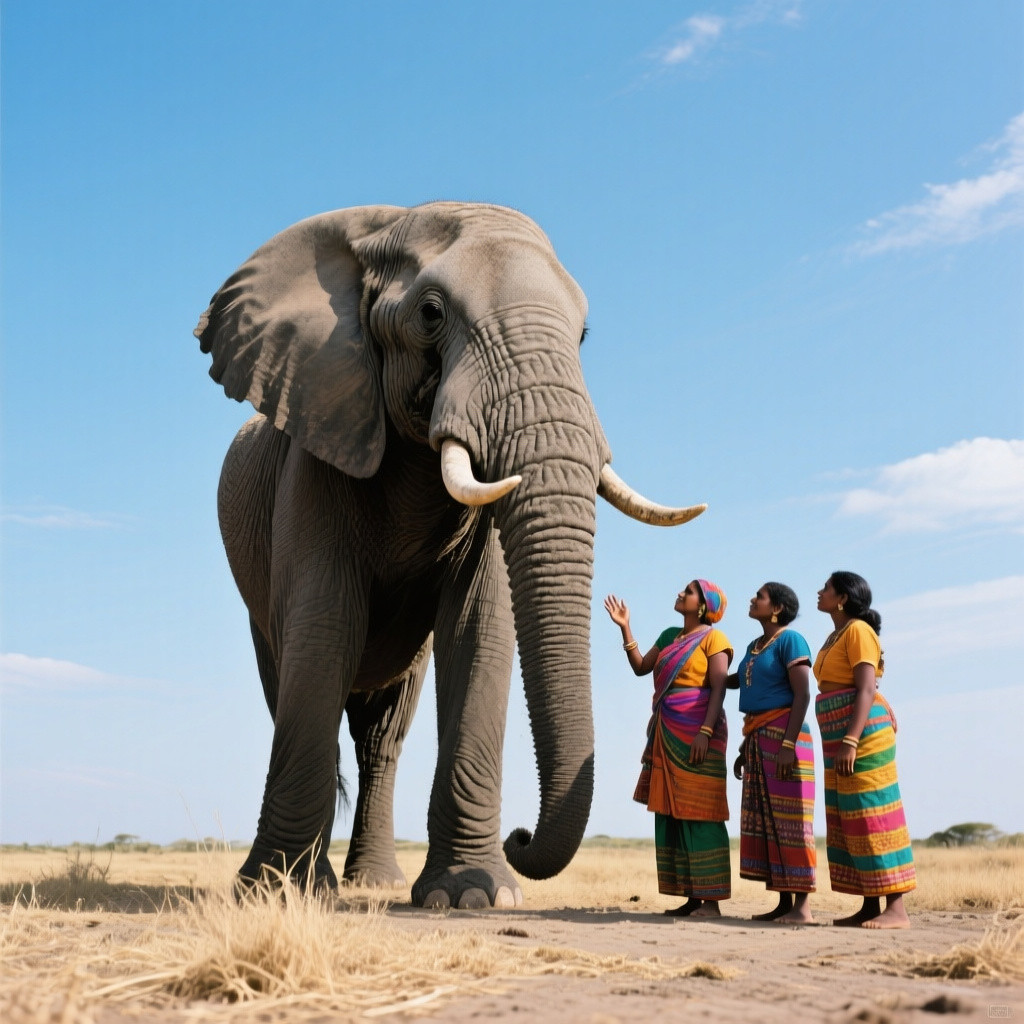}} \\
        \addlinespace[2pt]
        \QualRowLabel{\makecell{Qwen-Lightning\checkpointmark \\ {\tiny\citep{fan2025:qwenlightning}}}} &
        \adjustbox{valign=m}{\includegraphics[width=0.23\textwidth]{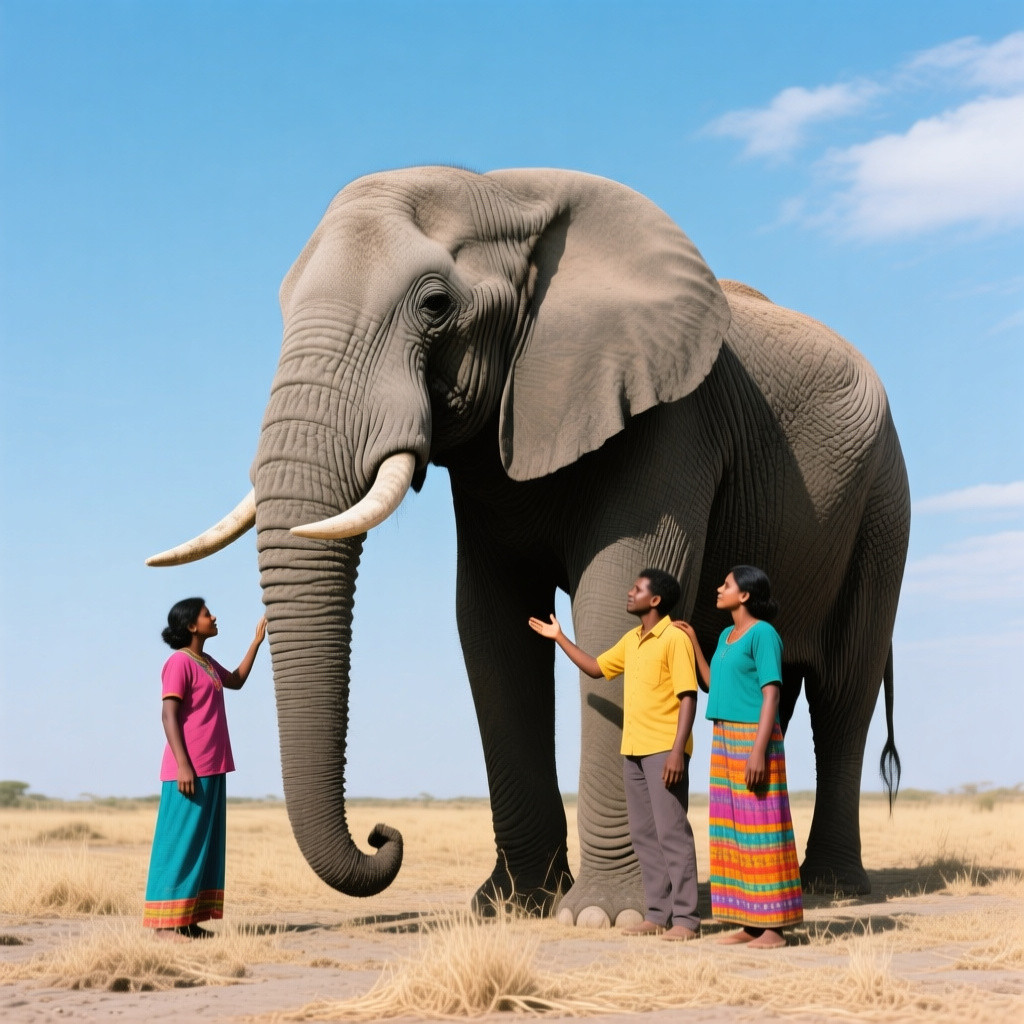}} & \adjustbox{valign=m}{\includegraphics[width=0.23\textwidth]{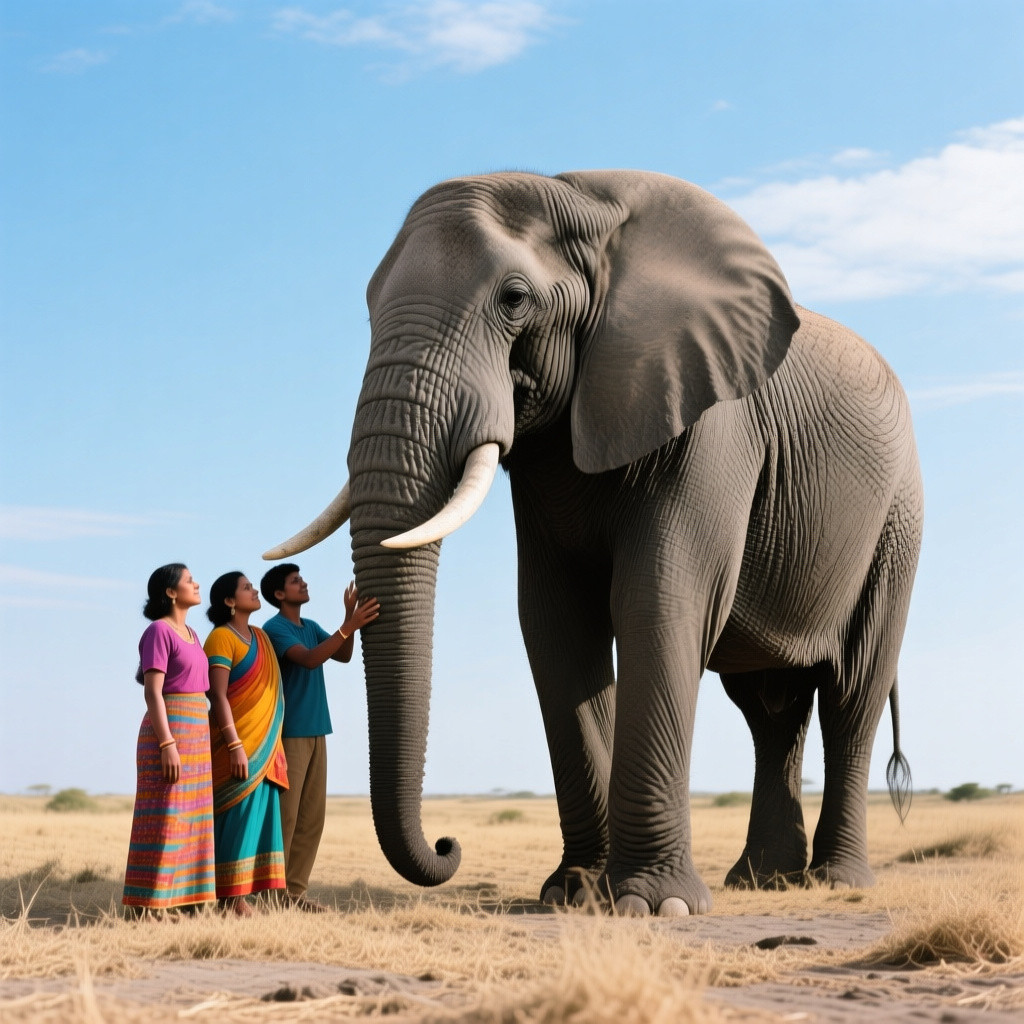}} & \adjustbox{valign=m}{\includegraphics[width=0.23\textwidth]{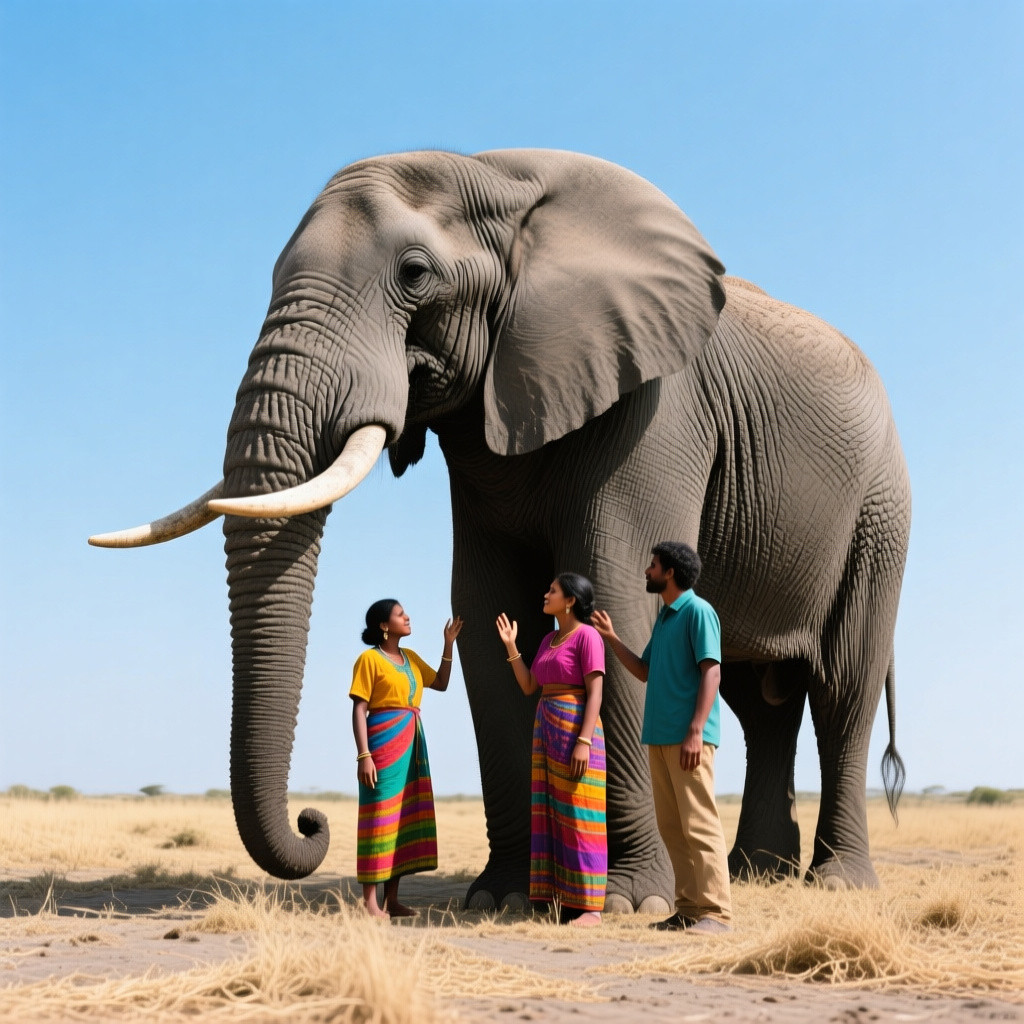}} & \adjustbox{valign=m}{\includegraphics[width=0.23\textwidth]{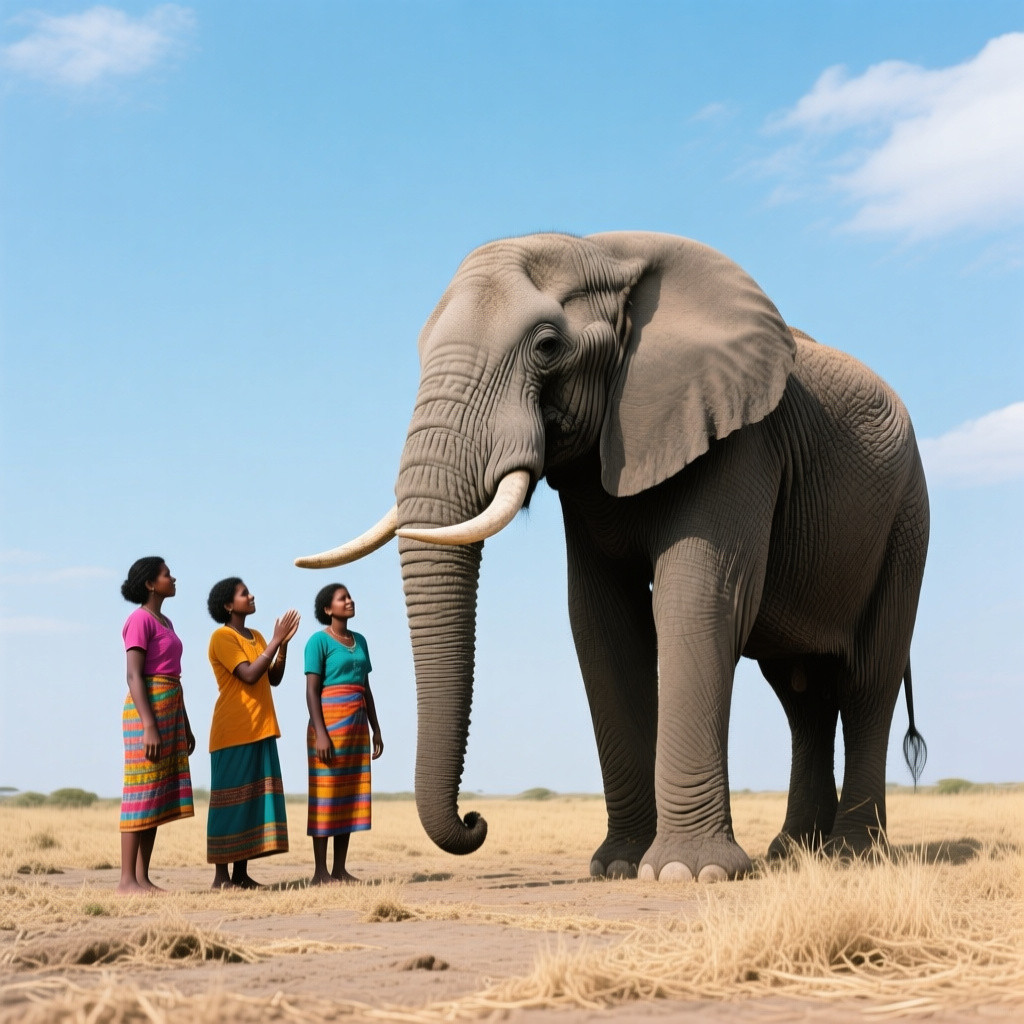}} \\
        \addlinespace[2pt]
        \QualRowLabel{\makecell{{\Oursbf{}} \\ {(Ours)}}} &
        \adjustbox{valign=m}{\includegraphics[width=0.23\textwidth]{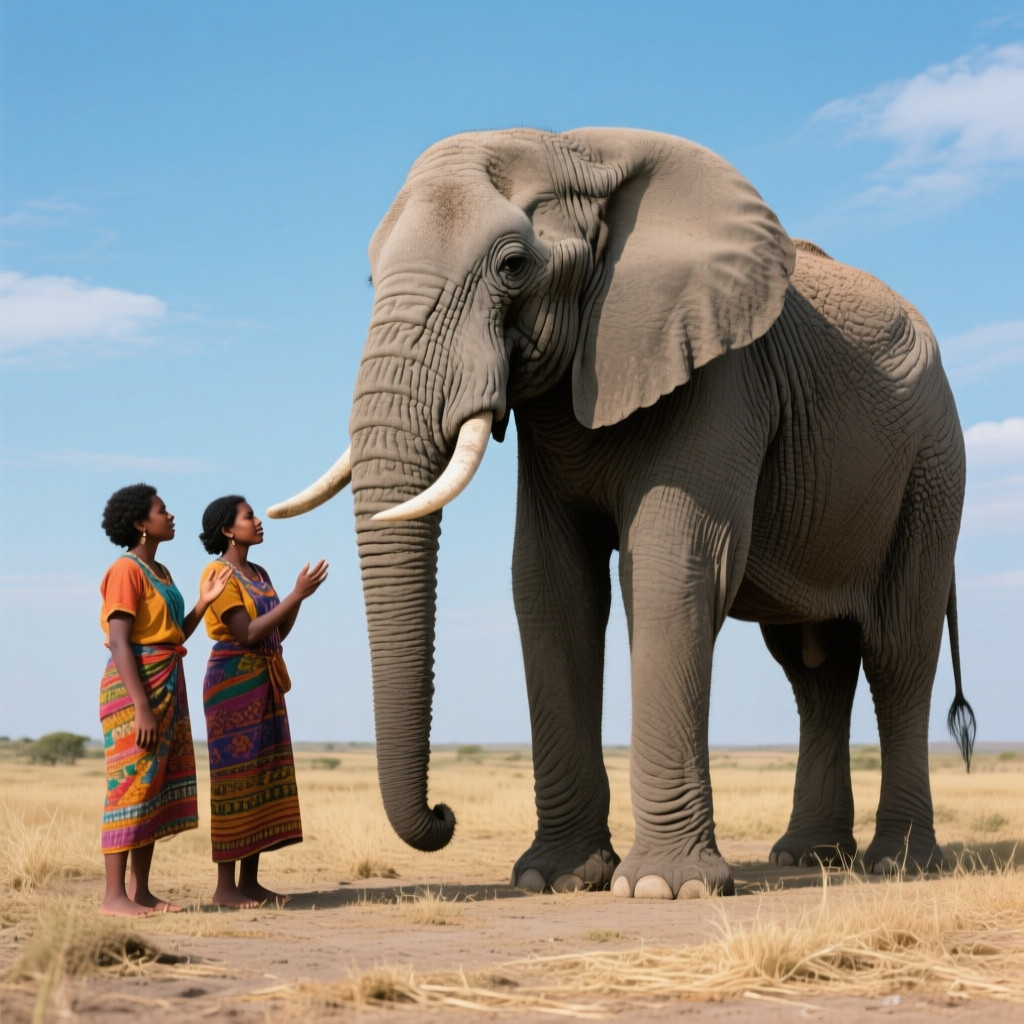}} & \adjustbox{valign=m}{\includegraphics[width=0.23\textwidth]{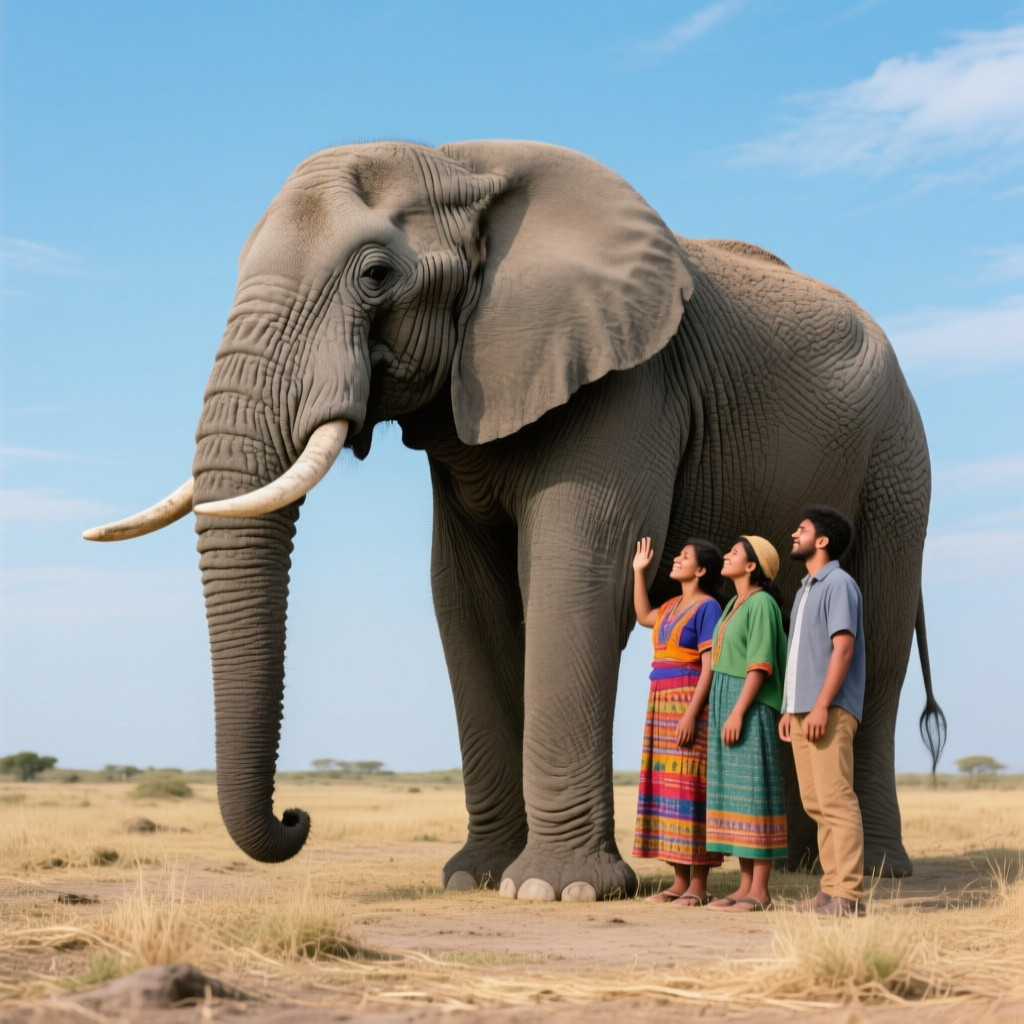}} & \adjustbox{valign=m}{\includegraphics[width=0.23\textwidth]{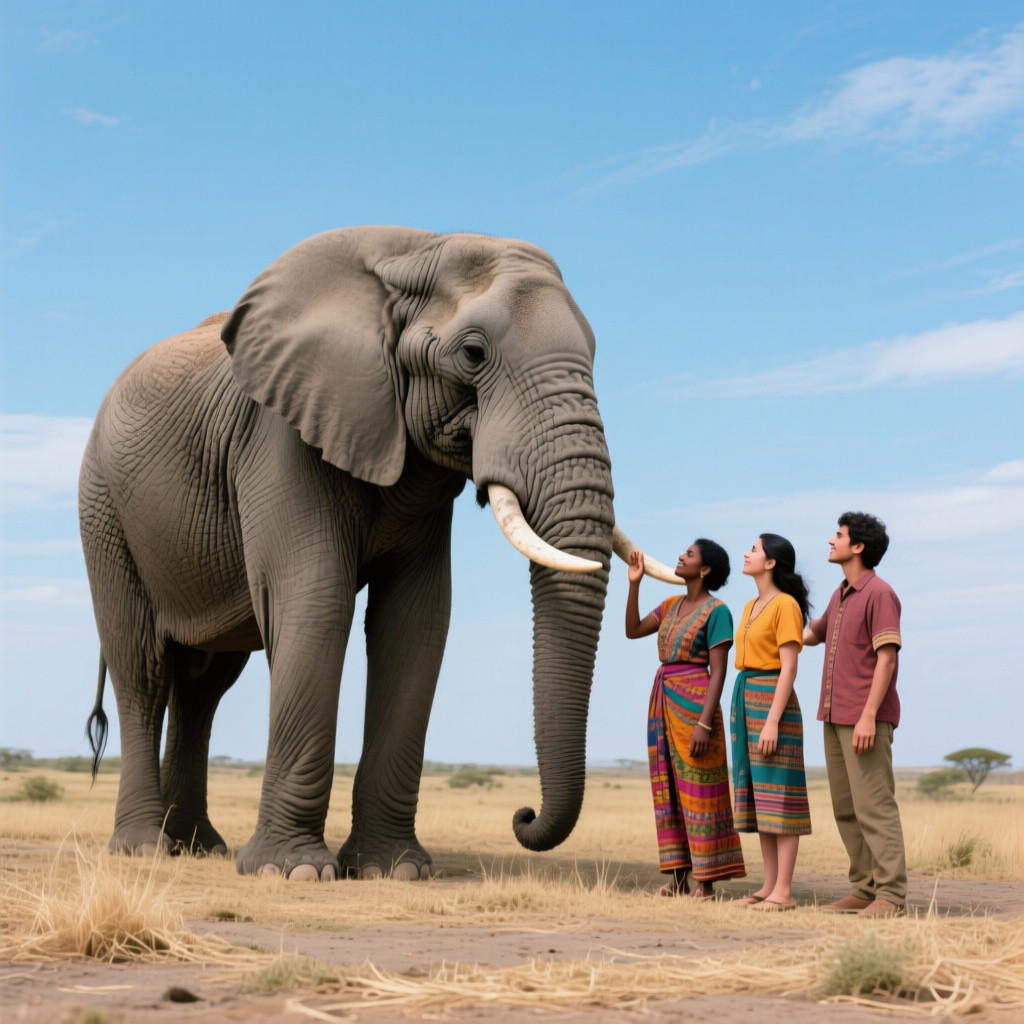}} & \adjustbox{valign=m}{\includegraphics[width=0.23\textwidth]{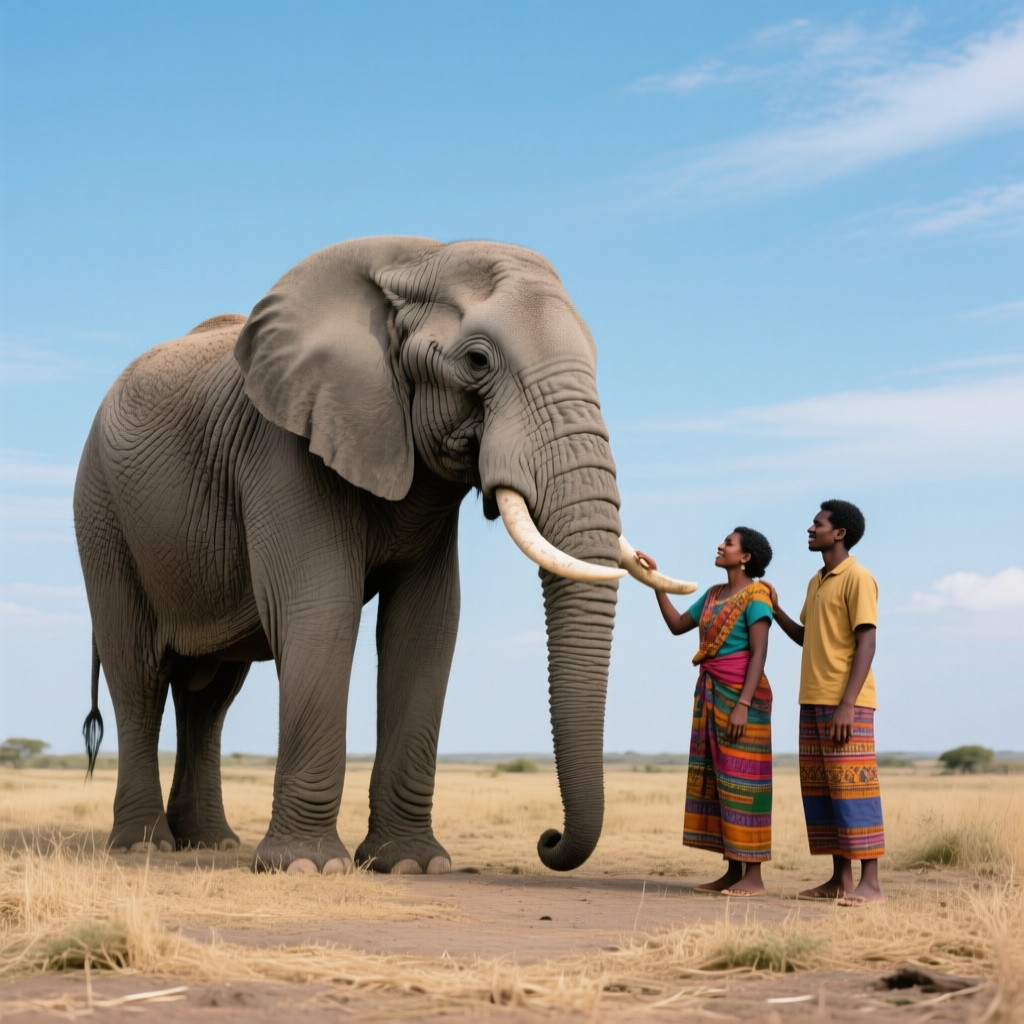}} \\
        \addlinespace[4pt]
        \multicolumn{5}{c}{\parbox{0.94\textwidth}{\centering\tiny\itshape ``A group of three individuals positioned beside a towering elephant in an open area. The elephant's gray skin contrasts with the colorful clothing of the people. They appear to be engaged in a moment of interaction, with the vast sky above them and the ground beneath scattered with dry grass.''}} \\[9pt]
    \end{tabular}
    \end{adjustbox}
\vspace{-3pt}\end{figure}

\providecommand{\QualRowLabel}[1]{\adjustbox{valign=m}{\rotatebox[origin=c]{90}{\scriptsize #1}}}
\begin{figure}[p]\vspace{-3pt}
    \centering
    \begin{minipage}{\textwidth}
    \centering
    \begin{threeparttable}
    \captionsetup{font=normalsize}
    \caption{
        \textbf{Additional Qualitative Comparison on Wan 14B.}
        Each method row shows two video samples for the prompt below.
        \checkpointmark{} denotes an officially released checkpoint.
    }
    \label{fig:wan_qual_video}
    \setlength{\tabcolsep}{1.5pt}
    \renewcommand{\arraystretch}{1.05}
    \scriptsize
    \begin{tabular}{@{}c ccc@{\hspace{8pt}}ccc@{}}
        \QualRowLabel{DMD2} &
        \adjustbox{valign=m}{\includegraphics[width=0.145\textwidth]{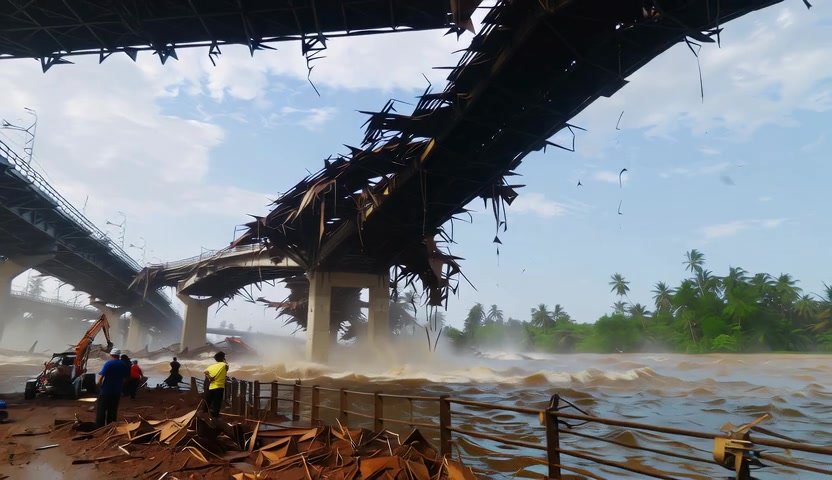}} & \adjustbox{valign=m}{\includegraphics[width=0.145\textwidth]{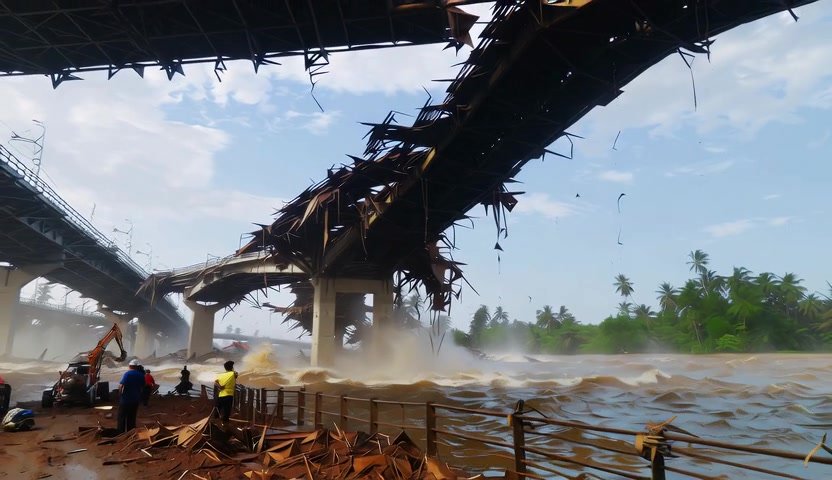}} & \adjustbox{valign=m}{\includegraphics[width=0.145\textwidth]{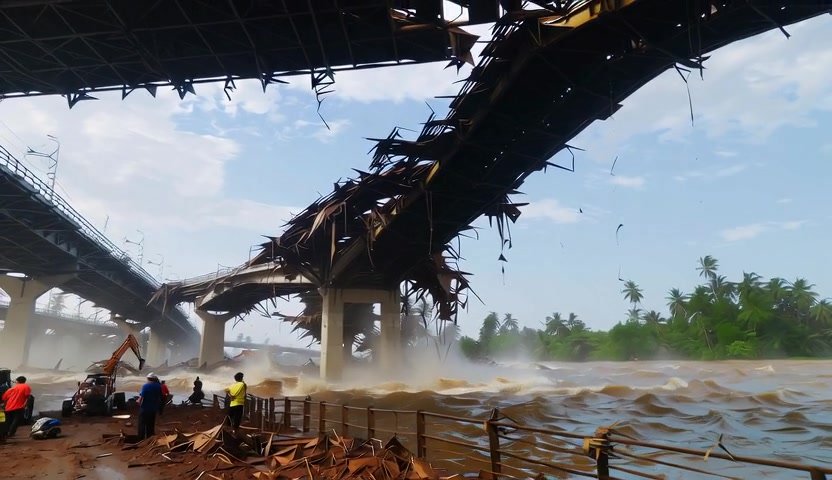}} & \adjustbox{valign=m}{\includegraphics[width=0.145\textwidth]{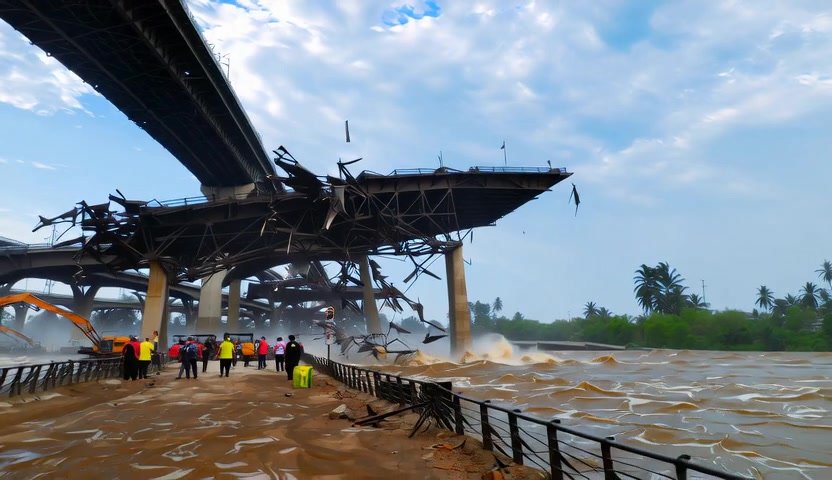}} & \adjustbox{valign=m}{\includegraphics[width=0.145\textwidth]{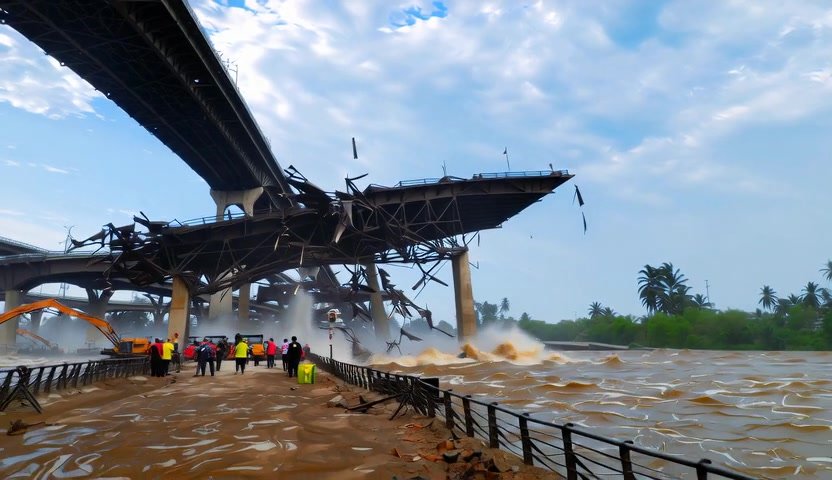}} & \adjustbox{valign=m}{\includegraphics[width=0.145\textwidth]{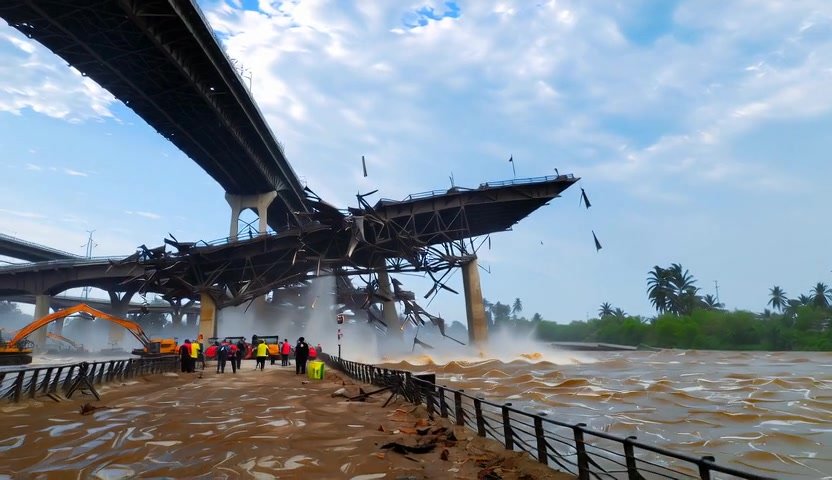}} \\[2pt]
        \QualRowLabel{AnyFlow\checkpointmark} &
        \adjustbox{valign=m}{\includegraphics[width=0.145\textwidth]{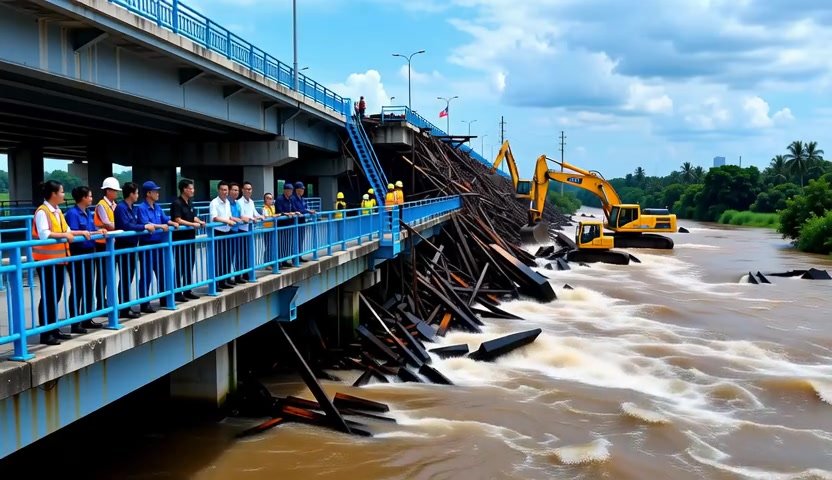}} & \adjustbox{valign=m}{\includegraphics[width=0.145\textwidth]{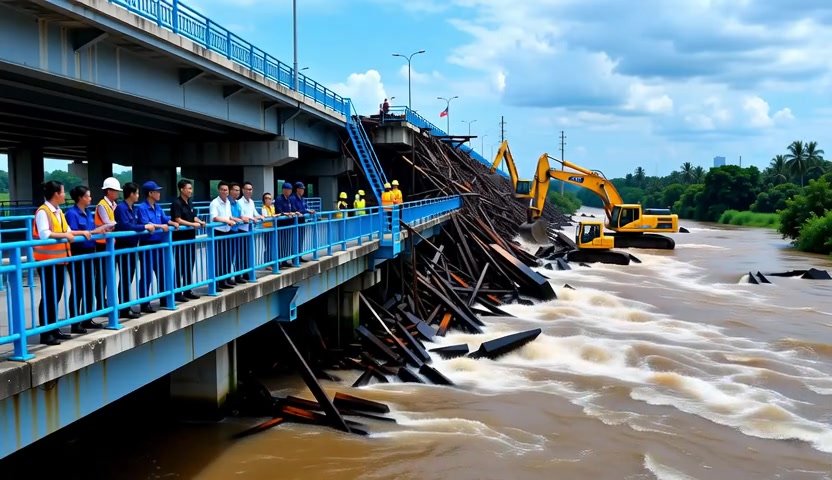}} & \adjustbox{valign=m}{\includegraphics[width=0.145\textwidth]{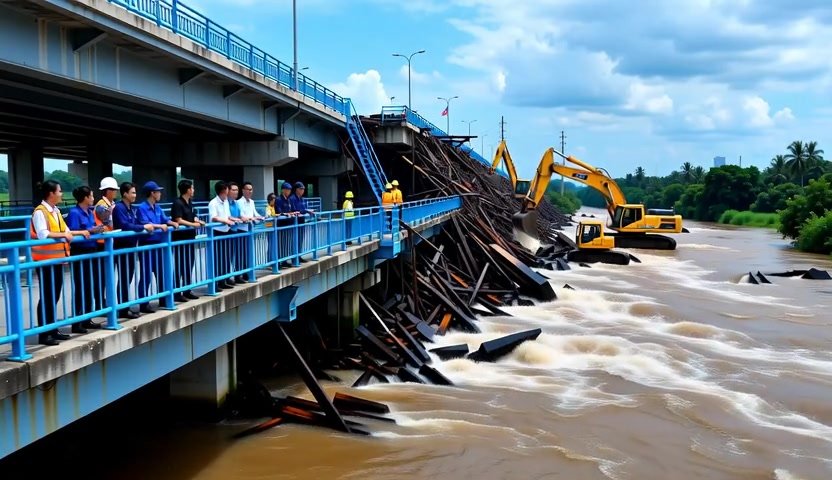}} & \adjustbox{valign=m}{\includegraphics[width=0.145\textwidth]{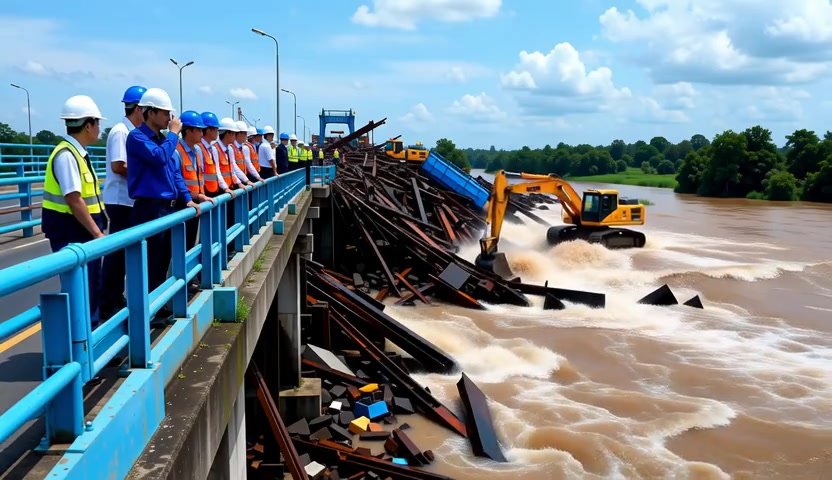}} & \adjustbox{valign=m}{\includegraphics[width=0.145\textwidth]{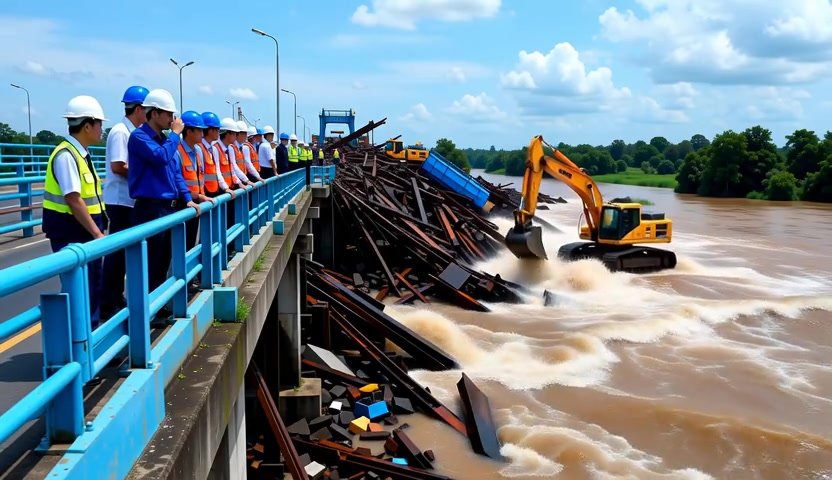}} & \adjustbox{valign=m}{\includegraphics[width=0.145\textwidth]{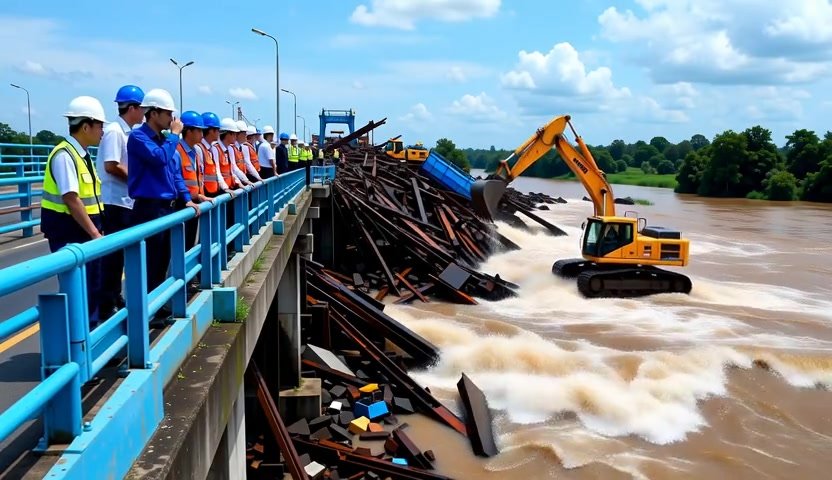}} \\[2pt]
        \QualRowLabel{\makecell{{\Oursbf{}} \\ {(Ours)}}} &
        \adjustbox{valign=m}{\includegraphics[width=0.145\textwidth]{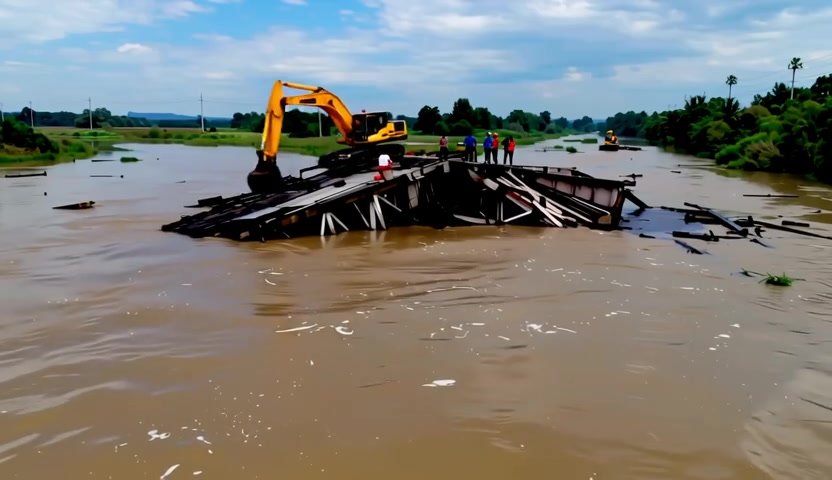}} & \adjustbox{valign=m}{\includegraphics[width=0.145\textwidth]{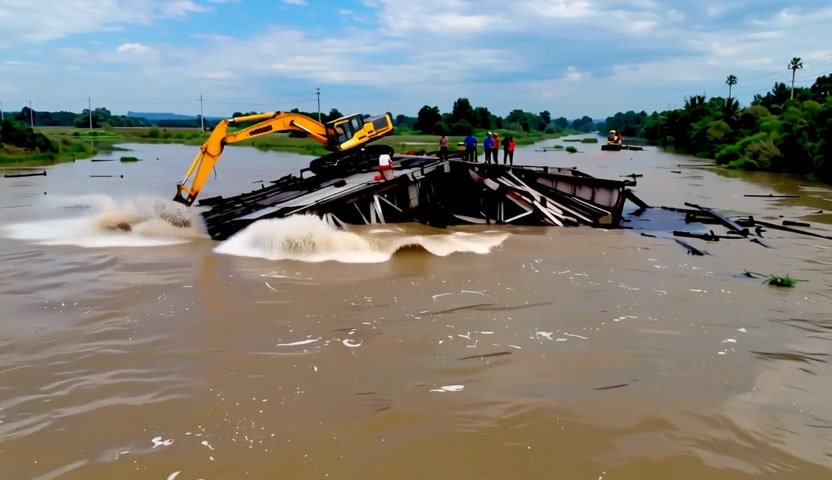}} & \adjustbox{valign=m}{\includegraphics[width=0.145\textwidth]{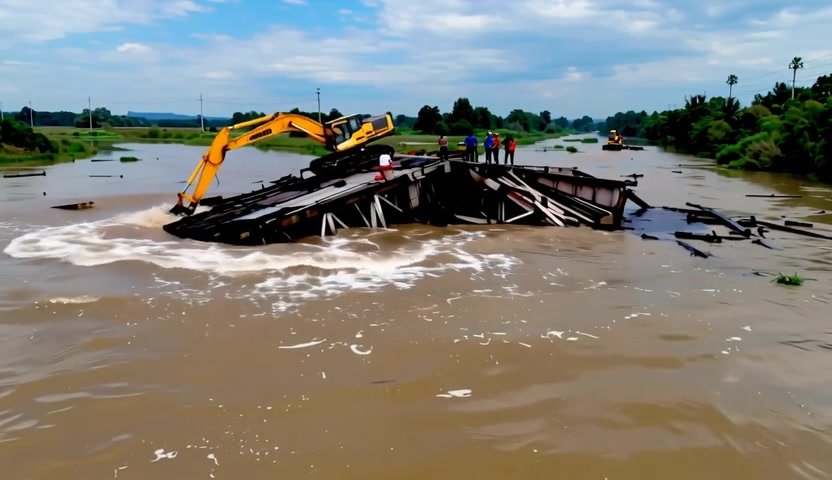}} & \adjustbox{valign=m}{\includegraphics[width=0.145\textwidth]{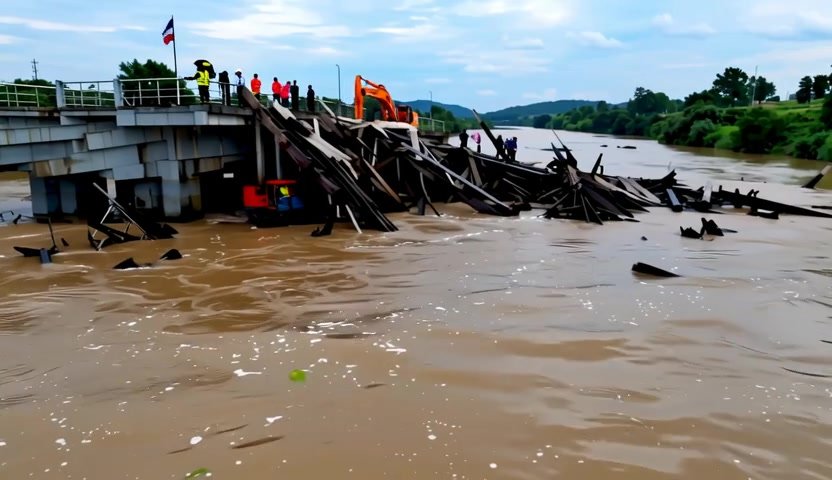}} & \adjustbox{valign=m}{\includegraphics[width=0.145\textwidth]{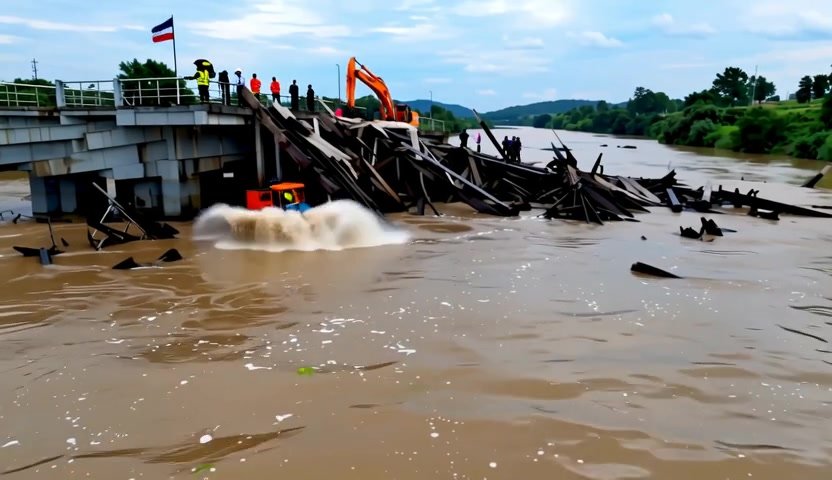}} & \adjustbox{valign=m}{\includegraphics[width=0.145\textwidth]{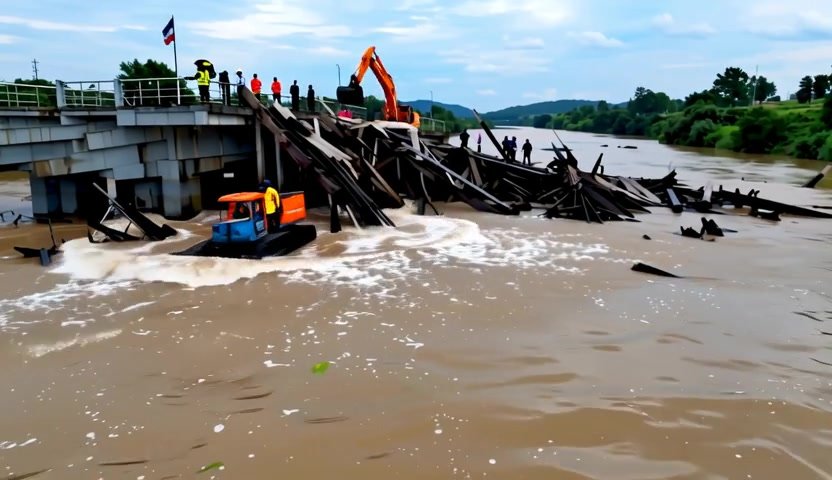}} \\[16pt]
        & \multicolumn{6}{c}{\parbox{0.90\textwidth}{\centering\tiny\itshape ``The video captures a scene of emergency response following the collapse of the Ji Rao Bridge in Thailand. A group of individuals, including officials and workers, is gathered on the bridge's railing, observing the damage. They appear concerned and engaged in discussion, possibly assessing the situation or coordinating rescue efforts. In the background, heavy machinery, likely excavators, is actively working to clear debris and assess the extent of the damage. The water below flows rapidly, indicating the severity of the flood conditions. The sky is partly cloudy, suggesting variable weather conditions during the incident. This collaborative effort highlights the urgency and scale of the disaster, with multiple parties working together to address the crisis.''}} \\[28pt]
        \QualRowLabel{DMD2} &
        \adjustbox{valign=m}{\includegraphics[width=0.145\textwidth]{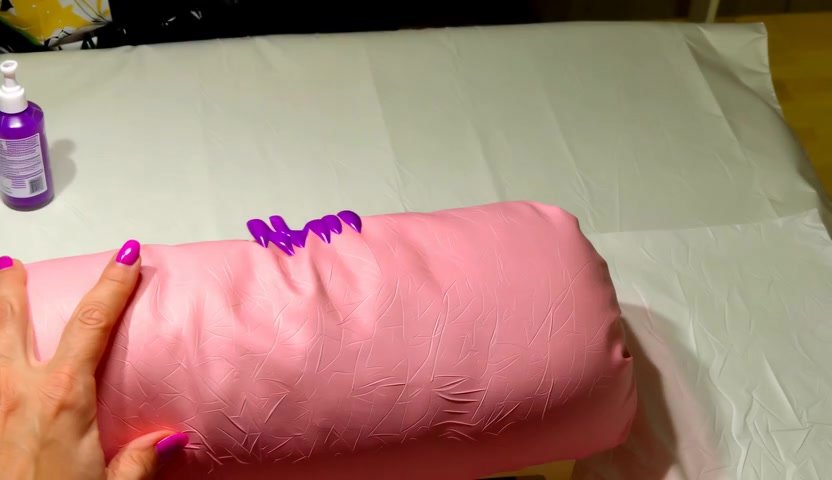}} & \adjustbox{valign=m}{\includegraphics[width=0.145\textwidth]{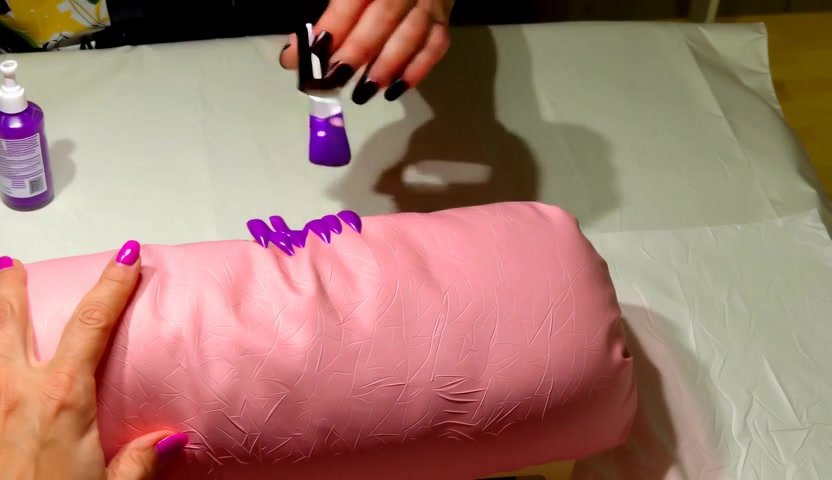}} & \adjustbox{valign=m}{\includegraphics[width=0.145\textwidth]{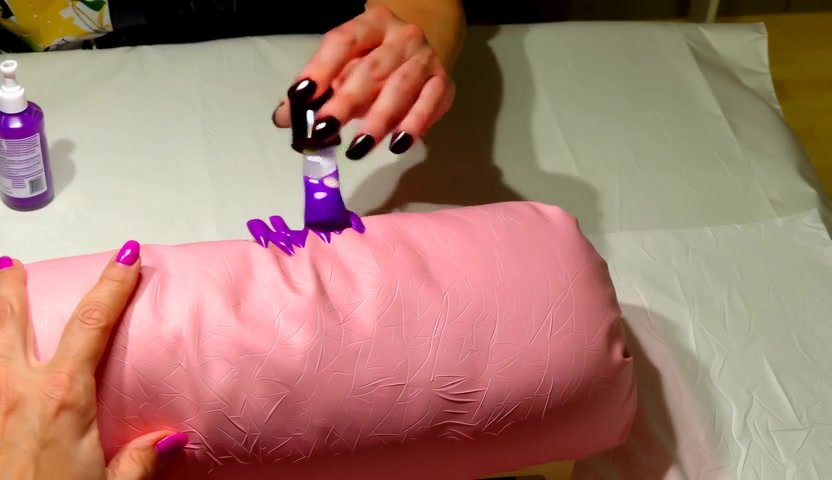}} & \adjustbox{valign=m}{\includegraphics[width=0.145\textwidth]{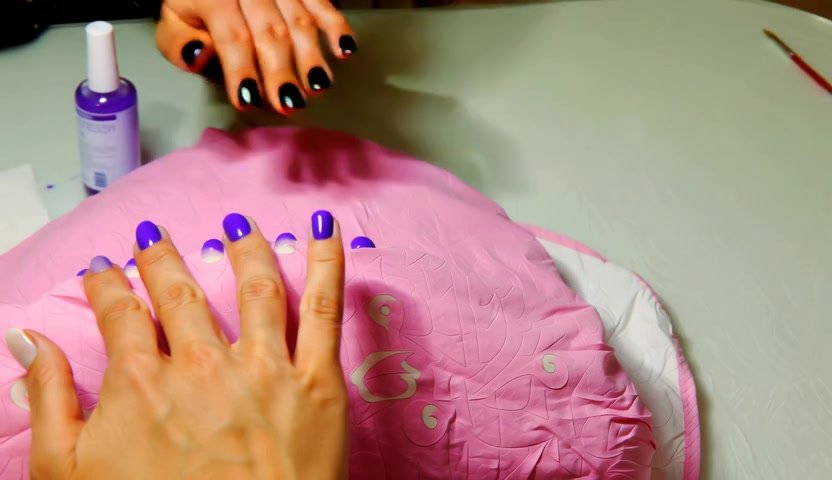}} & \adjustbox{valign=m}{\includegraphics[width=0.145\textwidth]{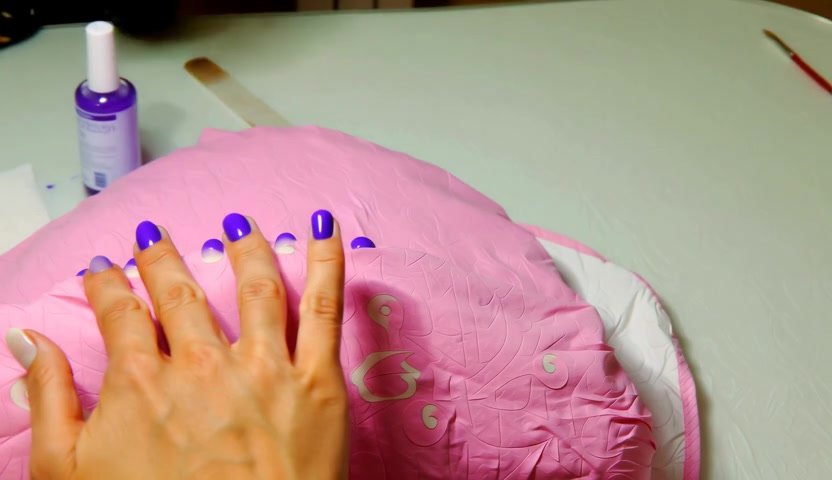}} & \adjustbox{valign=m}{\includegraphics[width=0.145\textwidth]{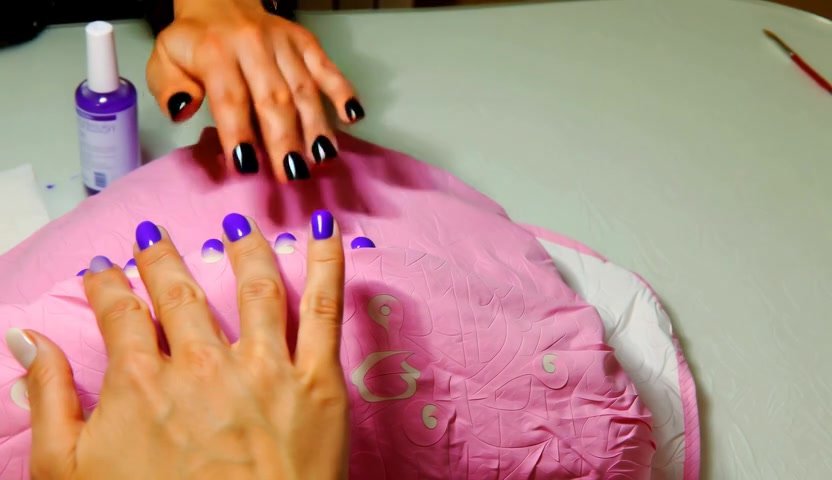}} \\[2pt]
        \QualRowLabel{AnyFlow\checkpointmark} &
        \adjustbox{valign=m}{\includegraphics[width=0.145\textwidth]{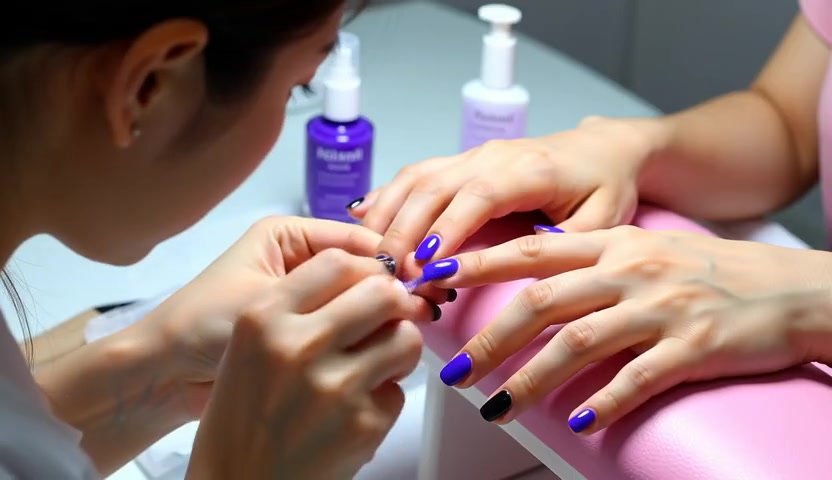}} & \adjustbox{valign=m}{\includegraphics[width=0.145\textwidth]{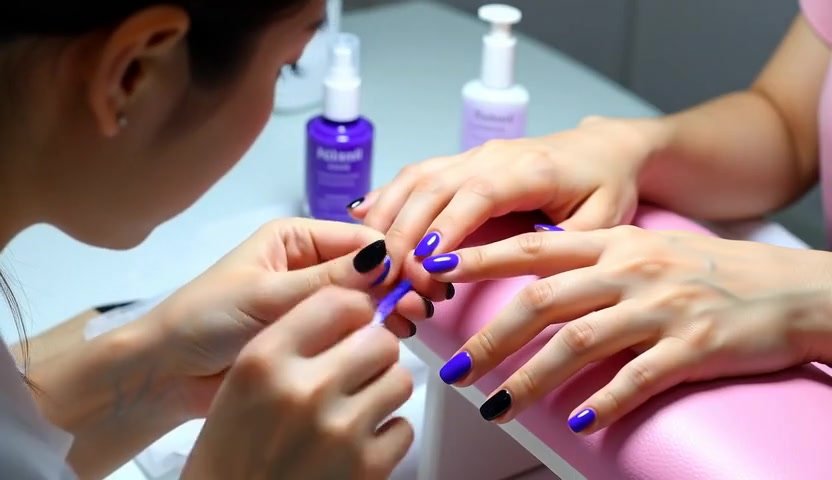}} & \adjustbox{valign=m}{\includegraphics[width=0.145\textwidth]{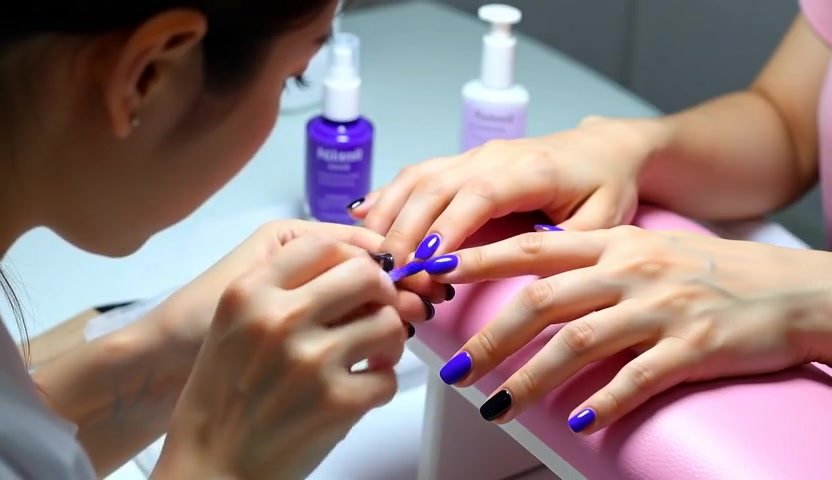}} & \adjustbox{valign=m}{\includegraphics[width=0.145\textwidth]{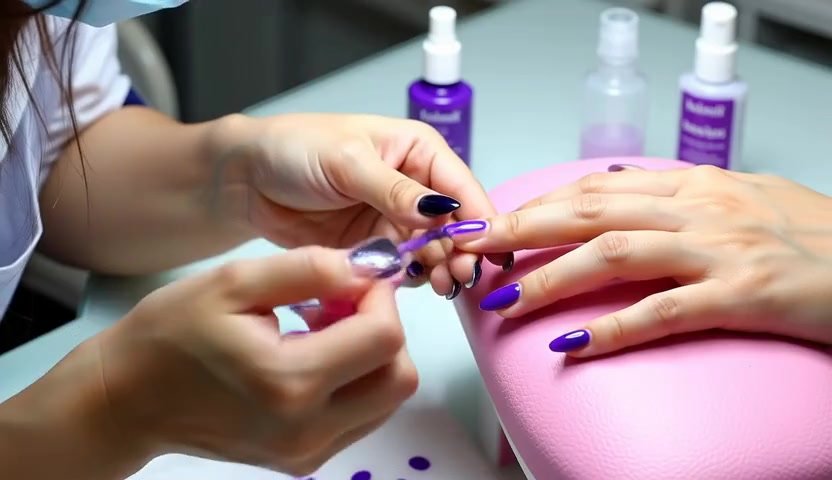}} & \adjustbox{valign=m}{\includegraphics[width=0.145\textwidth]{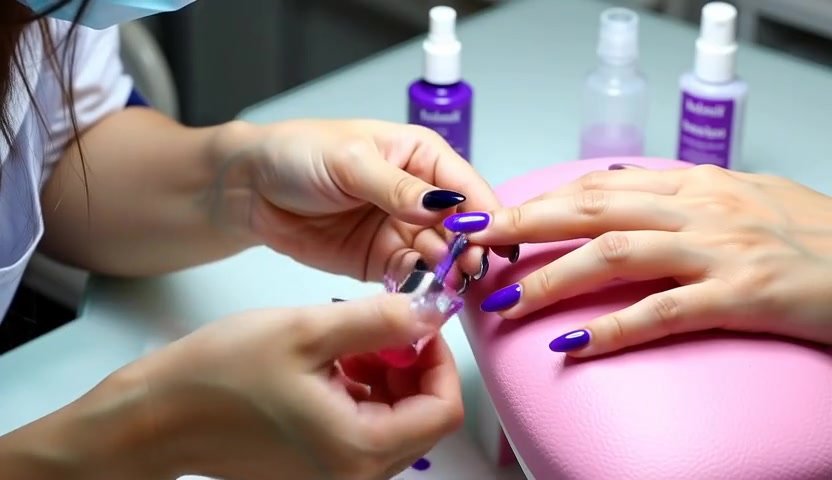}} & \adjustbox{valign=m}{\includegraphics[width=0.145\textwidth]{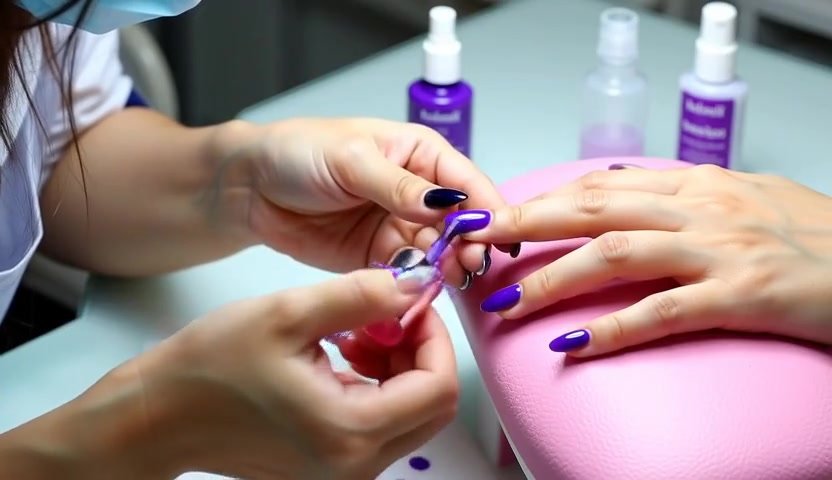}} \\[2pt]
        \QualRowLabel{\makecell{{\Oursbf{}} \\ {(Ours)}}} &
        \adjustbox{valign=m}{\includegraphics[width=0.145\textwidth]{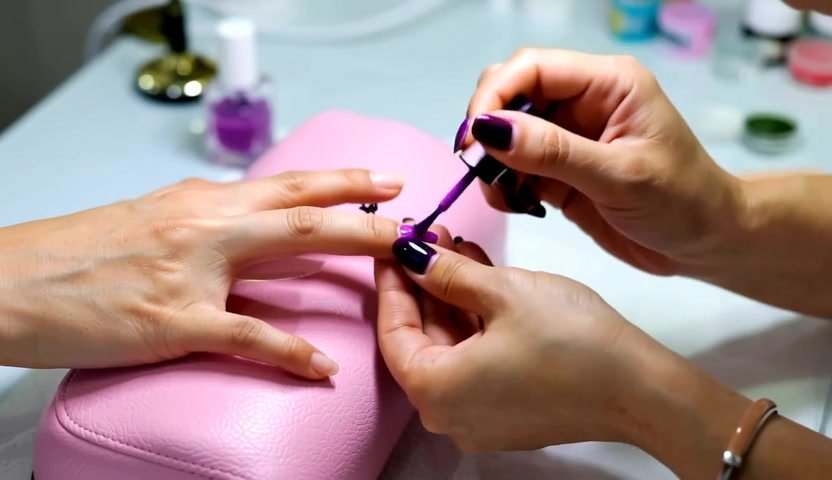}} & \adjustbox{valign=m}{\includegraphics[width=0.145\textwidth]{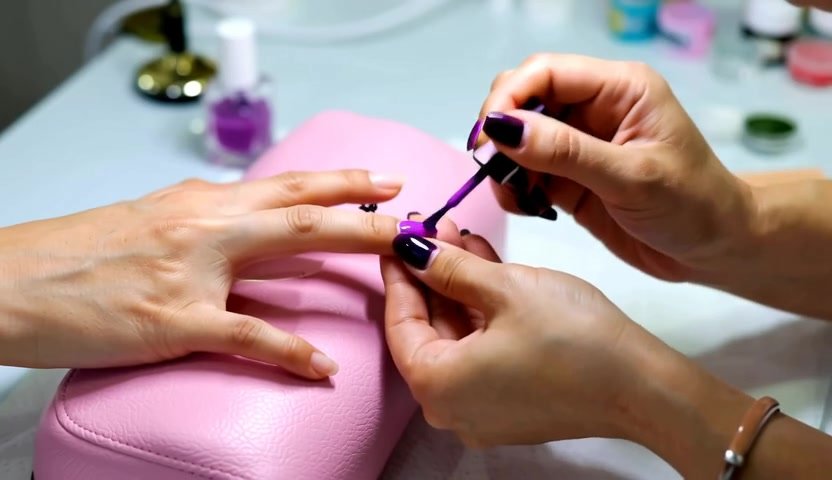}} & \adjustbox{valign=m}{\includegraphics[width=0.145\textwidth]{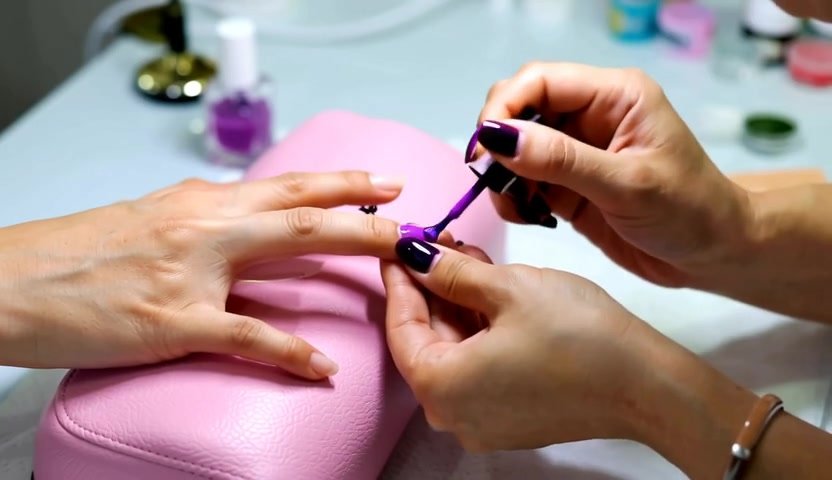}} & \adjustbox{valign=m}{\includegraphics[width=0.145\textwidth]{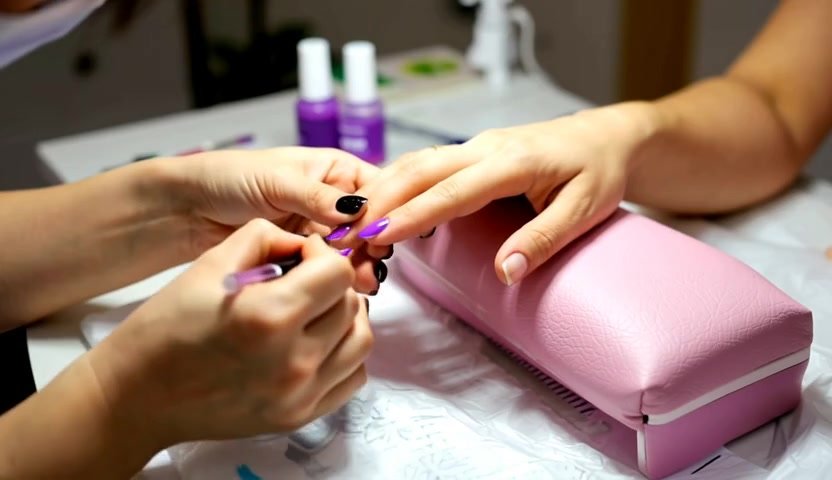}} & \adjustbox{valign=m}{\includegraphics[width=0.145\textwidth]{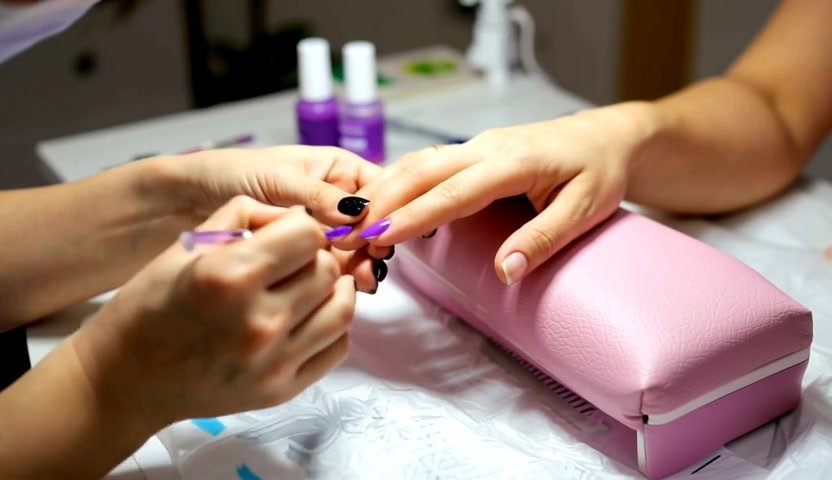}} & \adjustbox{valign=m}{\includegraphics[width=0.145\textwidth]{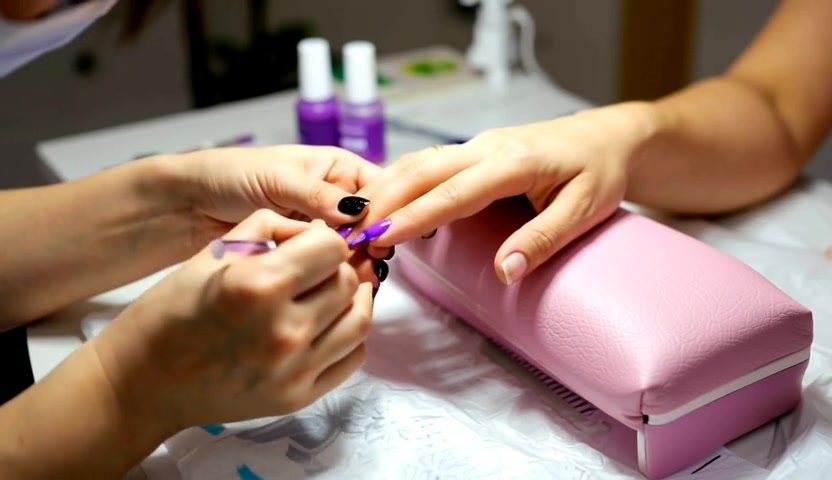}} \\[16pt]
        & \multicolumn{6}{c}{\parbox{0.90\textwidth}{\centering\tiny\itshape ``In this video clip, a Korean manicurist is meticulously working on a client's nails. The manicurist's hands, adorned with dark nail polish, are carefully applying a vibrant purple nail polish to the client's fingernails. The client's hands, resting on a pink cushion, are positioned to allow easy access for the manicurist. A bottle of nail polish remover is visible in the background, suggesting a professional setting. The interaction between the manicurist and the client is gentle and precise, highlighting the skill and care involved in the process. The scene captures the intimate and focused nature of a nail treatment session, emphasizing the attention to detail and the personal touch provided by the manicurist.''}} \\[28pt]
        \QualRowLabel{DMD2} &
        \adjustbox{valign=m}{\includegraphics[width=0.145\textwidth]{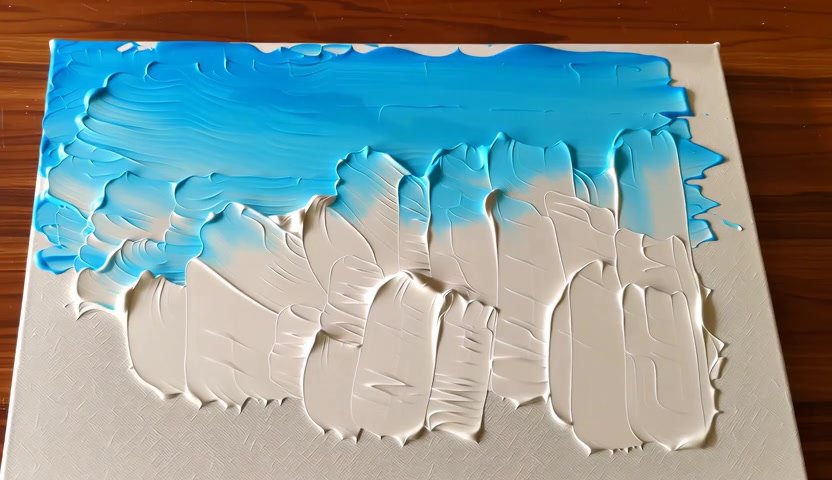}} & \adjustbox{valign=m}{\includegraphics[width=0.145\textwidth]{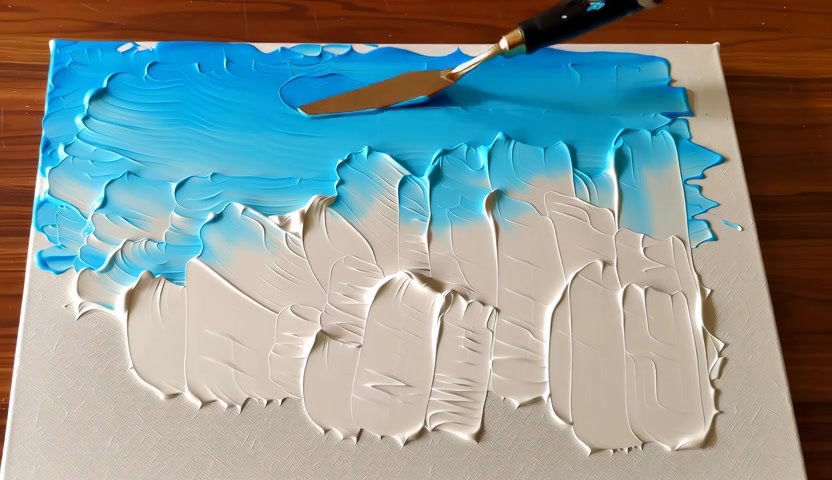}} & \adjustbox{valign=m}{\includegraphics[width=0.145\textwidth]{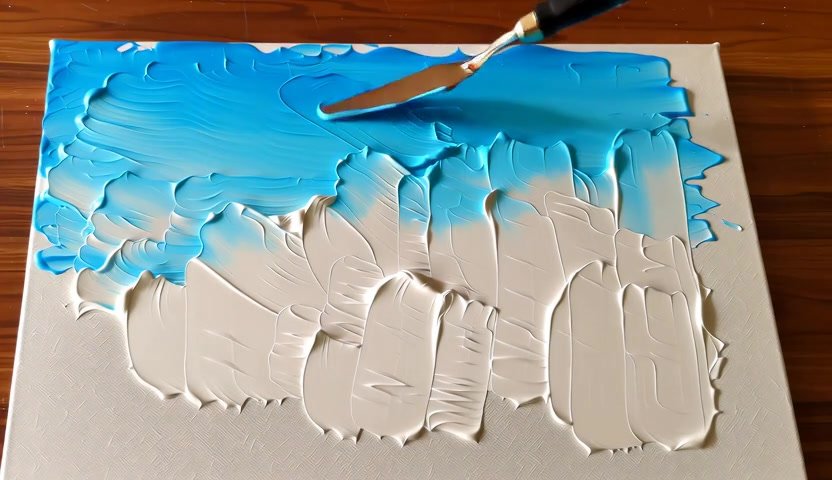}} & \adjustbox{valign=m}{\includegraphics[width=0.145\textwidth]{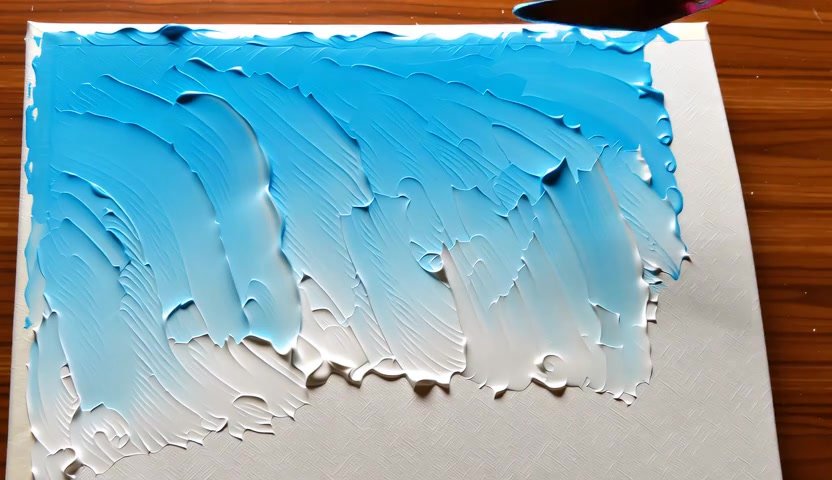}} & \adjustbox{valign=m}{\includegraphics[width=0.145\textwidth]{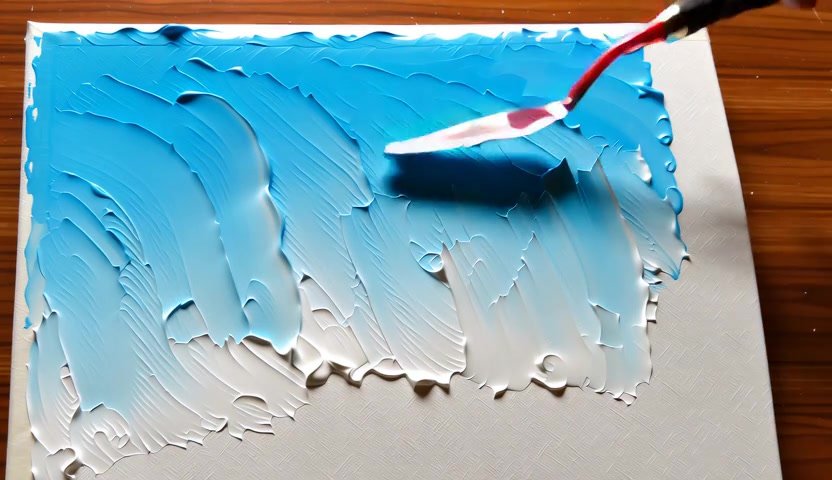}} & \adjustbox{valign=m}{\includegraphics[width=0.145\textwidth]{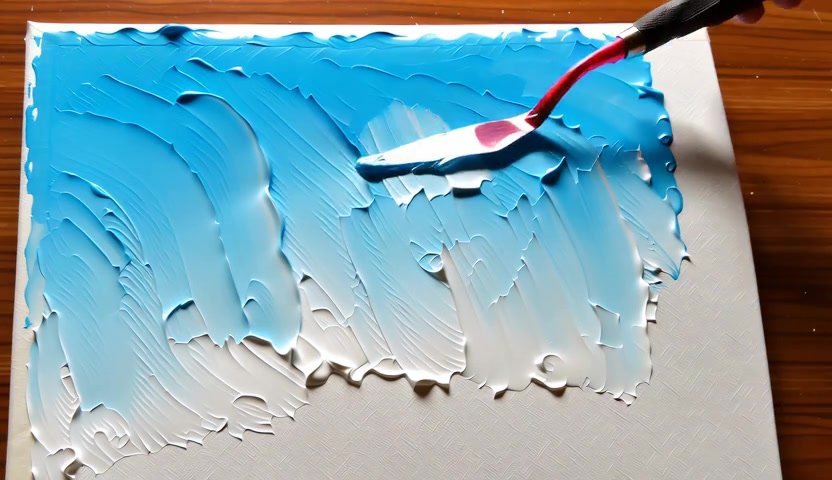}} \\[2pt]
        \QualRowLabel{AnyFlow\checkpointmark} &
        \adjustbox{valign=m}{\includegraphics[width=0.145\textwidth]{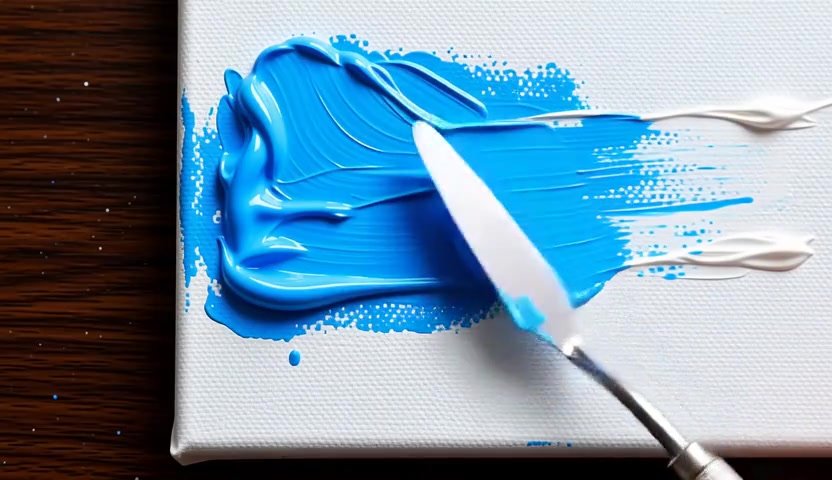}} & \adjustbox{valign=m}{\includegraphics[width=0.145\textwidth]{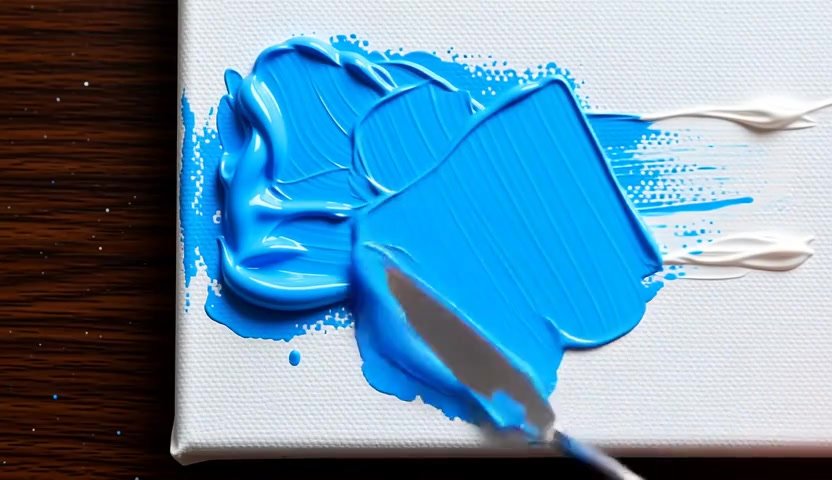}} & \adjustbox{valign=m}{\includegraphics[width=0.145\textwidth]{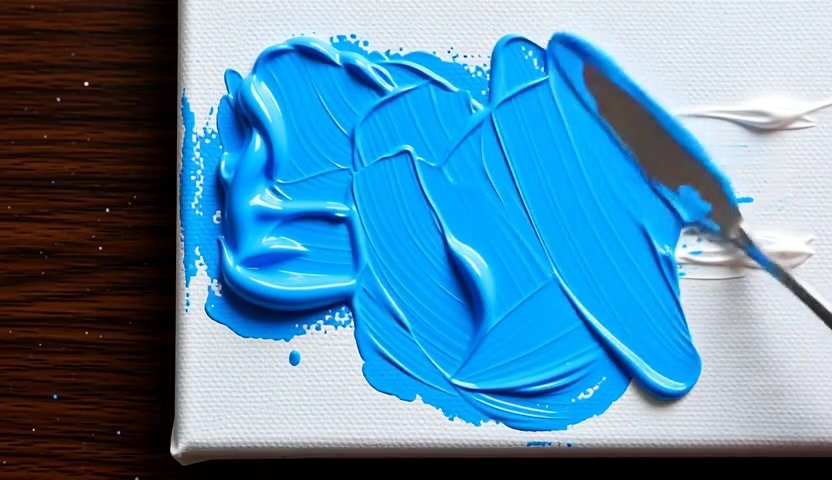}} & \adjustbox{valign=m}{\includegraphics[width=0.145\textwidth]{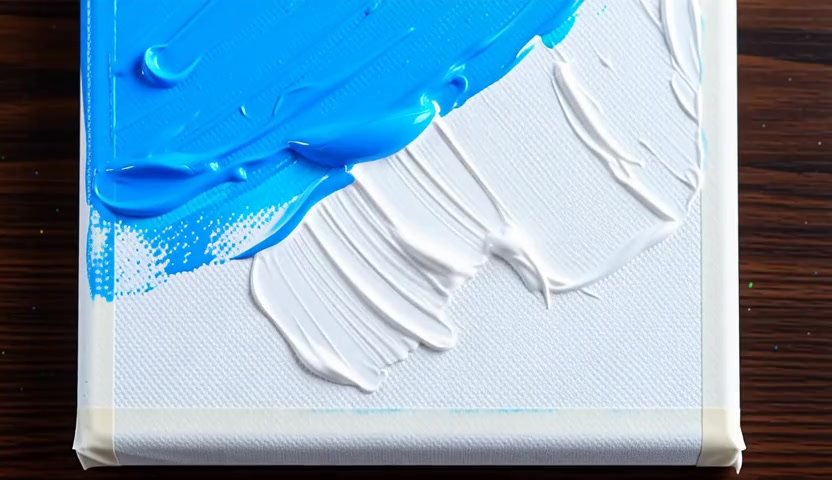}} & \adjustbox{valign=m}{\includegraphics[width=0.145\textwidth]{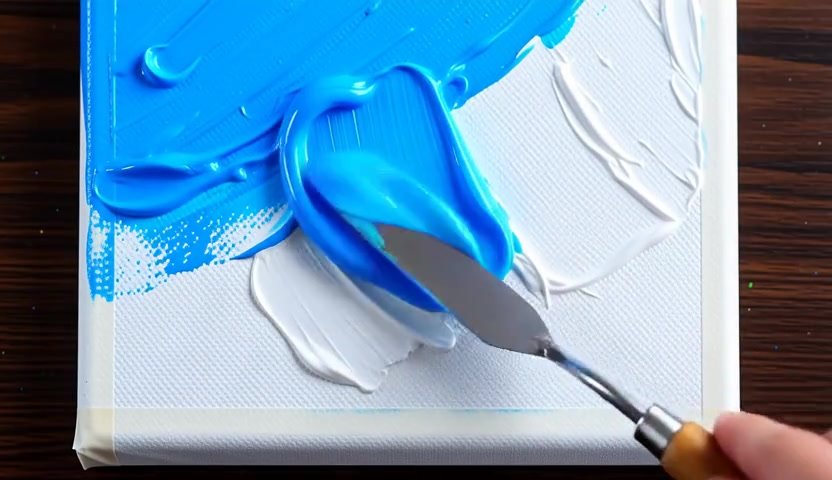}} & \adjustbox{valign=m}{\includegraphics[width=0.145\textwidth]{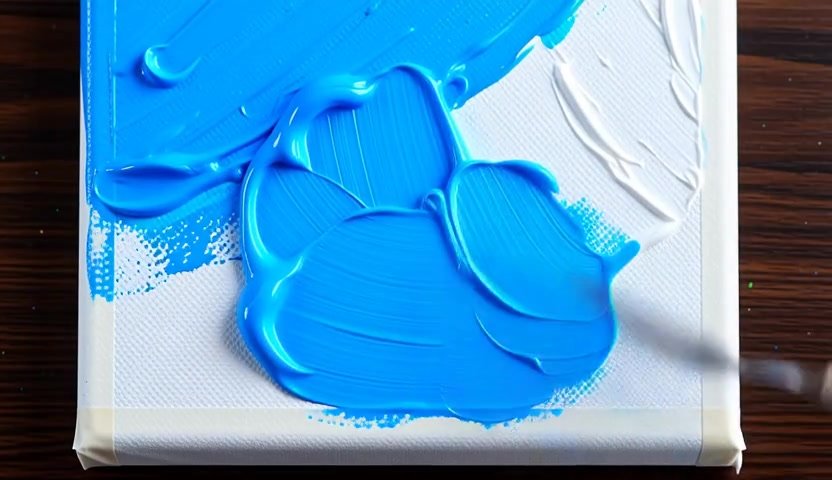}} \\[2pt]
        \QualRowLabel{\makecell{{\Oursbf{}} \\ {(Ours)}}} &
        \adjustbox{valign=m}{\includegraphics[width=0.145\textwidth]{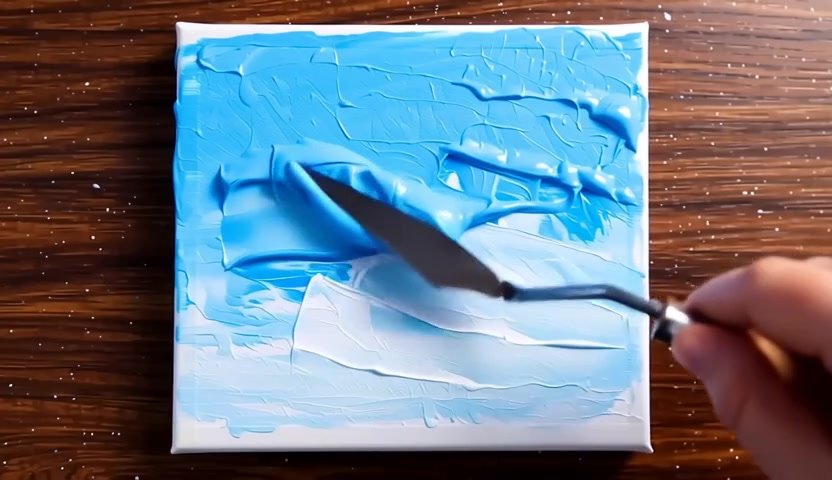}} & \adjustbox{valign=m}{\includegraphics[width=0.145\textwidth]{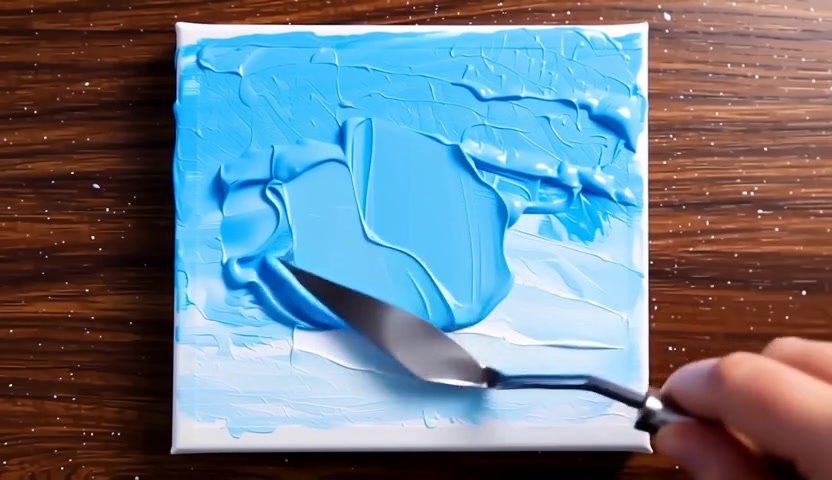}} & \adjustbox{valign=m}{\includegraphics[width=0.145\textwidth]{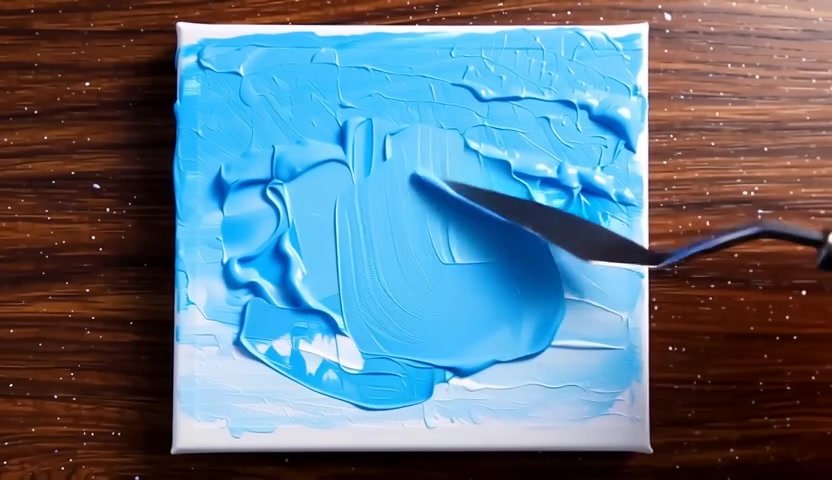}} & \adjustbox{valign=m}{\includegraphics[width=0.145\textwidth]{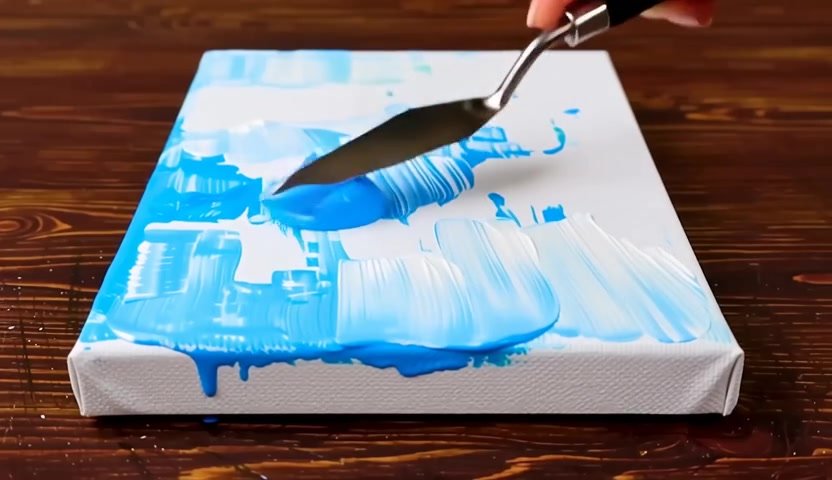}} & \adjustbox{valign=m}{\includegraphics[width=0.145\textwidth]{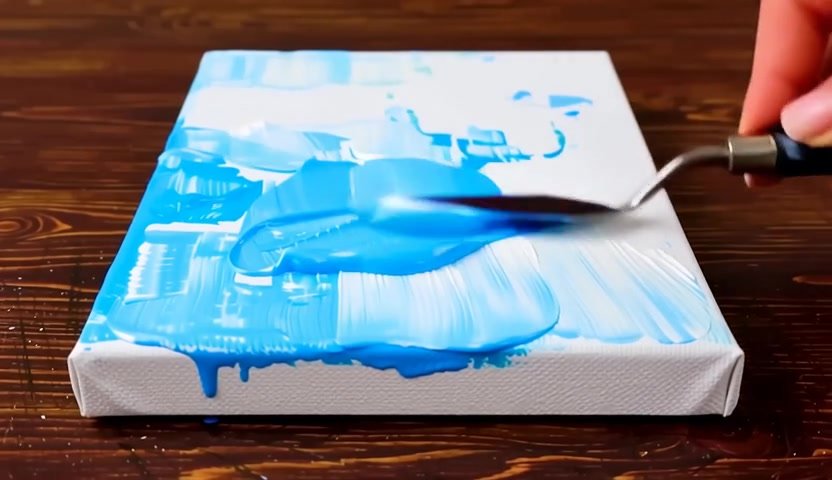}} & \adjustbox{valign=m}{\includegraphics[width=0.145\textwidth]{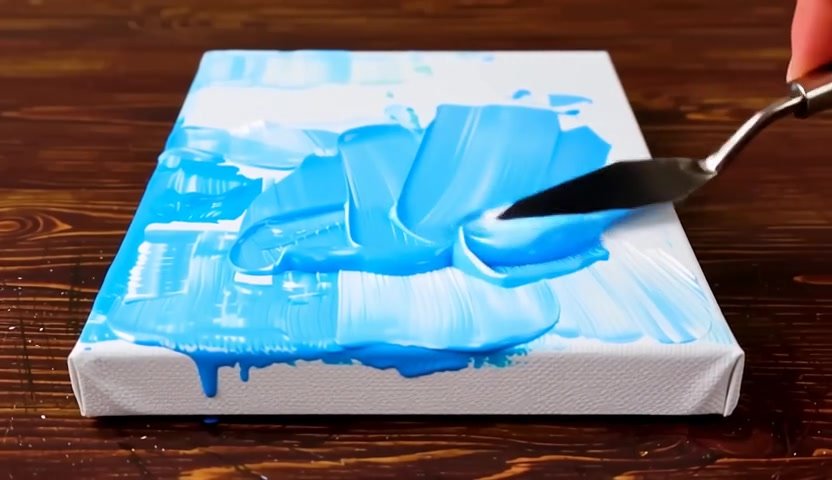}} \\[16pt]
        & \multicolumn{6}{c}{\parbox{0.90\textwidth}{\centering\tiny\itshape ``The video captures a close-up of a textured white canvas with blue paint being applied using a palette knife. The knife moves methodically across the canvas, spreading the paint in broad, sweeping strokes. The paint appears thick and creamy, creating a raised, almost sculptural effect on the surface. The canvas rests on a dark wooden table, which contrasts sharply with the bright white background of the canvas. The deliberate motion of the knife suggests an artist at work, carefully crafting the texture and form of the painting. The overall scene emphasizes the tactile nature of the painting process, highlighting the interplay between the tool, the paint, and the canvas.''}} \\[28pt]
    \end{tabular}
    \end{threeparttable}
    \end{minipage}
\vspace{-3pt}\end{figure}

\begin{figure}[p]\vspace{-3pt}
    \centering
    \begin{minipage}{\textwidth}
    \centering
    \begin{threeparttable}
    \captionsetup{font=normalsize}
    \caption{
        \textbf{Additional Qualitative Comparison on Wan 14B (Continued).}
        Each method row shows two video samples for the prompt below.
        \checkpointmark{} denotes an officially released checkpoint.
    }
    \label{fig:wan_qual_video_2}
    \setlength{\tabcolsep}{1.5pt}
    \renewcommand{\arraystretch}{1.05}
    \scriptsize
    \begin{tabular}{@{}c ccc@{\hspace{8pt}}ccc@{}}
        \QualRowLabel{DMD2} &
        \adjustbox{valign=m}{\includegraphics[width=0.145\textwidth]{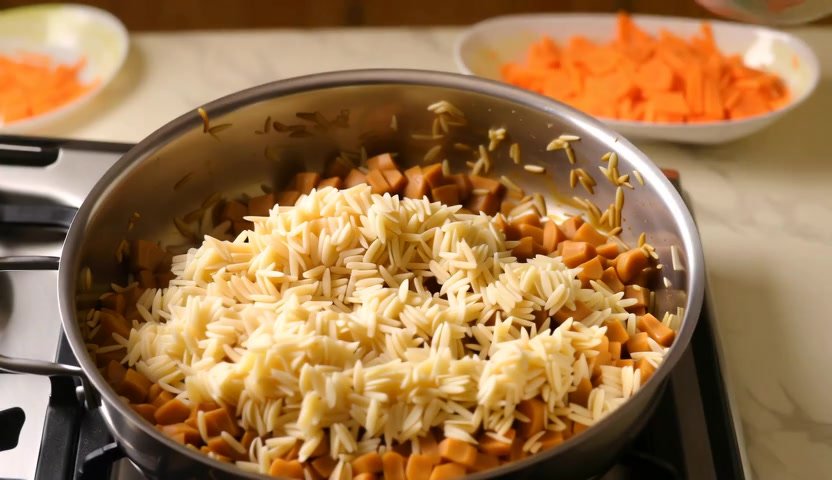}} & \adjustbox{valign=m}{\includegraphics[width=0.145\textwidth]{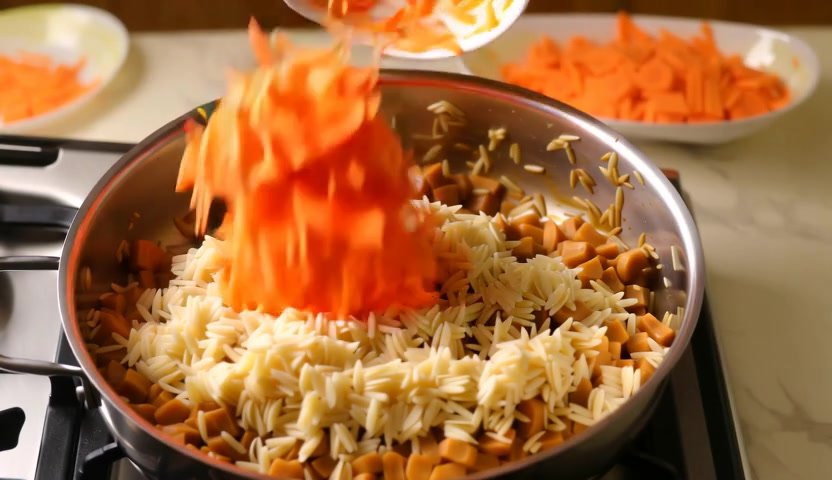}} & \adjustbox{valign=m}{\includegraphics[width=0.145\textwidth]{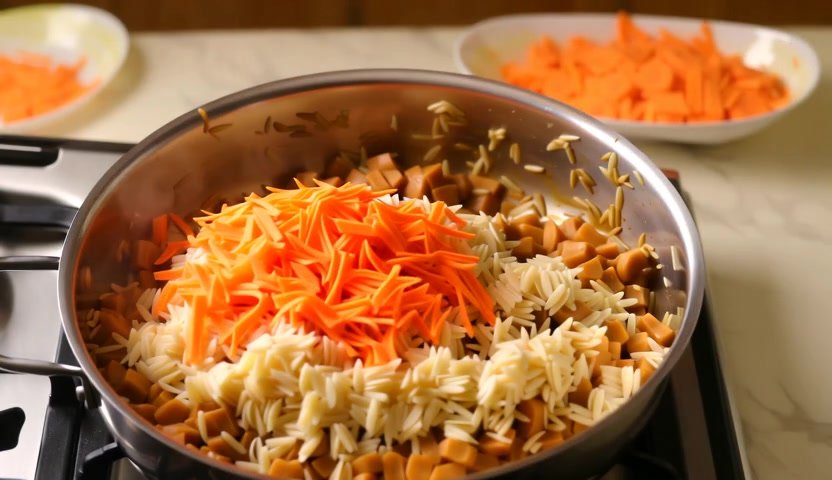}} & \adjustbox{valign=m}{\includegraphics[width=0.145\textwidth]{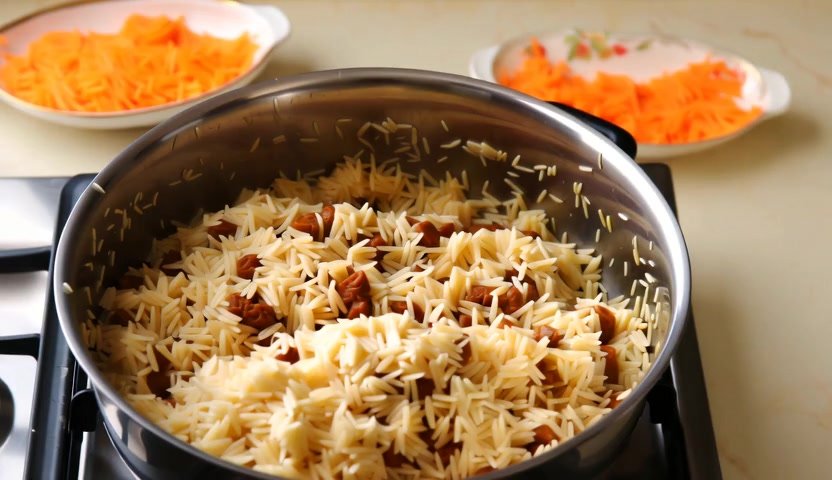}} & \adjustbox{valign=m}{\includegraphics[width=0.145\textwidth]{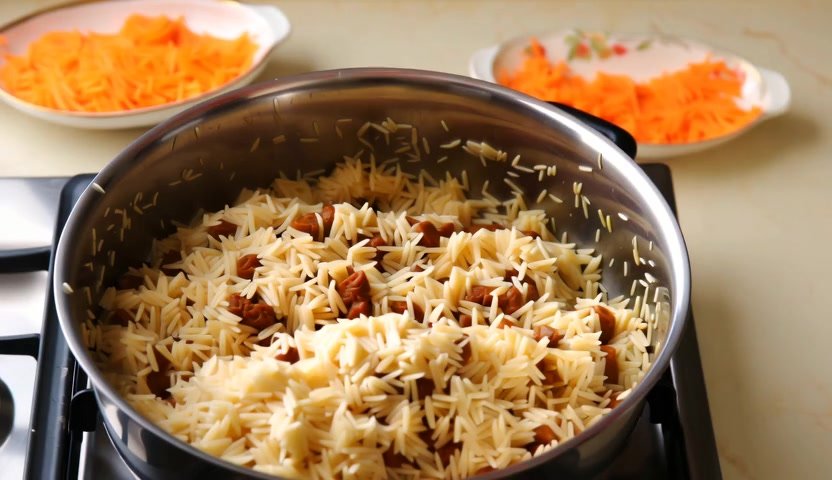}} & \adjustbox{valign=m}{\includegraphics[width=0.145\textwidth]{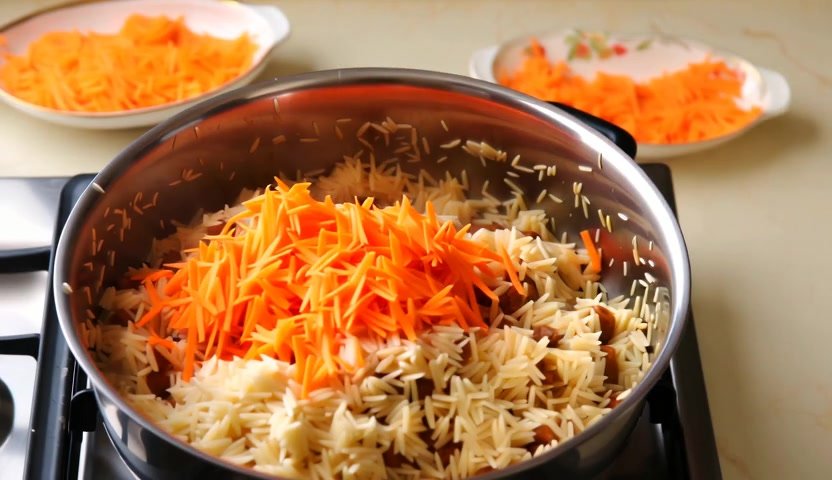}} \\[2pt]
        \QualRowLabel{AnyFlow\checkpointmark} &
        \adjustbox{valign=m}{\includegraphics[width=0.145\textwidth]{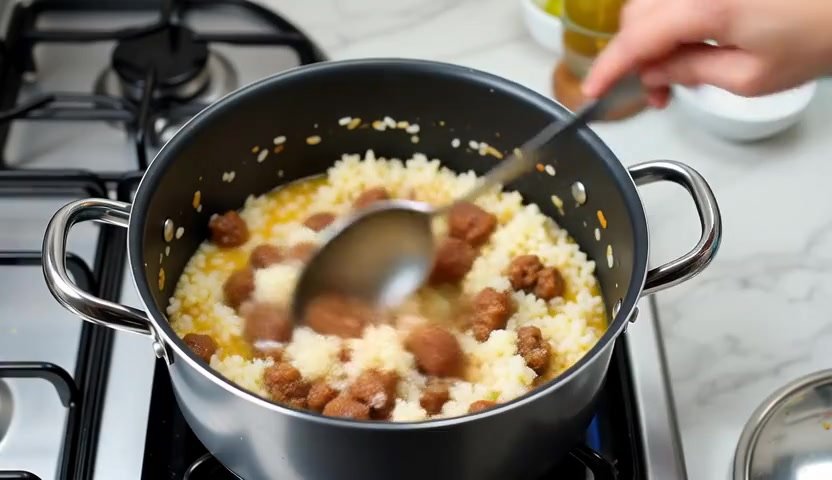}} & \adjustbox{valign=m}{\includegraphics[width=0.145\textwidth]{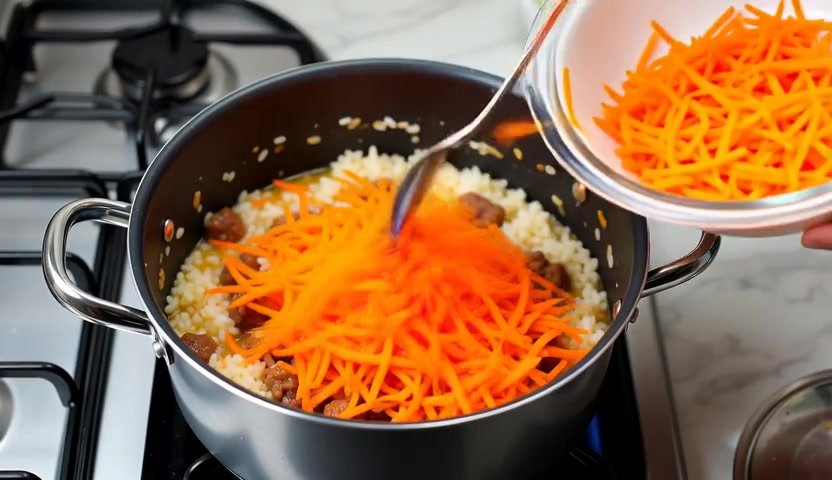}} & \adjustbox{valign=m}{\includegraphics[width=0.145\textwidth]{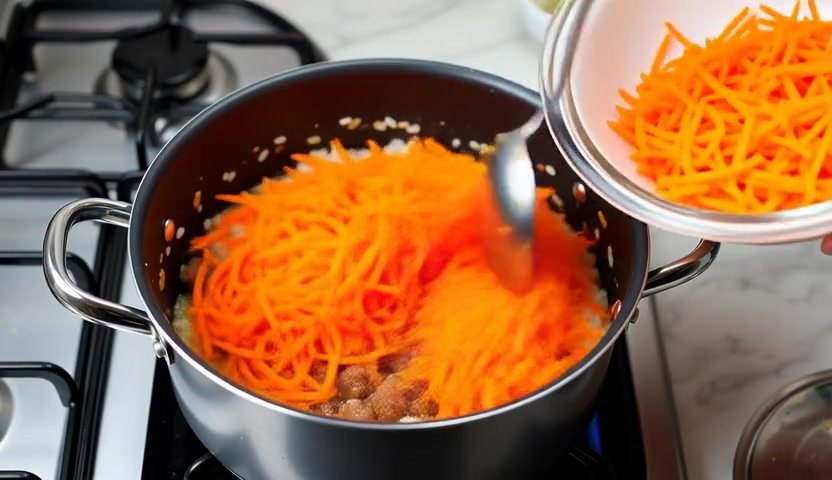}} & \adjustbox{valign=m}{\includegraphics[width=0.145\textwidth]{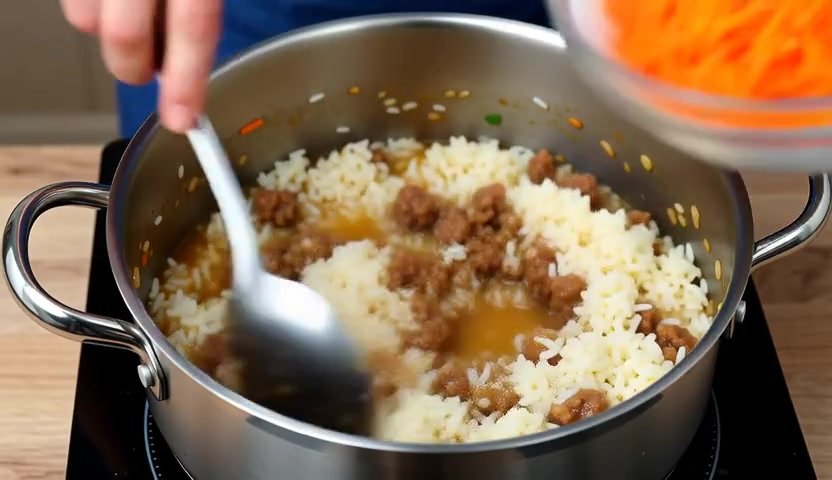}} & \adjustbox{valign=m}{\includegraphics[width=0.145\textwidth]{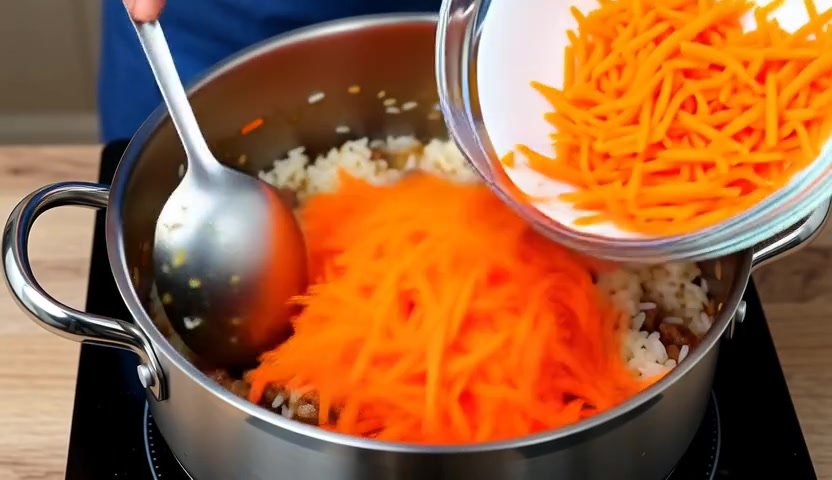}} & \adjustbox{valign=m}{\includegraphics[width=0.145\textwidth]{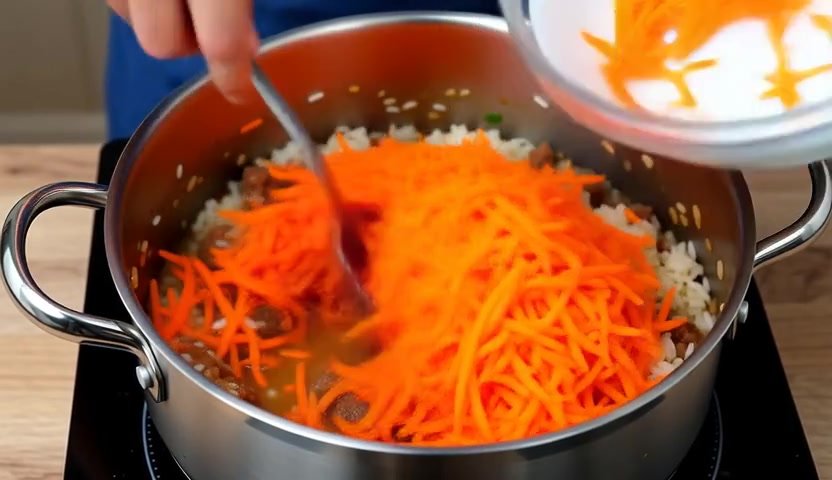}} \\[2pt]
        \QualRowLabel{\makecell{{\Oursbf{}} \\ {(Ours)}}} &
        \adjustbox{valign=m}{\includegraphics[width=0.145\textwidth]{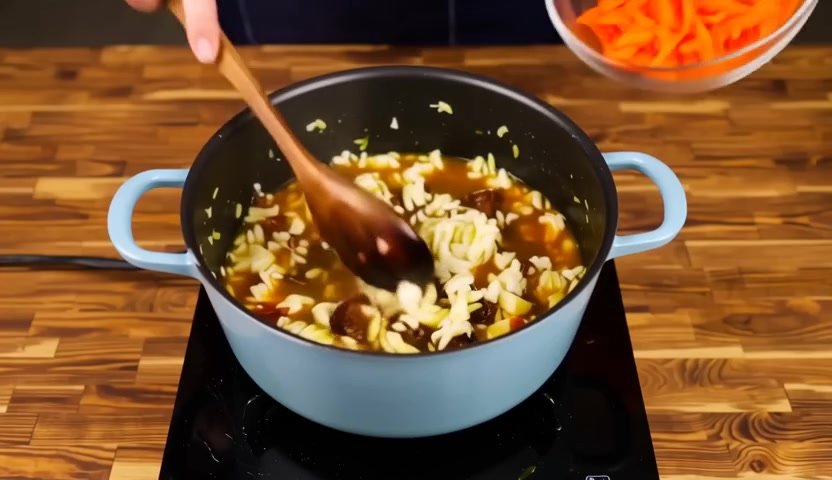}} & \adjustbox{valign=m}{\includegraphics[width=0.145\textwidth]{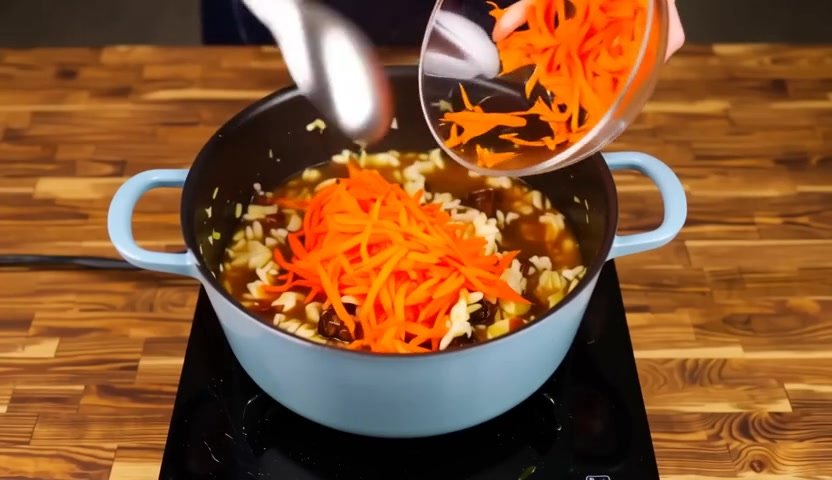}} & \adjustbox{valign=m}{\includegraphics[width=0.145\textwidth]{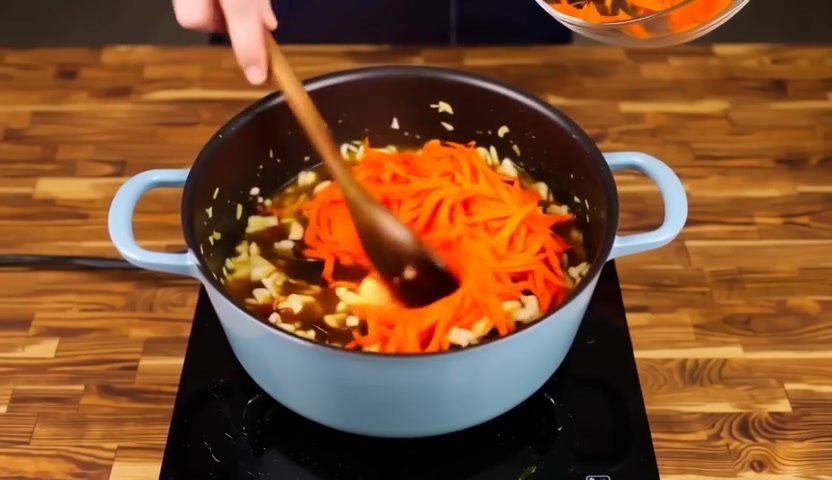}} & \adjustbox{valign=m}{\includegraphics[width=0.145\textwidth]{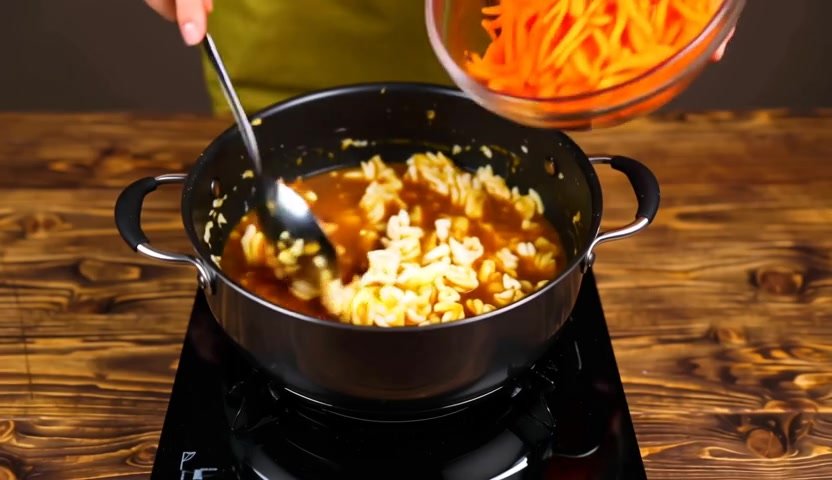}} & \adjustbox{valign=m}{\includegraphics[width=0.145\textwidth]{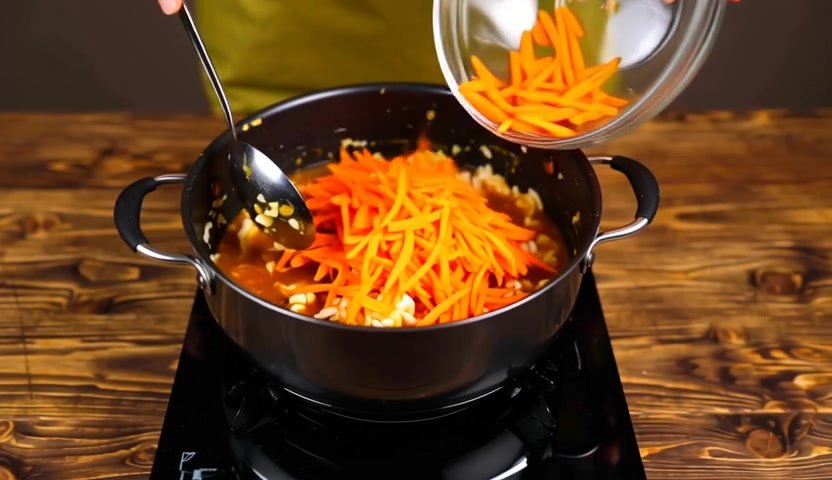}} & \adjustbox{valign=m}{\includegraphics[width=0.145\textwidth]{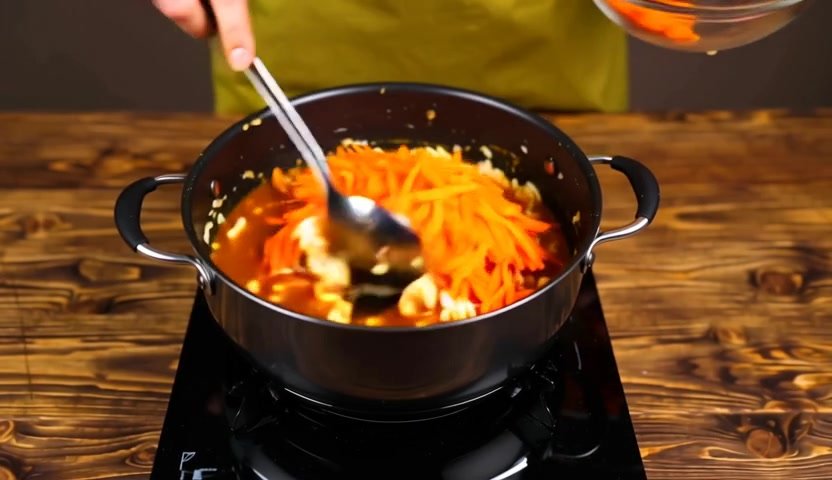}} \\[16pt]
        & \multicolumn{6}{c}{\parbox{0.90\textwidth}{\centering\tiny\itshape ``In the video, a person is seen stirring a pot on a stovetop with a spoon. The pot contains a mixture of what appears to be rice and possibly meat, creating a base for a dish. The individual then adds a generous portion of shredded carrots from a separate bowl into the pot. The carrots are bright orange and appear freshly grated, contrasting with the white rice and brownish hue of the meat. The person uses a large spoon to mix the ingredients thoroughly, ensuring the carrots are evenly distributed throughout the pot. This action suggests the preparation of a hearty, flavorful meal, combining different textures and colors to create a visually appealing and nutritious dish.''}} \\[28pt]
        \QualRowLabel{DMD2} &
        \adjustbox{valign=m}{\includegraphics[width=0.145\textwidth]{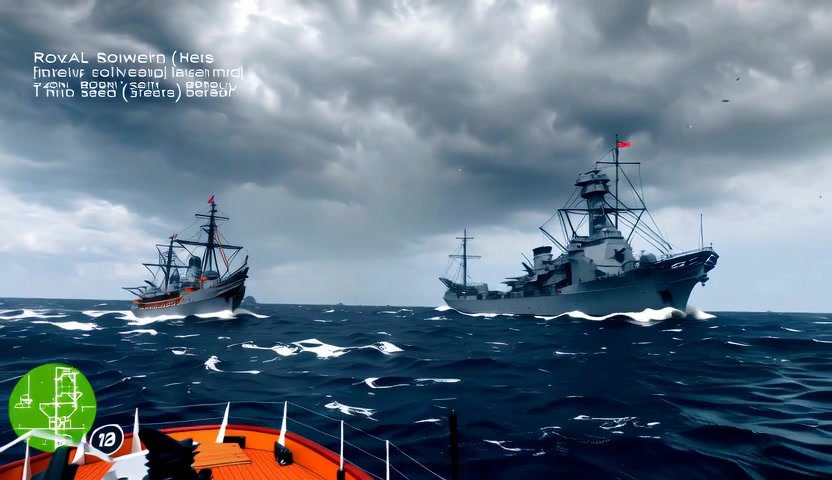}} & \adjustbox{valign=m}{\includegraphics[width=0.145\textwidth]{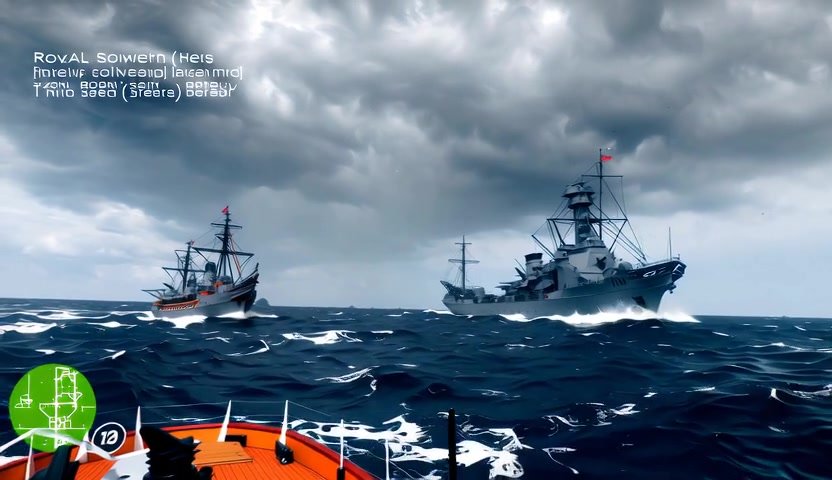}} & \adjustbox{valign=m}{\includegraphics[width=0.145\textwidth]{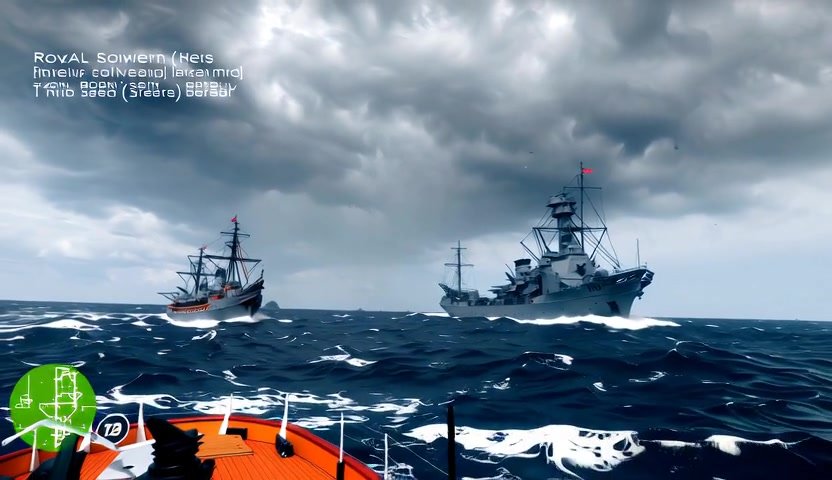}} & \adjustbox{valign=m}{\includegraphics[width=0.145\textwidth]{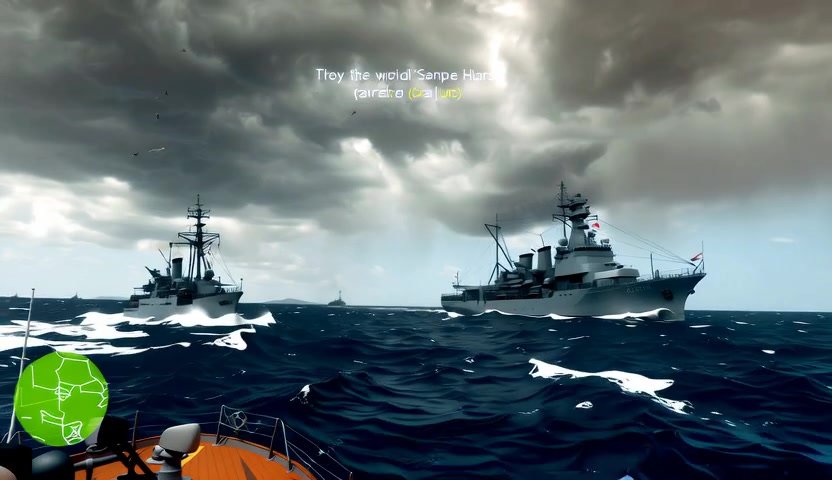}} & \adjustbox{valign=m}{\includegraphics[width=0.145\textwidth]{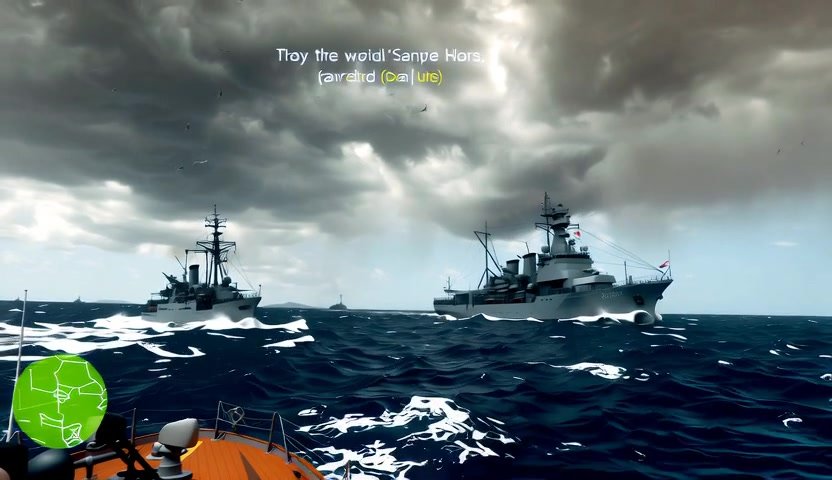}} & \adjustbox{valign=m}{\includegraphics[width=0.145\textwidth]{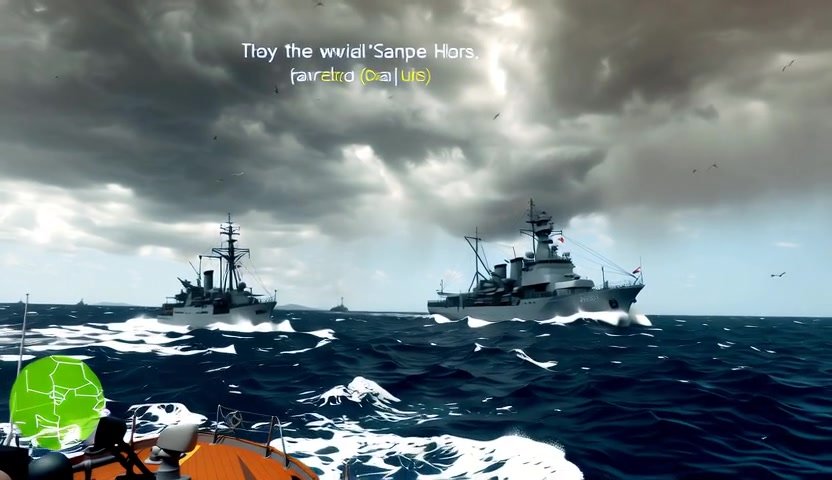}} \\[2pt]
        \QualRowLabel{AnyFlow\checkpointmark} &
        \adjustbox{valign=m}{\includegraphics[width=0.145\textwidth]{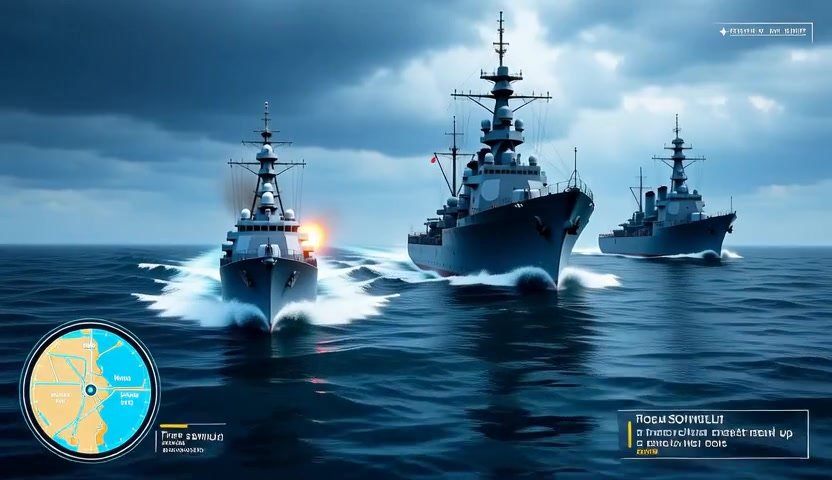}} & \adjustbox{valign=m}{\includegraphics[width=0.145\textwidth]{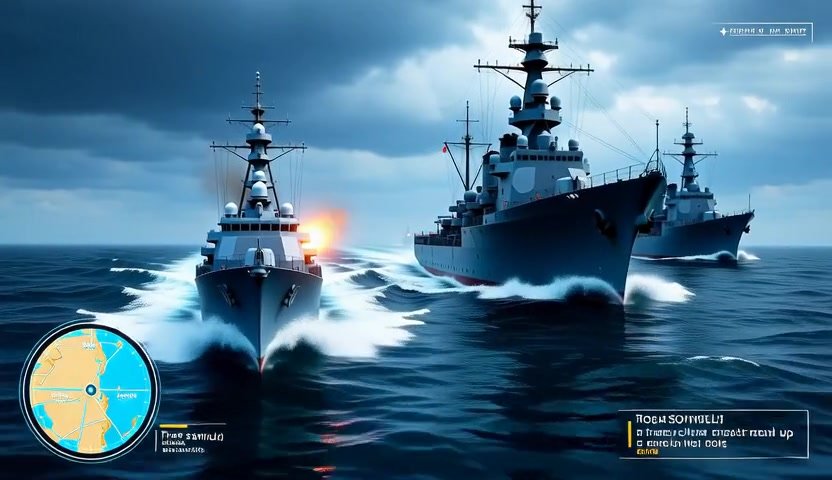}} & \adjustbox{valign=m}{\includegraphics[width=0.145\textwidth]{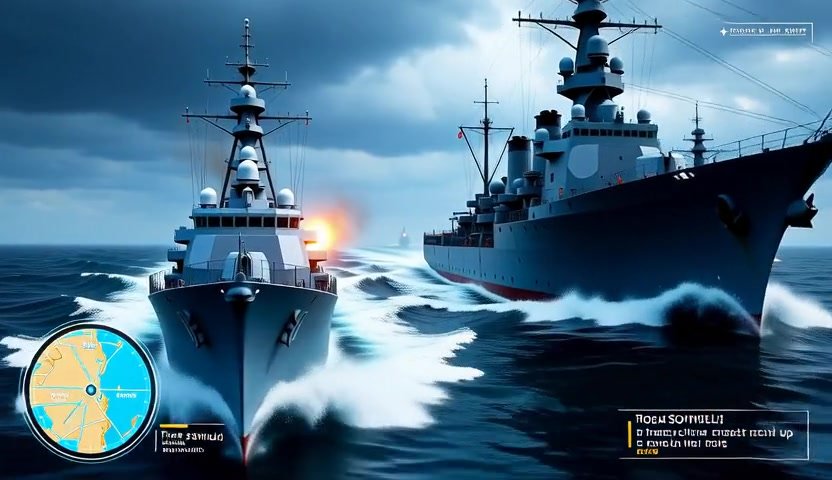}} & \adjustbox{valign=m}{\includegraphics[width=0.145\textwidth]{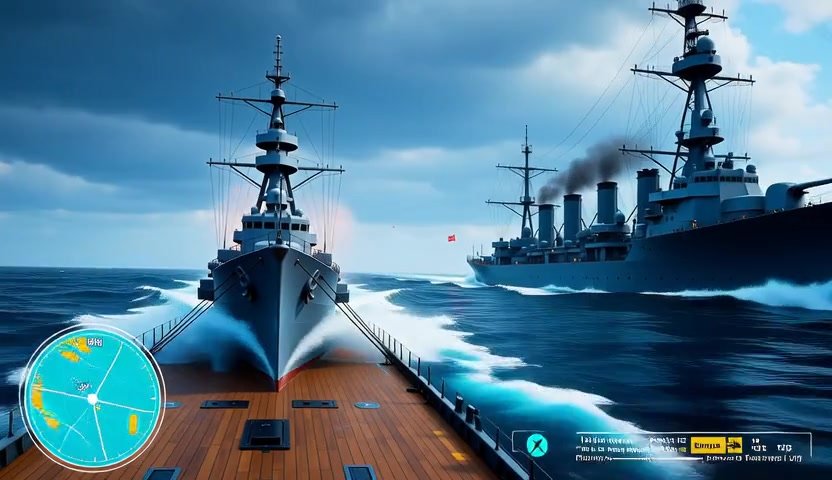}} & \adjustbox{valign=m}{\includegraphics[width=0.145\textwidth]{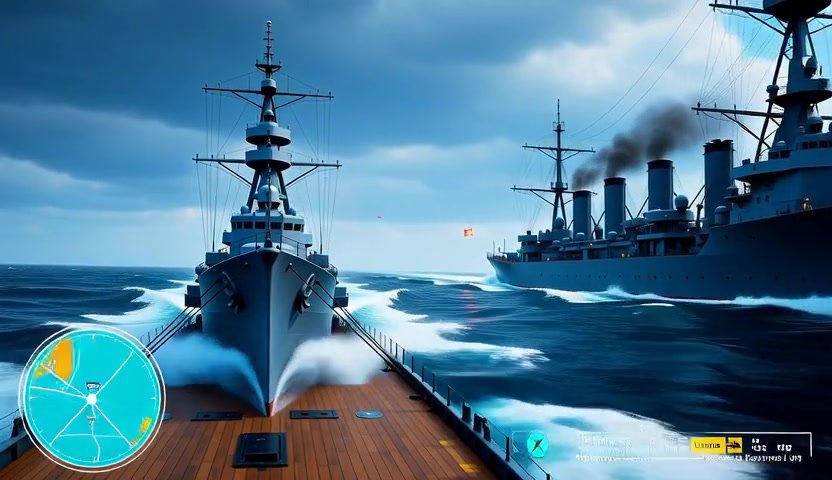}} & \adjustbox{valign=m}{\includegraphics[width=0.145\textwidth]{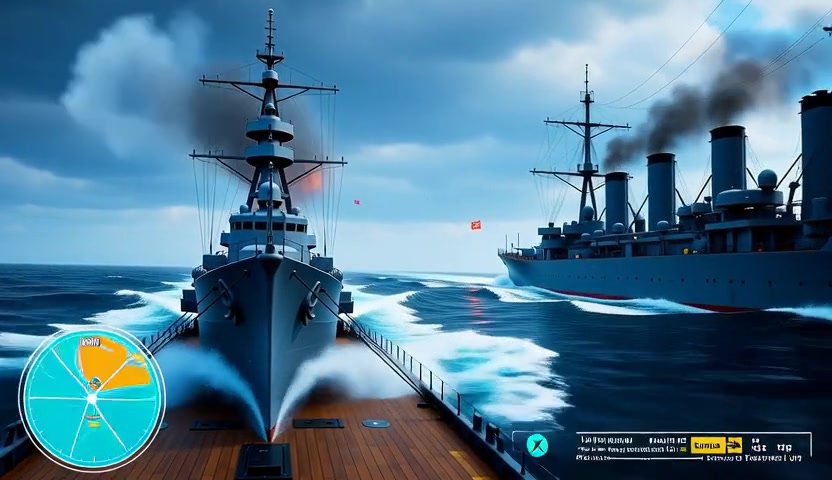}} \\[2pt]
        \QualRowLabel{\makecell{{\Oursbf{}} \\ {(Ours)}}} &
        \adjustbox{valign=m}{\includegraphics[width=0.145\textwidth]{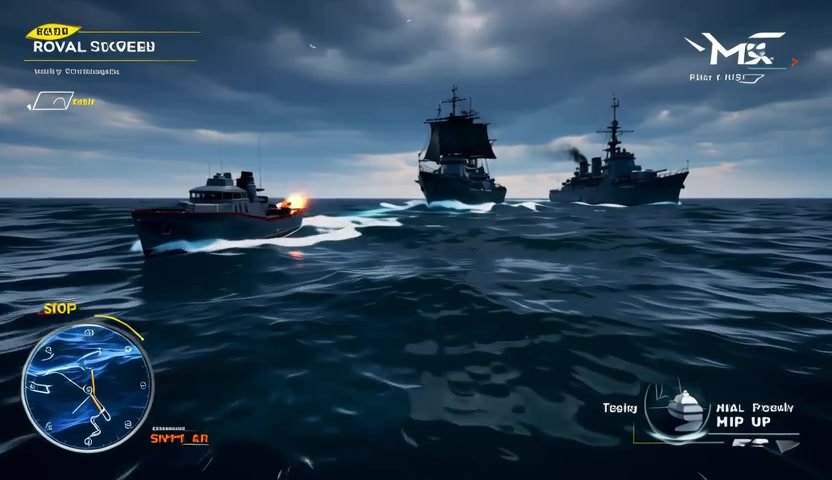}} & \adjustbox{valign=m}{\includegraphics[width=0.145\textwidth]{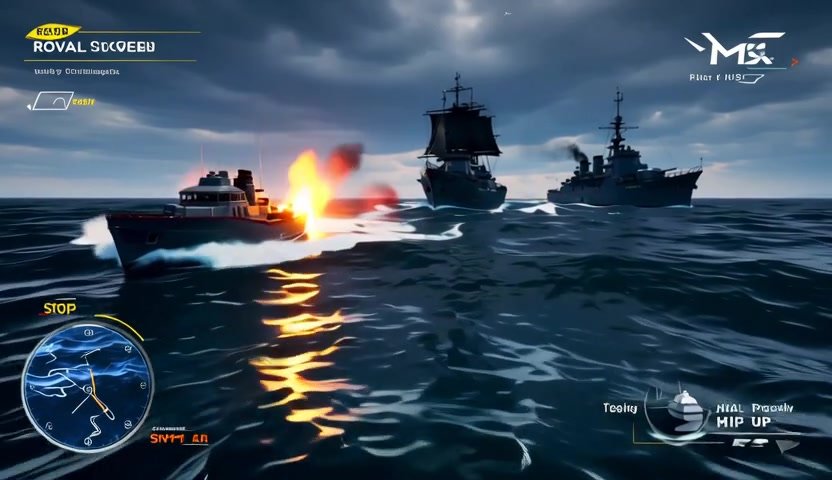}} & \adjustbox{valign=m}{\includegraphics[width=0.145\textwidth]{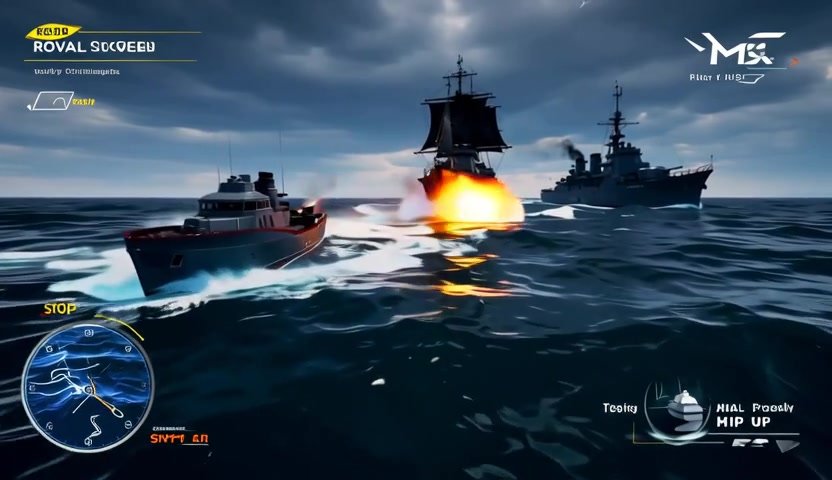}} & \adjustbox{valign=m}{\includegraphics[width=0.145\textwidth]{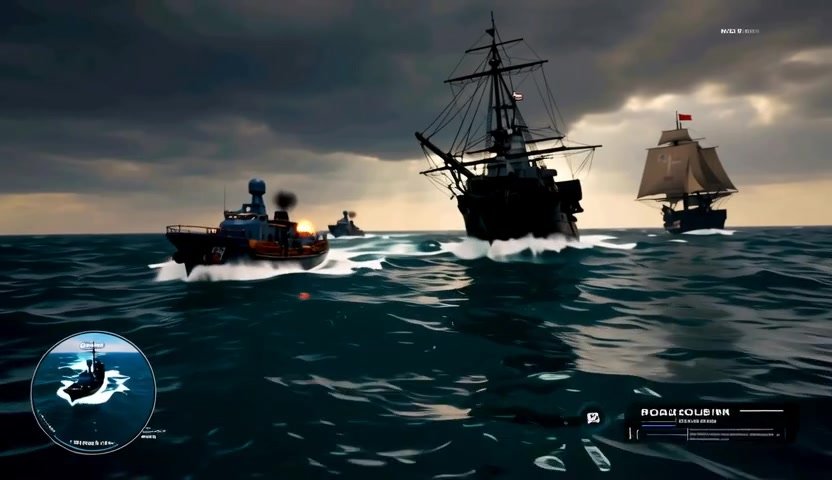}} & \adjustbox{valign=m}{\includegraphics[width=0.145\textwidth]{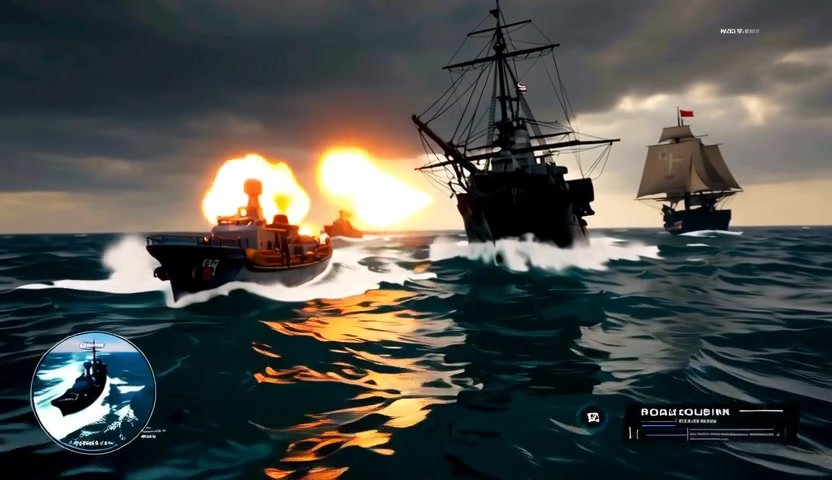}} & \adjustbox{valign=m}{\includegraphics[width=0.145\textwidth]{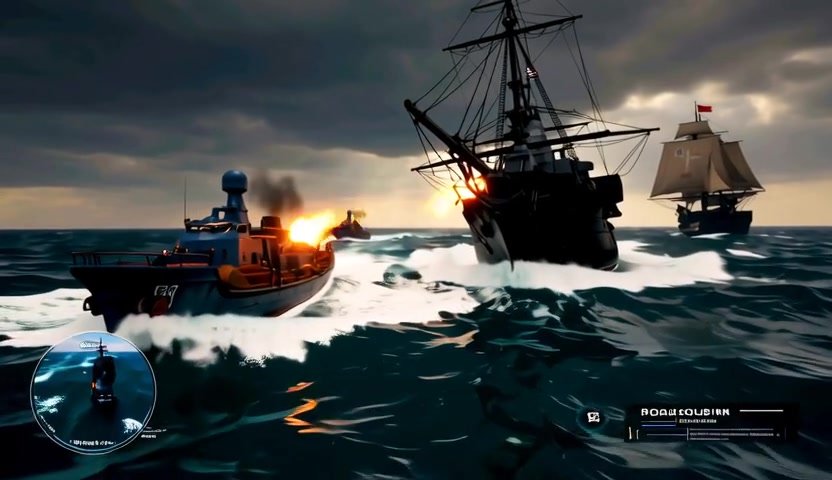}} \\[16pt]
        & \multicolumn{6}{c}{\parbox{0.90\textwidth}{\centering\tiny\itshape ``In this dynamic scene from a video game, a player-controlled ship is engaged in a naval battle against two formidable vessels: the Royal Sovereign and HMS Fearless. The player's ship, positioned on the left, maneuvers skillfully amidst turbulent waves, while the enemy ships, larger and more imposing, dominate the right side of the frame. The player's ship is equipped with various controls visible on-screen, including options for "fire swivels," "brace (hold)," "stop," and "speed up." A circular mini-map at the bottom left corner indicates the player's position and nearby objectives. The text overlay provides instructions on targeting weak points, guiding the player through the combat strategy. The overall atmosphere is intense, with the ships battling under a stormy sky, highlighting the strategic and action-packed nature of the gameplay.''}} \\[28pt]
        \QualRowLabel{DMD2} &
        \adjustbox{valign=m}{\includegraphics[width=0.145\textwidth]{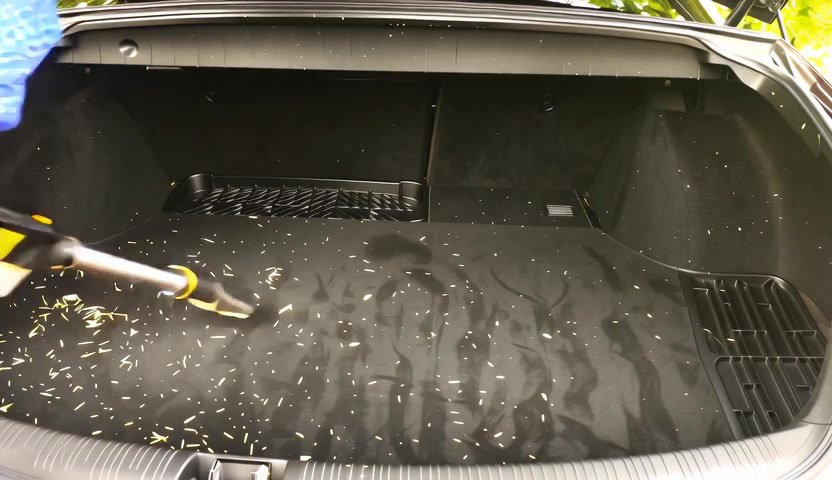}} & \adjustbox{valign=m}{\includegraphics[width=0.145\textwidth]{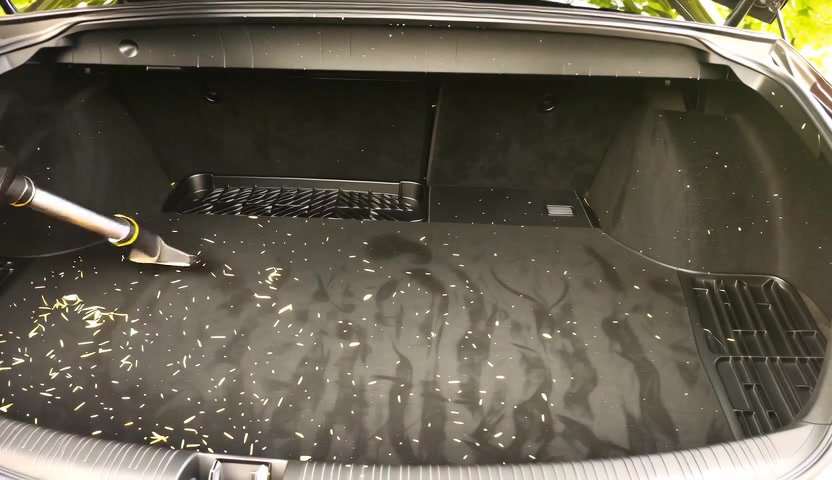}} & \adjustbox{valign=m}{\includegraphics[width=0.145\textwidth]{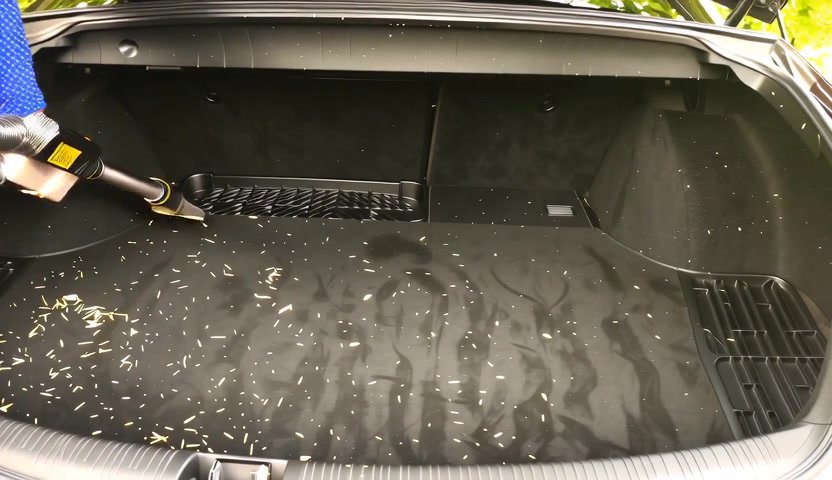}} & \adjustbox{valign=m}{\includegraphics[width=0.145\textwidth]{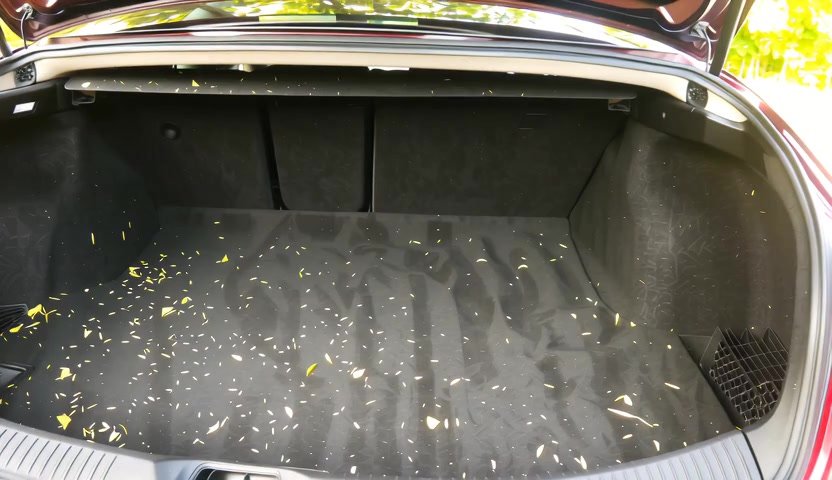}} & \adjustbox{valign=m}{\includegraphics[width=0.145\textwidth]{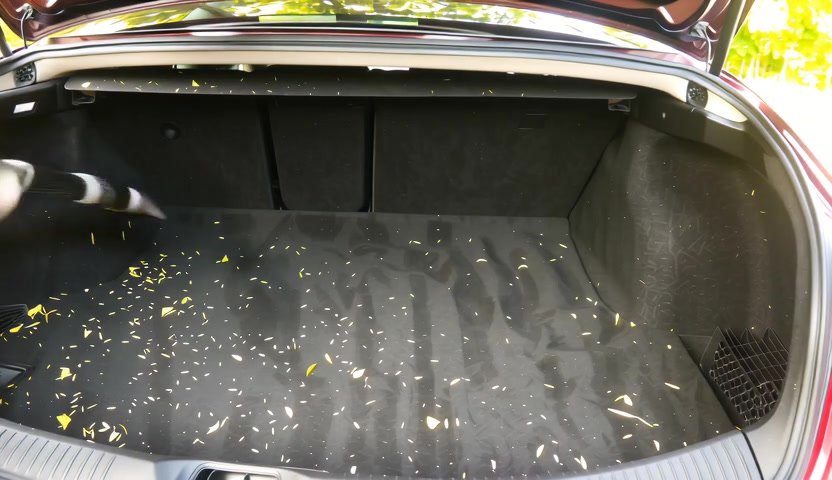}} & \adjustbox{valign=m}{\includegraphics[width=0.145\textwidth]{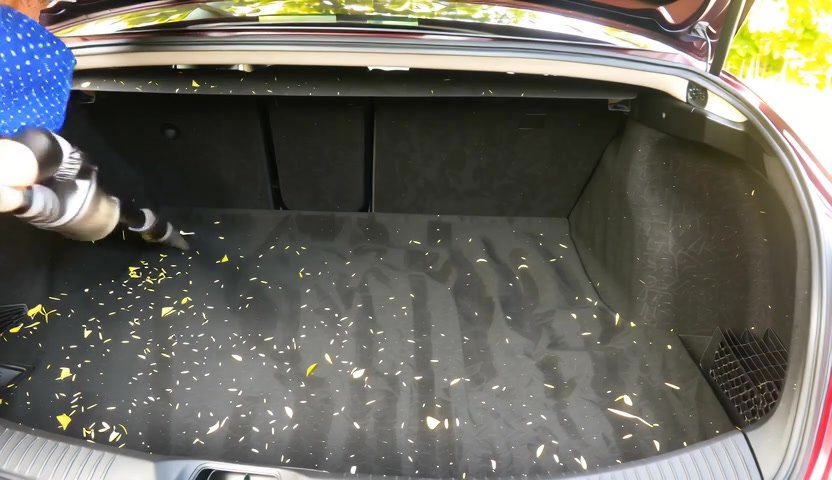}} \\[2pt]
        \QualRowLabel{AnyFlow\checkpointmark} &
        \adjustbox{valign=m}{\includegraphics[width=0.145\textwidth]{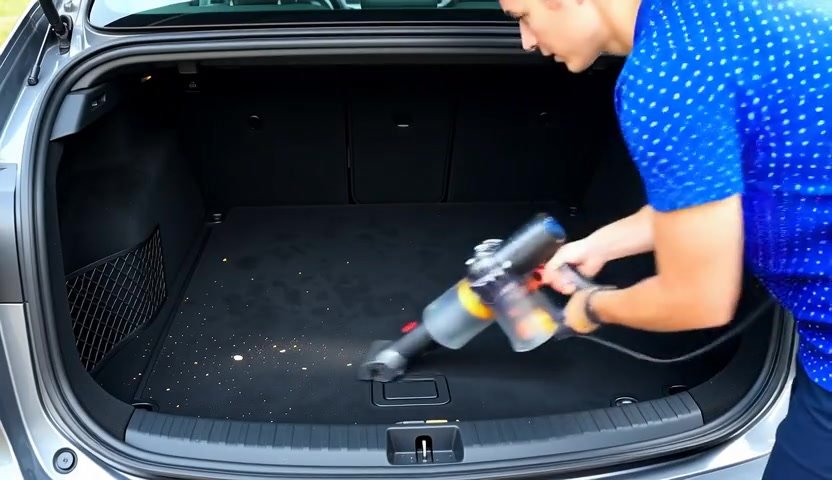}} & \adjustbox{valign=m}{\includegraphics[width=0.145\textwidth]{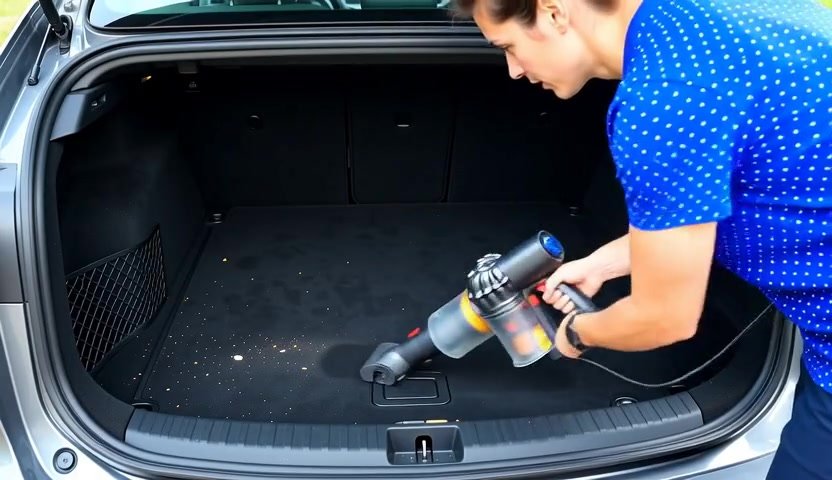}} & \adjustbox{valign=m}{\includegraphics[width=0.145\textwidth]{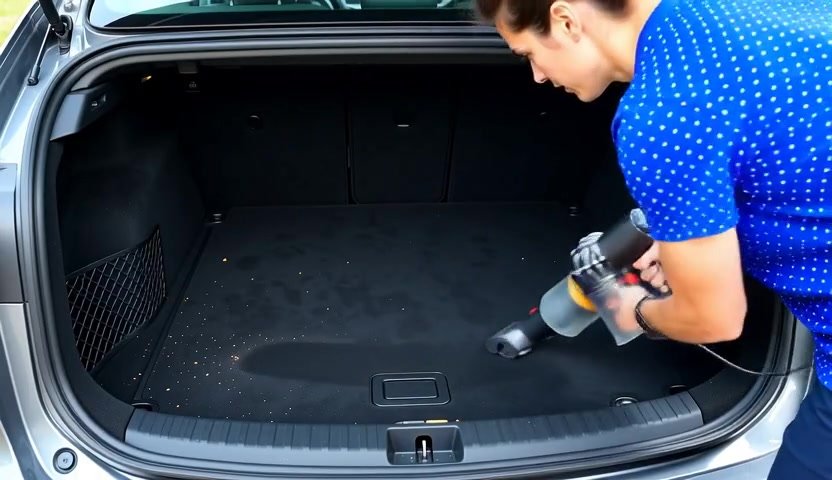}} & \adjustbox{valign=m}{\includegraphics[width=0.145\textwidth]{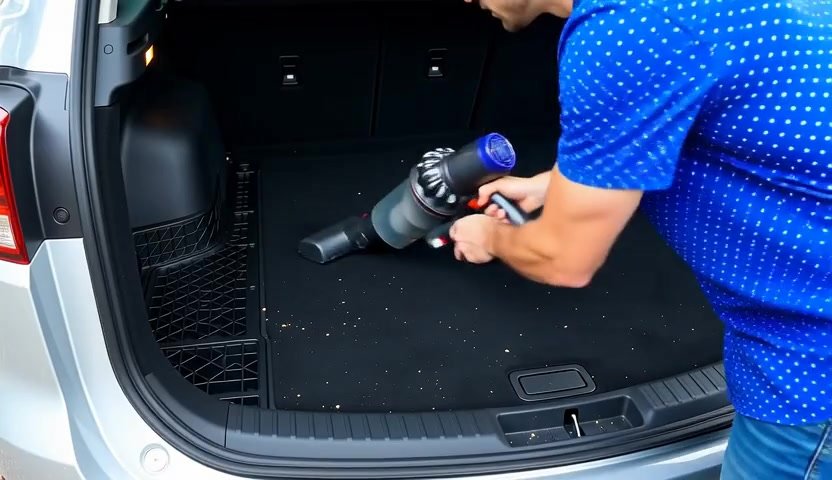}} & \adjustbox{valign=m}{\includegraphics[width=0.145\textwidth]{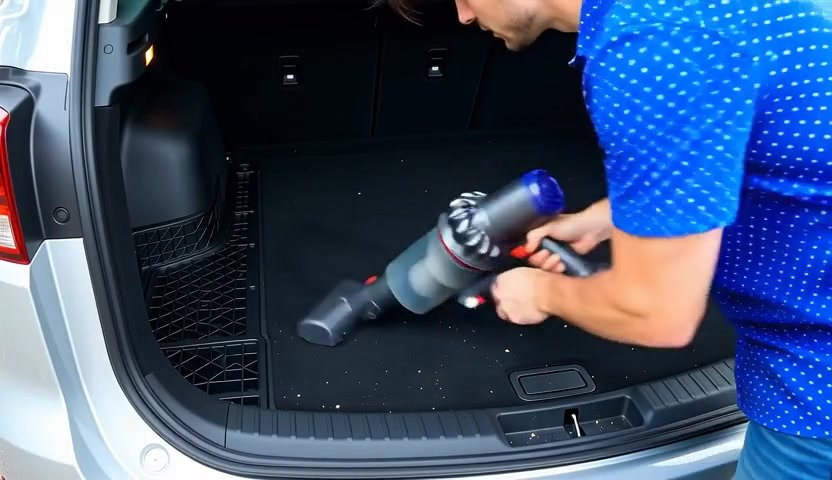}} & \adjustbox{valign=m}{\includegraphics[width=0.145\textwidth]{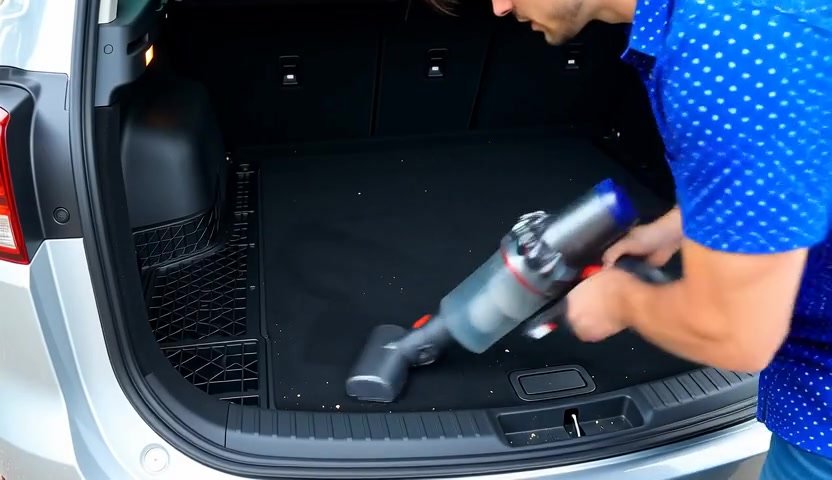}} \\[2pt]
        \QualRowLabel{\makecell{{\Oursbf{}} \\ {(Ours)}}} &
        \adjustbox{valign=m}{\includegraphics[width=0.145\textwidth]{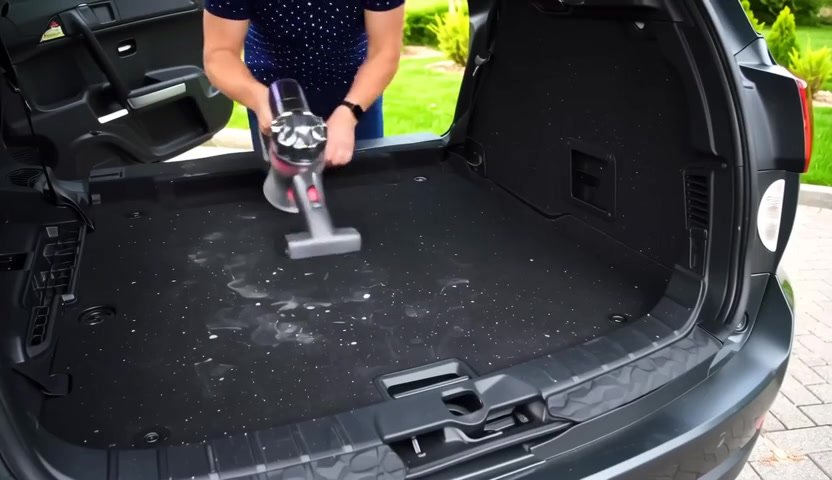}} & \adjustbox{valign=m}{\includegraphics[width=0.145\textwidth]{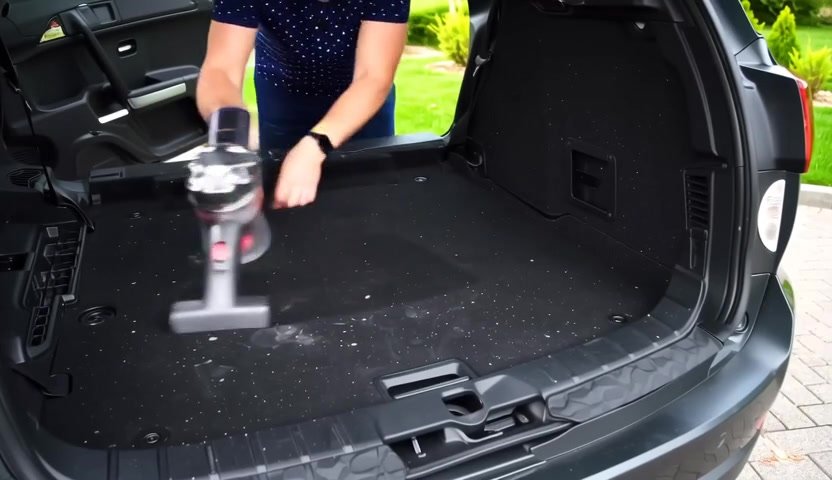}} & \adjustbox{valign=m}{\includegraphics[width=0.145\textwidth]{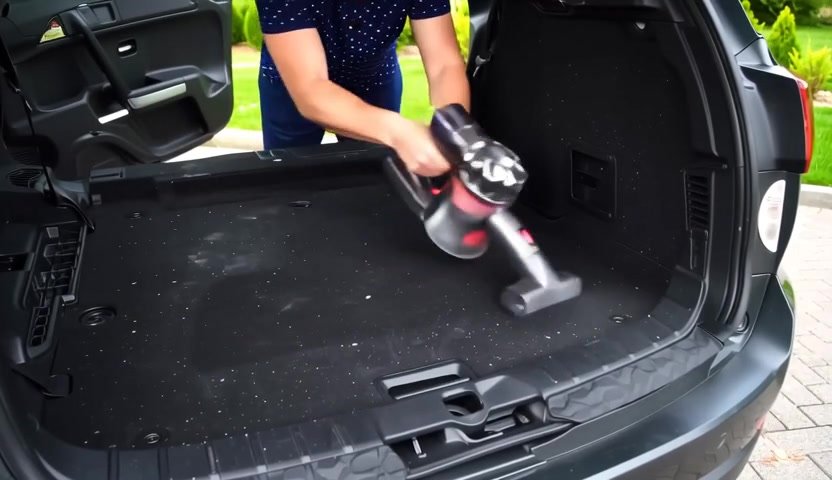}} & \adjustbox{valign=m}{\includegraphics[width=0.145\textwidth]{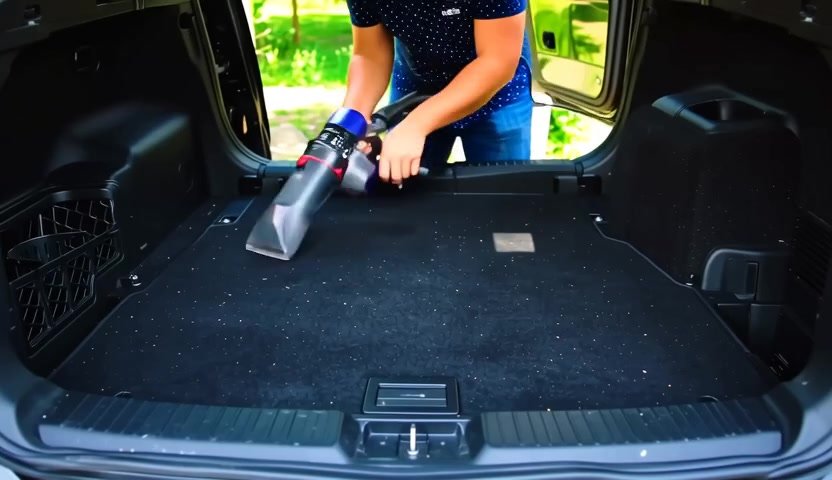}} & \adjustbox{valign=m}{\includegraphics[width=0.145\textwidth]{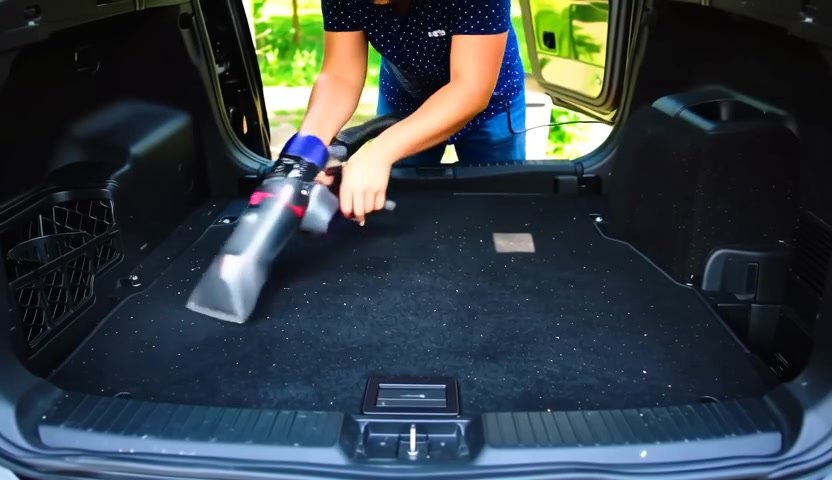}} & \adjustbox{valign=m}{\includegraphics[width=0.145\textwidth]{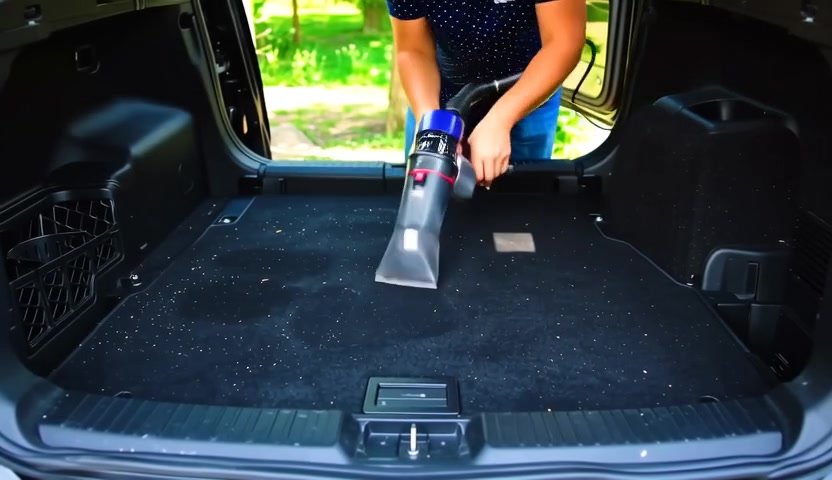}} \\[16pt]
        & \multicolumn{6}{c}{\parbox{0.90\textwidth}{\centering\tiny\itshape ``In the video, a person wearing a blue shirt with white polka dots is seen cleaning the trunk of a vehicle. They are using a handheld vacuum cleaner to remove debris from the black carpeted floor, which features a textured pattern. The individual bends down, maneuvering the vacuum to ensure thorough cleaning. The trunk's interior includes a black plastic tray with a grid-like design, likely for organizing items. The person's movements are deliberate and focused, indicating a meticulous approach to maintaining the cleanliness of the vehicle's storage area. The vacuum cleaner is small and portable, suitable for reaching into tight spaces within the trunk. The overall scene suggests a routine maintenance task aimed at keeping the vehicle tidy and functional.''}} \\[28pt]
    \end{tabular}
    \end{threeparttable}
    \end{minipage}
\vspace{-3pt}\end{figure}

\FloatBarrier

\fi

\end{document}